%% file: Causal_discovery_testbed_MAIN.tex
\documentclass[twoside,11pt]{article}

\usepackage{JMLR_style_modified}

\jmlrheading{1}{2015}{}{6/15}{}{Imme Ebert-Uphoff and Yi Deng}

\ShortHeadings{Using Causal Discovery to Track Information Flow}{Ebert-Uphoff and Deng}
\firstpageno{1}

\newcommand{\remove}[1]{}

\begin{document}

\title{Using Causal Discovery to Track Information Flow in Spatio-Temporal Data - A Testbed and Experimental Results Using Advection-Diffusion Simulations}

\author{\name Imme Ebert-Uphoff \email iebert@engr.colostate.edu \\
       \addr School of Electrical and Computer Engineering\\
       Colorado State University\\
       Fort Collins, CO, USA
       \AND
       \name Yi Deng \email yi.deng@eas.gatech.edu \\
       \addr School of Earth and Atmospheric Sciences\\
       Georgia Institute of Technology\\
       Atlanta, GA, USA}

\editor{}

\maketitle

\input Causal_discovery_testbed_TEXT_part1.tex

\input Causal_discovery_testbed_TEXT_part2.tex

\input Causal_discovery_testbed_TEXT_acknowledge.tex

\bibliography{./BN,./climate}

\input Causal_discovery_testbed_TEXT_appendix.tex

\vskip 0.2in

\end{document}

%% file: Causal_discovery_testbed_TEXT_part1.tex

\begin{abstract}

Causal discovery algorithms based on probabilistic graphical models have emerged in geoscience applications for the identification and visualization of dynamical processes. The key idea is to learn the structure of a graphical model from observed spatio-temporal data, which indicates information flow, thus pathways of interactions, in the observed physical system. Studying those pathways allows geoscientists to learn subtle details about the underlying dynamical mechanisms governing our planet. Initial studies using this approach on real-world atmospheric data have shown great potential for scientific discovery. However, in these initial studies no ground truth was available, so that the resulting graphs have been evaluated only by whether a domain expert thinks they seemed physically plausible. This paper seeks to fill this gap. We develop a testbed that emulates two dynamical processes dominant in many geoscience applications, namely advection and diffusion, in a 2D grid. Then we apply the causal discovery based information tracking algorithms to the simulation data to study how well the algorithms work for different scenarios and to gain a better understanding of the physical meaning of the graph results, in particular of instantaneous connections.  We make all data sets used in this study available to the community as a benchmark.

\end{abstract}




\section{Introduction}
\label{intro_sec}

Recent research has shown great potential for causal discovery algorithms 
to track information flow 
from observed data for geoscience applications.
The key idea for tracking information flow in geoscience is to interpret large-scale dynamical processes as information flow 
and to identify the pathways of this information flow by learning models from observational data.  
Since probabilistic graphical models are based on information-theoretical measures, they provide an ideal tool to track such information flow. 
%
We have obtained very promising results by applying structure learning of graphical models 
 to real-world atmospheric data. 
For example we compared information flow in two case studies, (1) boreal winter vs.\ summer
(\cite{EbDe:2012GRL})
and (2) current climate vs.\ projected climate in 100 years under global warming (\cite{DeEb:2014GRL}),
that provided new insights into the change of large-scale dynamics for these cases.
(Obviously, the latter comparison is based on data generated by climate models, rather than observed data.)

One gap in our analysis so far has been that there is never any exact ground truth available 
in climate data\footnote{Even 
  when using the output from climate models, we do not have information on the large-scale 
  dynamics, since the climate models utilize numerical equations localized 
  in both space and time, i.e.\ expressing the state of the system for each location 
  at the next time step based on that at the previous time step.  
  These equations themselves thus do not provide explicit information 
  on the large-scale interactions occurring in the climate system.},  
i.e.\ the only way to evaluate the results we obtained was to  
have the domain expert (second author of this paper) 
visually inspect the resulting graphs of information flow and consider 
whether they {\it seemed physically plausible} 
given the current knowledge in climate science about interactions in the atmosphere.
While this evaluation confirmed the potential of this new methodology, it leaves much 
to be desired.  
In particular, we did not have the tools to evaluate the accuracy of the method or to know 
how {\it exactly} to interpret the resulting networks.

The same type of gap existed, until recently, for a different type of network learned from climate data, namely {\it complex networks}.
Complex networks, also known as {\it climate networks}, 
were first proposed by \cite{TsRo:2004} and are a much simpler concept, 
exclusively based on Pearson correlation.
Namely, any two nodes are connected if and only if the Pearson-correlation of the corresponding data is above a chosen threshold.
(Note that the purpose of complex networks in geoscience applications
is to identify {\it similarities} between different locations, while 
the purpose of the structure-learning networks is to identify {\it interactions}
between different locations - a distinctly different purpose.)
Complex networks have been applied to climate data for over a decade
(\cite{TsRo:2004,TsSwRo:2006,YaGoHa:2008,TsSwKr:2007,DoZoMaKu:2009,StChGa:2010}),
and many insights have been drawn from them over the years, but
they had never actually been tested on simulation data until very recently. 
\cite{MoReMaKu:2014} finally filled this gap by testing complex networks on 
simulated data developed for that purpose
and then comparing the results to the known physics of the simulation data. 

Here we seek to achieve the same goal for structure-learning networks, 
which requires its own set of specific scenarios for testing.
Namely, we develop a simulation framework that models the two most important 
dynamical processes in the atmosphere, diffusion and advection, 
allowing us to generate simulation data for a great variety of different conditions and 
for which the exact dynamics are known.
We devise scenarios to test specific properties
of the information tracking method for such dynamical processes.
Each scenario consists of a choice of advection velocity field, advection and diffusion 
parameters, numerical parameters, spatial and temporal resolution, etc.,
and results in a data set.
We apply information tracking algorithms to those data sets and 
report the results, then  
we use those results as a guide to the interpretation 
of results obtained from real-world data. 

In addition 
we have created a supplemental website for this article 
that makes all of the resulting data sets, along with a description of the dynamic parameters, available to the community (see
{\small
\verb+http://www.engr.colostate.edu/~iebert/DATA_SETS_CAUSAL_DISCOVERY/+} \footnote{Should this site ever become unavailable, please send email to the first author at ebert@stups.com.}).
We hope that these datasets can become benchmarks 
for researchers to explore and compare a variety of methods for information flow tracking, 
including graphical models, but also Gaussian models (\cite{ZeFrLeHe:2014}) and Granger causal models (\cite{ArLiAb:2007}).
Having such benchmarks is important 
because (1) tracking information flow in spatio-temporal data 
generated from physical processes has not yet received much attention in literature;
and (2) we believe that this area has significant applications in a large range 
of geoscience applications and thus will gain in importance in coming years.


\subsection{Organization of This Article}

The remainder of this article is organized as follows. 
The remainder of Section \ref{intro_sec} describes the causal discovery 
algorithm used throughout 
this article and provides the motivating application for this study, 
namely tracking new insights about dynamical processes of our planet's atmosphere.
Section \ref{simulation_sec} describes how we generate data sets for testing, 
namely by using advection diffusion simulations.
Section \ref{simple_scenarios_sec} presents results for three simple scenarios and
Section \ref{complex_scenarios_sec} for three complex scenarios.
Section \ref{computational_effort_sec} briefly discusses computational time.
Section \ref{future_sec} discusses the results and suggests future work.


\subsection{Algorithm Details}

This section discusses the causal discovery method we have used in the past 
to track information flow in climate data 
and that is used here as well.
We employ the well known framework of 
{\it constraint-based structure learning} of graphical models 
(\cite{Pearl:1988,SGS:1993,Ne:2003,KoFr:2009}). 
We use the {\it PC stable} algorithm (\cite{CoMa:2012,CoMa:2014}), 
which is a variation of 
the classic {\it PC} algorithm (\cite{SpGl:1991,SGS:1993}).
{\it PC stable} 
has several advantages, namely
(1) it is order-independent, i.e.\ the order of variables does not affect the results;
(2) it is more robust, i.e.\ mistakes early on cause less follow-up mistakes in the graphs;
(3) it is easy to parallelize, thus reducing execution time.

The constraint-based structure learning yields independence graphs and 
we need to consider under which conditions these graphs can be interpreted 
to identify direct physical interactions, i.e.\ direct cause-effect relationships.
Going from probability distribution (data) to independence graph, we have to make 
sure that the obtained independence graph actually models the data well, 
i.e.\ that it is {\it faithful} to the probability distribution.
In our applications so far that is not a major concern.  
Even if faithfulness is violated
we still seem to get decent results.
Going from independence graph to causal interpretation, however, is a significant challenge,
since we must ordinarily ensure that the nodes 
in the graph are {\it causally sufficient}, i.e.\ if any two nodes $X, Y$ of the graph 
have a common cause, $Z$, then $Z$ must also be included in the graph.
In practice this condition is rarely satisfied in the geosciences, 
typically because some common causes may be unknown, hard to observe or including them
all would make the model too complex.
Some algorithms have been developed that can identify the existence of 
many hidden common causes (\cite{SGS:1993}), but are of high computational
complexity and currently not feasible for large graphs. 
Recent advances (\cite{Colombo:2012}) may change that in the near future.
Our approach is to simply {\it not} assume causal sufficiency, and to interpret the results 
accordingly.  
We accept that then we {\it cannot prove} causal connections, 
but we can {\it disprove} causal connections. 
Thus we can use an {\it elimination procedure} by first assuming that all variables 
are connected to each other, then disproving most of those connections until 
typically only few 
{\it potential} causal relationships are left at the end. 
Each one of those relationships may present a true causal connection, be due to a
common cause or a combination of both.

{\bf Evaluation step:}
Thus we include a final evaluation step in our analysis. 
In the final graph, every link (or group of links) must be checked by a domain expert. 
If we can find a mechanism that explains it (e.g.\ from literature), the causal 
connection is confirmed. 
Otherwise, the link presents a {\it new hypothesis} to be investigated. 

{\bf Scientific discovery:}
When seeking to learn new knowledge from data one interesting and common 
scenario is  
to have most links in the final graph confirmed from literature, thus confirming that 
the overall approach is correct, but also having a few unconfirmed links.
The unconfirmed links are the ones that provide {\it new} hypotheses of causal connections
and thus potentially {new knowledge}.
Another scenario is to have the results confirm known mechanisms, but to provide 
quantitative information to the extend of the mechanism.  
For example, it is known that storm tracks move poleward in a warming climate, 
but we may be able to provide additional information on which locations are affected the most
and how strong the effect is going to be, based on analyzing climate model data with this method. 

{\bf Incorporating Spatial Dimensions:}
We use a grid to incorporate spatial dimensions.
Any atmospheric field in the data, e.g.\ $X$, is represented by different variables,
$X_i$, that represent its values at the $i$th grid point.
While setting up the problem seems trivial at first, 
we showed in (\cite{EbDe:2014ICMLA}) that the spacing between the measurement locations 
can result in artifacts in the resulting graphs, so proper grid spacing 
- or at least understanding the potential problems if proper spacing is not possible - 
is critical.

{\bf Incorporating Time:}
For many applications in geosciences temporal information  
plays a crucial role. 
For example, the climate system is very dynamic, with states at individual locations 
changing from day to day,  
but interactions often also taking days to travel from one location to another,
while the strength of many signals decays significantly within days.
Therefore for our applications taking time into account provided much stronger causal signals.
In fact, for the climate science applications we considered, static models were unable 
to provide robust results, and we had to move to temporal models to be able to identify 
strong, robust causal signals (\cite{EbDe:2012JCLI,EbDe:2012GRL}).  We believe the same holds for many 
physical systems in which temporal order is important. 
Another advantage of temporal models is that 
temporal information helps to establish causal {\it directions}.

To incorporate time explicitly into the modeling we use the approach first introduced by 
\cite{ChDaGl:2005}, which adds lagged variables to the model.
Since this approach does not seem to be widely known, 
we briefly outline it in Appendix A.
Using that approach standard algorithms can be used to provide a temporal graphical model, 
but
the price we pay for this is much higher computational complexity, 
since we are now dealing with $(N \cdot S)$, rather than $N$ variables,
where $S$ is the number of time steps included in the model.
Furthermore, as discussed in \cite{EbDe:2014ICMLA}, proper initialization of the first time slices 
is a critical issue, but one that can be resolved easily by calculating the model 
for more time slices than needed and then discarding the first few time slices in the results.

\subsection{Sample Application: Tracking Information Flow in the Earth' Atmosphere}

A very promising application of causal discovery in climate science is 
to track the pathways of physical interaction around the globe.
In order to do that we define a grid around the globe and evaluate an atmospheric
field (such as temperature or geopotential height) at all grid points, 
which provides time series data at all grid points.
Our approach is to   
then use the temporal version of {\it PC stable} 
to identify the strongest {\it pathways of interactions} around the globe based on the time series data (\cite{EbDe:2012GRL}). 
(Gaussian graphical models present an alternative approach for this
purpose (\cite{ZeFrLeHe:2014}) and \cite{Runge:2014dis} investigates this and 
other approaches.)
No matter which method is used, 
the key idea is to interpret large-scale atmospheric dynamical processes as information flow 
around the globe and to identify the pathways of this information flow (physical interactions)
using causal discovery.

%
\begin{figure*}
%
\centerline{ 
\includegraphics[width=5.0cm,angle=0]
{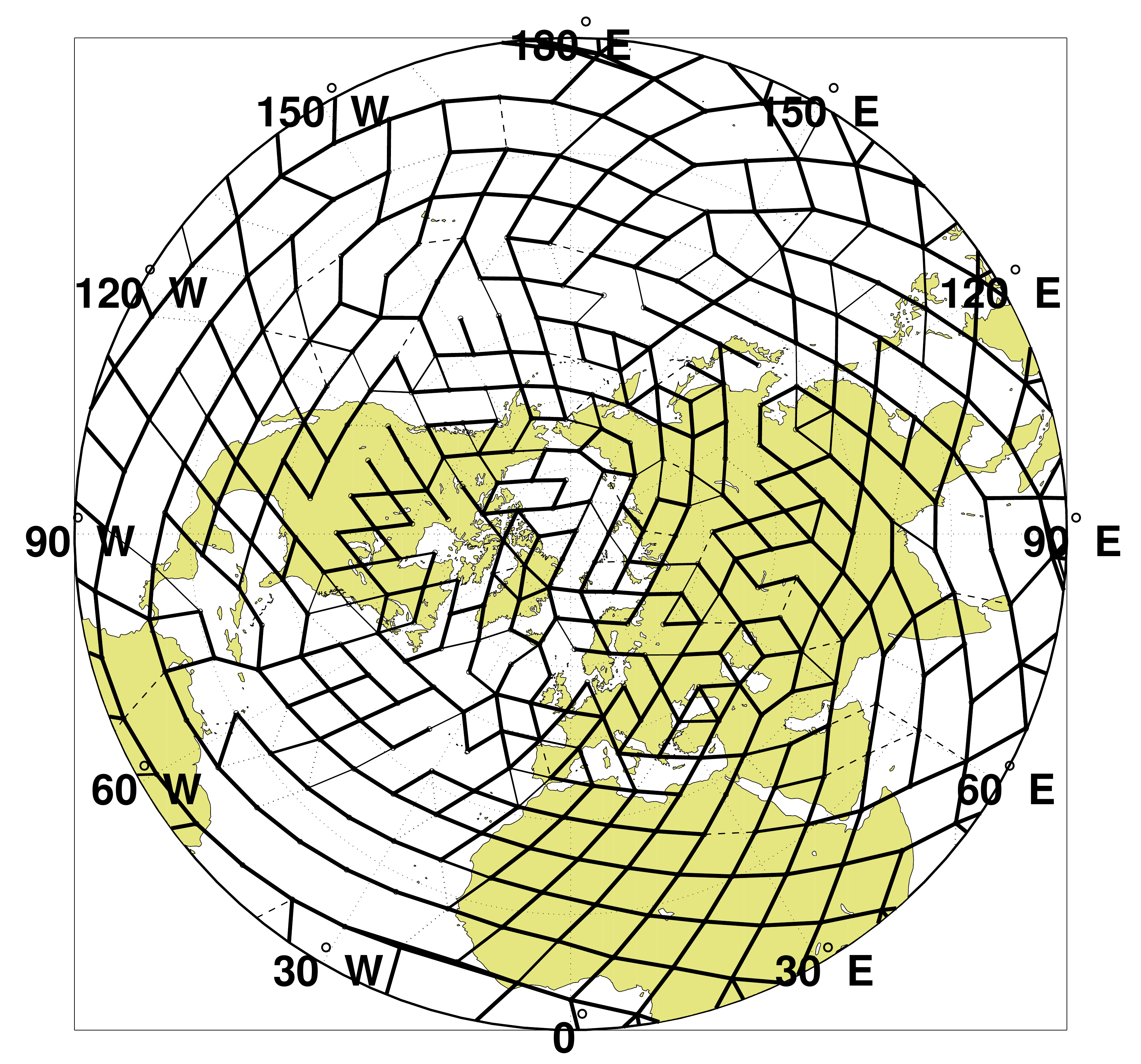}
\hspace{0.0cm}
\includegraphics[width=5.0cm,angle=0]
{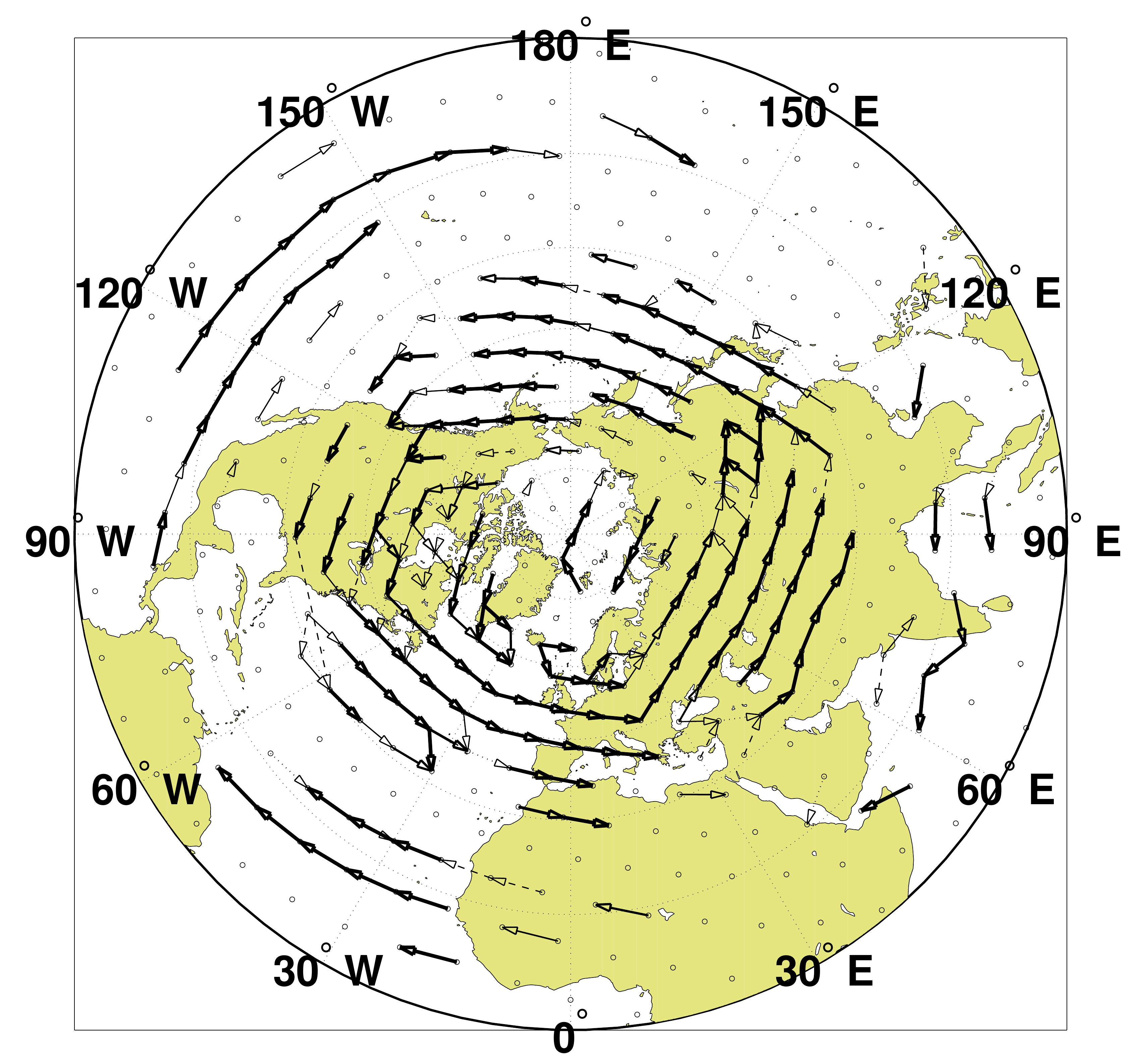}
\hspace{0.0cm}
\includegraphics[width=5.0cm,angle=0]
{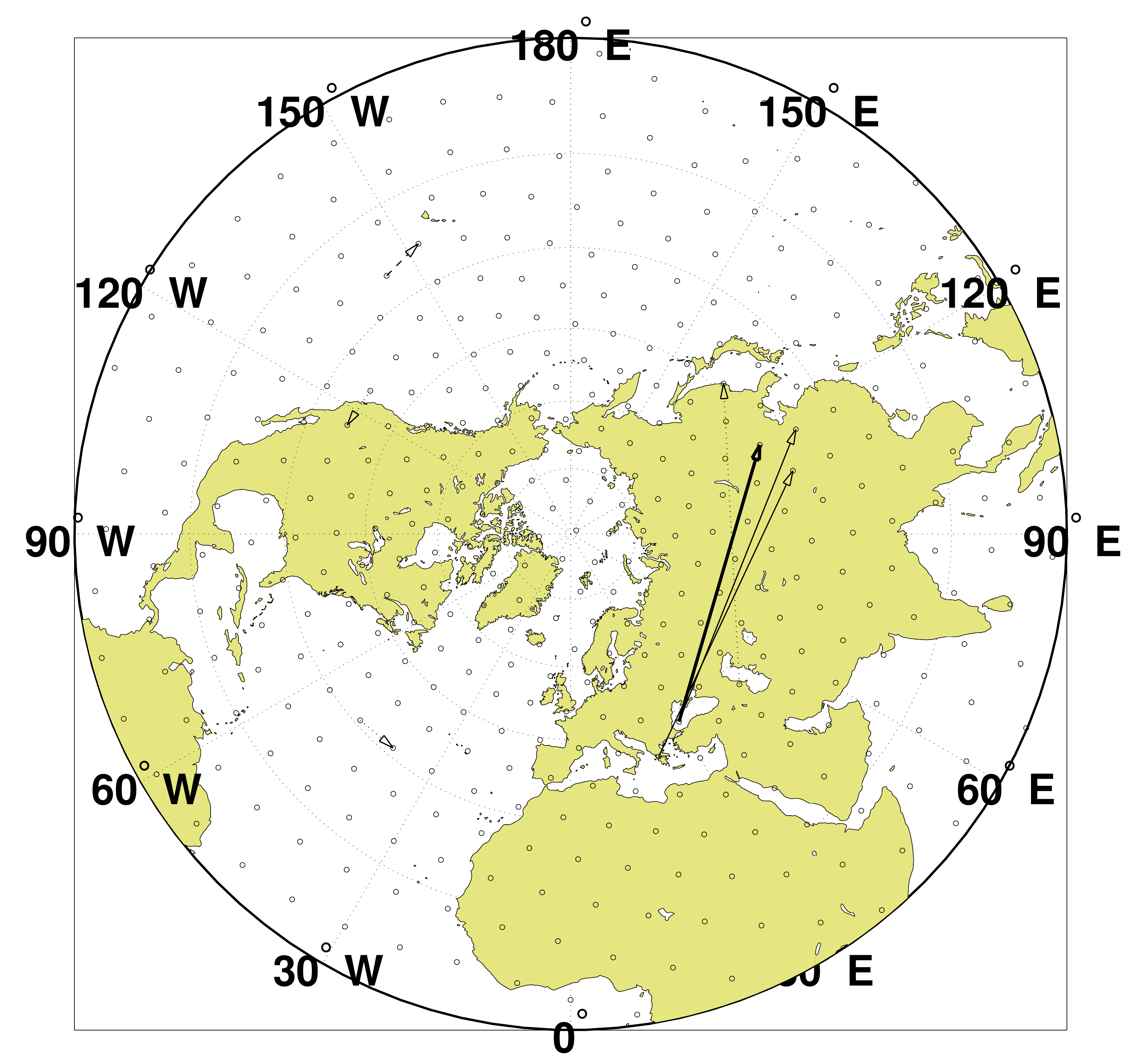}
}
\centerline{(a) $T = 0$ days \hspace*{2.2cm} (b) $T = 1$ day \hspace*{2.2cm} (c) $T = 2$ days  }

\caption{Network plots for Northern hemisphere from {\it PC stable} ($D=1$ day, $\alpha=0.1$)
   based on Fekete grid with 800 grid points. \label{Fekete_800_fig}}
\end{figure*}
%

Figure \ref{Fekete_800_fig} shows sample network plots obtained 
from atmospheric data
using {\it PC stable} with $D=1$ day between time slices and significance level 
$\alpha=0.1$.
The data used is daily NCEP-NCAR reanalysis data (\cite{Ka:1996,Ki:2001})
for geopotential height at 500mb 
for boreal winter months (Dec-Feb) of years 1950-2000.
Fig.\ \ref{Fekete_800_fig}(a), (b) and (c)
show the strongest direct connections identified that take $0, 1$ and $2$ days, 
respectively, to travel from source to target. 
%
%
These graphs were obtained by first generating temporal graphs with lagged variables, 
then converting them to 
summary graphs 
summarizing the strongest connections.
It turns out that the interactions captured in Figures \ref{Fekete_800_fig}(b) and (c)
are storm 
tracks\footnote{Which 
   physical processes are tracked in the network 
   depends primarily on two factors, the atmospheric field used and 
   the time scale (e.g.\ daily data vs. monthly data).
   Thus using 
   a variety of different atmospheric fields we can track the causal pathways 
   of a variety of different dynamical processes around the globe or in 
   specific locations of interest.}.
However, we were never able to fully understand the many 
interactions identified in Figure \ref{Fekete_800_fig}(a).
{\it What exactly do the apparently instantaneous connections 
between neighboring locations in Figure \ref{Fekete_800_fig}(a) represent?}  

This question was a strong motivation for 
the work reported in this article, namely to gain a better understanding 
of how {\it exactly} the underlying dynamical processes are represented in the network plots
we obtain, with special emphasis on concurrent edges ($T=0$), i.e.\ those represented 
in graphs such as Figure \ref{Fekete_800_fig}(a).
Furthermore, causal discovery algorithms have 
rarely been used (1) for spatio-temporal systems and 
(2) for physical systems.
Studying advection and diffusion processes from simulations serves as an excellent 
benchmark to test such algorithms for those types of applications.







\section{Generating Datasets for Testing}
\label{simulation_sec}

\begin{figure}
\centerline{ 
\includegraphics[width=2.3cm,angle=270,clip=]{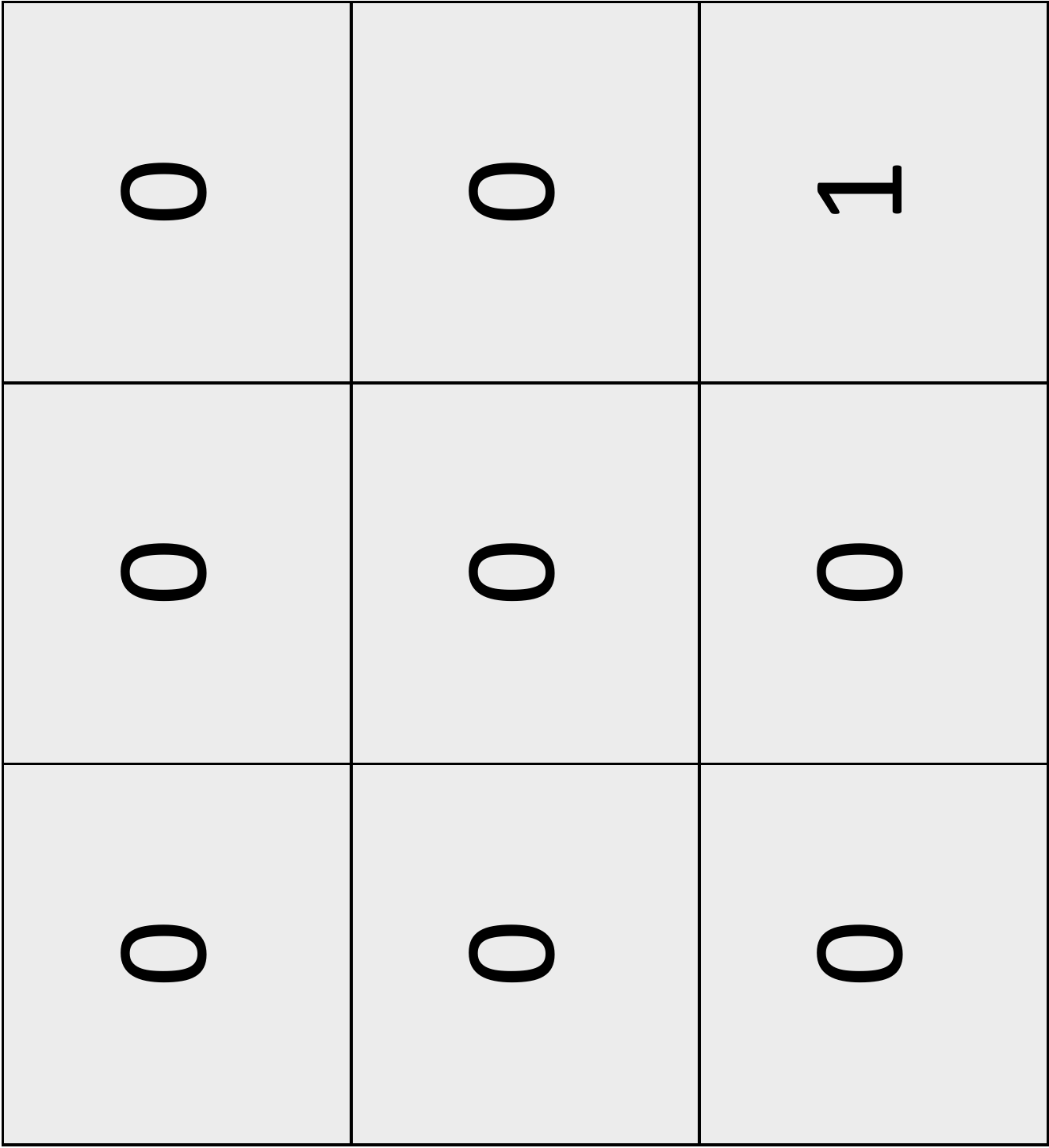}
\hspace{3.0cm}
\includegraphics[width=2.3cm,angle=270,clip=]{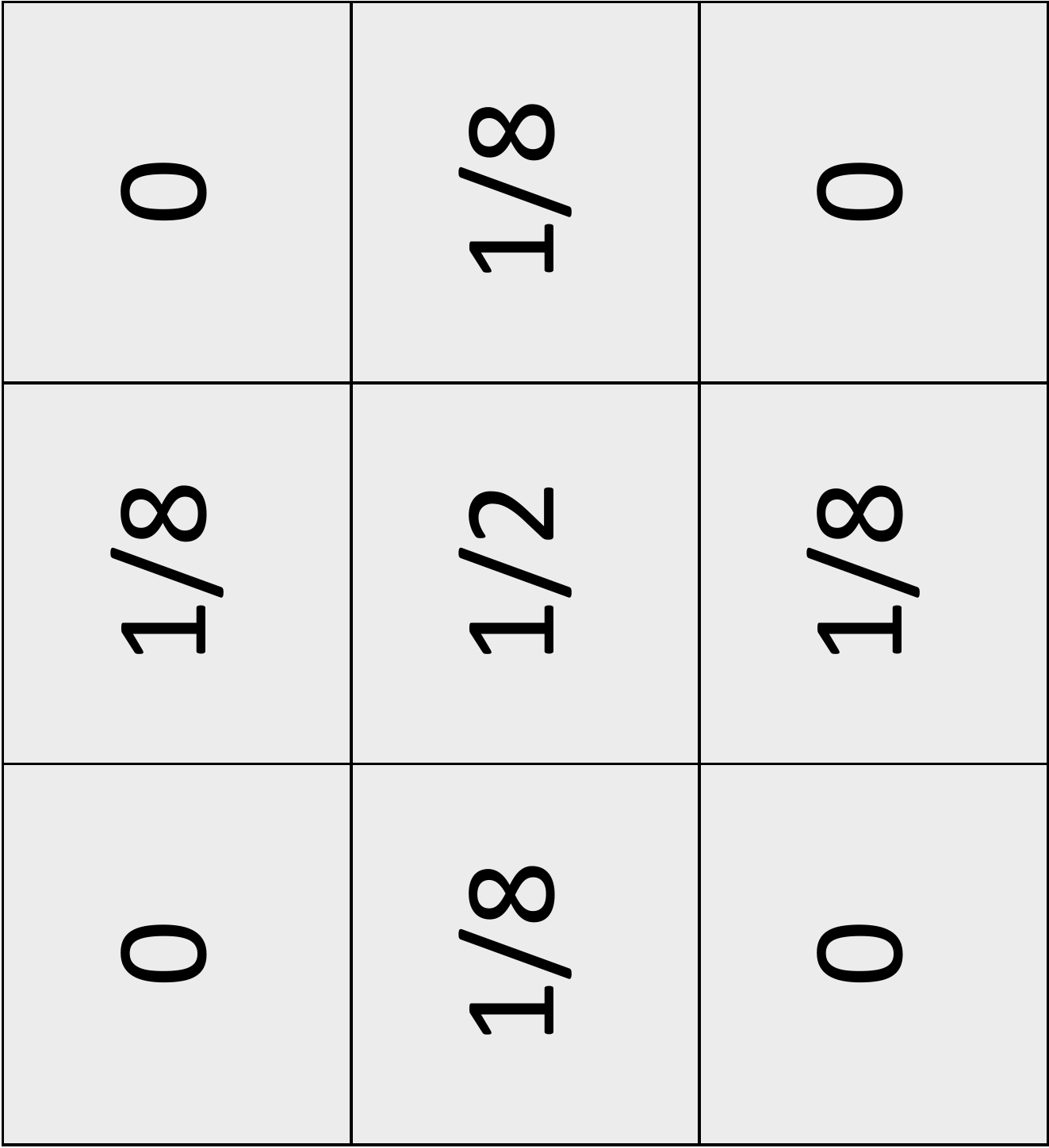}
}
\centerline{(a) Advection-like filter \hspace*{1.7cm}  (b) Diffusion-like filter \hspace*{0.2cm}}
\caption{Examples of convolution filters corresponding to advection and diffusion processes. \label{convolution_filters}}
\end{figure}

In our simulations we model two processes, advection and diffusion, 
in a two-dimensional grid.
Advection and diffusion are common - and often dominant - 
processes in many dynamical processes in nature, especially in the geosciences. 
%
Advection is often described as a transport mechanism of a substance or property 
by a fluid (or air) due to the fluid's bulk motion.  
An example is the transport of {\it heat} by a moving fluid.
The motion of the fluid is described by a vector field that is constant over time, 
while the temperature is described by a scalar field that changes over time.
In the context of this study, where we interpret changes of properties 
- such as temperature, pressure, etc.\ - as {\it signals},
we can think of an advection process
as shifting a signal without changing the shape of the signal.
In the language of image processing or digital filters,
this can be understood as applying a pure translation filter, 
such as the convolution operator shown in Figure \ref{convolution_filters}(a).

In contrast, diffusion causes a signal to {\it spread} while the peak of the signal
remains in place, e.g.\ any narrow wave of high amplitude 
is spread out into a wide wave with much lower amplitude.
An example of diffusion would be to inject a small amount of hot water 
within a large amount of resting cold water, 
then diffusion would slowly spread the heat throughout the water 
until a new equilibrium (constant temperature throughout) is reached.
In the language of digital filters, Fig.\ \ref{convolution_filters}(b) 
provides a sample convolution filter for a diffusion process.
The dominant processes in many geoscience applications 
can be modeled as a combination of both processes,
advection to transport a signal and diffusion to spread it. 
Therefore we selected those two processes for our simulation framework.

\subsection{Simulation Concepts and Parameters}

While advection and diffusion can represent many processes in nature, 
here we focus on one physical scenario for illustrative purposes.
We assume that we are modeling a moving fluid 
and the property of interest is the temperature at different locations over time.
We denote as $f(x,y)$ the temperature at any point $(x,y)$. 
The motion of the fluid is described by a vector field, ${\bf V}(x,y)$,
that specifies a velocity vector for any location $(x,y)$. 
$\kappa_x$ and $\kappa_y$ specify the diffusion coefficients in $x$ and $y$ direction.
For $\kappa_x = \kappa_y=0$ there is no spreading of the signal, while 
increasing values of $\kappa_x, \kappa_y$ indicate increased spreading of the signal.
Appendix B reviews the corresponding partial differential equations 
governing the advection diffusion process and the numerical implementation thereof
that we use to generate the simulation data.  
In this section we discuss only those concepts and parameters of the simulation
that are necessary to understand the resulting data sets.

%

{\bf Numerical Grid:}
We use a rectangular grid for the numerical calculations.
Its primary parameters are
   $\Delta t$, the time step for numerical calculation, and
   $\Delta x, \Delta y$, the distance between neighboring grid points 
   in $x$ and $y$-directions. 

{\bf Periodic Boundary Conditions:}
We use periodic boundary conditions to emulate the behavior of 
a large (infinite) system using just a small area.
This means that we use a wrap-around in both x- and y-direction.  
For example, when reaching the right-most grid point in the x-direction, its neighbor to the 
right is defined to be the left-most grid point with the same y-coordinate, i.e. we jump from 
the last point in a row to the first point in the same row.
The same applies in the upward (y-) direction.

The governing differential equation (Equation (\ref{adv_dif_pde}) in Appendix B), 
along with the periodic boundary conditions, and a set of 
initial conditions describing the 
temperature distribution at a time $t_0$, defines the temperature distribution over time.
We can approximate this temperature distribution using a discrete grid, 
resulting in a numerical version of Equation (\ref{adv_dif_pde}).
We use the first-order upwind scheme for the numerical implementation, which
is described in detail in Appendix B.

{\bf Guaranteeing numerical stability:}
To keep the numerical calculations 
stable there are several conditions on how to choose the time step used in the 
numerical calculations,
which depend on the advection velocity and diffusion coefficients used.
For example, for pure advection and the 1-dimensional case
the Courant-Friedrichs-Lewy (CFL) condition requires that the time step must be chosen
such that
\begin{equation}
   \Delta t \le C \cdot \frac{\Delta x}{V_{max}},
   \label{CFL_eq}
\end{equation}
where $C$, the Courant number, is between 0 and 1 ($0 < C \le 1$),
and $V_{max}$ is the maximal absolute value of the velocity field.
For the two-dimensional case we use the condition above with
$V_{max}$ as the maximal magnitude of the velocity field and instead of 
$\Delta x$ we use the diagonal pixel length ($\sqrt{\Delta x^2 + \Delta y^2}$) as a first guideline.
The diffusion term has its own requirements for the time step, but those are not discussed here.

{\bf Numerical diffusion:}
In addition to the diffusion from the actual diffusion term, the numerical implementation creates additional diffusion effects, the amount of which depends on 
$C$ and other factors.  
In the special case where the advection velocity aligns with the 
vertical or horizontal grid directions, there is no numerical diffusion for $C=1$.
Generally, numerical diffusion increases for decreasing values of $C$.

{\bf Signal speed and temporal resolution:}
$V_{max}$ is the maximal speed of the velocity field, and thus also the maximal 
signal speed - signal meaning in this context a change of temperature in the fluid -, 
since all signals travel with the fluid.  
As we have to obey Eq.\ (\ref{CFL_eq}) for numerical stability,
it follows that we always have $\Delta t V_{max} \le \Delta x$, 
i.e.\ the maximal distance traveled in one time step of the numerical calculations
is always smaller than the width (diagonal length in 2D) of a grid pixel.  
In other words, the signal can never travel 
across more than one grid point at a time in the numerical calculations.
Thus, in order to be able to experiment with higher signal speed for information tracking, 
i.e.\ signals
crossing more than one grid point at a time, we may choose to only save 
data for every $M$th sample when generating data sets, i.e.\ the time step 
in the data file is $\widehat{\Delta t}= M \Delta t$.
This procedure thus reduces the sample time {\it after} the numerical calculations are completed.

\subsection{Information Sent to the System (Initial Conditions and Forcings)}

The equilibrium state is for all grid points to have the same temperature. 
We send information (messages) to the system by injecting signals that disturb
that equilibrium, either as initial conditions (IC) or as external forcings,
then let the message pass through the system.
The type of initial condition we use is as follows:
\begin{itemize}
\item
   {\bf Message Type 1: Single-point peak initial conditions ({\it IC peak})}\\
   First the temperature of a {\it single grid point} is set to a much higher 
   value {\it at a single time step}, then we let the resulting 
   signal propagate throughout the system until it dissipates.
   We send IC single-point peaks  
   sequentially to all grid points, waiting for each signal to propagate, before 
   restarting the system with initial conditions for the next grid point, 
   thus creating as many consecutive runs as there are grid points.
\end{itemize}
Injecting a signal only at one point at a time assures that 
there is only a single signal propagating in the system at any given time.
However, one potential problem with the peak approach is that for systems 
with high diffusion the signal may dissipate quickly and then there is no more 
signal left to track.
Furthermore, real-world data sets are likely to include also other types of messages,
such as continuous external forcing.
Therefore we also include a second type of signal, continuous noise, as external 
forcing.
We inject noise in two different ways:
\begin{itemize}
\item
   {\bf Message Type 2: Continuous single-point noise forcing ({\it single-point noise})}\\
   We feed normally distributed noise continuously to a single grid point.  
   We do that for one run,
   then repeat for the next grid point.  That assures that information is continuously 
   fed to one grid point at a time. Even though the message (noise) is less crisp than 
   a single high peak at that point,  
   this method guards against the quick decay problem that 
   can arise when feeding only a single peak.
\item
   {\bf Message Type 3: Continuous all-point noise forcing ({\it all-point noise})}\\
   To make things more realistic, we can inject 
   normally distributed noise continuously and simultaneously 
   at all grid points.
\end{itemize}
Both the single-point and all-point noise   
are injected at every time step while the simulation is running, i.e.\ 
whatever noise is added at a grid point is transferred along with the rest of the signal
at the next step.  We call this {\it prior} noise, since it is added before the signal 
propagates in the simulation.
An example of prior noise sources in real-world data is an un-modeled physical source
of disturbances at grid points, such as spontaneously excited atmospheric convection over tropical oceans due to local convective instability, and large-scale disturbances that grow in midlatitudes as a result of hydrodynamic instability.
Prior noise can be used as background noise in addition to a peak signal,
or by itself as the primary signal.

In contrast, noise that is added to the time series {\it after} the simulation has completed 
we call {\it posterior noise}. That noise is {\it not} transferred according to 
the advection-diffusion equations and thus does {\it not} result in information transfer. 
An example of posterior noise sources in real-world data could be measurement errors
that are fairly erratic. 
Posterior noise is not included in the simulations here, but can be easily added to 
the data sets provided by simply adding noise to the final signals.

\subsection{Types of output edges and when we {\it expect} to see them}

When interpreting the output of the causal discovery algorithm 
from spatio-temporal data, we distinguish three types of edges,
(1) {\it intra} edges, (2) {\it concurrent inter} edges and (3) {\it nonconcurrent inter} edges. 
These edge types are explained below only for the advection-diffusion case, 
where the quantity of interest is temperature $f$. 
For the general case $f$ is simply substituted by {\it any} type 
of physical quantity considered or combinations thereof. 
In the following we denote as $f_i(x_j,y_k)$ the temperature at time $i$
at the grid point with coordinates $(x_j,y_k)$.

An {\bf intra edge} connects the temperature $f_i$ at 
some location $(x_j,y_k)$
to the temperature at the same location at a later time:
   \begin{displaymath} 
   \qquad f_i(x_j,y_k) \longrightarrow f_{{i}^{'}}(x_j,y_k),
   \quad \mbox{where} \; i^{'} > i.
   \end{displaymath} 
Intra edges encode the {\it local memory} (aka persistence in climate science) 
of a variable, i.e.\ for how long the current temperature at a location 
significantly affects the temperature at the {\it same} location.
Note that by definition intra edges can only be nonconcurrent, i.e.\ not within 
the same time step. 

An {\bf inter edge} connects the temperature 
at some location to the temperature at a {\it different} location.
Inter edges encode the information flow between different locations 
and thus track the {\it remote impact} of any location, i.e.\ its role in 
long-distance information transfer.
%
{\bf Concurrent inter connections}
connect temperatures between different locations, but {\it within the same time step},
and thus generally end up being undirected (no arrow head) in our type of analysis
(since we use temporal order to determine edge directions):
      \begin{displaymath} 
         f_i(x_j,y_k) \; \hbox{---} \; f_{i}(x_{j^{'}},y_{k^{'}}), \;
         \mbox{where } \; 
         j^{'} \neq j \; \mbox{or} \; k^{'} \neq k.
      \end{displaymath} 
{\bf Nonconcurrent inter connections} connect temperature between different locations 
{\it and} different times, and are directed from earlier time to later time.
     \begin{displaymath} 
         f_{i}(x_j,y_k) \longrightarrow f_{i^{'}}(x_{j^{'}},y_{k^{'}}), \;
         \mbox{where} \; i^{'} > i \; \mbox{and} \; 
         (j^{'} \neq j \; \mbox{or} \; k^{'} \neq k).
      \end{displaymath}

{\bf Expectations:}
Before we delve into the simulation results, we take a moment to formulate when we had 
{\it expected} the different types of edges to occur.
\begin{enumerate}
\item  
   {\bf Intra edges:} 
   We expected intra edges to be dominant only for stationary processes 
   and nearly stationary processes, i.e.\ in this context 
   for pure diffusion or wherever advection velocities are very small.
\item
   {\bf Nonconcurrent inter edges:}
   We expected nonconcurring inter edges to dominate for processes with significant 
   velocities, i.e.\ in this context whenever there is significant advection velocity.
\item
   {\bf Concurrent inter edges:}
   We expected concurrent inter edges to occur only for extremely high speeds, 
   namely in cases where the signal moves across many grid points in a single time step,
   thus connections being so fast that they appear to be almost instantaneous.
\end{enumerate}
As we will see in the following sections, while our expectations were 
confirmed for the first two edge types, 
the simulations provided some 
surprising new results concerning the occurrence and role of the third type,
concurrent inter edges.

%% file: Causal_discovery_testbed_TEXT_part2.tex
\section{Simulation Results for Three Simple Scenarios}
\label{simple_scenarios_sec}

\begin{figure*}
\centerline{ 
\includegraphics[width=5.0cm,angle=0]{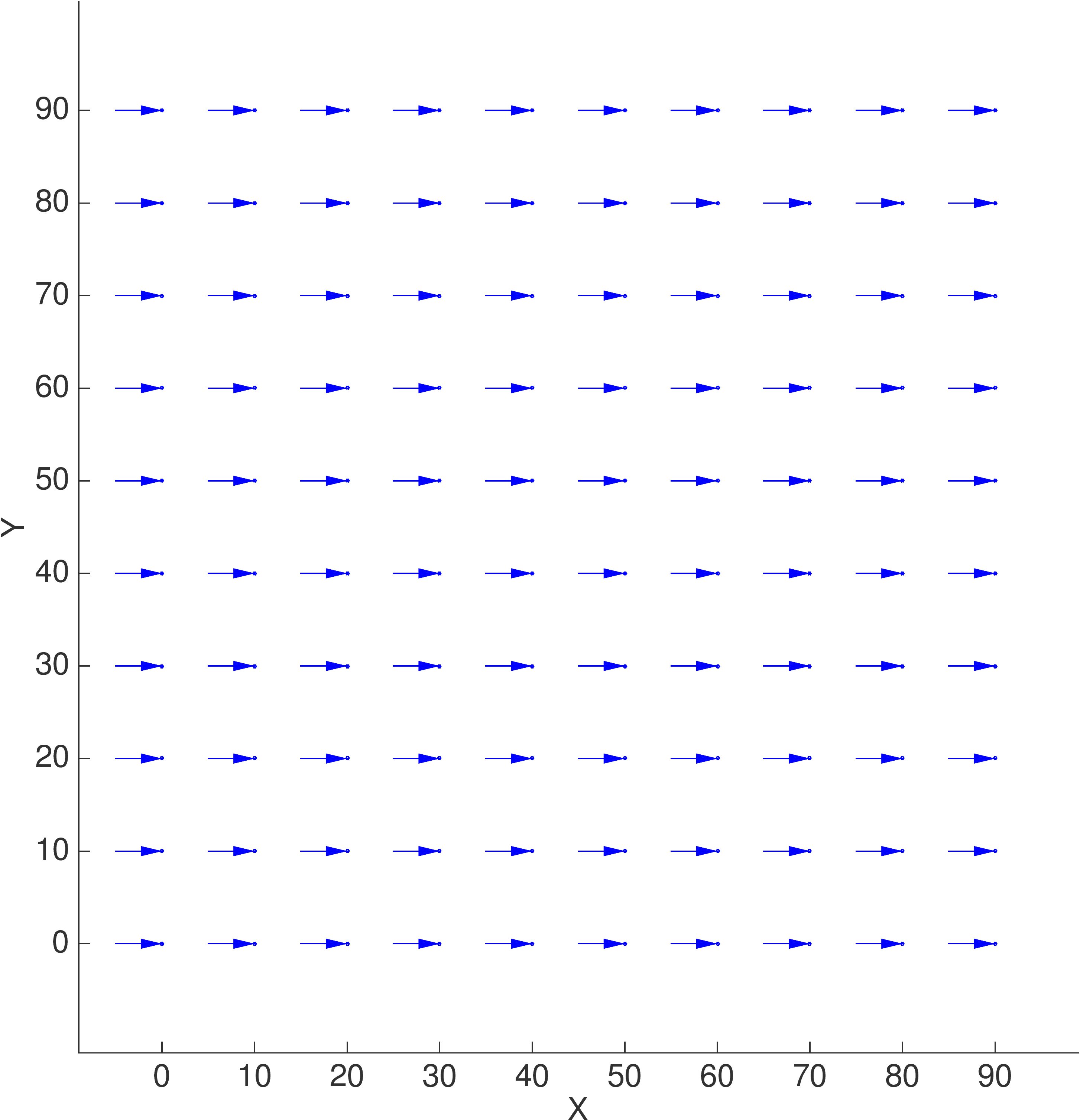}
}
\caption{Vector field for advection displayed as displacement (in meters) within 
   $\Delta t = 5$ sec.  \label{vector_field_straight_to_right_fig}}
\end{figure*}

In this section we show results for three highly simplified scenarios, 
designed to test the information tracking 
algorithms when only one single type of process, or a very simple combination thereof, is present.
The three simple scenarios are (a) pure diffusion, (b) pure advection and (c) mixed advection and diffusion. In those two cases involving advection the advection velocity field is trivial, 
namely, as shown in Fig.\ \ref{vector_field_straight_to_right_fig},
the velocity is identical at each point and points straight to the right.
%
%
%
Once we understand the {\it information flow signatures} for these simple
scenarios we move on to three more complex scenarios in Section \ref{complex_scenarios_sec}
that incorporate both advection and 
diffusion and use more interesting and realistic
advection velocity fields, representing different types of 
circular and cross current motion.

Parameters that are identical for all simulations are listed in Appendix C, while
parameters specific to each scenario are listed in the corresponding sections.
The supplemental website provides the corresponding data sets for all scenarios
discussed in this article.

\subsection{Types of Figures Generated}

\begin{figure*}
\centerline{ 
\includegraphics[width=3.5cm,angle=0]{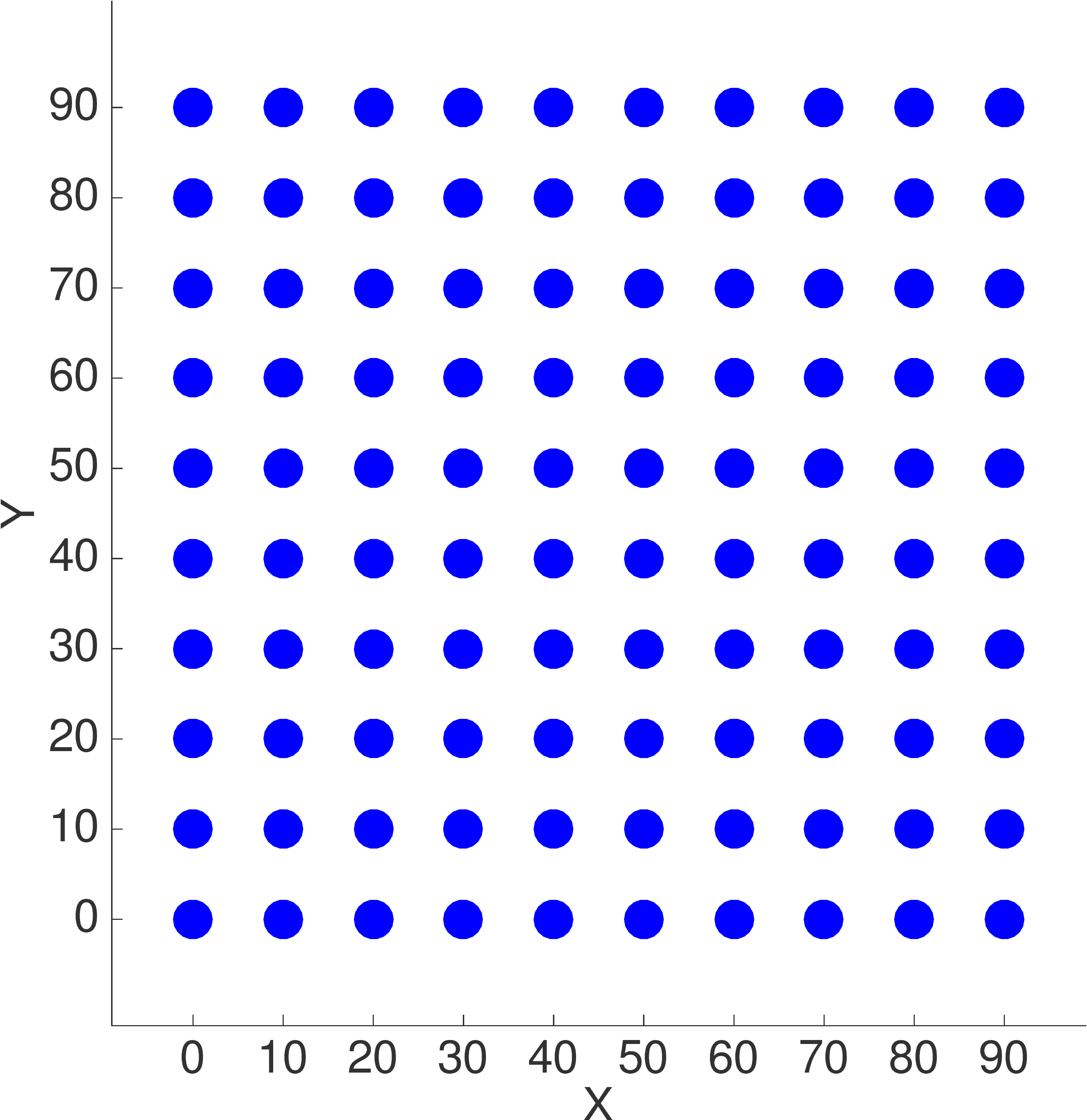}
\hspace*{0.5cm}
\includegraphics[width=3.5cm,angle=0]{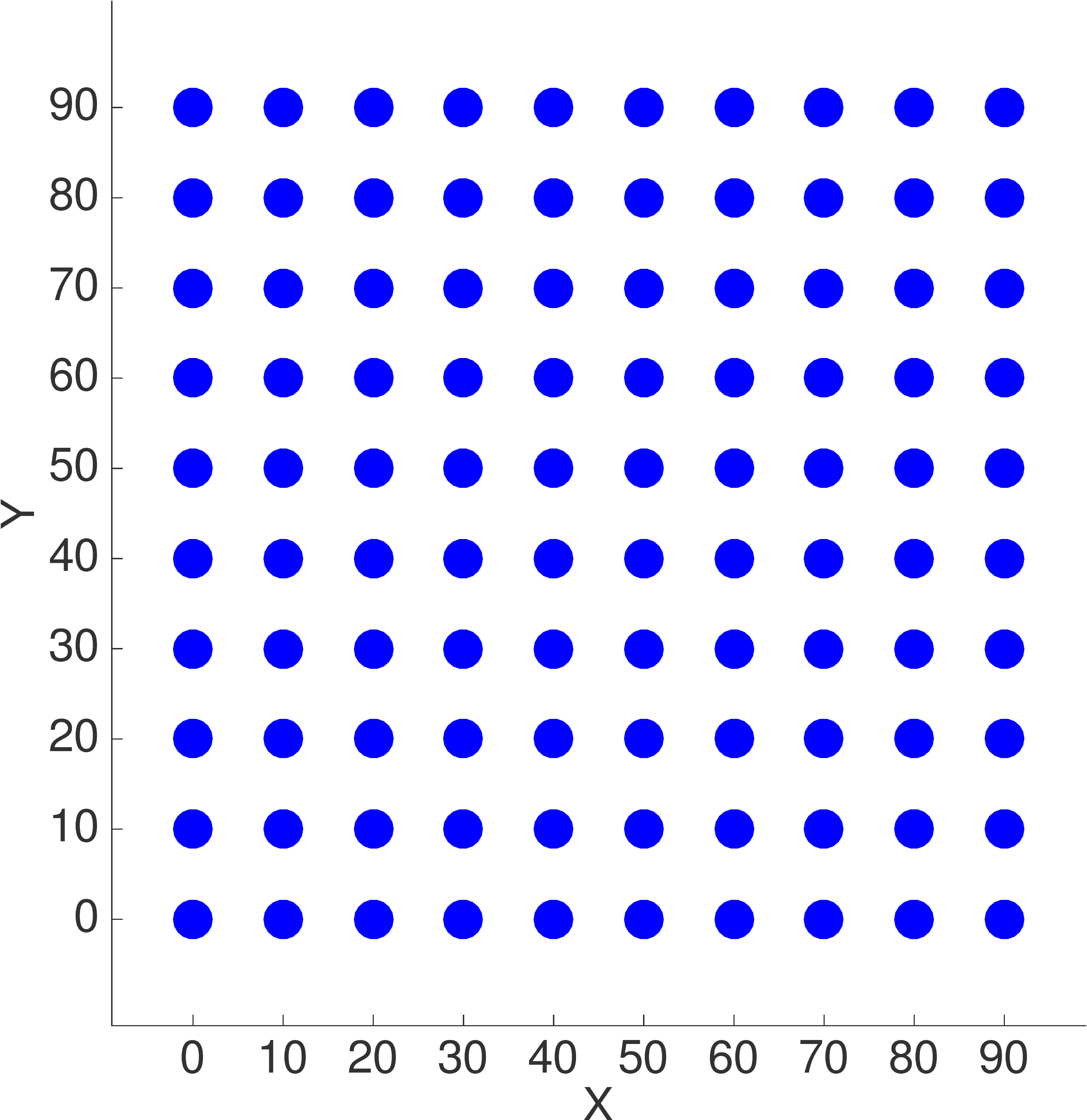}
\hspace*{0.5cm}
\includegraphics[width=3.5cm,angle=0]{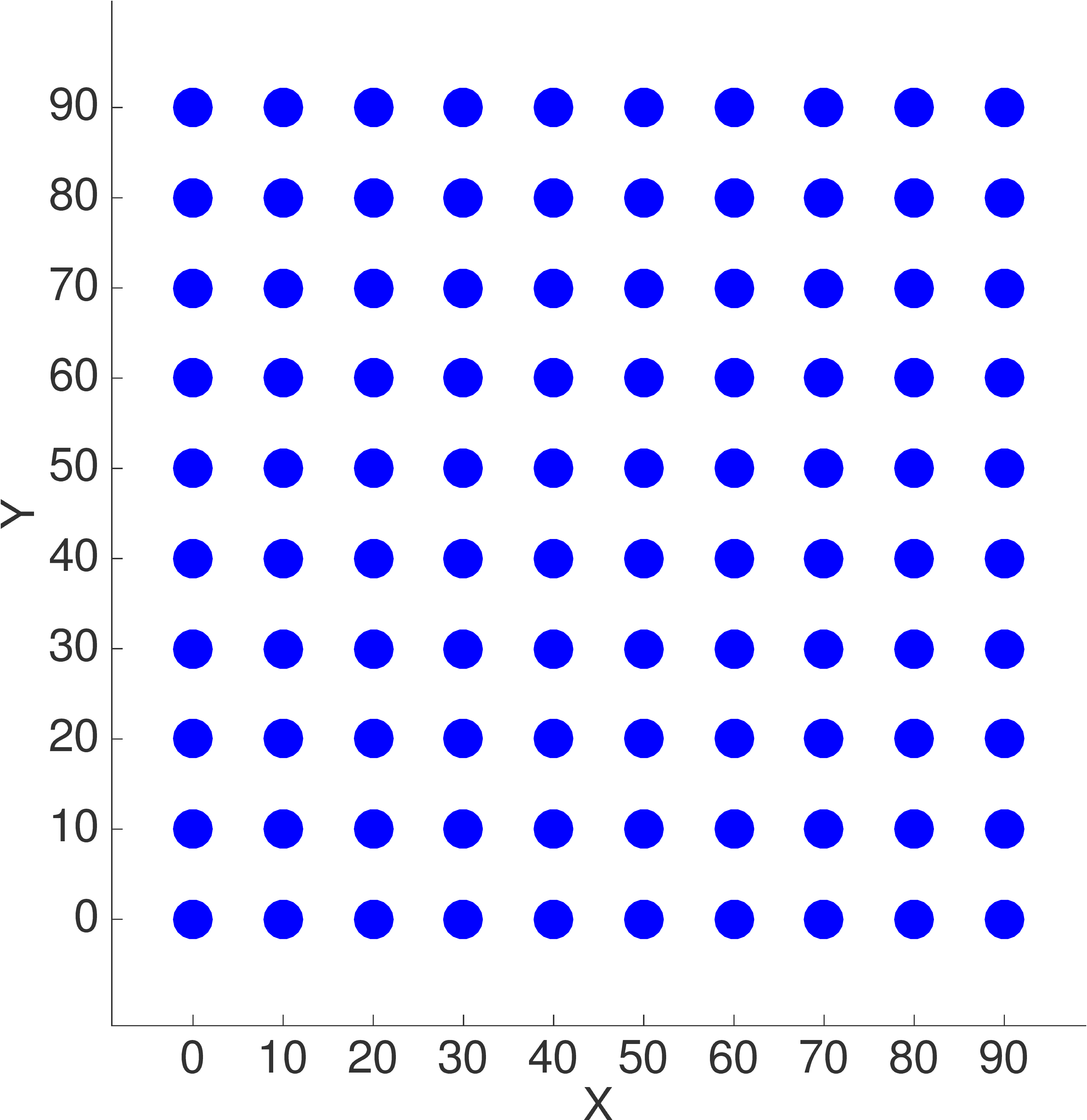}
\hspace*{0.5cm}
\includegraphics[width=3.5cm,angle=0]{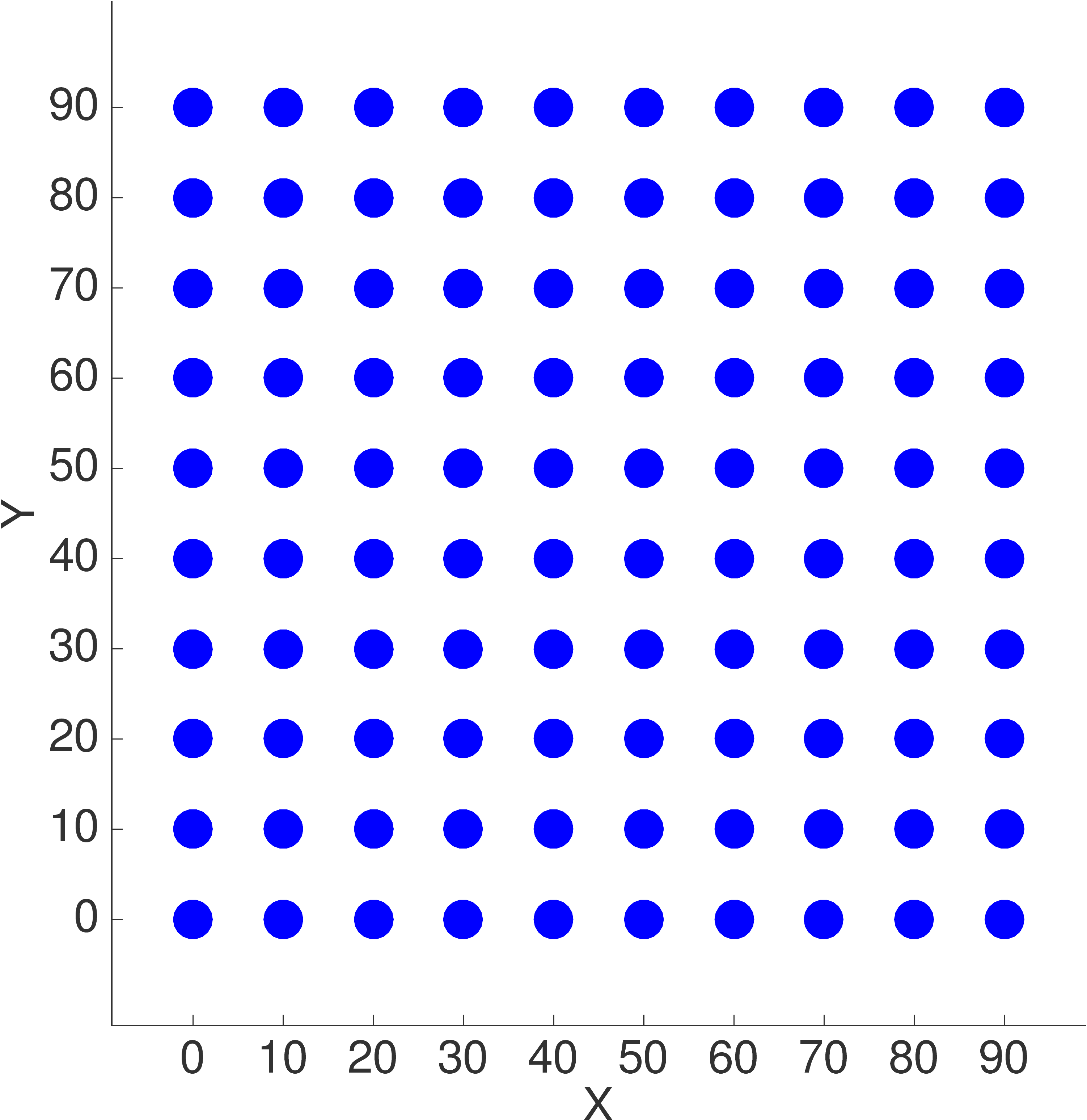}
}
\centerline{$T = \Delta t$  \hspace*{2.5cm} $T = 2 \Delta t$  \hspace*{2.5cm} $T = 3 \Delta t$ \hspace*{2.5cm} $T = 4 \Delta t$}
\centerline{(a) Intra edges}
\vspace*{0.3cm}
\centerline{ 
\includegraphics[width=3.5cm,angle=0]{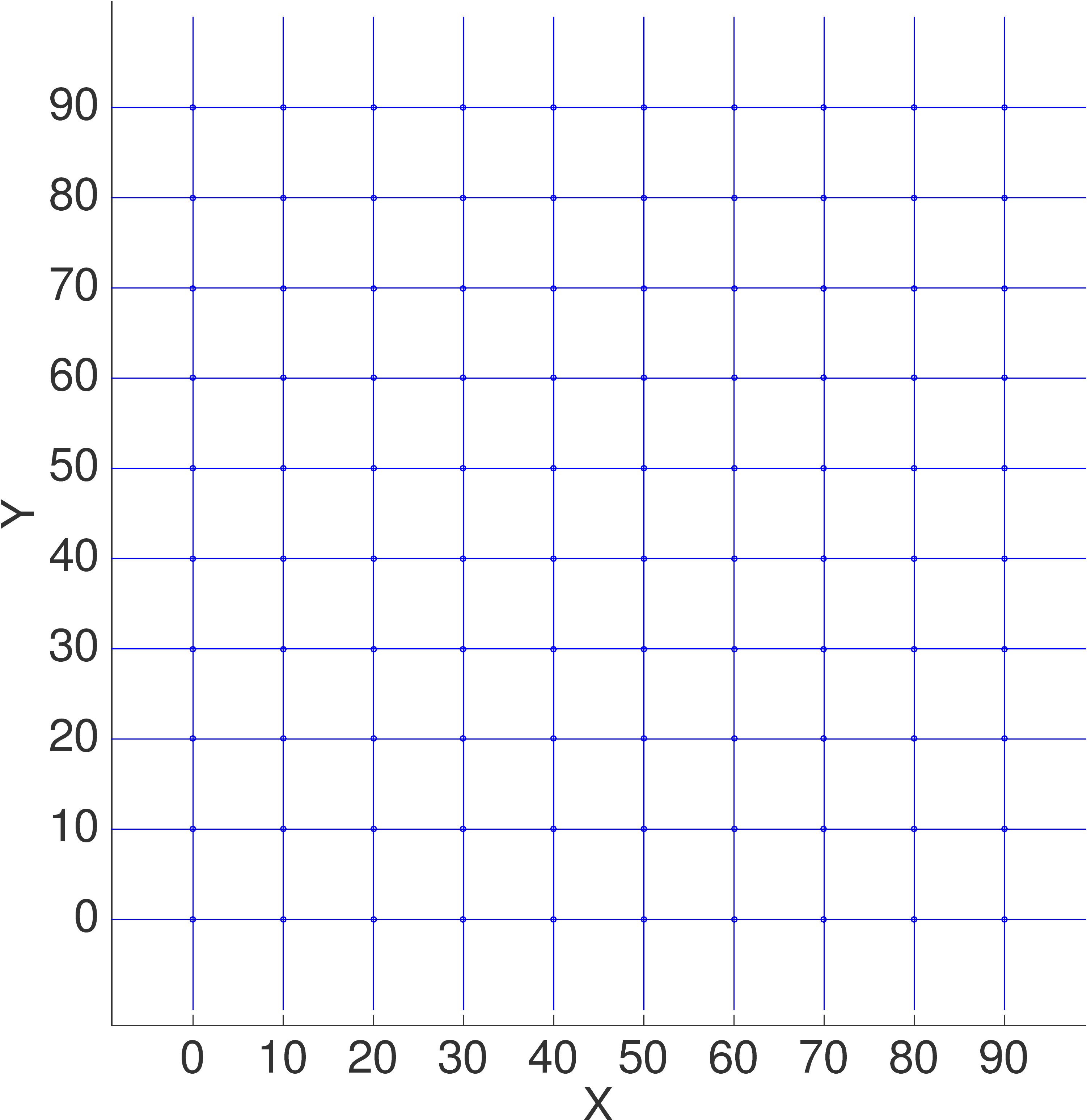}
\hspace*{0.5cm}
\includegraphics[width=3.5cm,angle=0]{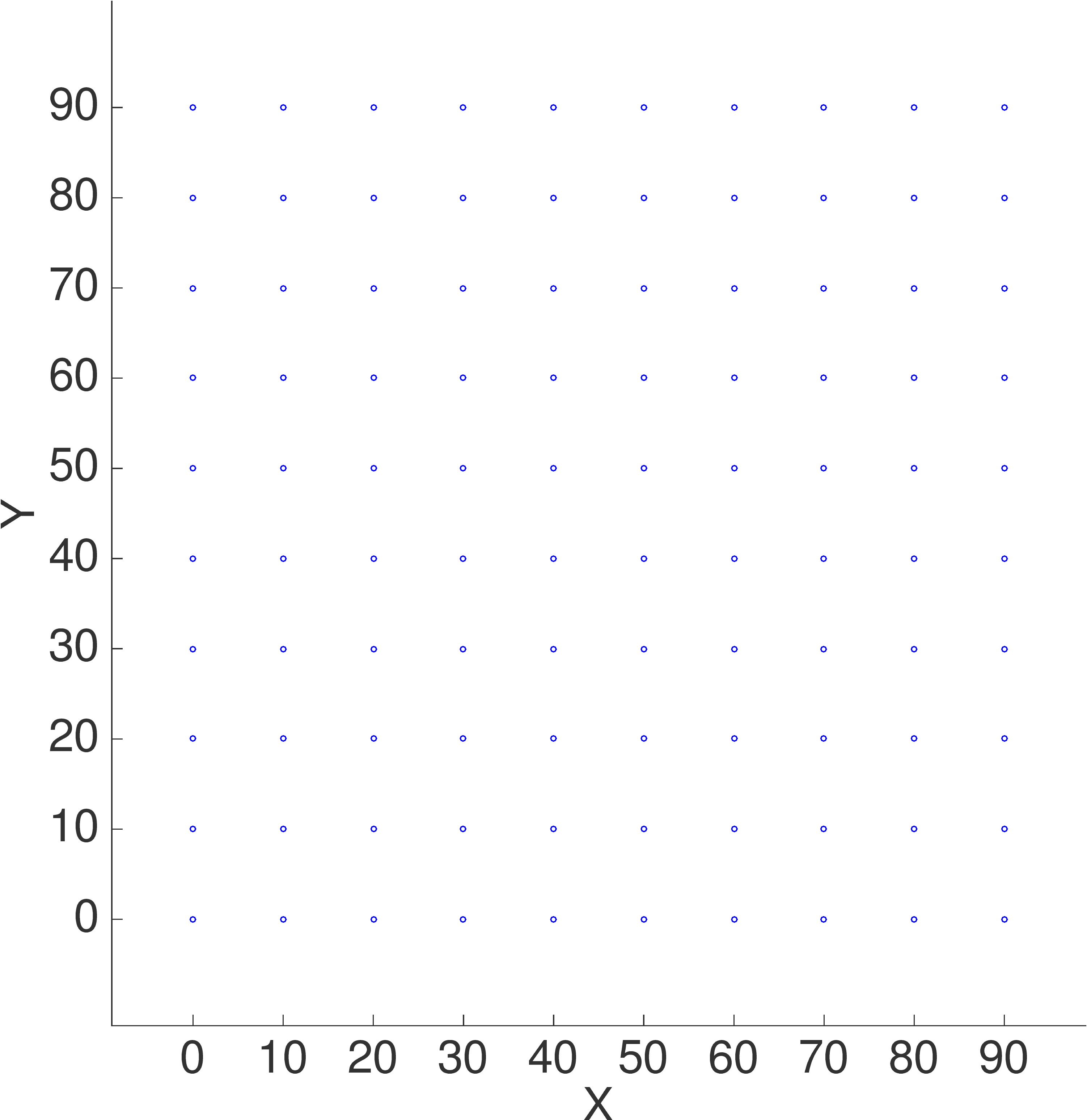}
\hspace*{0.5cm}
\includegraphics[width=3.5cm,angle=0]{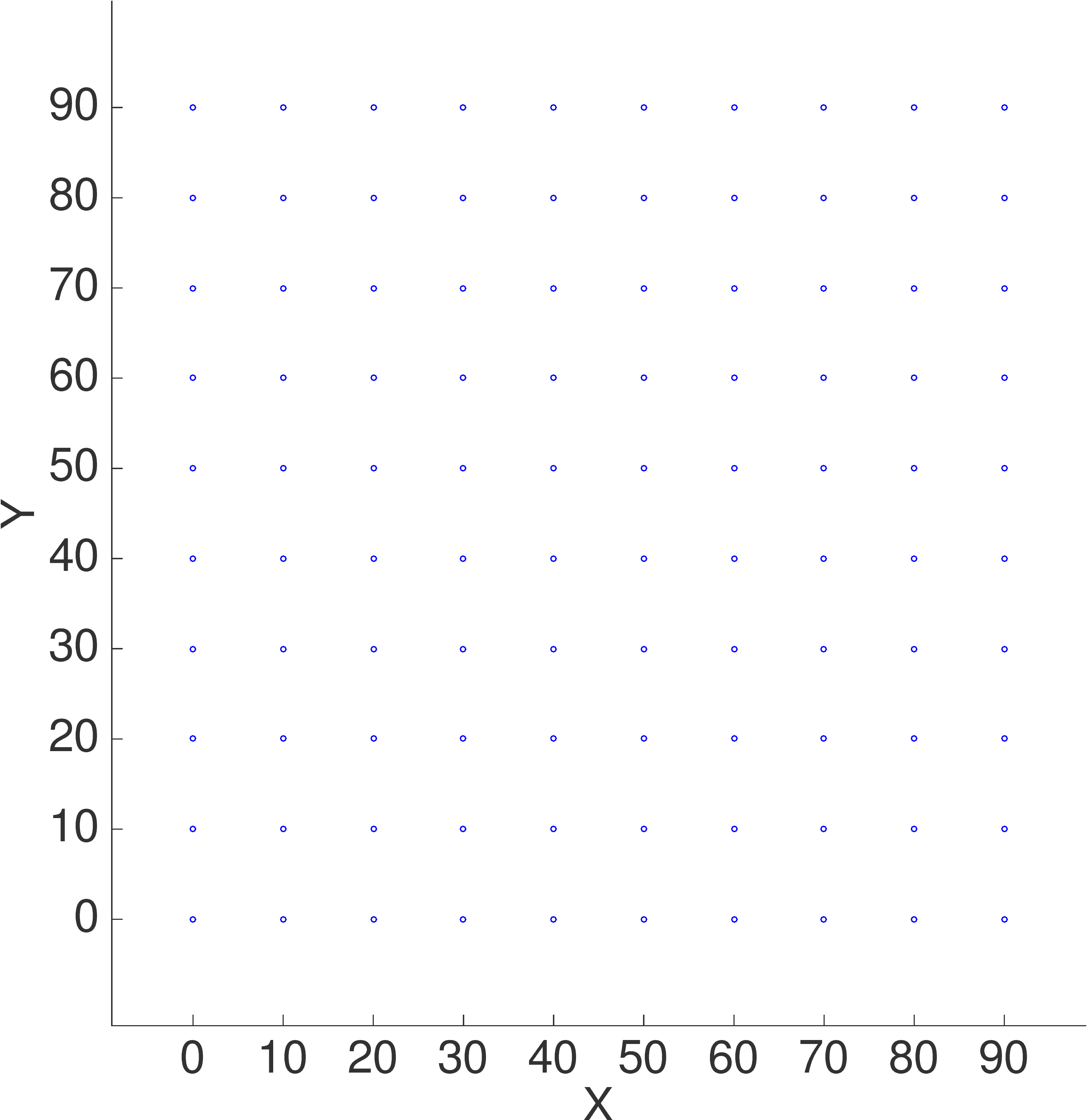}
\hspace*{0.5cm}
\includegraphics[width=3.5cm,angle=0]{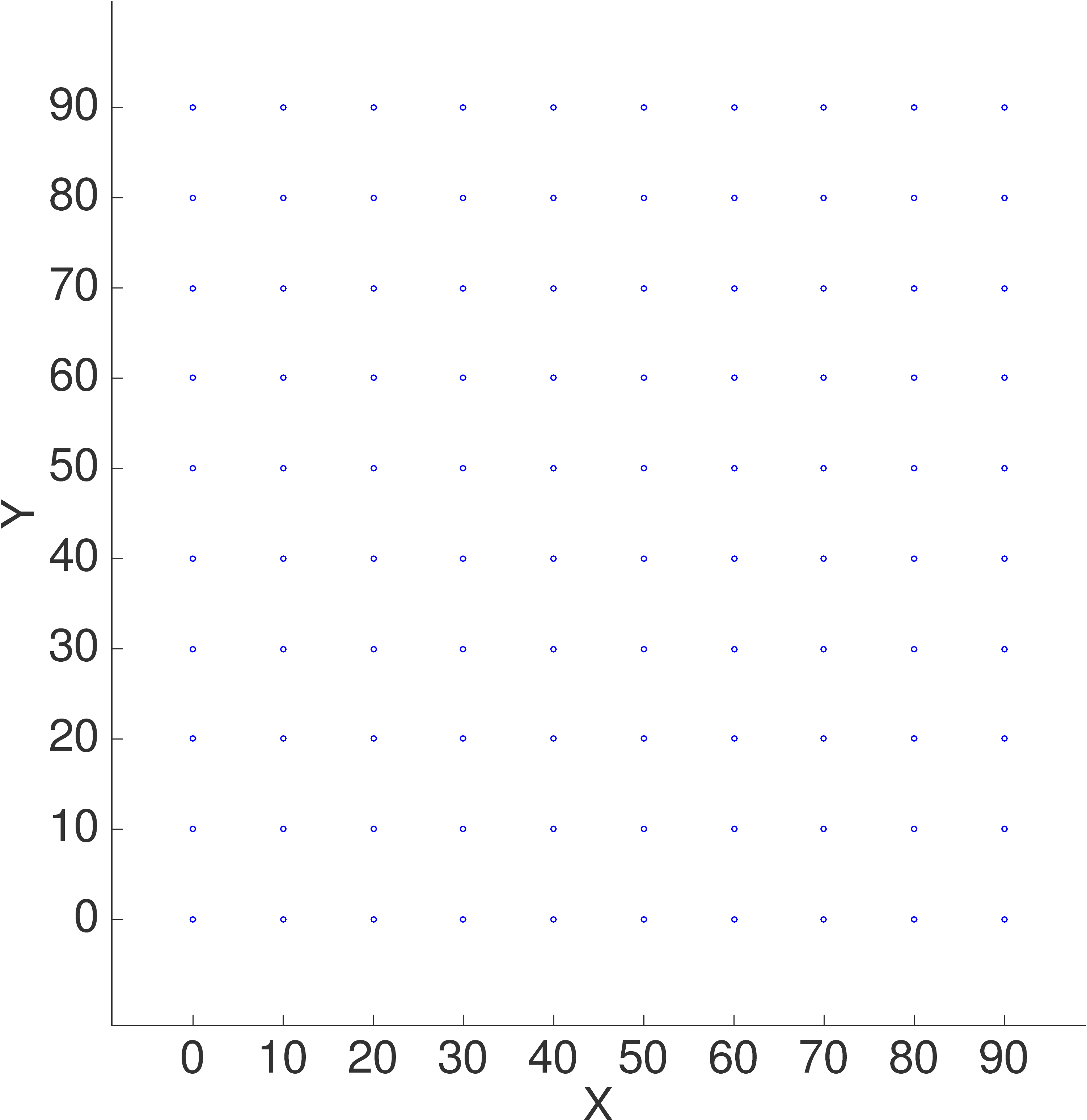}
}
\centerline{$T = 0$  \hspace*{2.5cm} $T = \Delta t$  \hspace*{2.5cm} $T = 2 \Delta t$  \hspace{2.5cm} $T = 3 \Delta t$}
\centerline{(b) Inter edges}
\vspace*{0.3cm}
\centerline{ 
\includegraphics[width=5.0cm,angle=0]{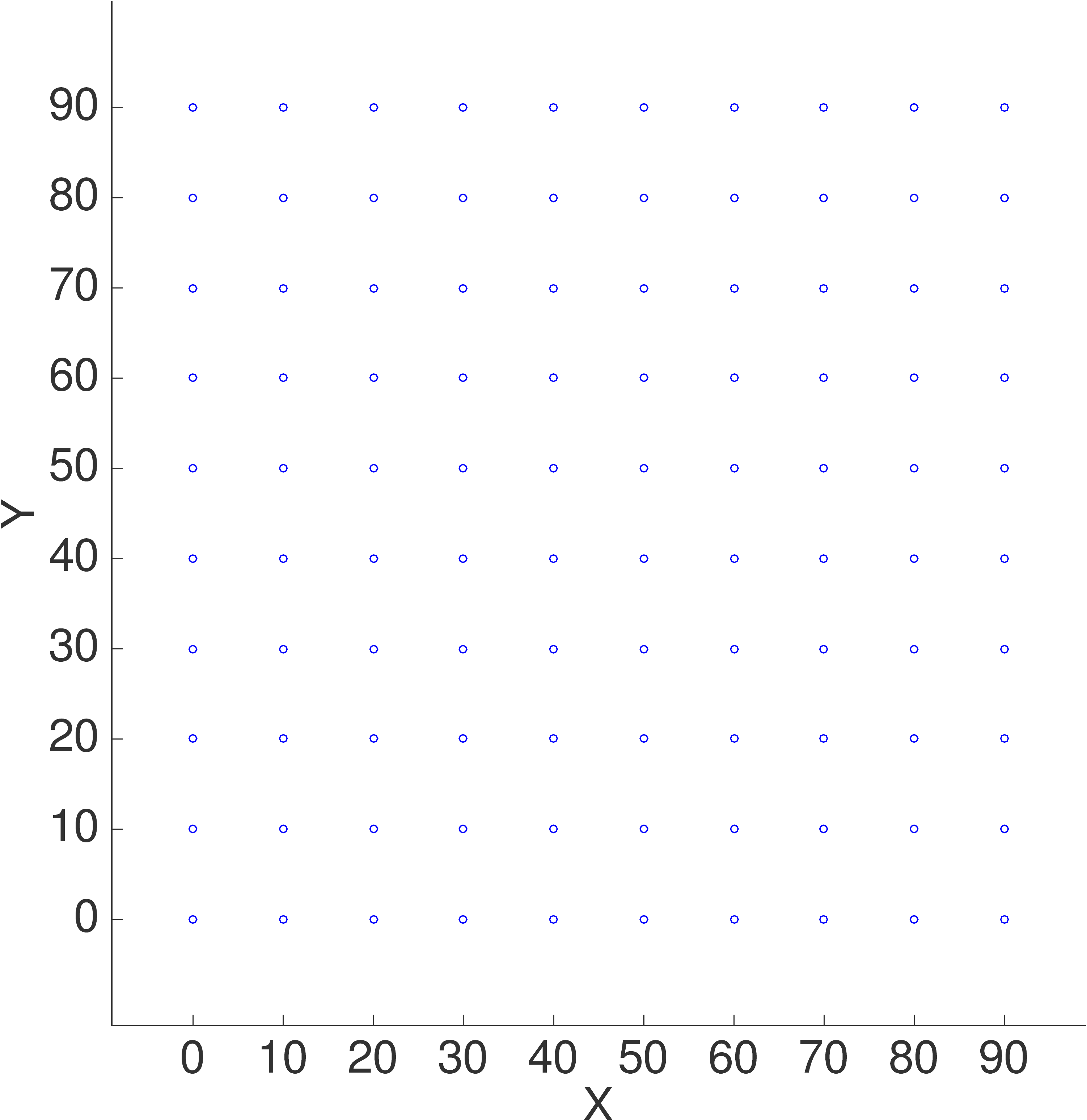}
\hspace*{3.0cm}
\includegraphics[width=5.0cm,angle=0]{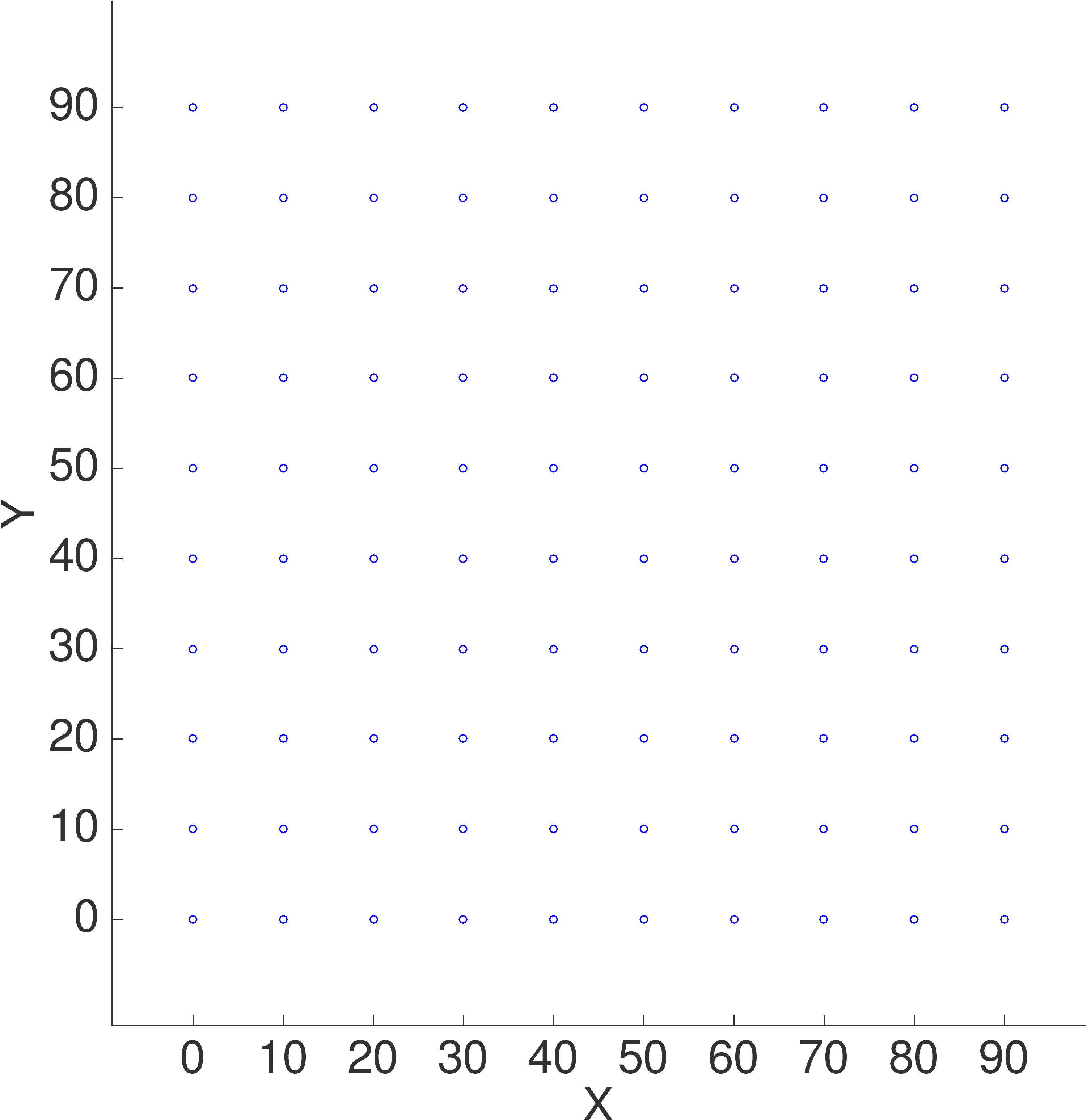}
}
\centerline{(c) Type 1 velocity estimate  \hspace*{3.0cm} (d) Type 2 velocity estimate}
\caption{Results for pure diffusion, 10x10 grid, $\kappa_x = \kappa_y = \kappa$, 
$M=1$.
\label{pure_diffusion_fig_1}}
\end{figure*}

The results for each scenario are visualized in a single figure
consisting of three types of plots, 
namely {\it intra edge plots} (top panels), 
{\it inter edge plots} (center panels) 
and {\it velocity plots} (bottom panels),
see for example Figure \ref{pure_diffusion_fig_1}.
In intra edge plots (e.g.\ Figure \ref{pure_diffusion_fig_1}(a)) 
each grid point is represented by a circle, which is empty if there is no intra edge,
and filled if there is an intra edge. 
The filled circle is grey if the connection is very 
weak, otherwise it is blue.  
There are separate plots for different travel times, e.g.\ $T=\Delta t$ indicates
connections across one time step, etc.  
For example, Fig.\ \ref{pure_diffusion_fig_1} (a) indicates that there are strong intra edges
at all grid points, connecting each location to itself after 1, 2, 3 and 4 time steps,
i.e.\ the local memory at each point is at least 4 time steps.

The inter edge plots (e.g.\ Fig.\ \ref{pure_diffusion_fig_1} (b)) show
the connections between different grid points.  
Concurrent edges are shown in the plot for $T=0$, 
and those edges are always {\it undirected} (no arrow heads), 
as seen for example on the left of 
Fig.\ \ref{pure_diffusion_fig_1} (b).
Inter edge plots for $T>0$ show the nonconcurrent inter edges which are always
directed (for an example see Fig.\ \ref{pure_advection_fig_1}(b,c), 
as there are no such inter edges present in Fig.\ \ref{pure_diffusion_fig_1} (b)).
Very weak inter edges are shown in grey, 
all other edges are shown in blue.
Note that the scale in the inter plots is according to the time step in the data sets 
($M \Delta t$), which may differ 
from the time step shown in the original advection velocity plots ($\Delta t$).

We generate two types of velocity estimate plots that combine the results 
of all {\it directed} edges found, taking the strengths of the connections into account. 
(See \cite{EbDe:2012JCLI} for how we define the strength of each edge.)
Note that the concurrent edges are ignored for now, since they have no direction and are thus 
hard to incorporate in velocity calculations.
(We will return to the topic of concurrent edges in Section \ref{complex_scenarios_sec}.)
Velocity estimate plots show for each grid point the {\it average direction} 
of the {\it incoming nonconcurring} edges at that grid point, 
weighted by the strength of each edge. 
The difference is that Type 1 velocity plots use {\it only} nonconcurrent inter edges
in the estimate,
while Type 2 velocity include both nonconcurrent inter {\it and} intra edges.
In these plots weak connections (i.e.\ those that result from only weak edges) 
are indicated by dashed lines, rather than grey color.

The purpose of Type 1 velocity plots 
is to highlight all pathways of information flow {\it between} different locations
and those are often better visible when the intra edges are ignored.
The idea behind the Type 2 velocity plot is 
that it can be compared to the original 
advection velocity plot used to generate the data.
Note that there are subtle differences in the meaning of the original advection velocity and 
Type 2 velocity.  
The advection velocity field by definition only represents advection, not diffusion, 
while the Type 2 velocity plot represents the result for advection combined with 
diffusion.  However, in the great majority of scenarios considered below, diffusion 
actually acts equally in all directions, so it should not impact the {\it direction}
of information travel, but only magnitude.  Furthermore, as we will see later, 
advection is the dominant process, thus the impact of diffusion on the magnitude 
of velocity is expected to be small wherever advection velocity is significant.
Thus Type 2 velocity plots can be seen as an approximation of the input 
advection velocity fields.
Both Type 1 and Type 2 velocity fields are scaled to be on the same scale 
as the original advection fields to allow for easy comparison.

We generally provide the full set of plots for each scenario, 
even if some of them show trivial or 
repetitive results, such as all intra edges being identical (Fig.\ \ref{pure_diffusion_fig_1}(a)), no inter edges being present (three of the plots in Fig.\ \ref{pure_diffusion_fig_1}(b)),
or vanishing velocities (Fig.\ \ref{pure_diffusion_fig_1}(c,d)).  
We provide those redundant plots nevertheless because it is much easier to 
visualize and compare results from different scenario, if all results
are provided in a similar format, in this case a consistent set of plots.

\subsection{Why Are No Error Measures Provided?}

Eventually we plan to develop error measures that quantitatively evaluate the 
accuracy of the results for the different scenarios.  
However, at this point any effort in that direction appears premature, as we are 
just starting to learn about the behavior of the algorithm and even about 
which types of output figures to use in which case.
Thus at this time qualitative visual representations of the output, 
i.e.\ figures, are much more useful than 
condensing the results to a single number.  
Furthermore, a primary use of the figures  
is for geoscientists to use them for scientific discovery of underlying dynamical
mechanisms, 
and it is not trivial to assign a number to how well the output is suited to achieve that goal.
Thus such measures will need to be carefully developed in the future in close collaboration with geoscientists, but that is a complex research topic of its own. 
  


\subsection{Results for Pure Diffusion}
\label{idealized_diffusion_results_sec}

To study pure diffusion we temporarily drop\footnote{We 
	need to drop the advection term and its corresponding stability criterion,
        rather than just setting the advection velocity to zero, because otherwise 
        the numerical stability condition (Eq.\ (\ref{CFL_eq})) would impose $\Delta t=0$.}
the advection term in the diffusion advection equation, 
leaving only the diffusion term.
Furthermore, we use $C=1$, so that there is no {\it numerical} diffusion. 
%
First we use Message Type 1, single peak initial conditions, without noise. 
We calculated results for
two different grid resolutions, 10x10 and 20x20 grids, with 
varying temporal resolutions, $M=1,2,4$.
In most cases we use $\kappa_x=\kappa_y=\kappa$, i.e.\ equal amount of diffusion in 
$x$ and $y$ direction.  
In some cases we use $\kappa_x= \kappa, \kappa_y=0$, to simulate diffusion only in 
$x$-direction.

\begin{figure}
\begin{center}
\begin{tabular}{|p{2.5cm}||p{3.1cm}|p{3.1cm}|p{3.1cm}|r|}
\hline
Scenario & \multicolumn{3}{c|}{Concurrent inter edges for} \\
   & $M=1$ & $M=2$ & $M=4$\\
\hline
10x10 grid, $\kappa_x=\kappa$, $\kappa_y=\kappa$ & Connections to $\pm x, \pm y$ & 
Connections to $\pm x, \pm y$ & Connections to $\pm x, \pm y$ and diagonals 
\\
& 
\includegraphics[width=3.0cm,angle=0,clip]{./FIG_DIFFUSION_ONLY_EXP_1_INTER_D_0.pdf} &
\includegraphics[width=3.0cm,angle=0,clip]{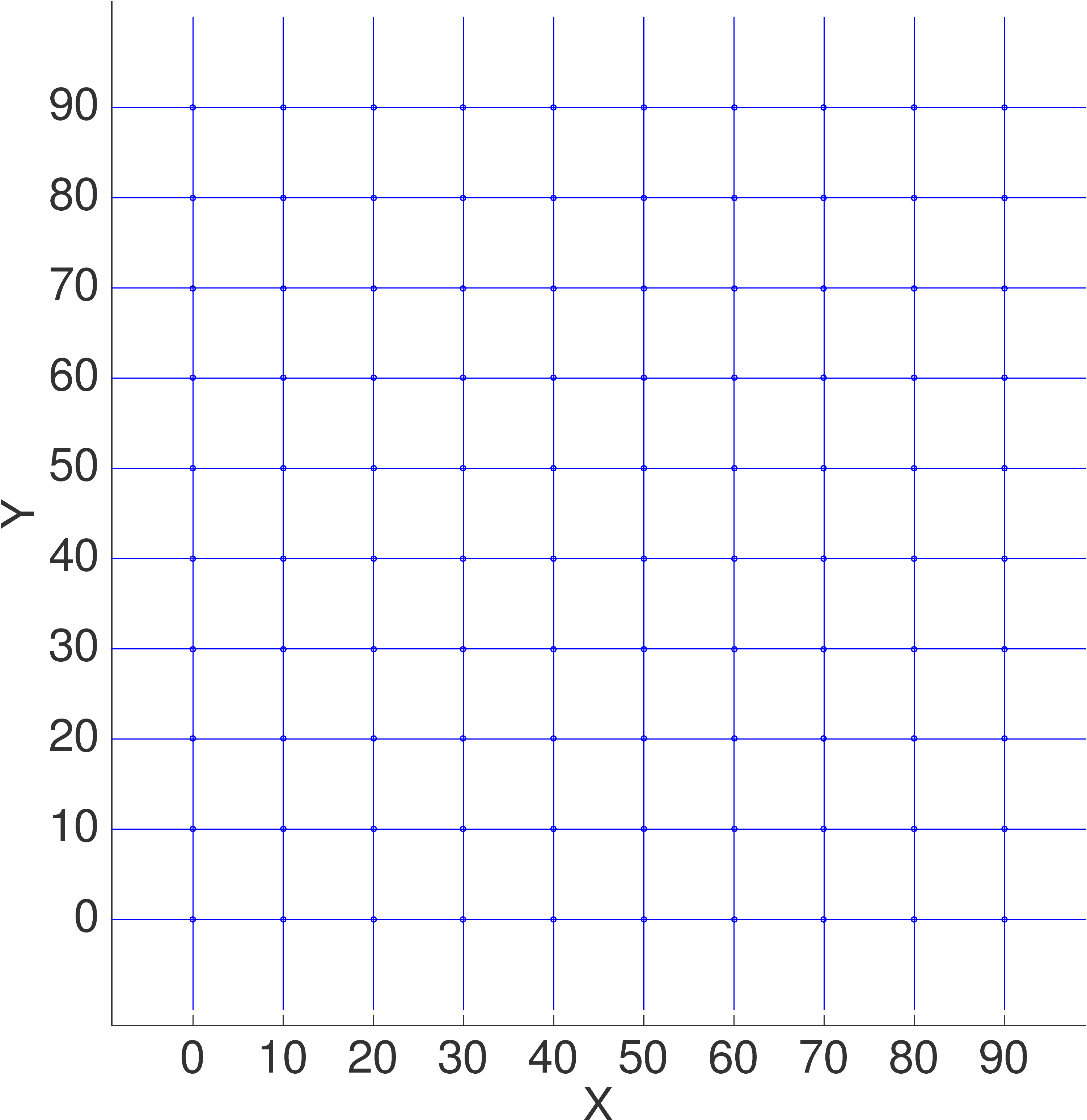} &
\includegraphics[width=3.0cm,angle=0,clip]{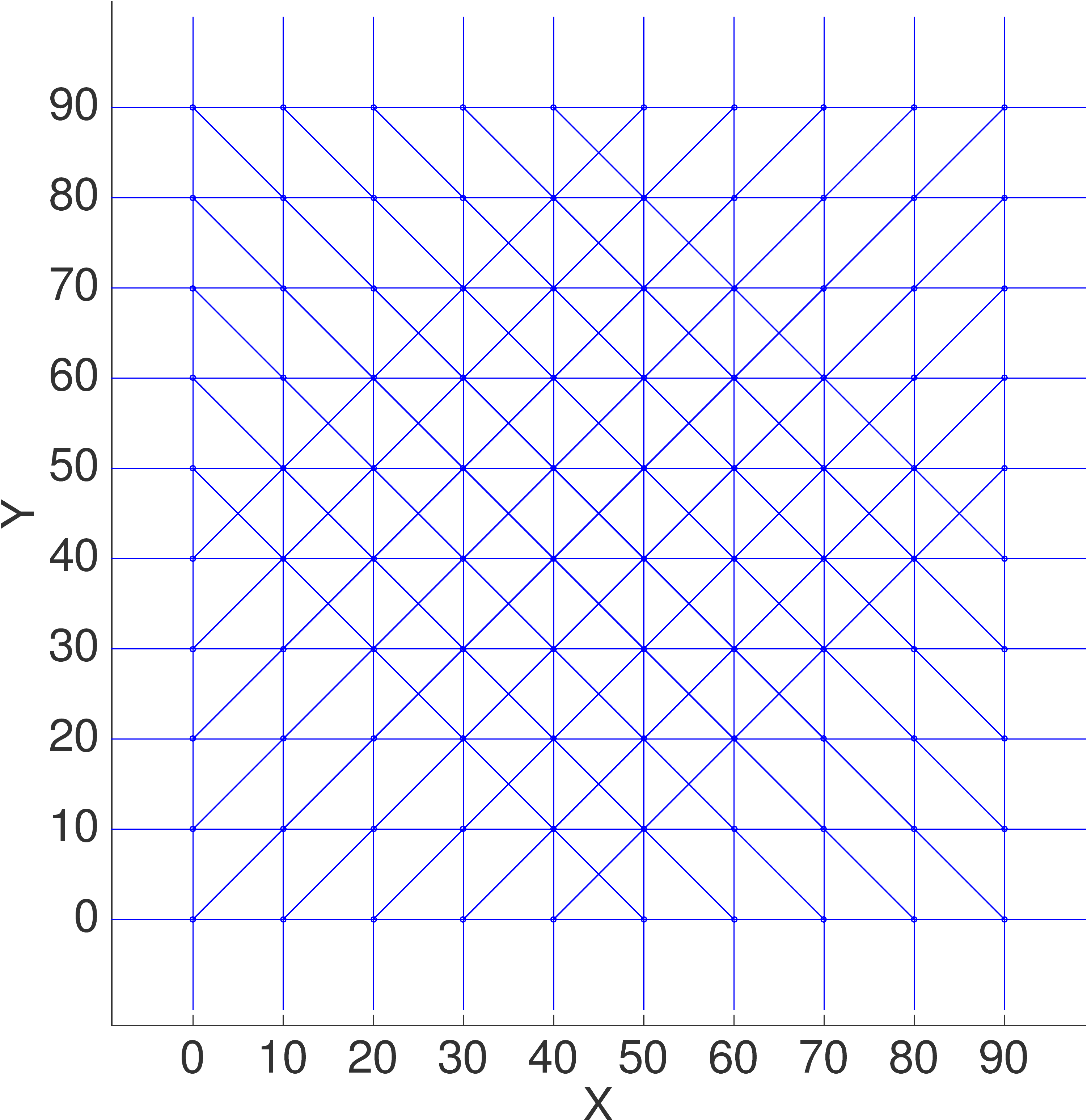}
\\
\hline
10x10 grid, $\kappa_x= \kappa$, $\kappa_y=0$ & Connections to $\pm x$
\\
& 
\includegraphics[width=3.0cm,angle=0,clip]{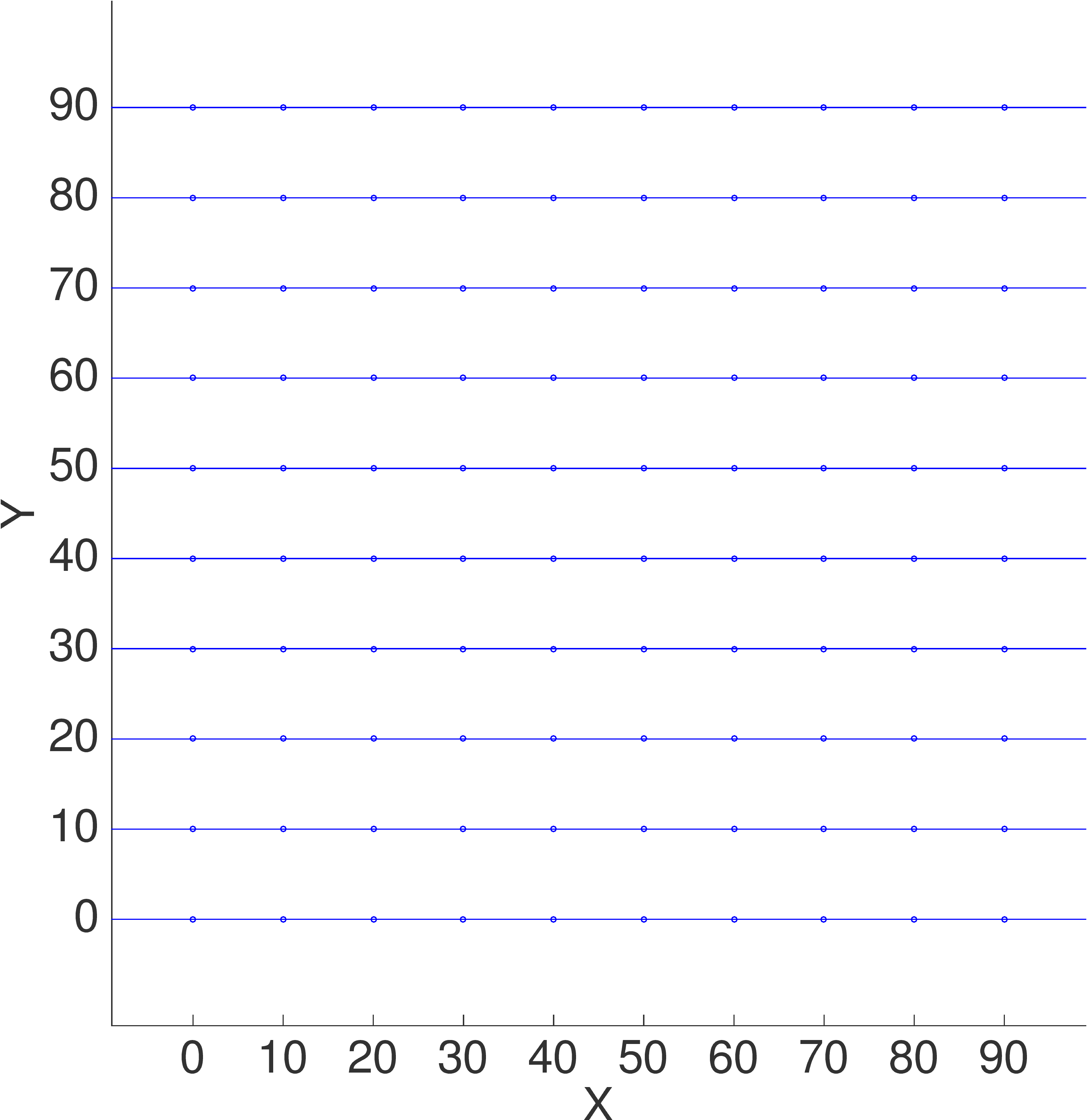}
\\
\hline
\hline
20x20 grid, $\kappa_x=\kappa$, $\kappa_y=\kappa$ & 
No connections & Connections to $\pm x, \pm y$ & Weak connections to $\pm x, \pm y$
\\
& 
\includegraphics[width=3.0cm,angle=0,clip]{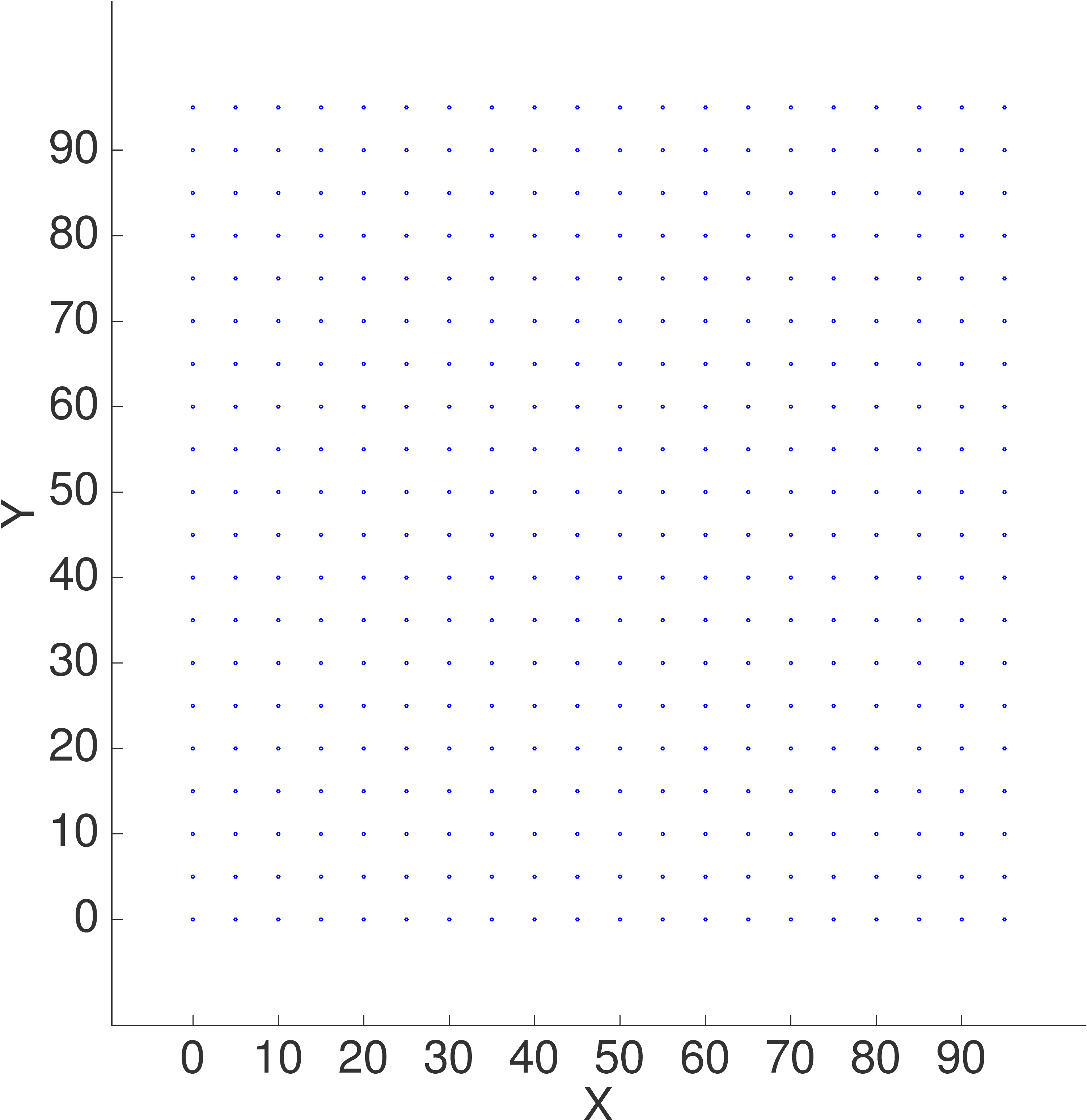} &
\includegraphics[width=3.0cm,angle=0,clip]{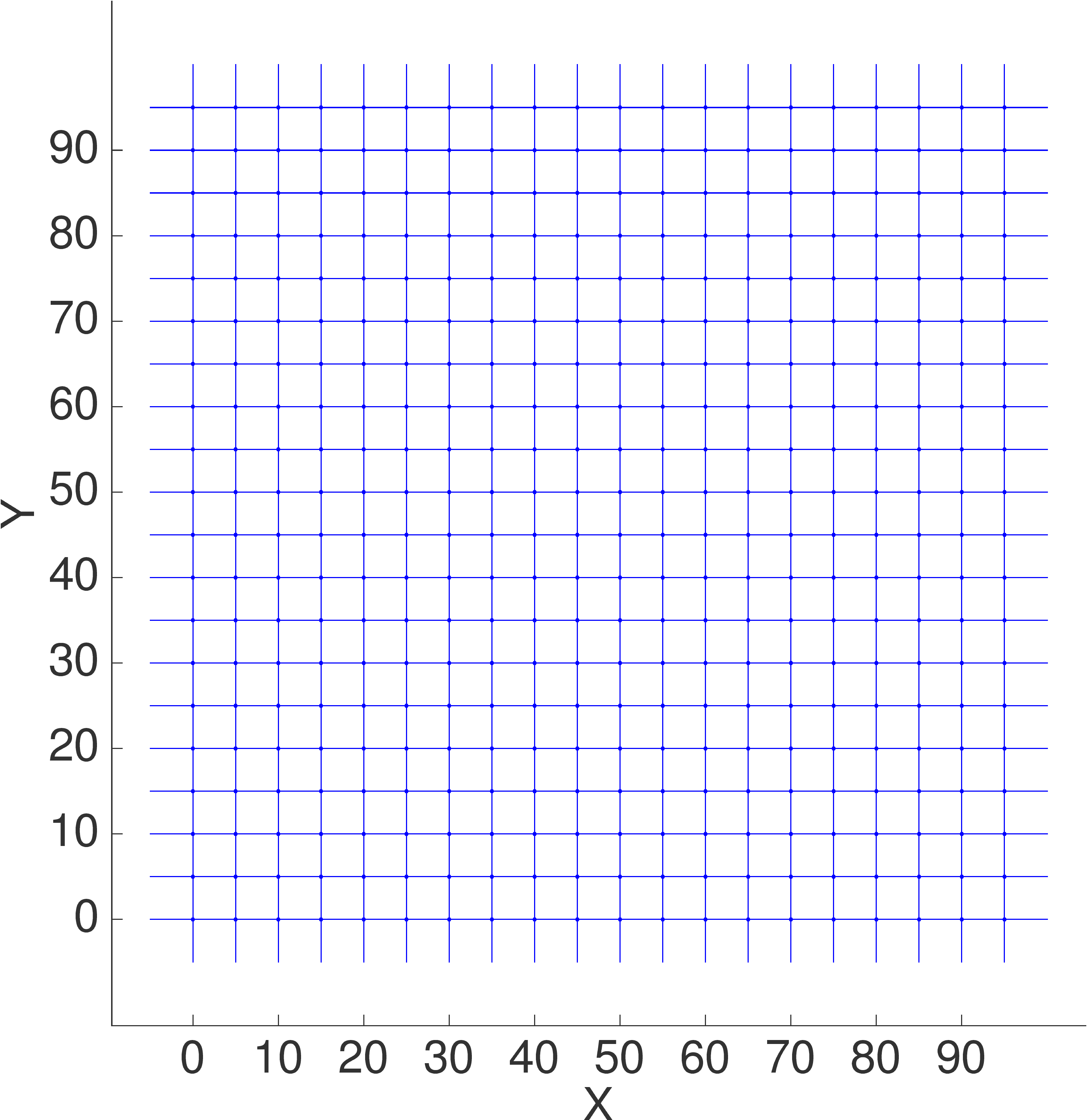} &
\includegraphics[width=3.0cm,angle=0,clip]{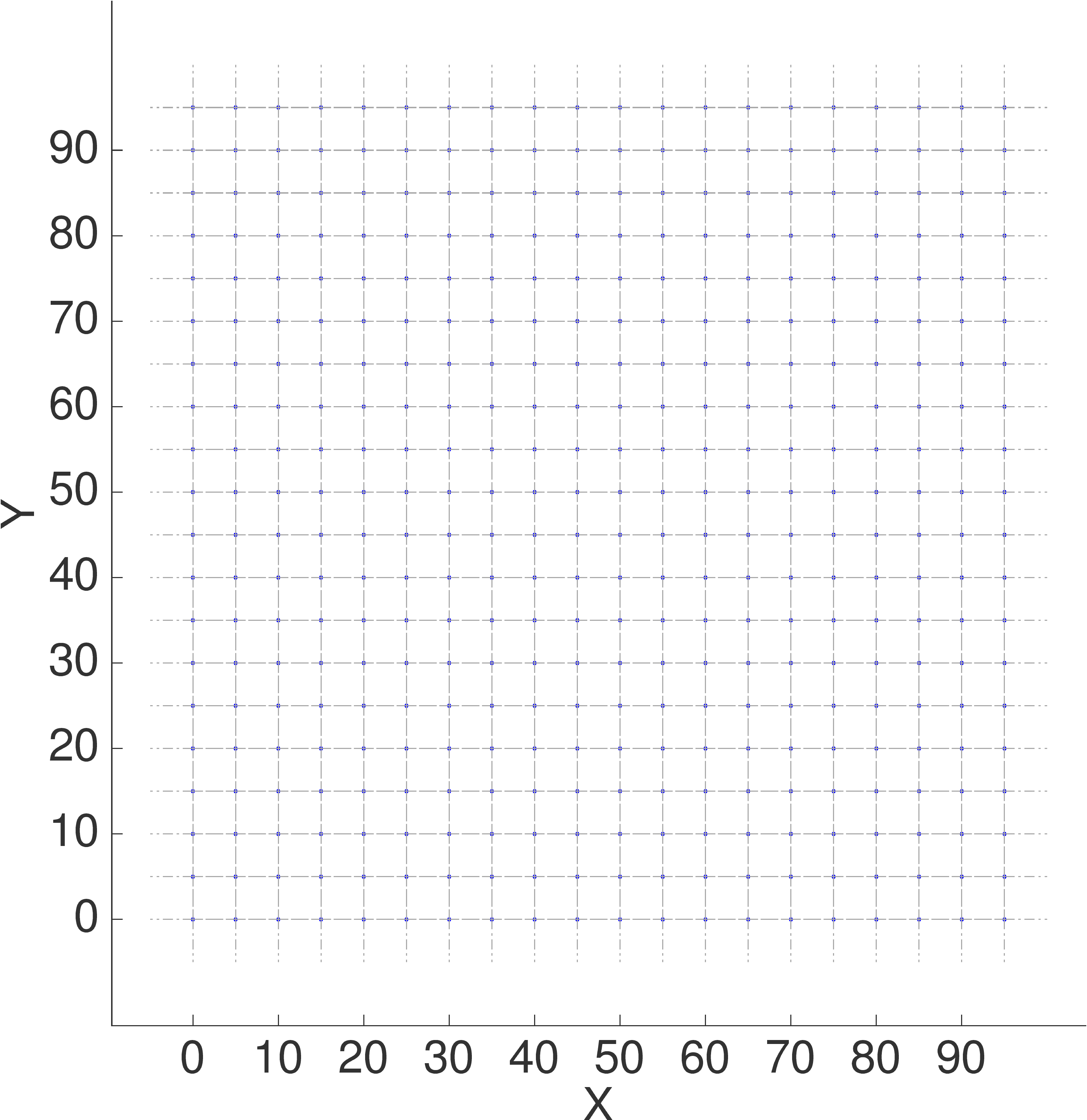}
\\
\hline
20x20 grid, $\kappa_x= \kappa$, $\kappa_y=0$ & Weak connections to $\pm x$ 
\\
& 
\includegraphics[width=3.0cm,angle=0,clip]{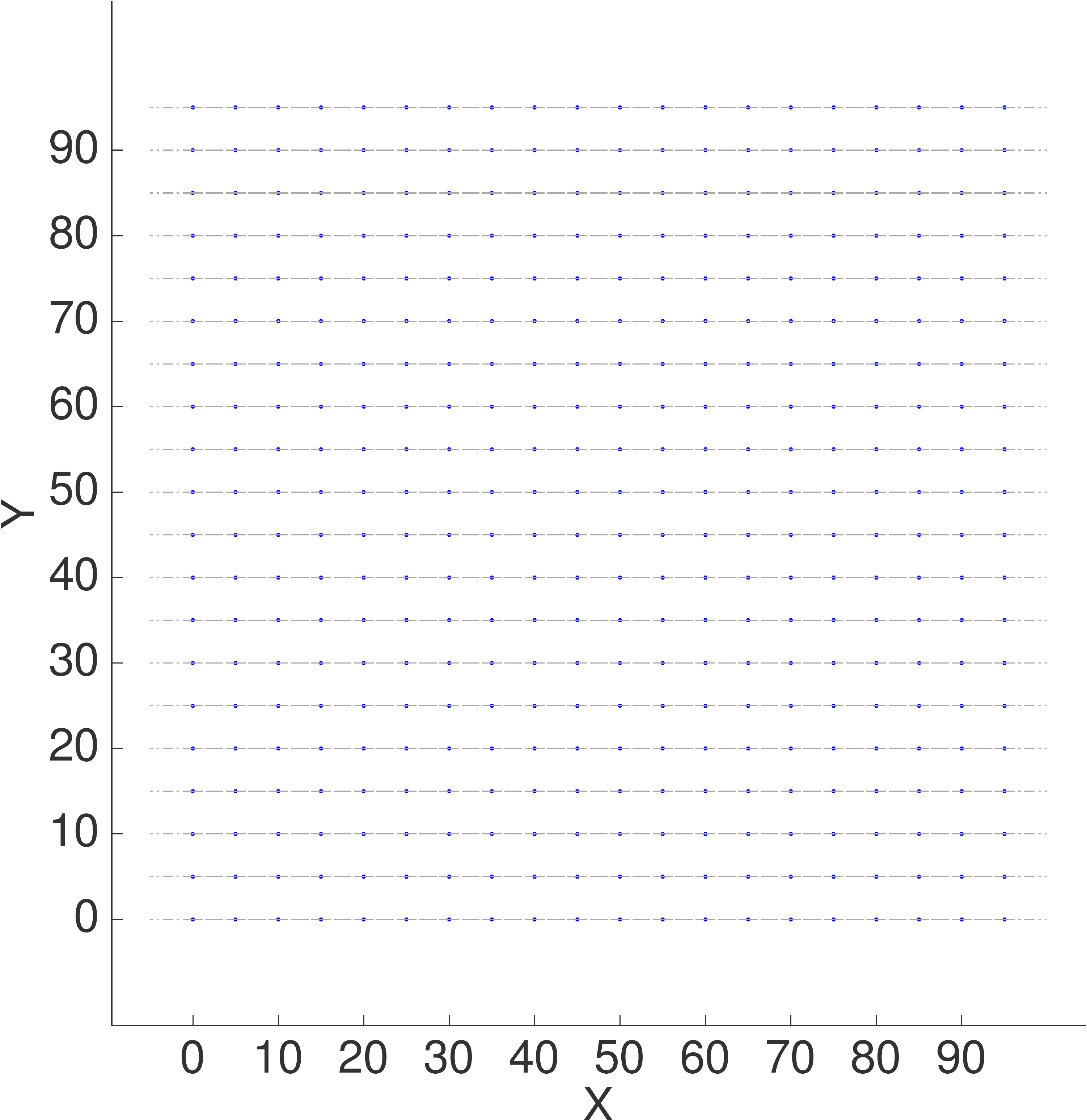}
\\
\cline{1-2}
\end{tabular}
\end{center}
\caption{Concurrent inter edges for several pure 
  diffusion experiments.  Each point is connected to its 0, 2, 4, 5, 6, 7 or 8 closest neighbors.
  \label{idealized_diffusion_results_fig}}
\end{figure}

\begin{figure*}
\centerline{ 
\includegraphics[width=3.5cm,angle=0]{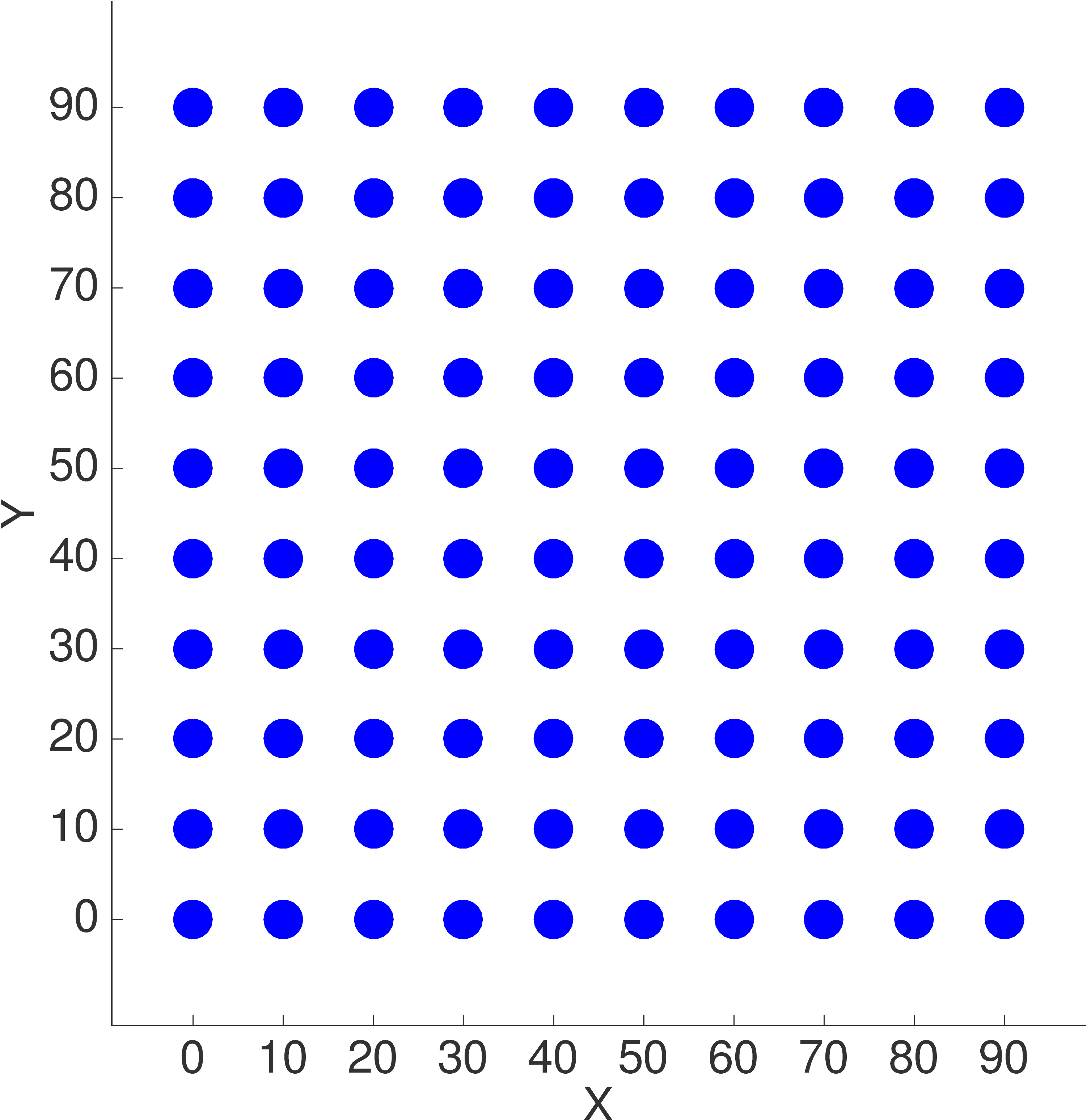}
\hspace*{0.5cm}
\includegraphics[width=3.5cm,angle=0]{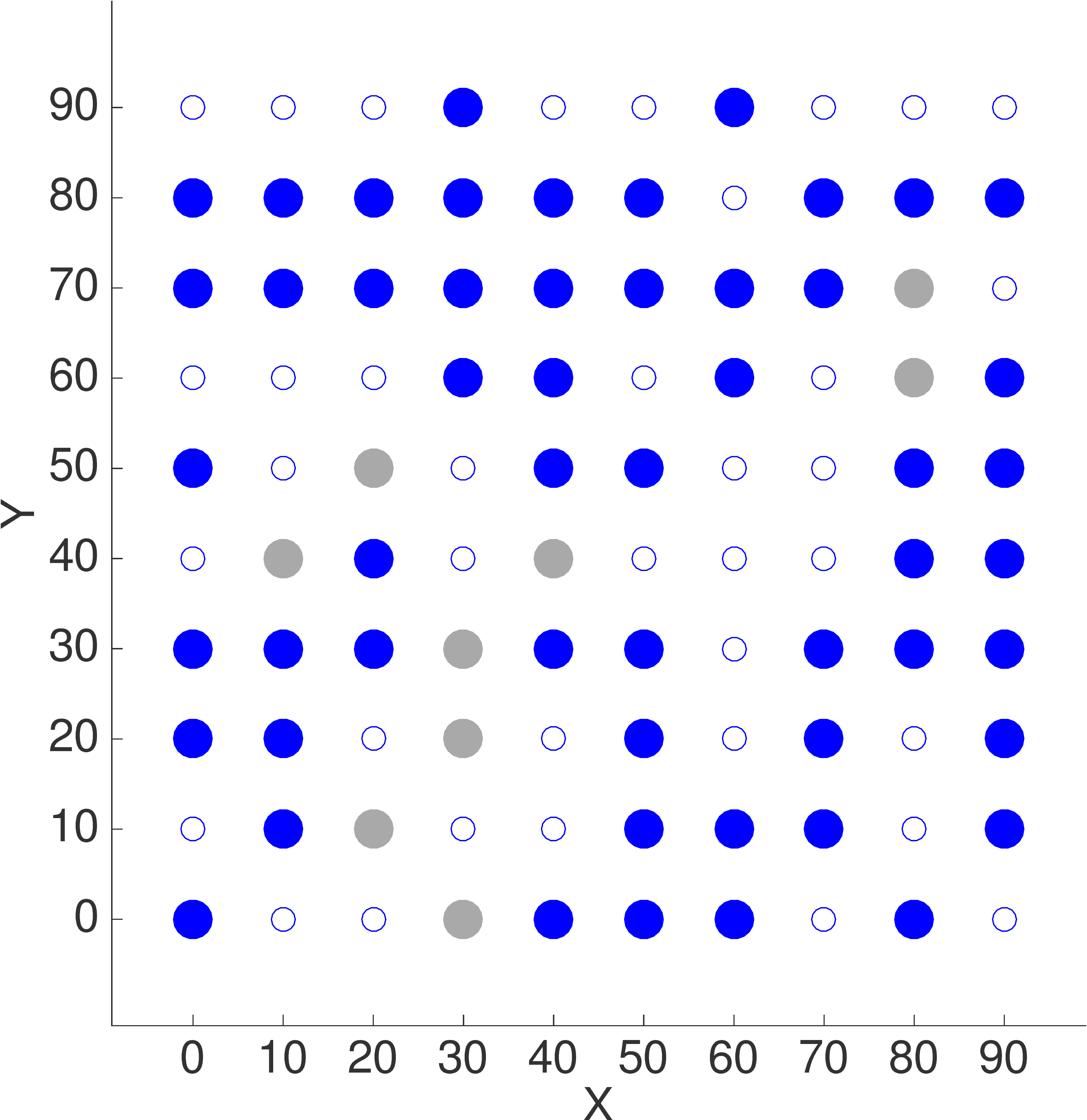}
\hspace*{0.5cm}
\includegraphics[width=3.5cm,angle=0]{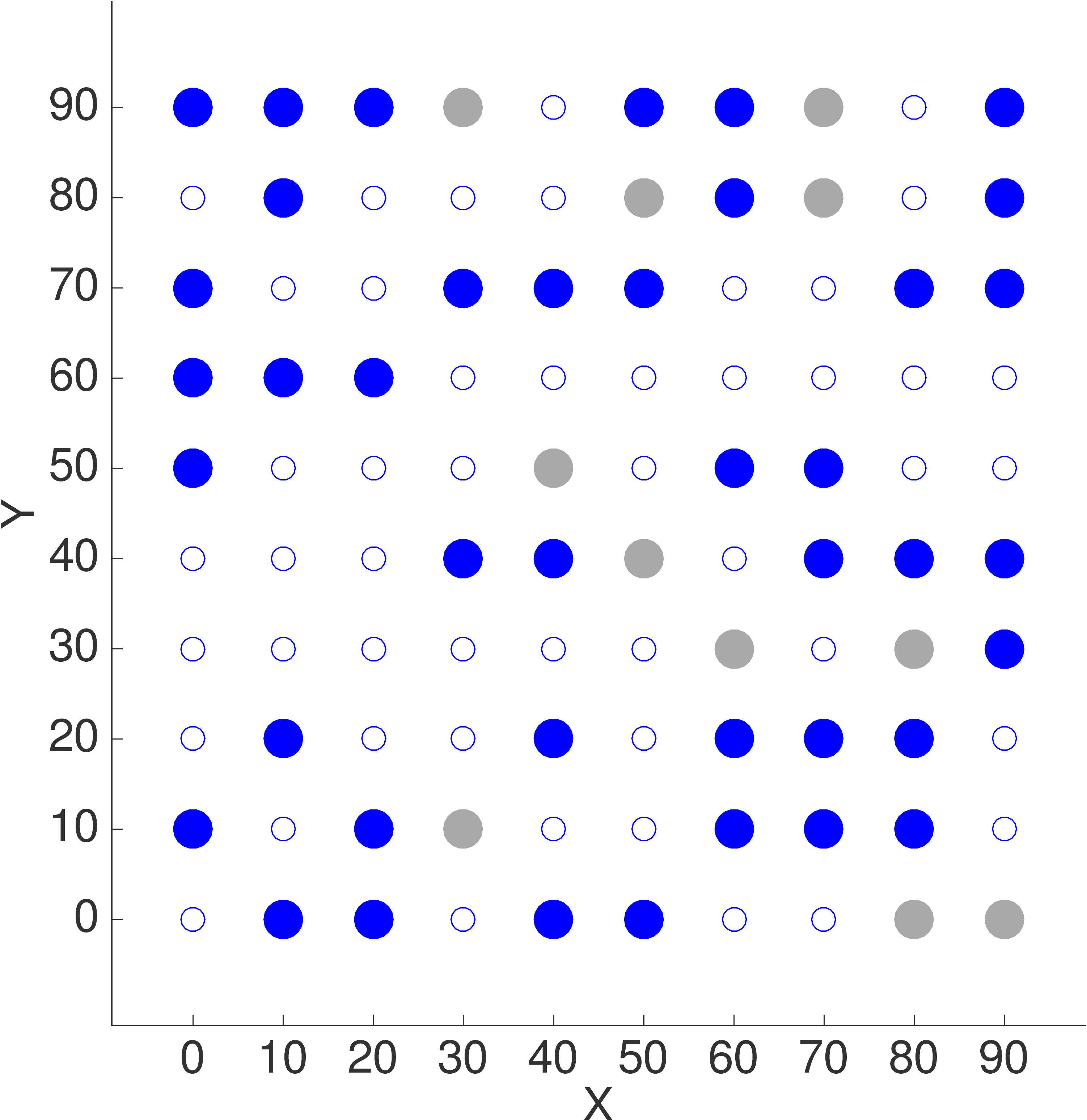}
\hspace*{0.5cm}
\includegraphics[width=3.5cm,angle=0]{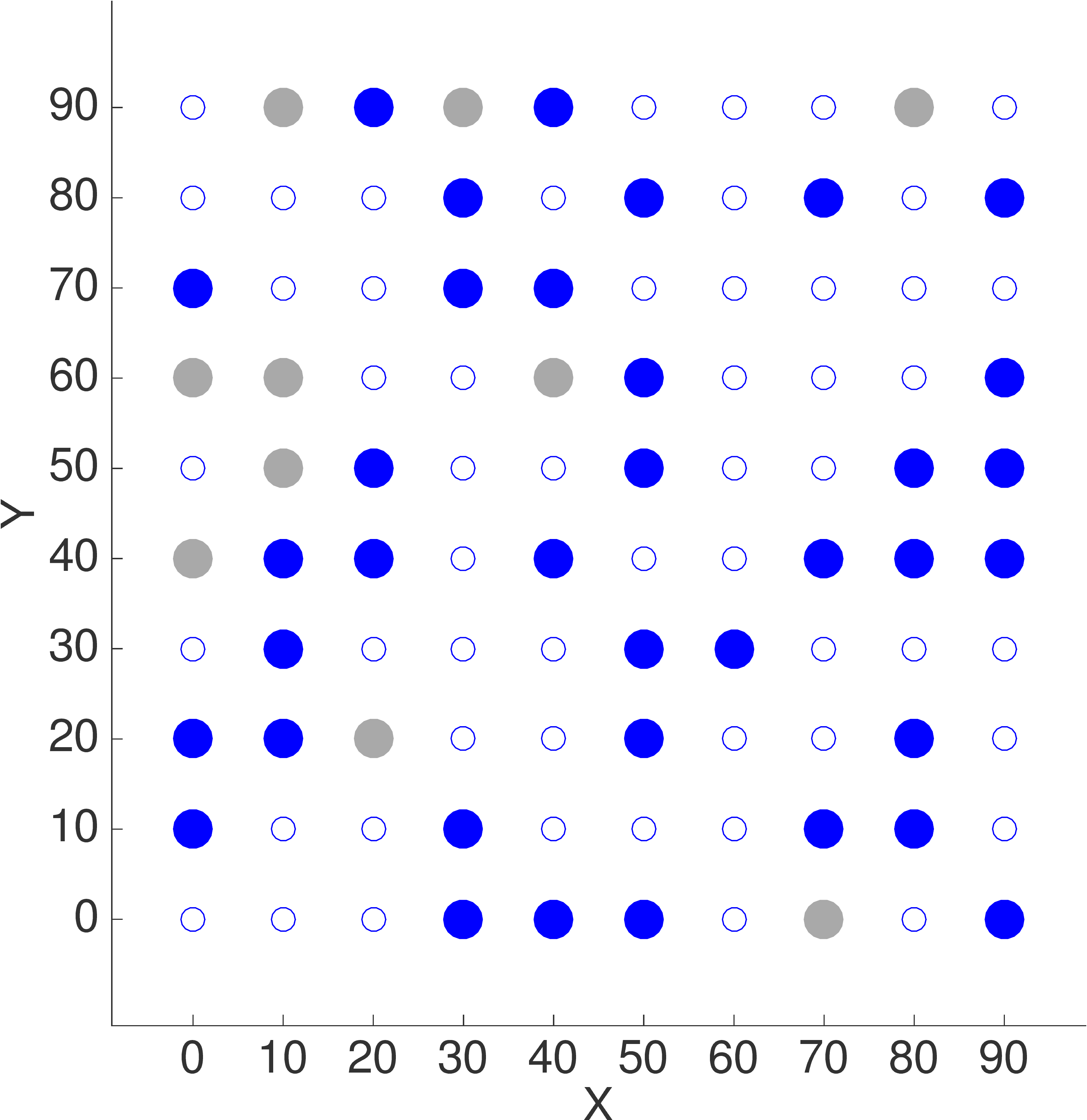}
}
\centerline{$T = \Delta t$  \hspace*{2.5cm} $T = 2 \Delta t$  \hspace*{2.5cm} $T = 3 \Delta t$ \hspace*{2.5cm} $T = 4 \Delta t$}
\vspace*{0.3cm}
\centerline{ 
\includegraphics[width=3.5cm,angle=0]{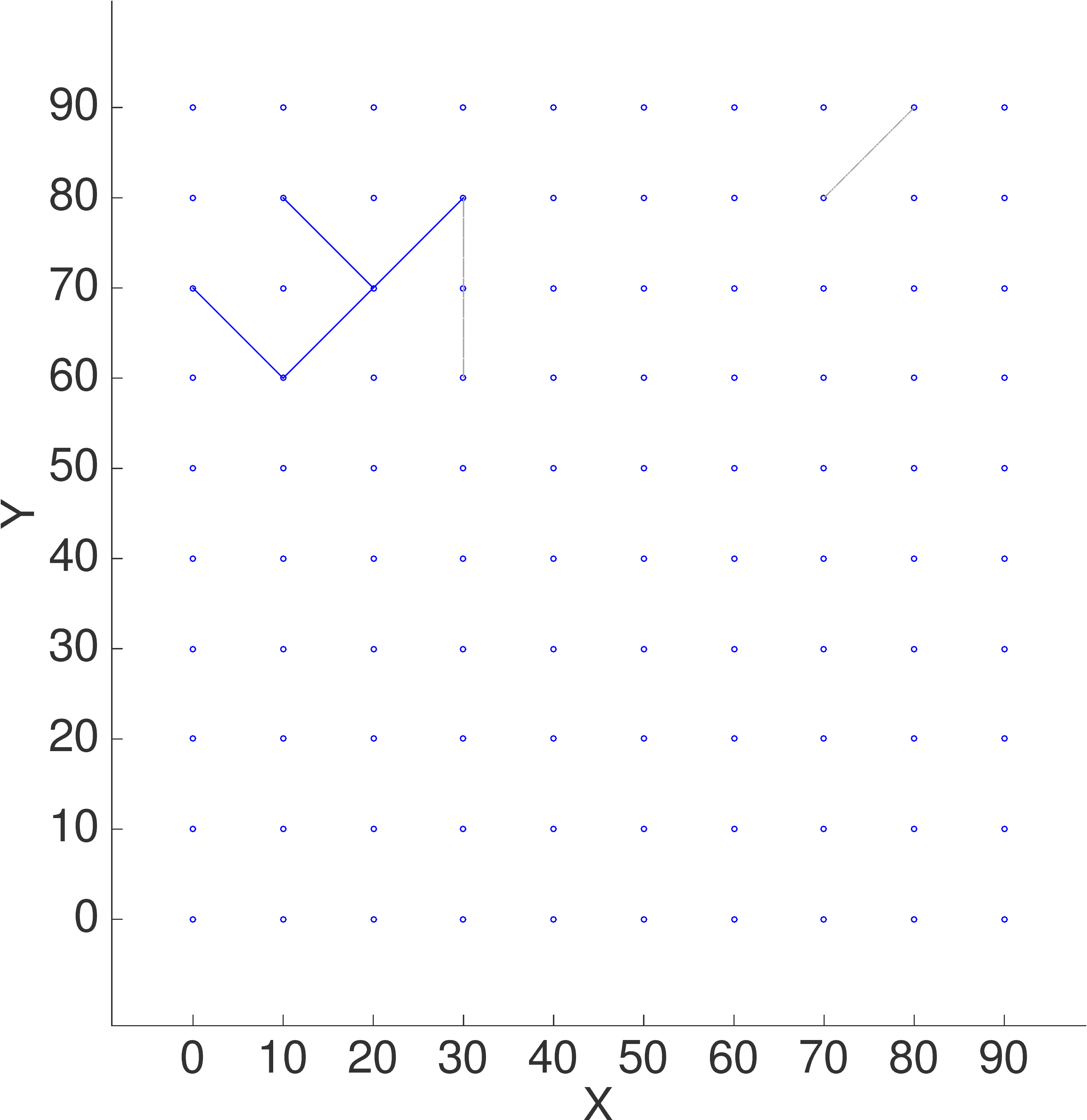}
\hspace*{0.5cm}
\includegraphics[width=3.5cm,angle=0]{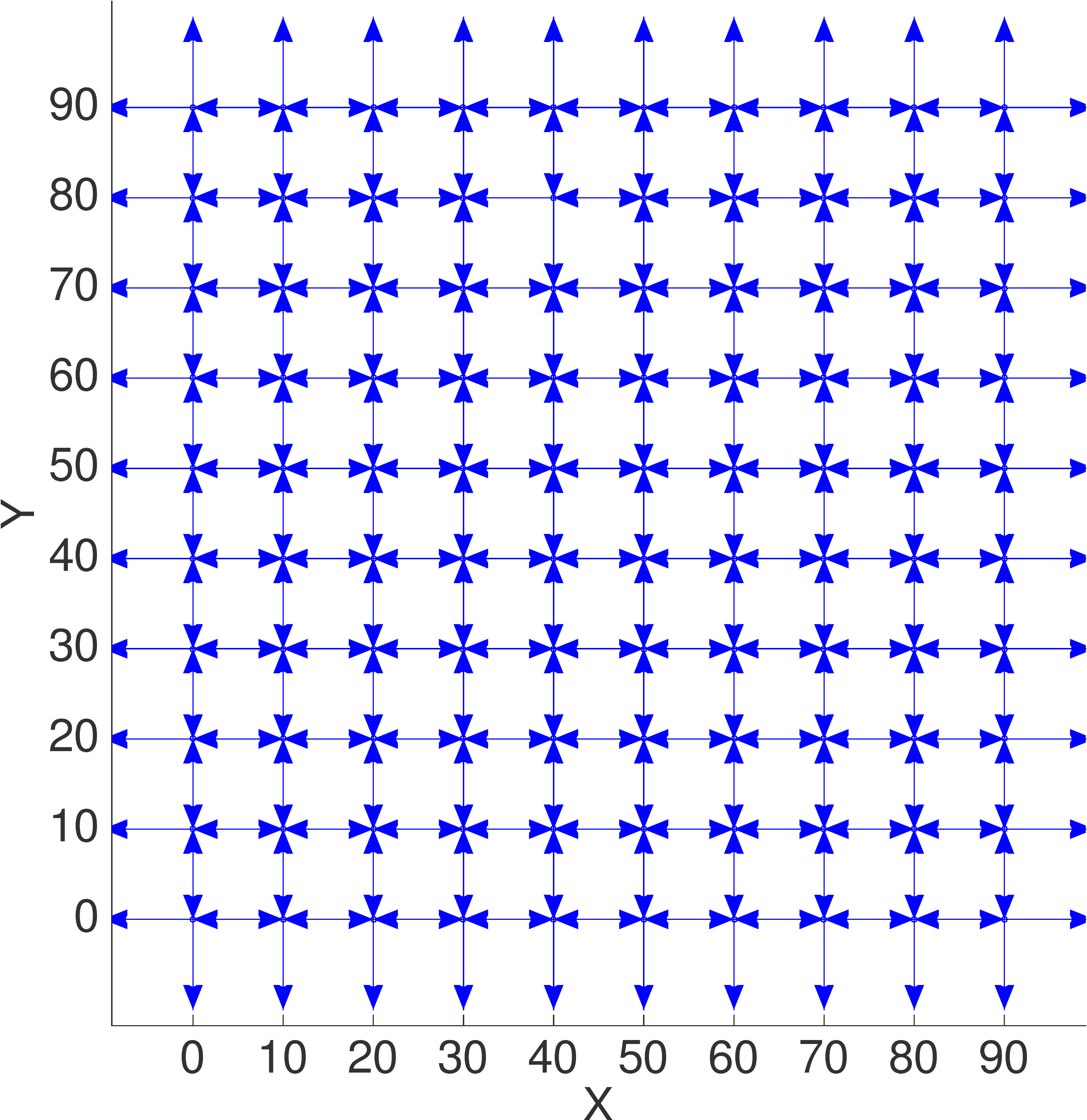}
\hspace*{0.5cm}
\includegraphics[width=3.5cm,angle=0]{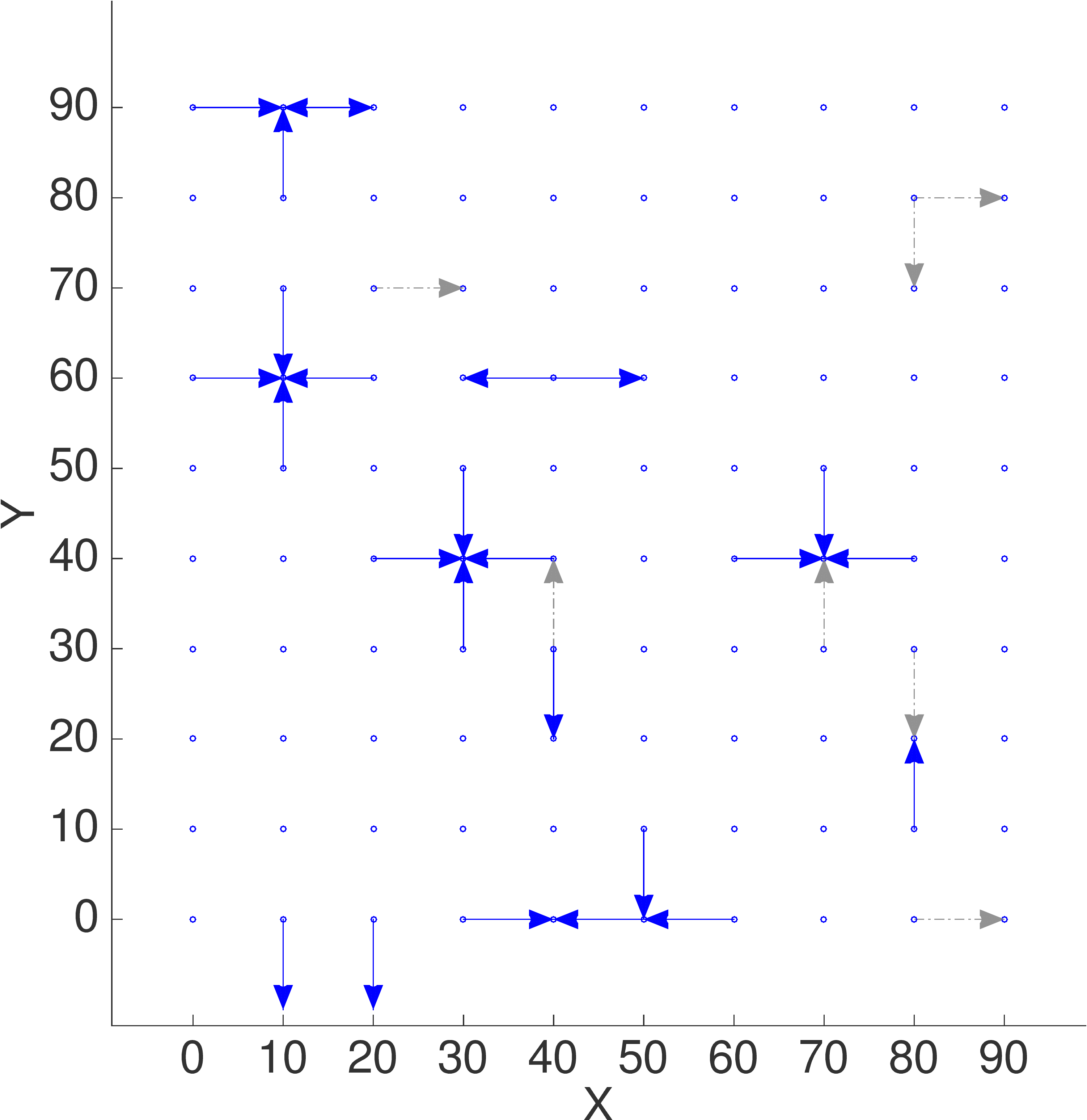}
\hspace*{0.5cm}
\includegraphics[width=3.5cm,angle=0]{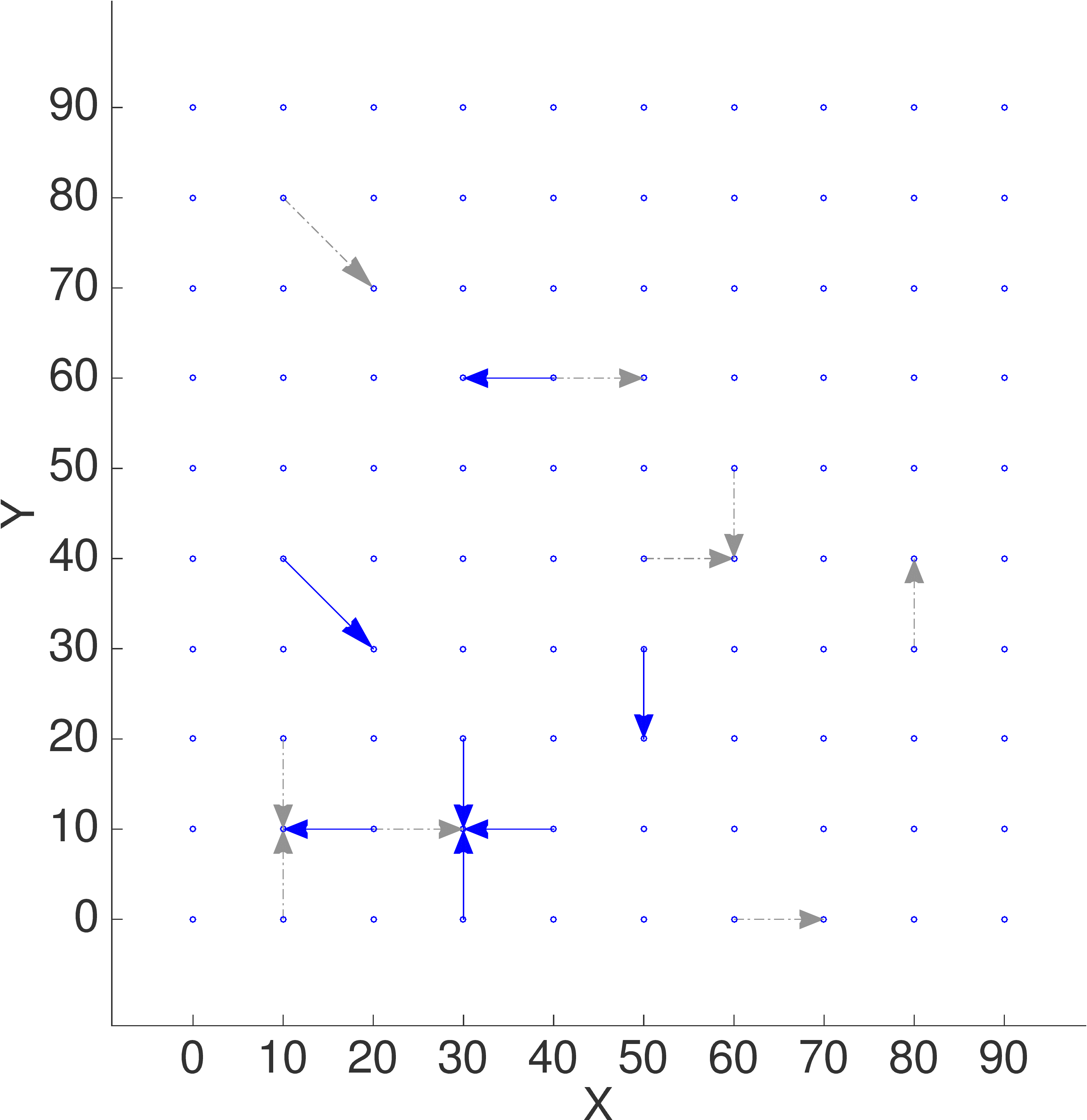}
}
\centerline{$T = 0$  \hspace*{2.5cm} $T = \Delta t$  \hspace*{2.5cm} $T = 2 \Delta t$  \hspace{2.5cm} $T = 3 \Delta t$}
\centerline{(a) Intra (top) and inter (bottom) edges for single-point noise}
\vspace*{0.8cm}
\centerline{ 
\includegraphics[width=3.5cm,angle=0]{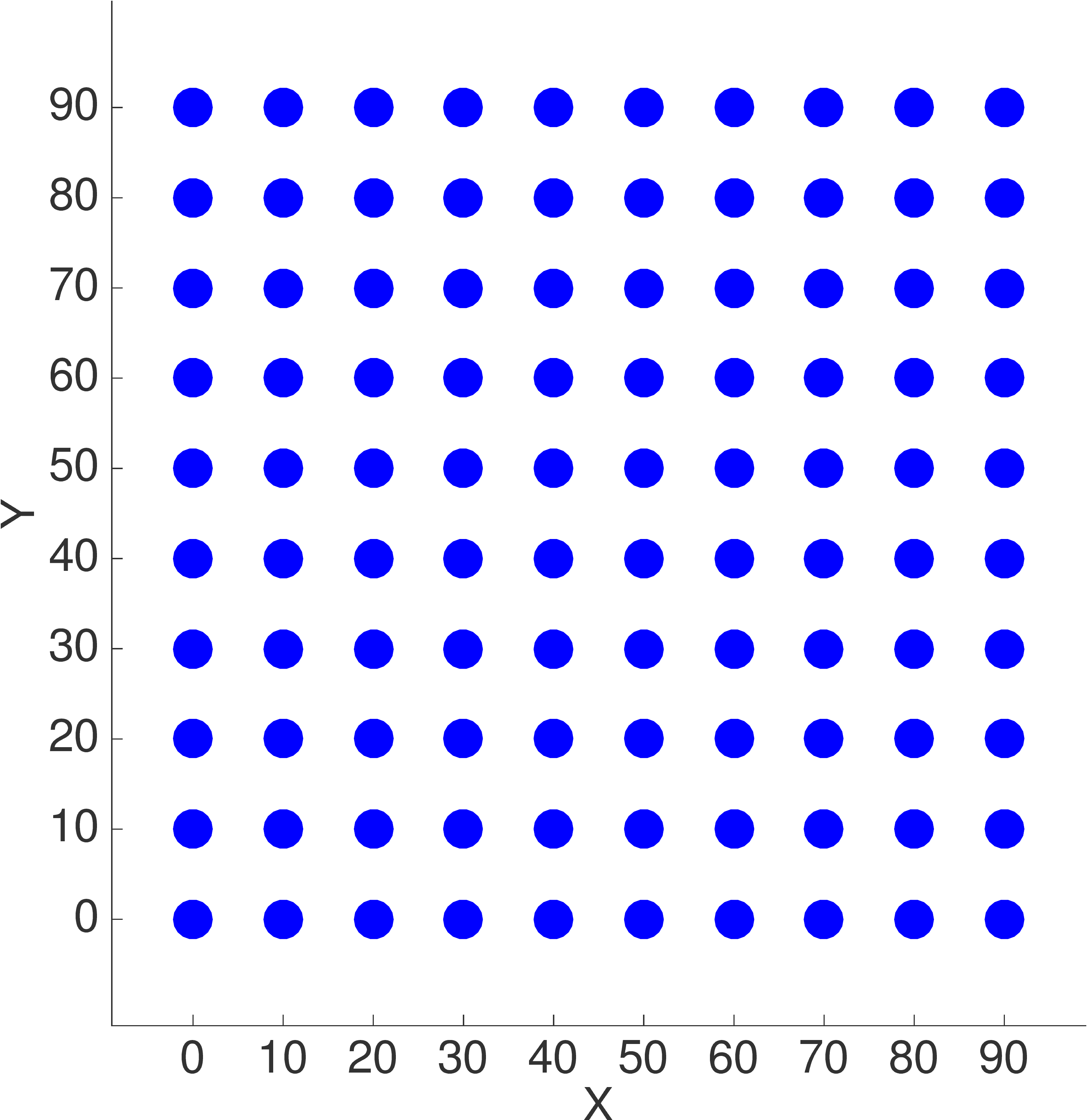}
\hspace*{0.5cm}
\includegraphics[width=3.5cm,angle=0]{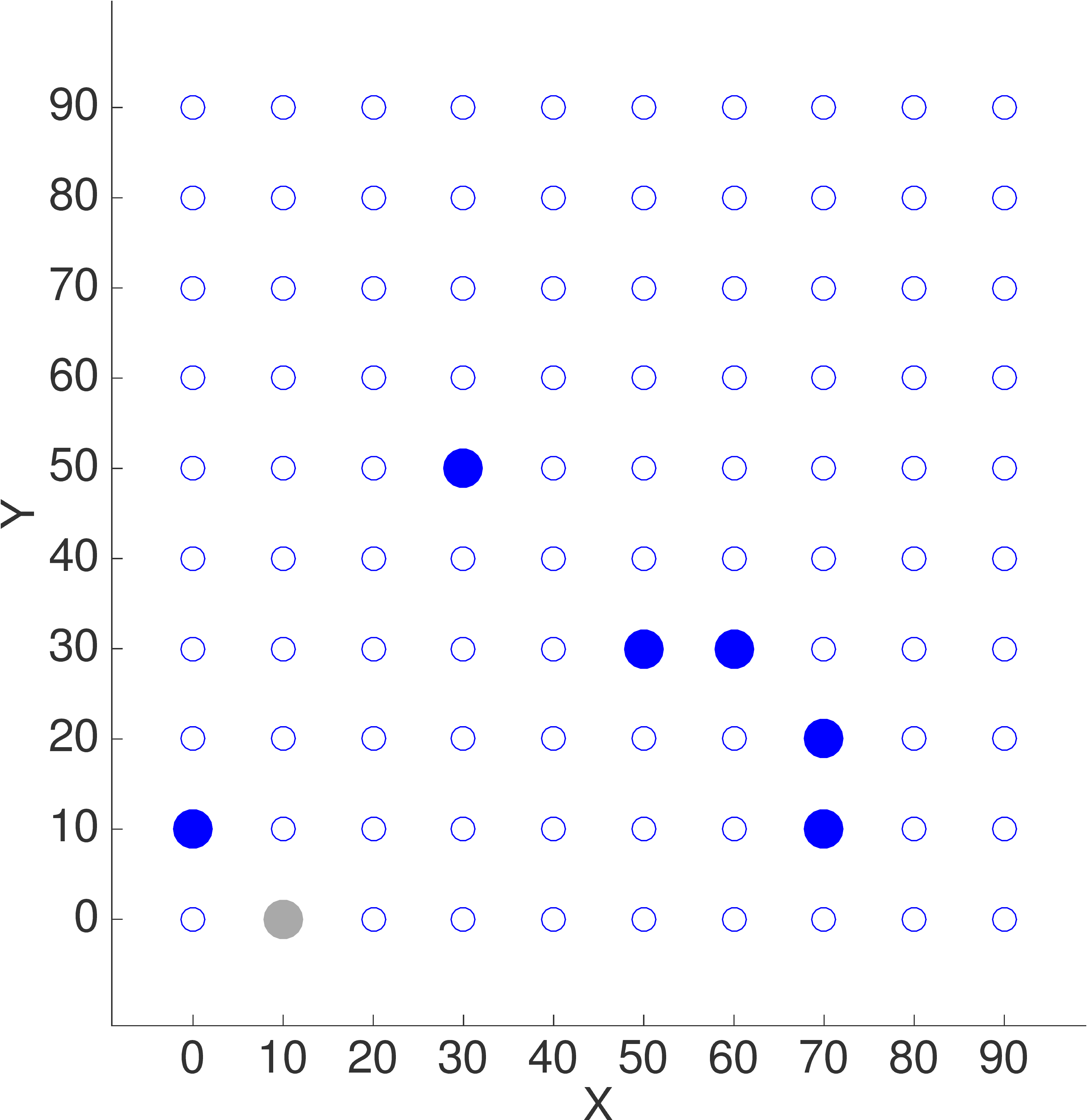}
\hspace*{0.5cm}
\includegraphics[width=3.5cm,angle=0]{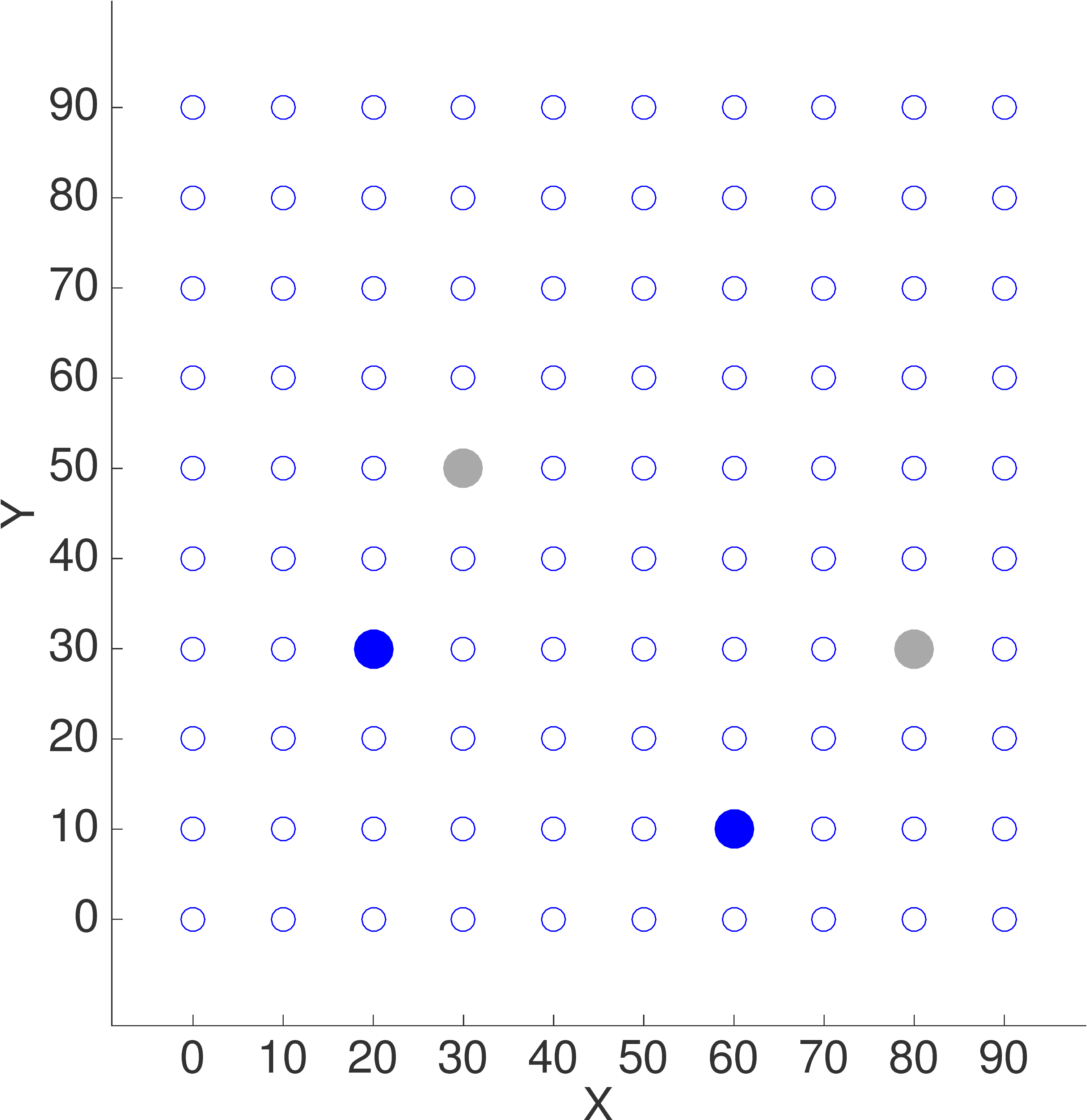}
\hspace*{0.5cm}
\includegraphics[width=3.5cm,angle=0]{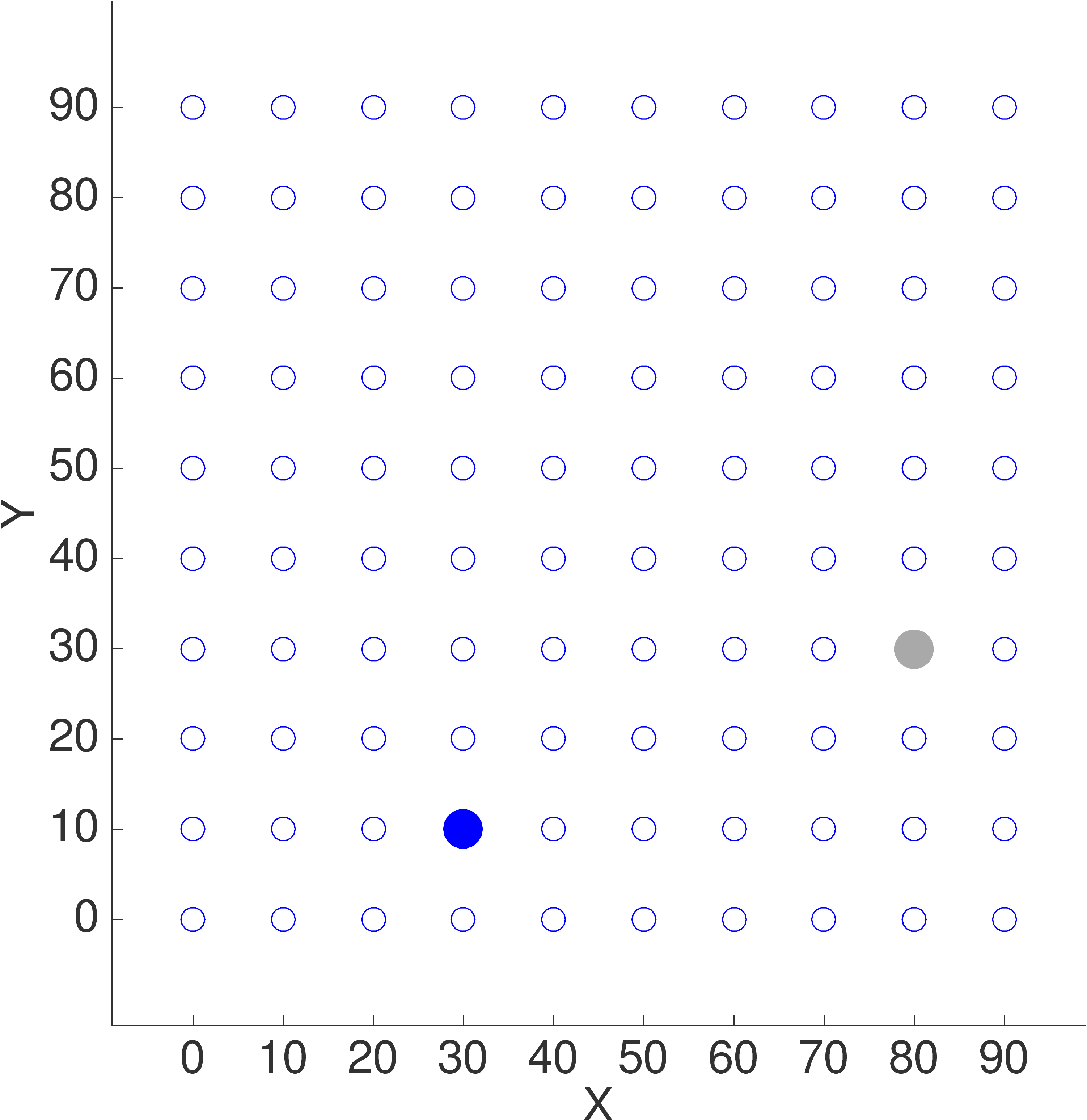}
}
\centerline{$T = \Delta t$  \hspace*{2.5cm} $T = 2 \Delta t$  \hspace*{2.5cm} $T = 3 \Delta t$ \hspace*{2.5cm} $T = 4 \Delta t$}
\vspace*{0.3cm}
\centerline{ 
\includegraphics[width=3.5cm,angle=0]{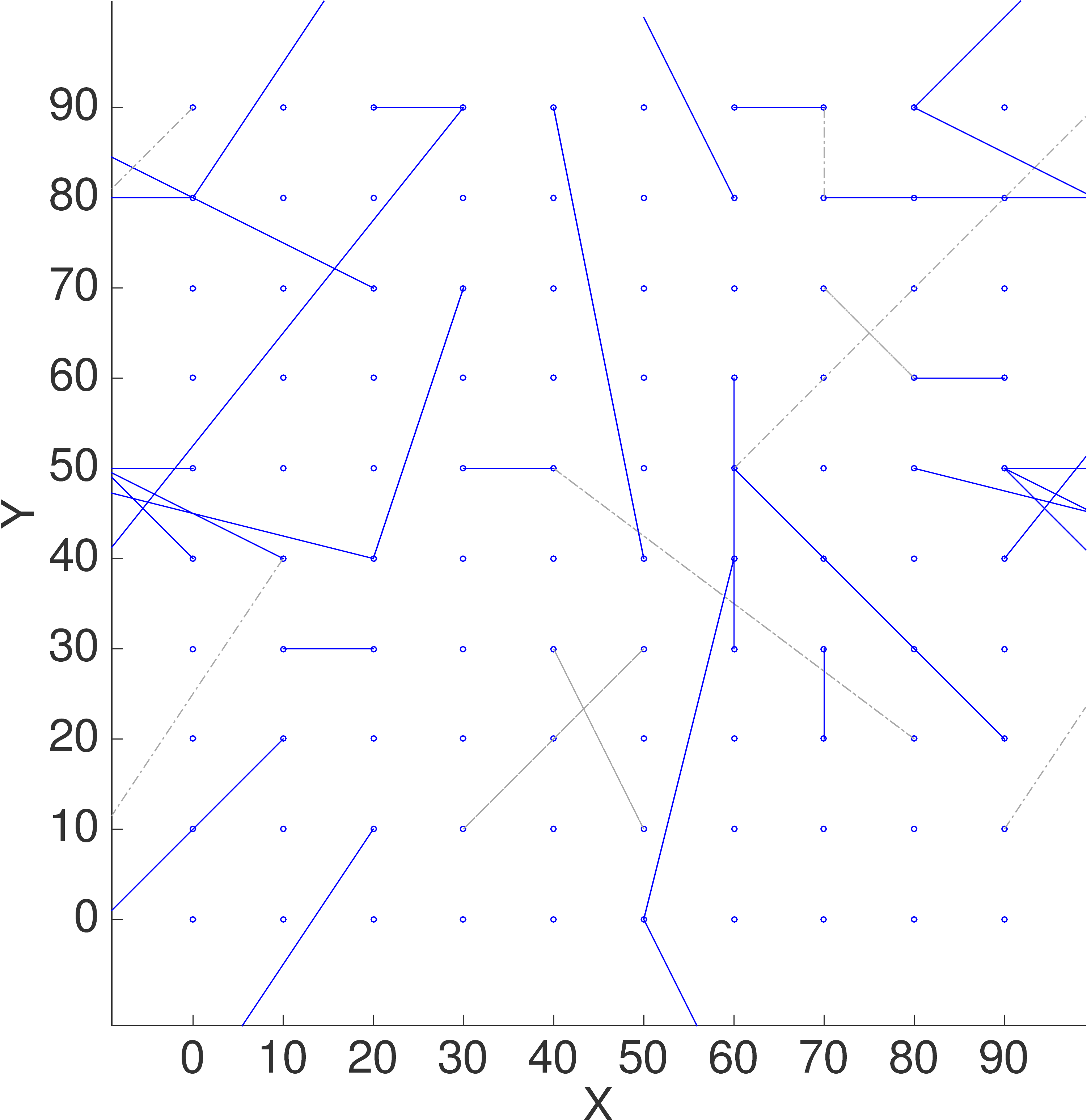}
\hspace*{0.5cm}
\includegraphics[width=3.5cm,angle=0]{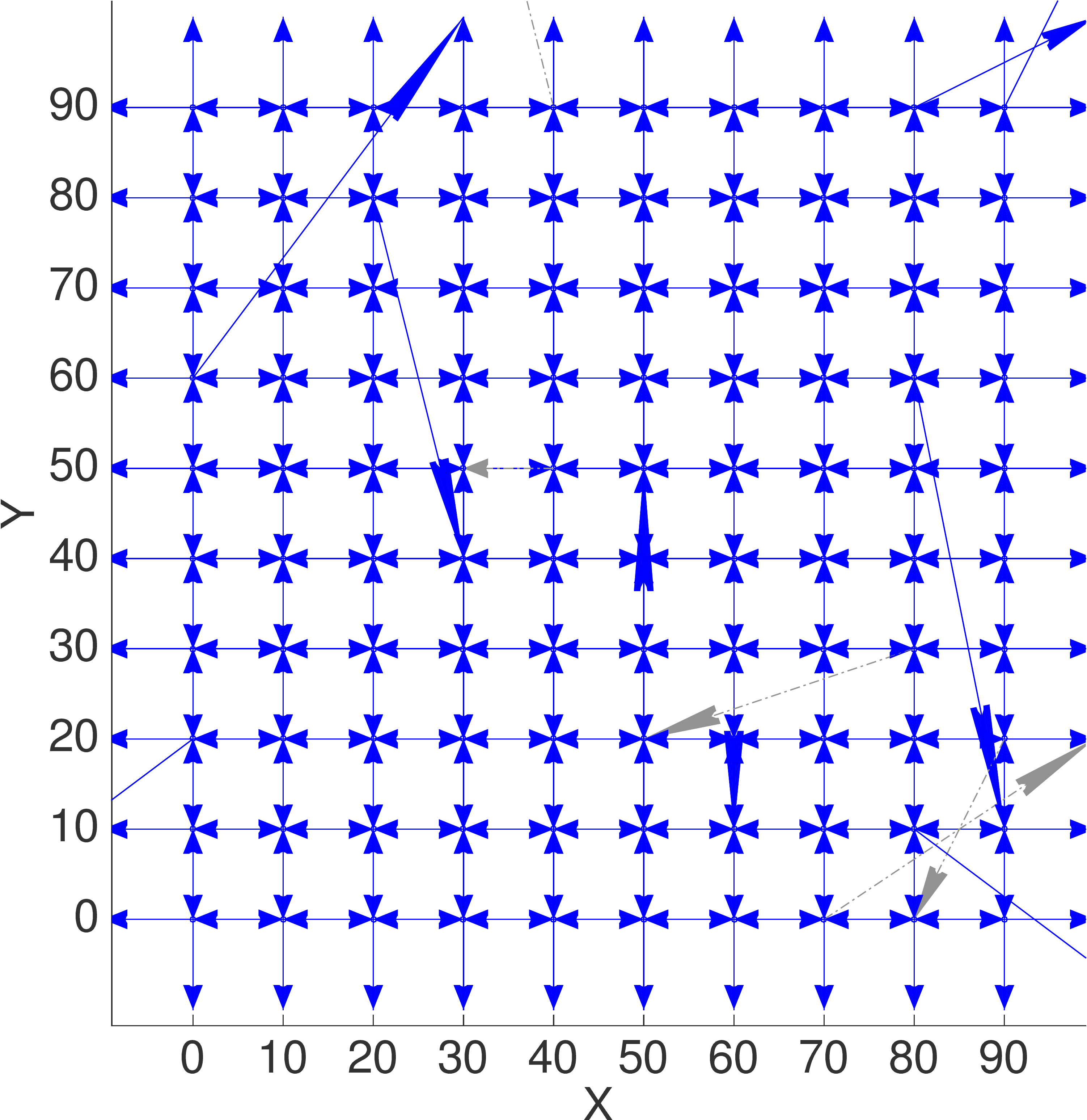}
\hspace*{0.5cm}
\includegraphics[width=3.5cm,angle=0]{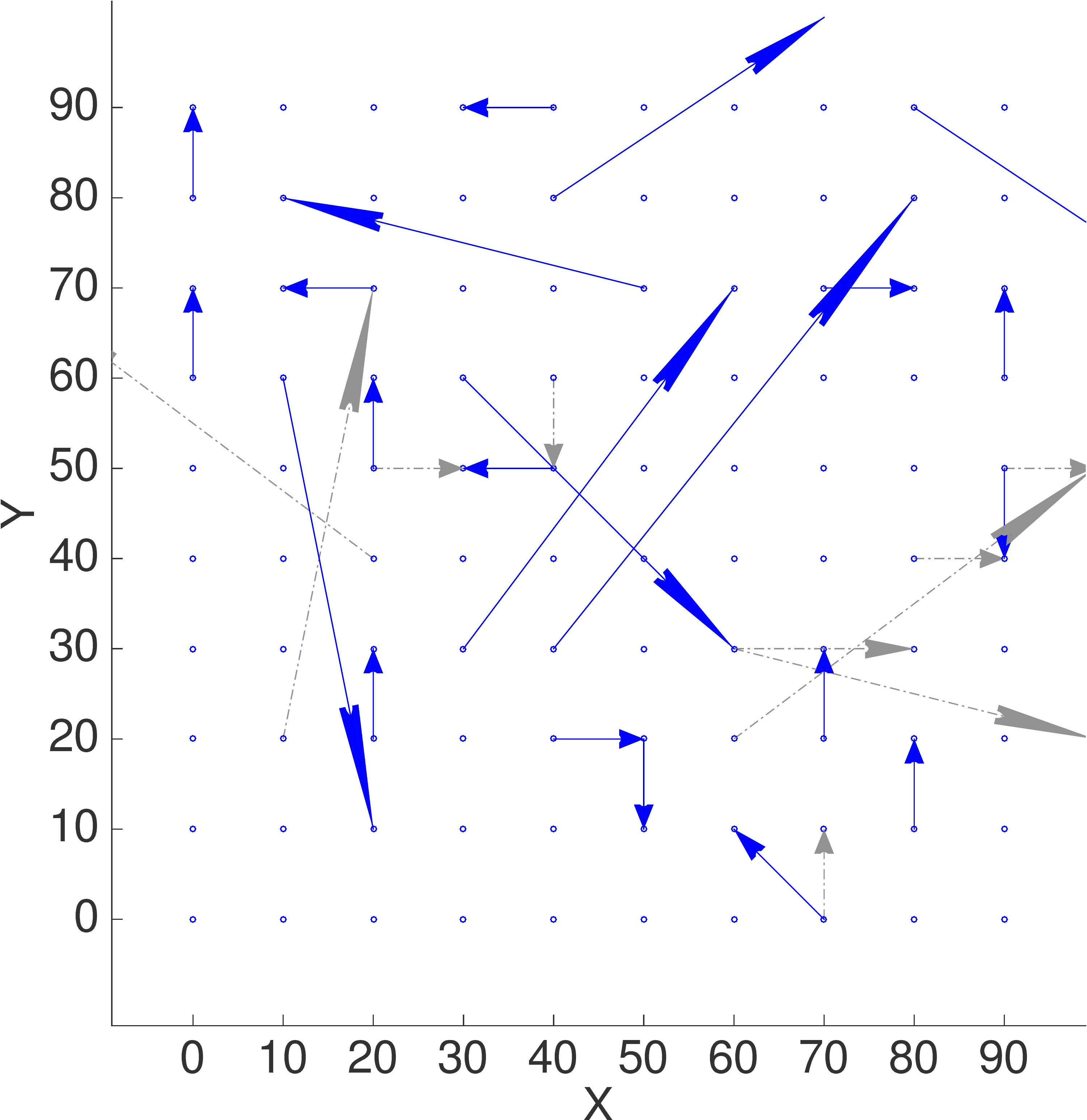}
\hspace*{0.5cm}
\includegraphics[width=3.5cm,angle=0]{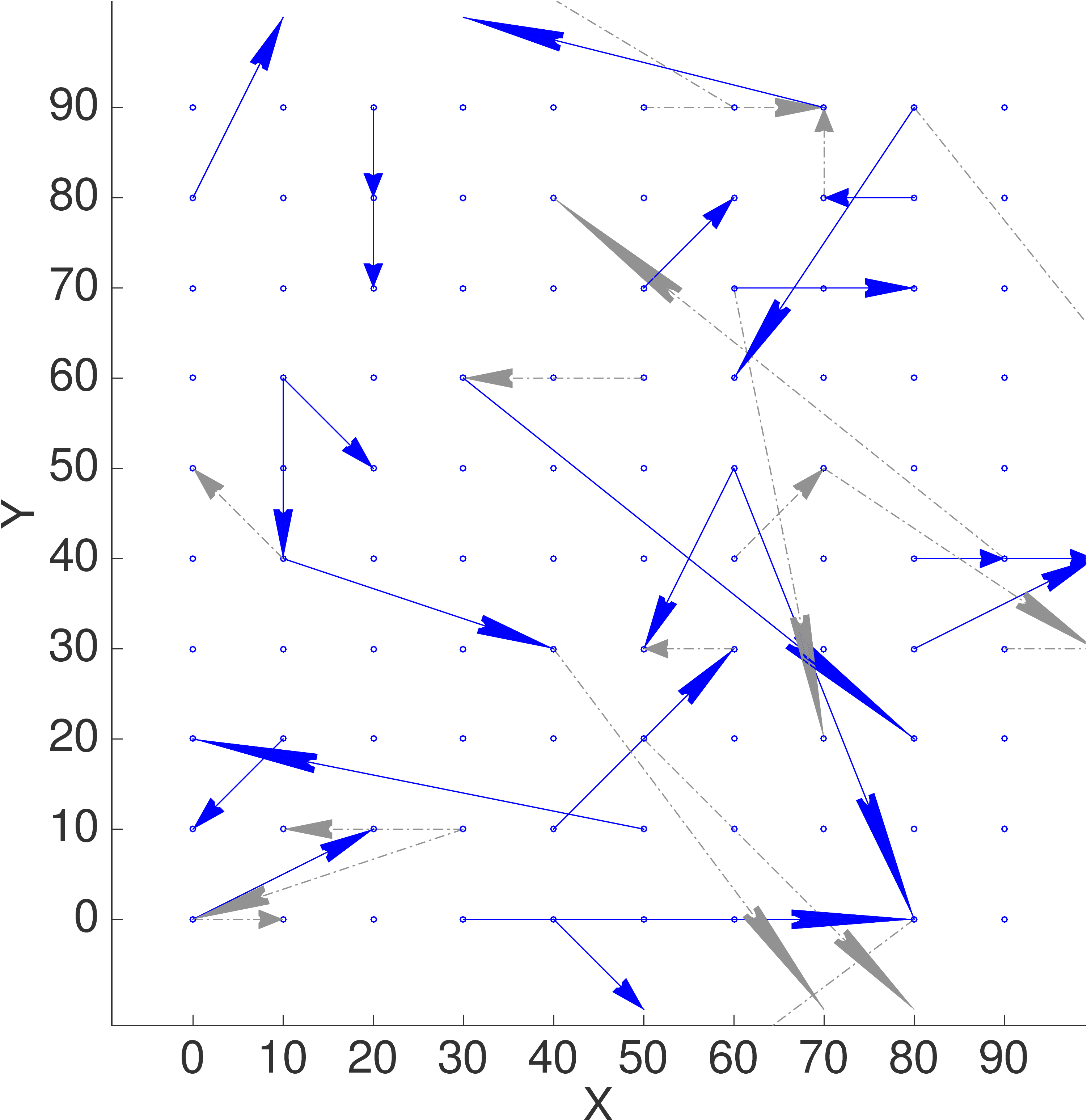}
}
\centerline{$T = 0$  \hspace*{2.5cm} $T = \Delta t$  \hspace*{2.5cm} $T = 2 \Delta t$  \hspace{2.5cm} $T = 3 \Delta t$}
\centerline{(b) Intra (top) and inter (bottom) edges for all-point noise}
\caption{Inter and intra edges for pure diffusion experiment, same parameters as in Fig.\ \ref{pure_diffusion_fig_1}, but using single-point or all-point noise instead of single-point peaks.
Type 1 and Type 2 velocity estimates (not shown) are close to zero throughout in both cases. 
\label{pure_diffusion_noise_fig}}
\end{figure*}

Figure \ref{pure_diffusion_fig_1} shows results for a 10x10 point grid,
$\kappa_x=\kappa_y=\kappa$, and $M=1$, i.e.\ full temporal resolution, while
Figure \ref{idealized_diffusion_results_fig} focuses on the concurrent inter connections 
for different values of $M$ and $\kappa_y$.  Figures \ref{pure_diffusion_fig_1} and \ref{idealized_diffusion_results_fig} use only Message Type 1.
Figure \ref{pure_diffusion_noise_fig} shows results when Message Types 2 and 3 are used instead.

{\bf When using Message Type 1 (IC peak)}, we found that\\ 
(1)
   	Intra edges are dominant, with local memory lasting several time steps, i.e.\ all locations have intra edges extending across $T=\Delta t, 2 \Delta t, 3 \Delta t$, etc.
\\
(2)   	No inter edges occur for $T>0$.  
\\
(3)
	There may or may not be concurrent edges ($T=0$), see Figure 	
	\ref{idealized_diffusion_results_fig}.
        While for some grids no edges occurred,
        the more common scenario was the four-neighbor pattern, 
        where each location connects to its 4 closest neighbors in the $\pm$x and $\pm$y 
        directions.  
	This occurred in most experiments for $k_x = k_y > 0$, i.e.\ 
        for equal diffusion in all directions.
        When diffusion was only in x-direction, i.e.\
        $k_x>0, k_y=0$, each location connects only to its 2 closest neighbors 
        in the $\pm$x direction.
\\
(4)
	Both types of velocity plots report zero average velocities, which makes sense, since diffusion  moves equally in all directions, i.e.\ the contributions cancel each other out.


{\bf When using Message Types 2 \& 3 (prior noise) instead} we found that
\\
(1)
  	For the majority of locations the local memory decreased (especially for all-point noise) to just one time step (see Fig.\ \ref{pure_diffusion_noise_fig}), but the intra edges are still strong for the first time step, $T= \Delta t$.
\\
(2)	The undirected edges found previously for $T=0$ shift completely to $T= \Delta t$, i.e.\ now there are neighboring directed edges pointing from each point to each neighboring point in x and y directions.  These edges make sense, since diffusion also causes information transfer when a peak spreads to neighboring locations.  That information transfer is just very slow and not traveling very far before decaying.  That is probably the reason why it was 
not identified as directed edges in the IC peak case. 
\\
(3)	Using noise as the only input signal creates several stray edges, especially for all-point noise.
\\
(4)	Type 1 and 2 velocity plots (not shown here) indicate that the estimated velocity is close to zero throughout, since the directed inter edges cancel each other out.  
There are only small non-zero values arising from the randomness in the noise.

{\bf Summary:} For pure diffusion strong intra edges dominate as expected, 
while inter connections can be found between neighboring grid points, 
either as concurrent edges (Message Type 1) or 
as nonconcurring edges (Message Types 2 \& 3).  
The estimated velocity throughout is near zero, which closely matches the 
zero advection velocity in this scenario.

\subsection{Results for Pure Advection}

\begin{figure*}
\centerline{ 
\includegraphics[width=3.5cm,angle=0]{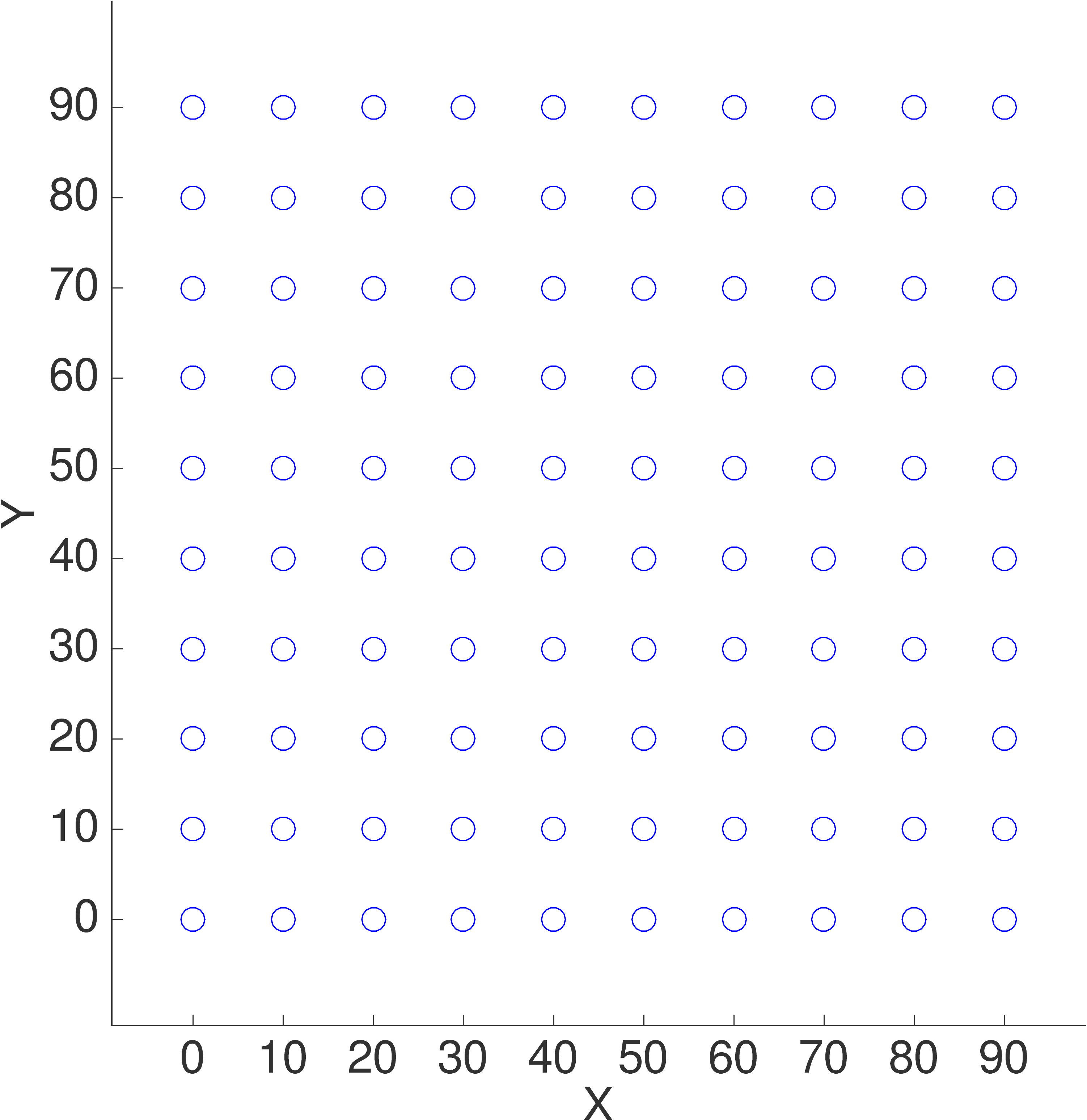}
\hspace*{0.5cm}
\includegraphics[width=3.5cm,angle=0]{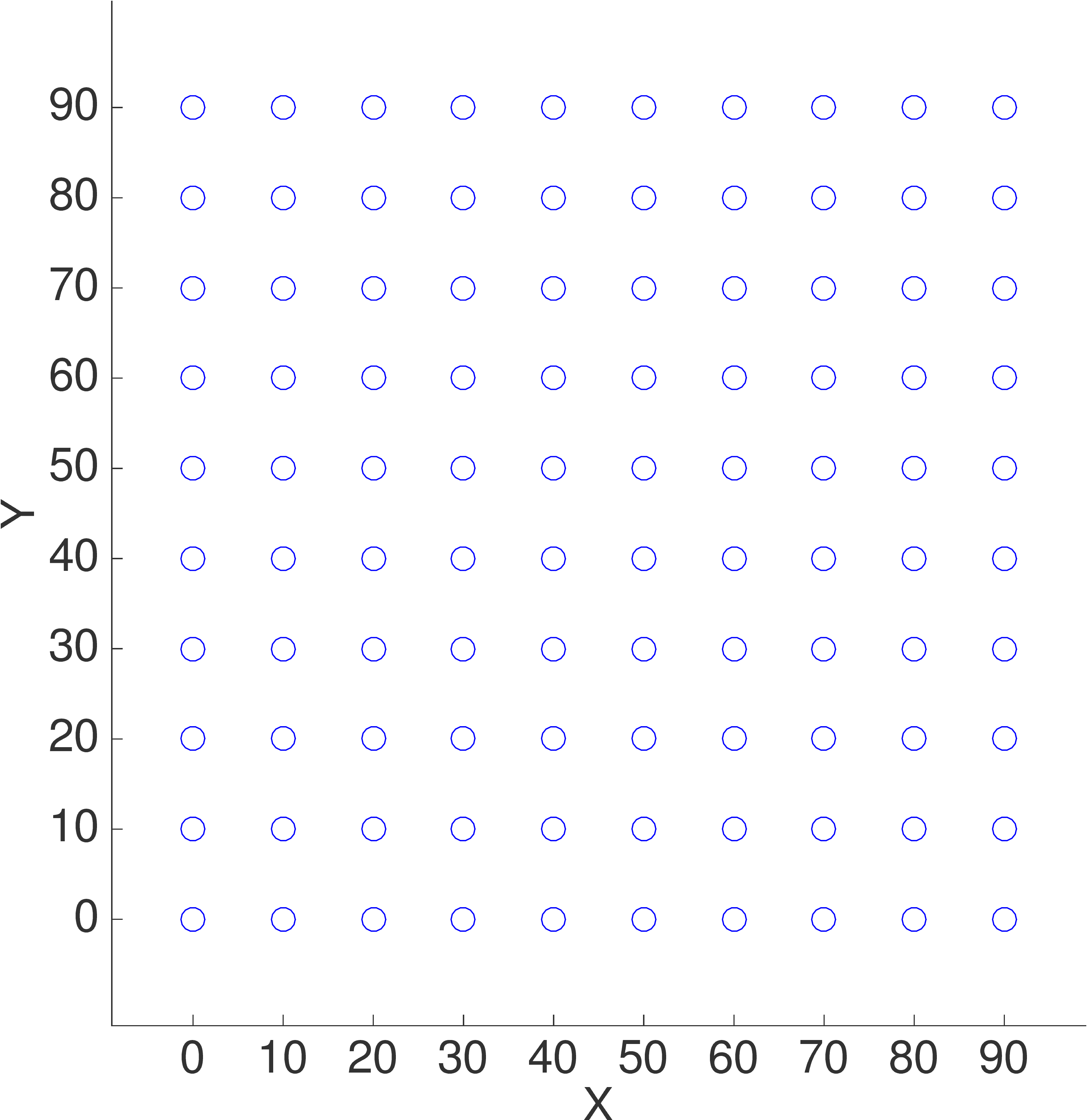}
\hspace*{0.5cm}
\includegraphics[width=3.5cm,angle=0]{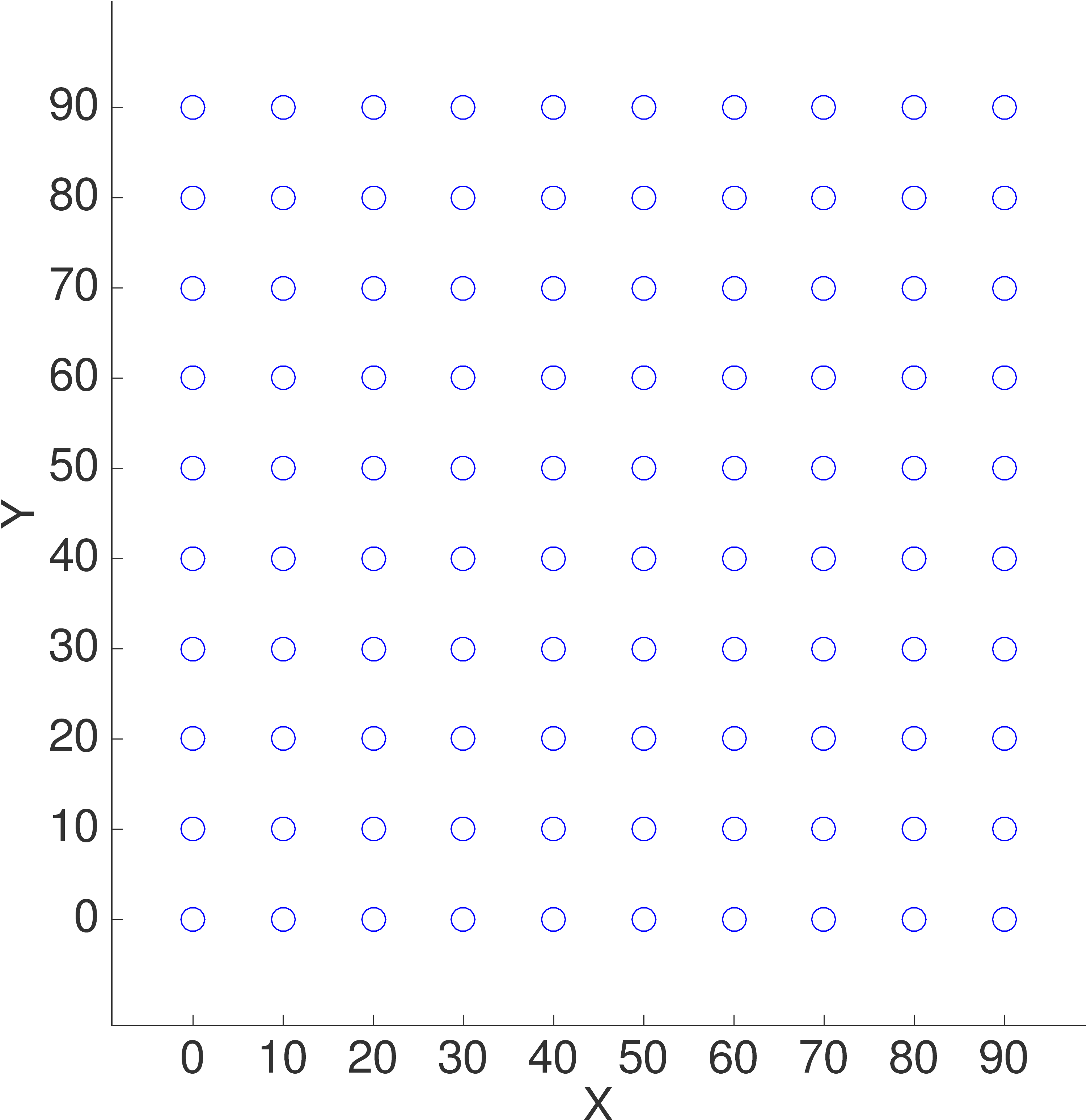}
\hspace*{0.5cm}
\includegraphics[width=3.5cm,angle=0]{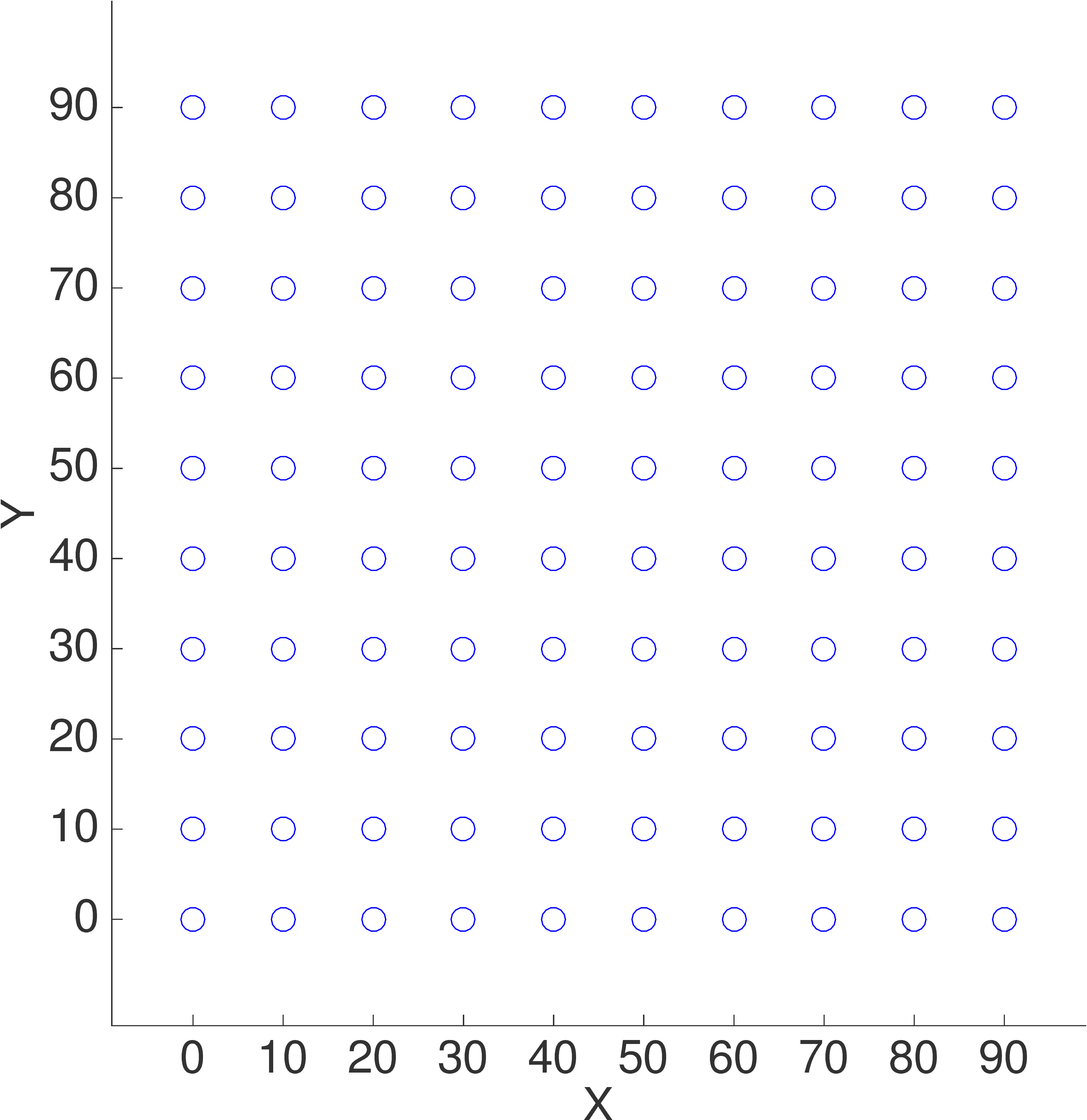}
}
\centerline{$T = \Delta t$  \hspace*{2.5cm} $T = 2 \Delta t$  \hspace*{2.5cm} $T = 3 \Delta t$ \hspace*{2.5cm} $T = 4 \Delta t$}
\centerline{(a) Intra edges for pure advection and $M=1,2,4$: none.}
\vspace*{0.3cm}
\centerline{ 
\includegraphics[width=3.5cm,angle=0]{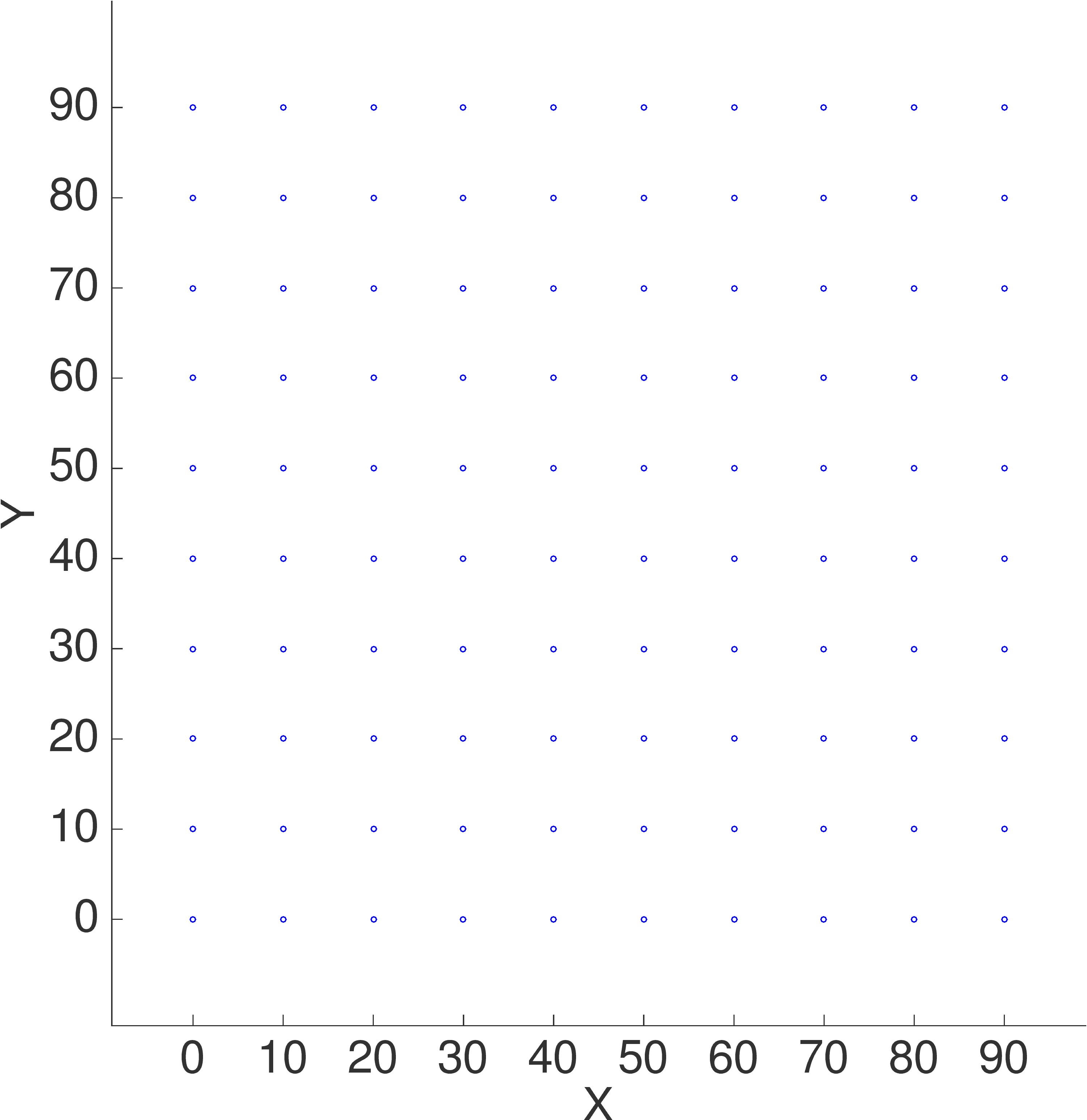}
\hspace*{0.5cm}
\includegraphics[width=3.5cm,angle=0]{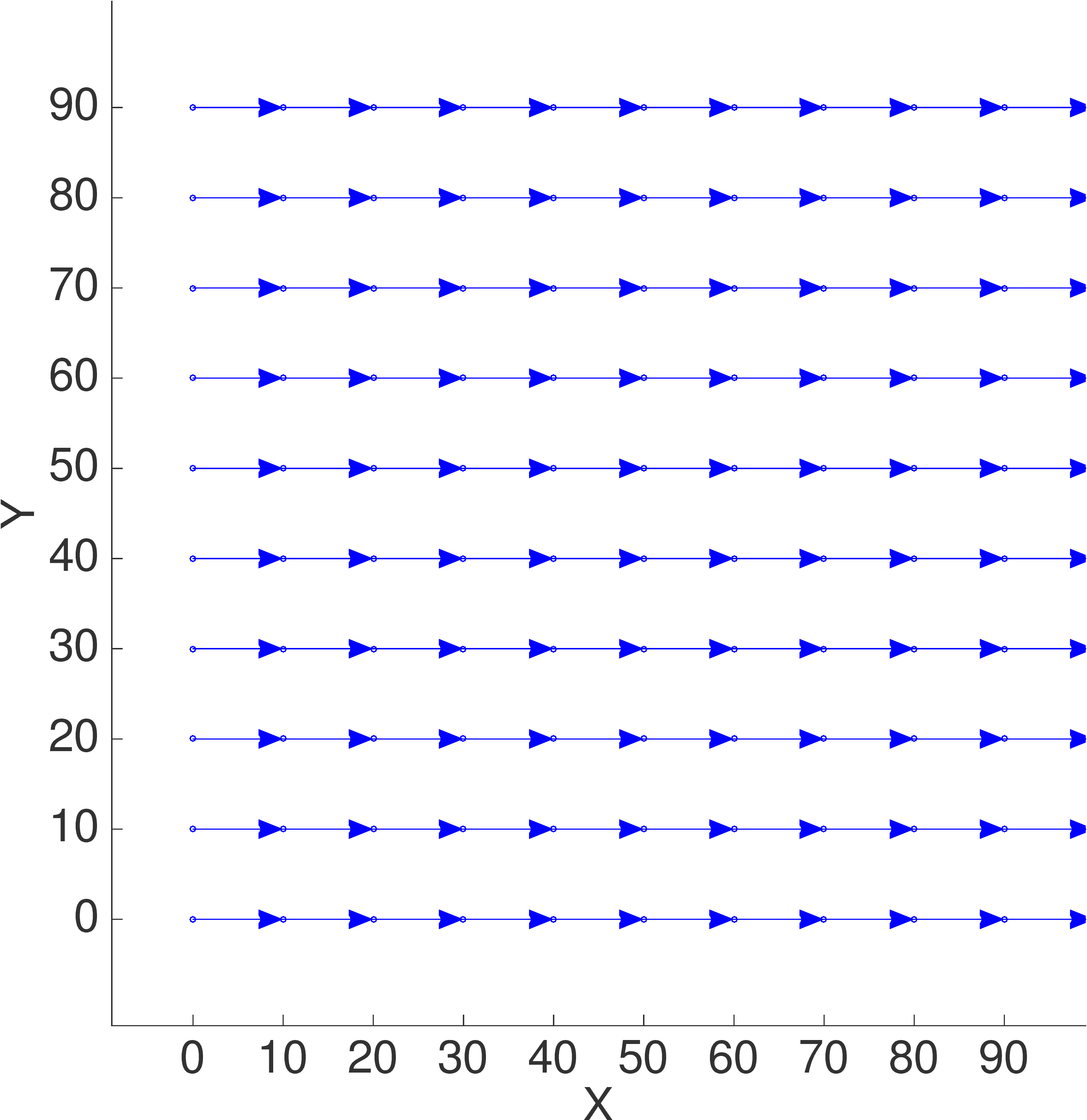}
\hspace*{0.5cm}
\includegraphics[width=3.5cm,angle=0]{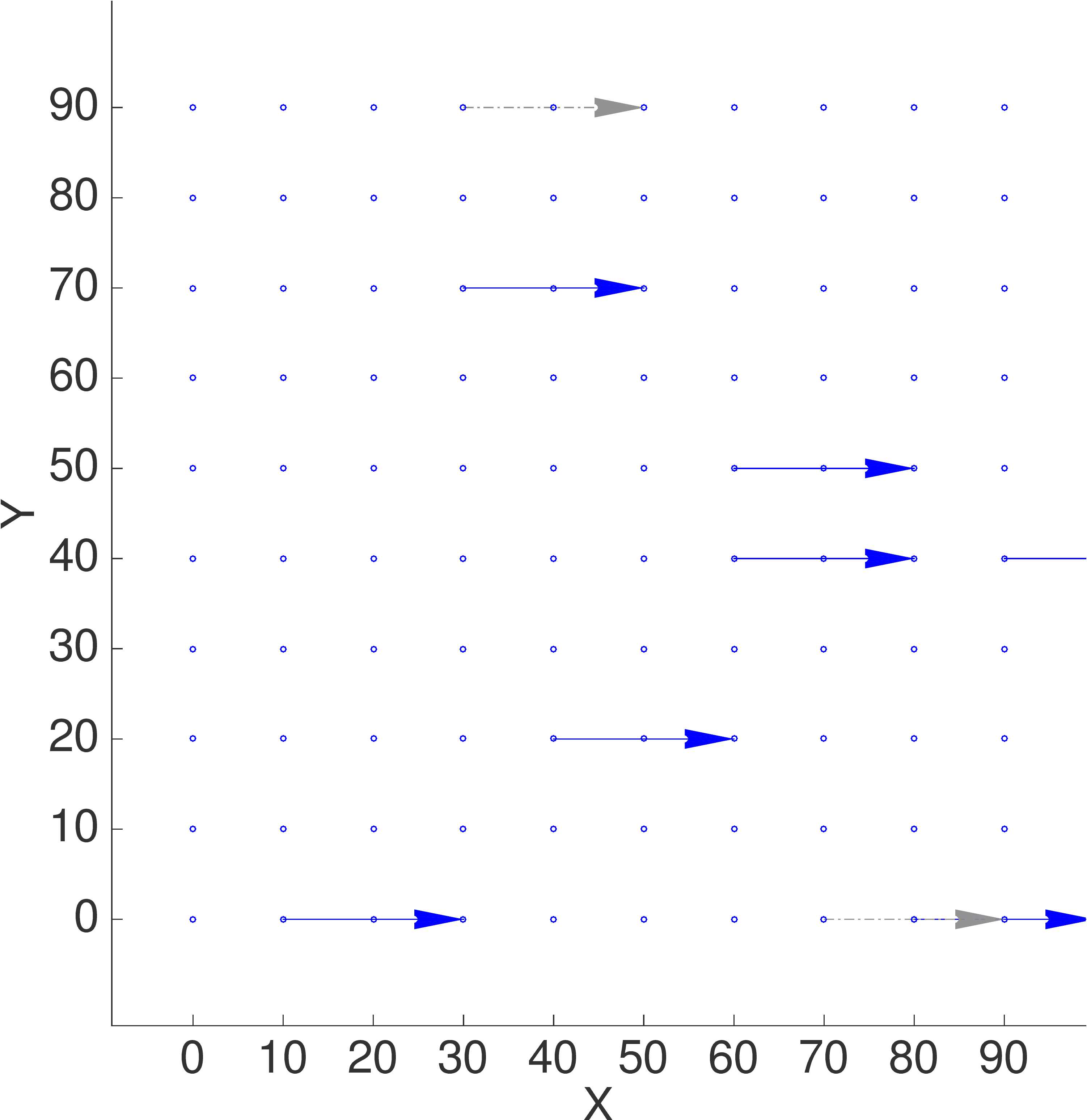}
\hspace*{0.5cm}
\includegraphics[width=3.5cm,angle=0]{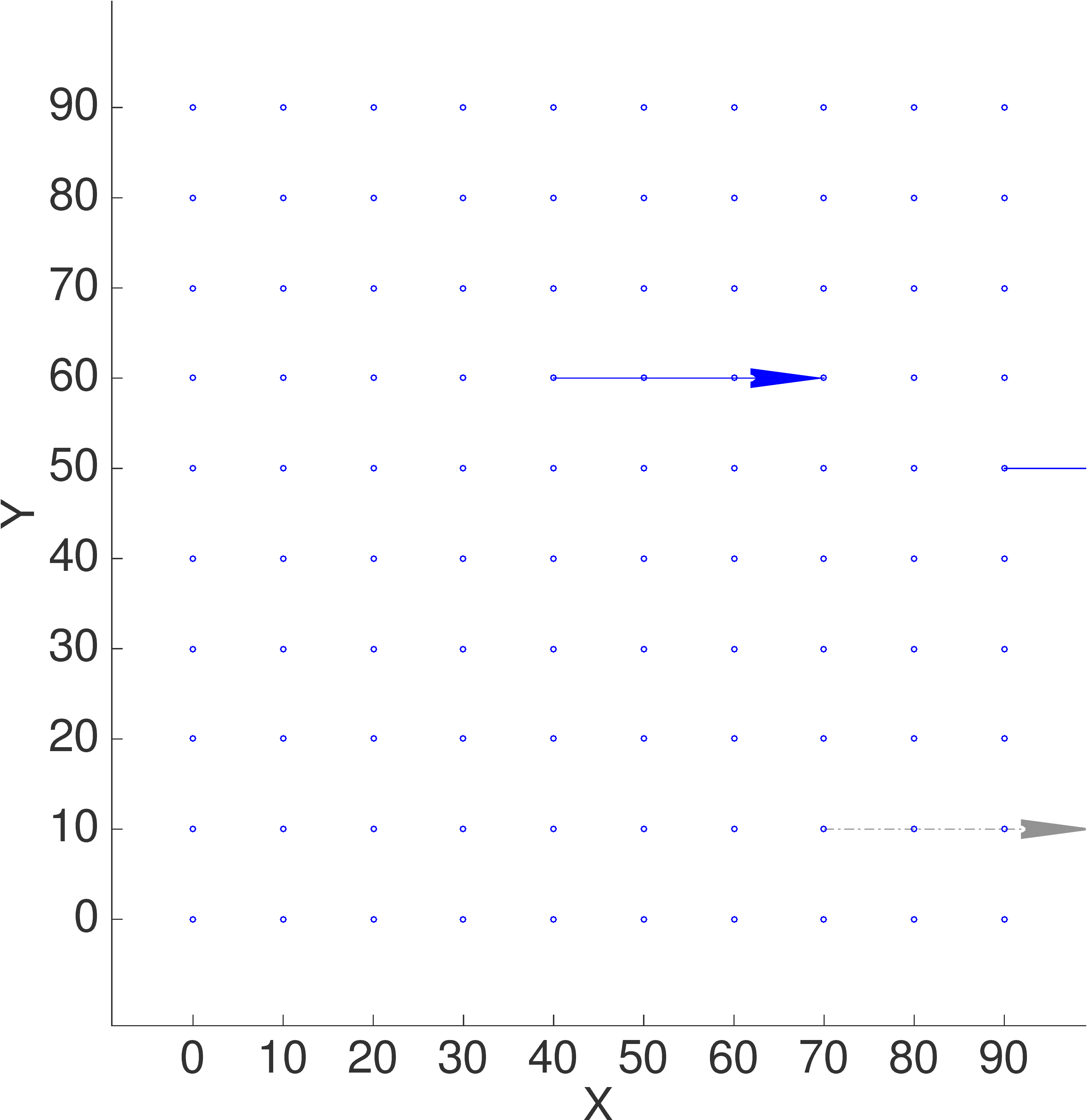}
}
\centerline{$T = 0$  \hspace*{2.5cm} $T = \Delta t$  \hspace*{2.5cm} $T = 2 \Delta t$  \hspace{2.5cm} $T = 3 \Delta t$}
\centerline{(b) Inter edges for pure advection and $M=1$}
\vspace*{0.3cm}
\centerline{ 
\includegraphics[width=3.5cm,angle=0]{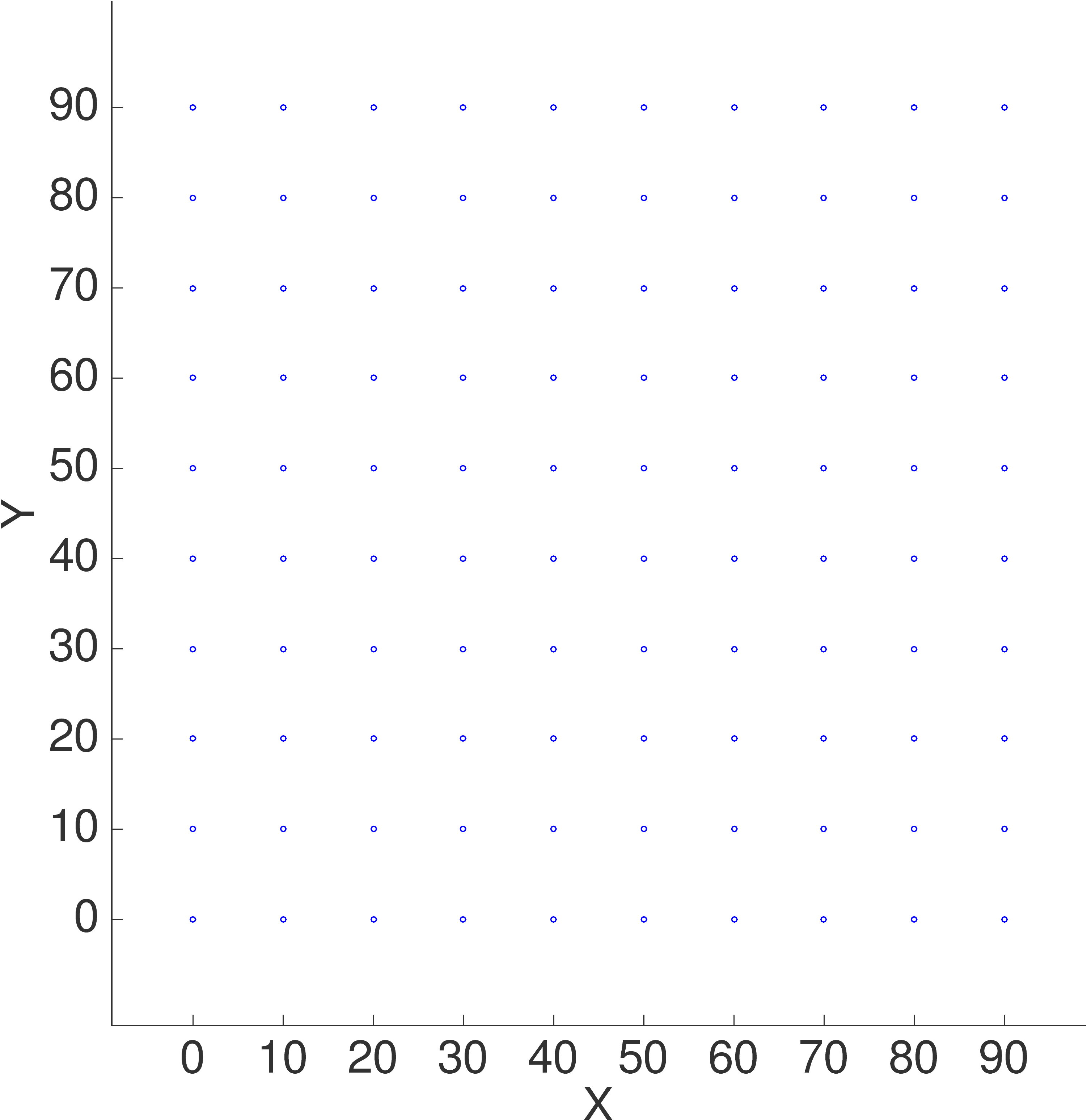}
\hspace*{0.5cm}
\includegraphics[width=3.5cm,angle=0]{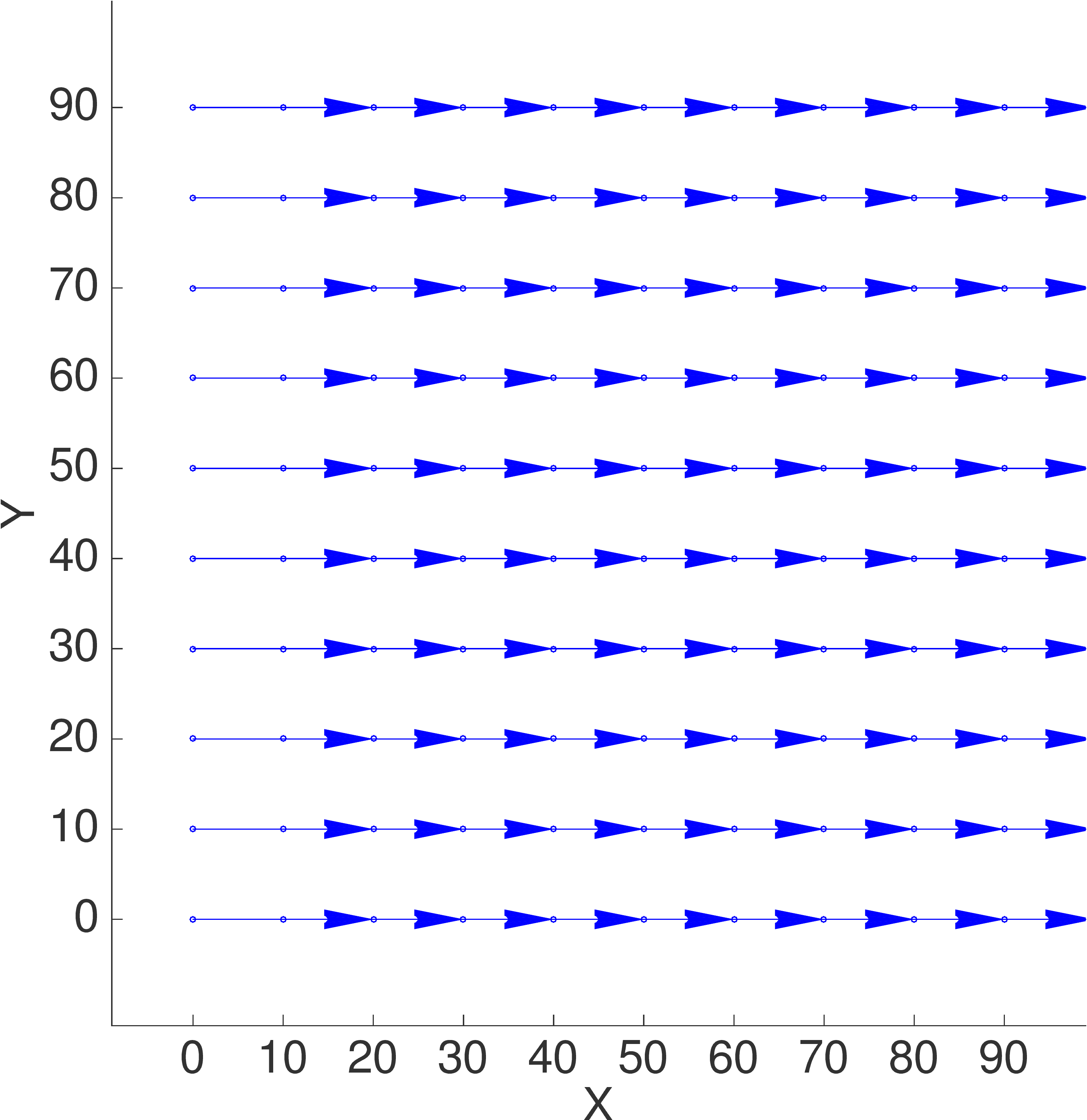}
\hspace*{0.5cm}
\includegraphics[width=3.5cm,angle=0]{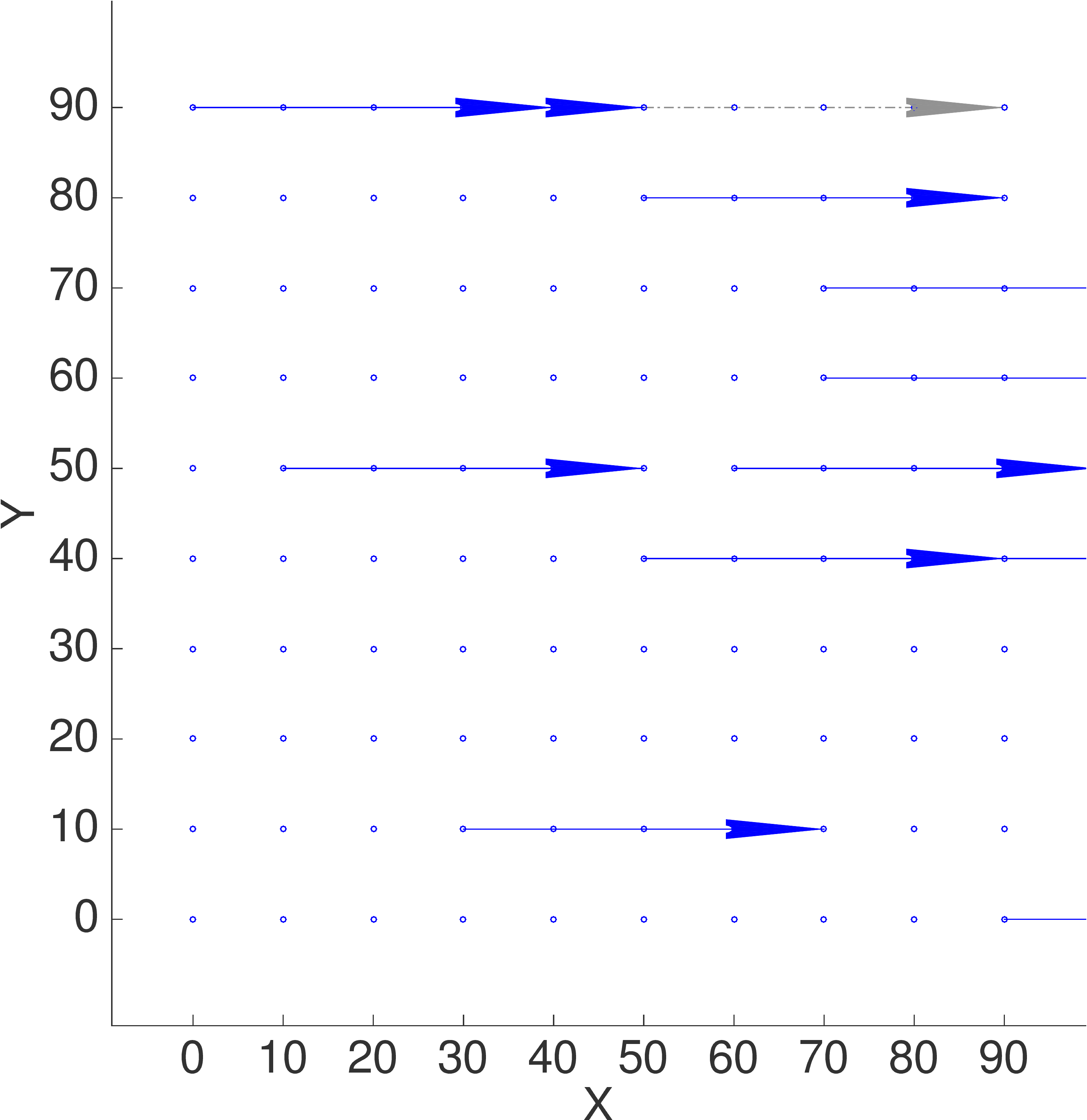}
\hspace*{0.5cm}
\includegraphics[width=3.5cm,angle=0]{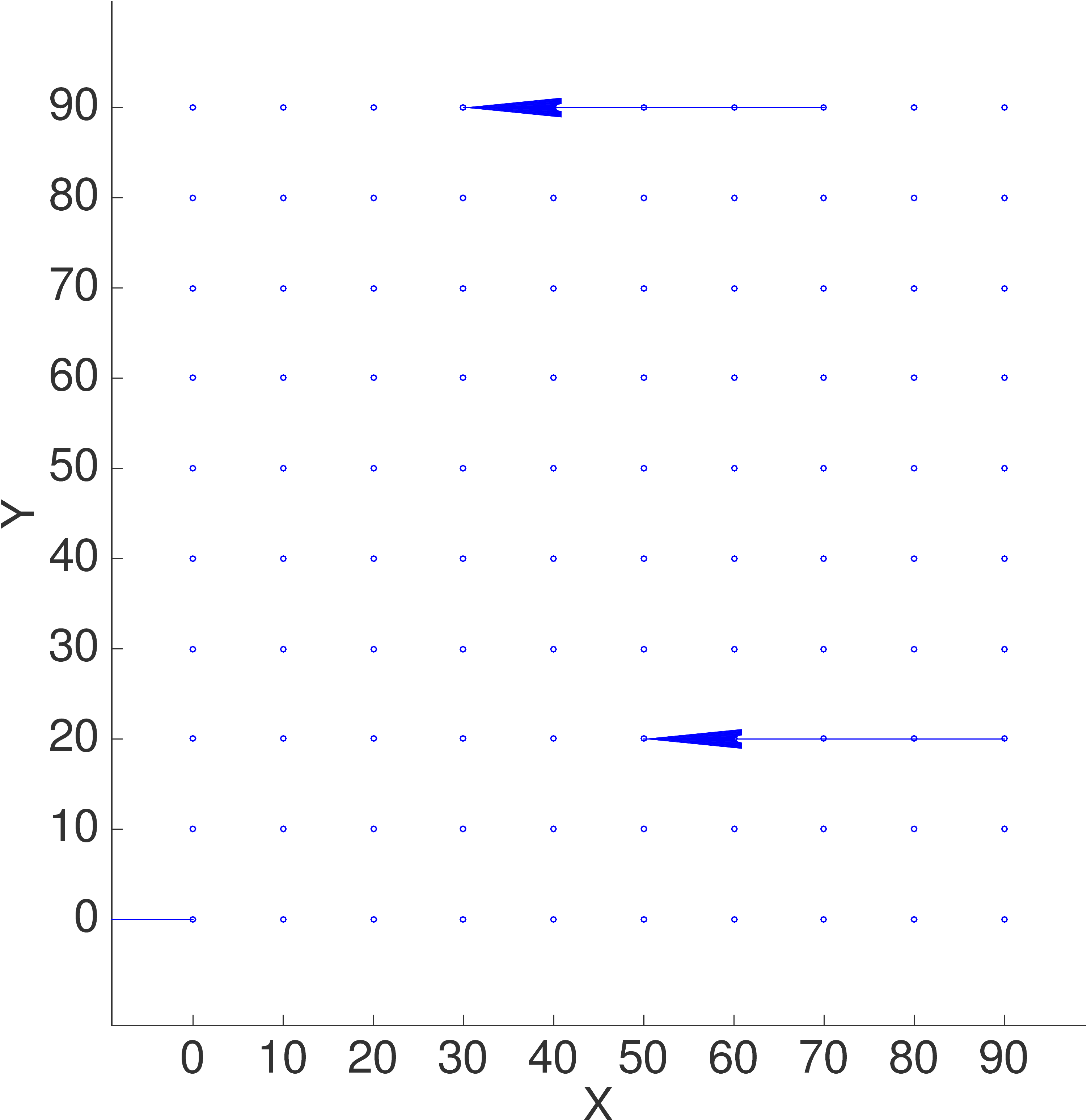}
}
\centerline{$T = 0$  \hspace*{2.5cm} $T = 2 \Delta t$  \hspace*{2.5cm} $T = 4 \Delta t$  \hspace{2.5cm} $T = 6 \Delta t$}
\centerline{(c) Inter edges for pure advection and $M=2$}
\vspace*{0.3cm}
\centerline{ 
\includegraphics[width=4.5cm,angle=0]{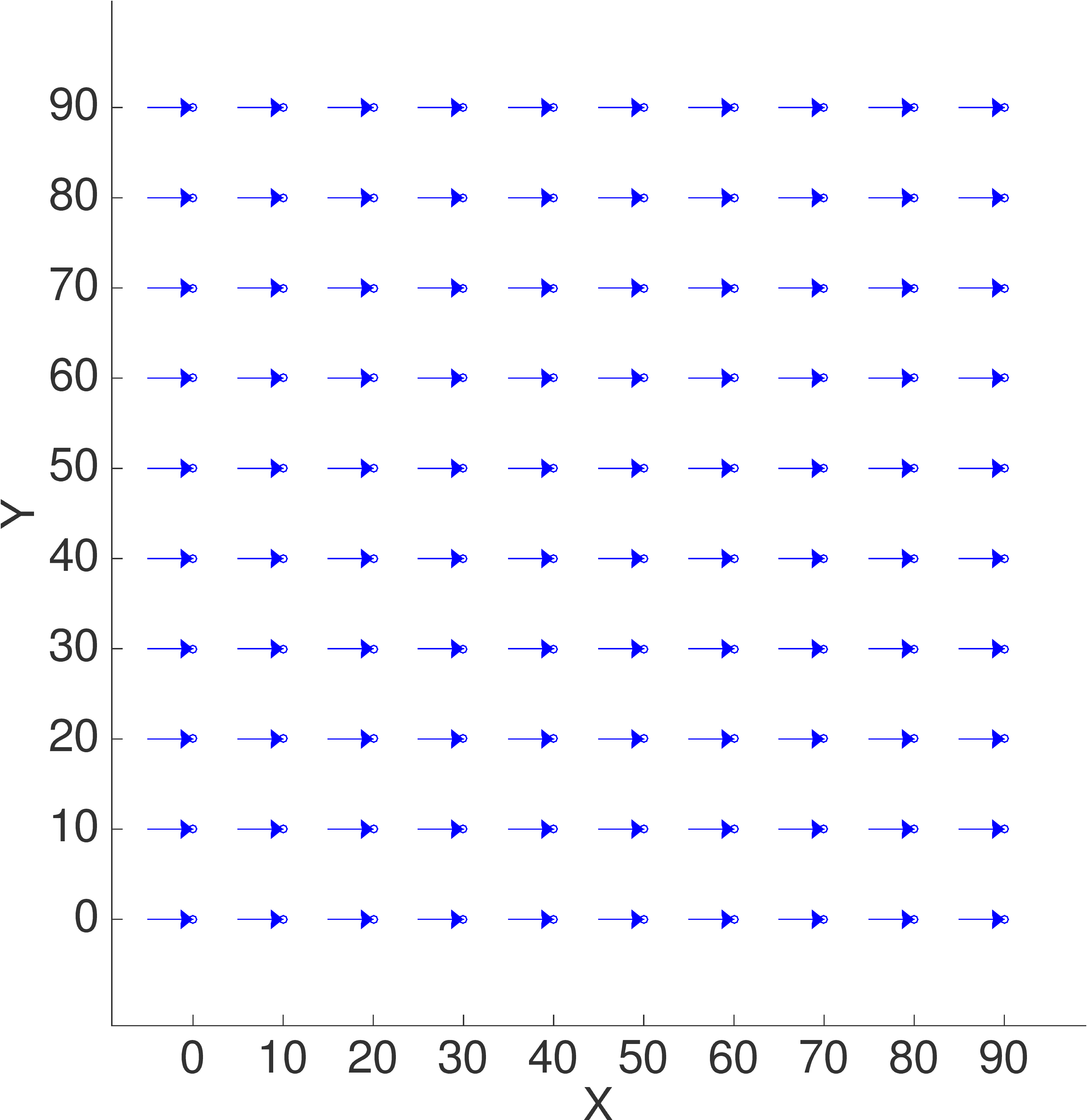}
\hspace*{0.8cm}
\includegraphics[width=4.5cm,angle=0]{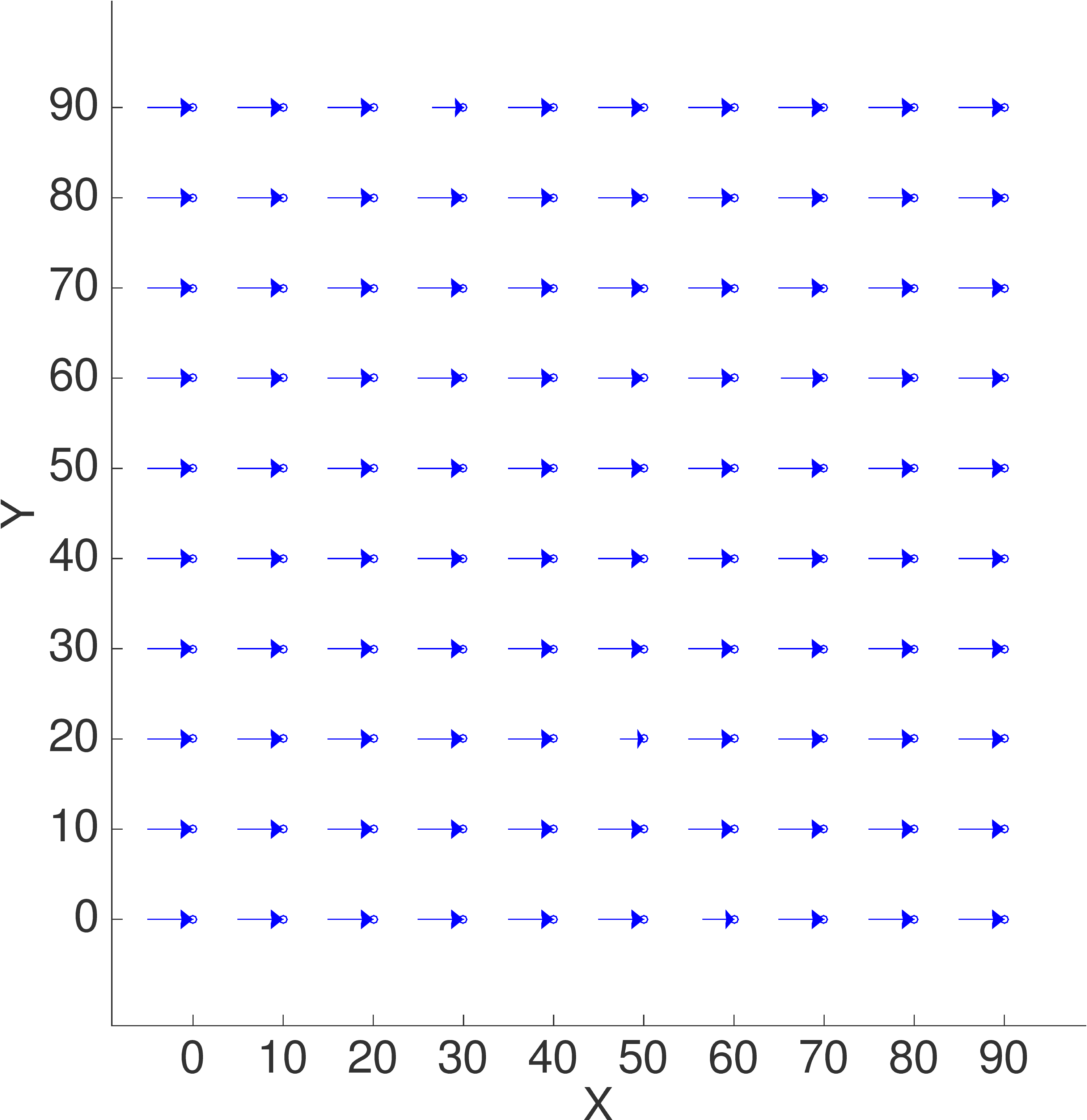}
\hspace*{0.8cm}
\includegraphics[width=4.5cm,angle=0]{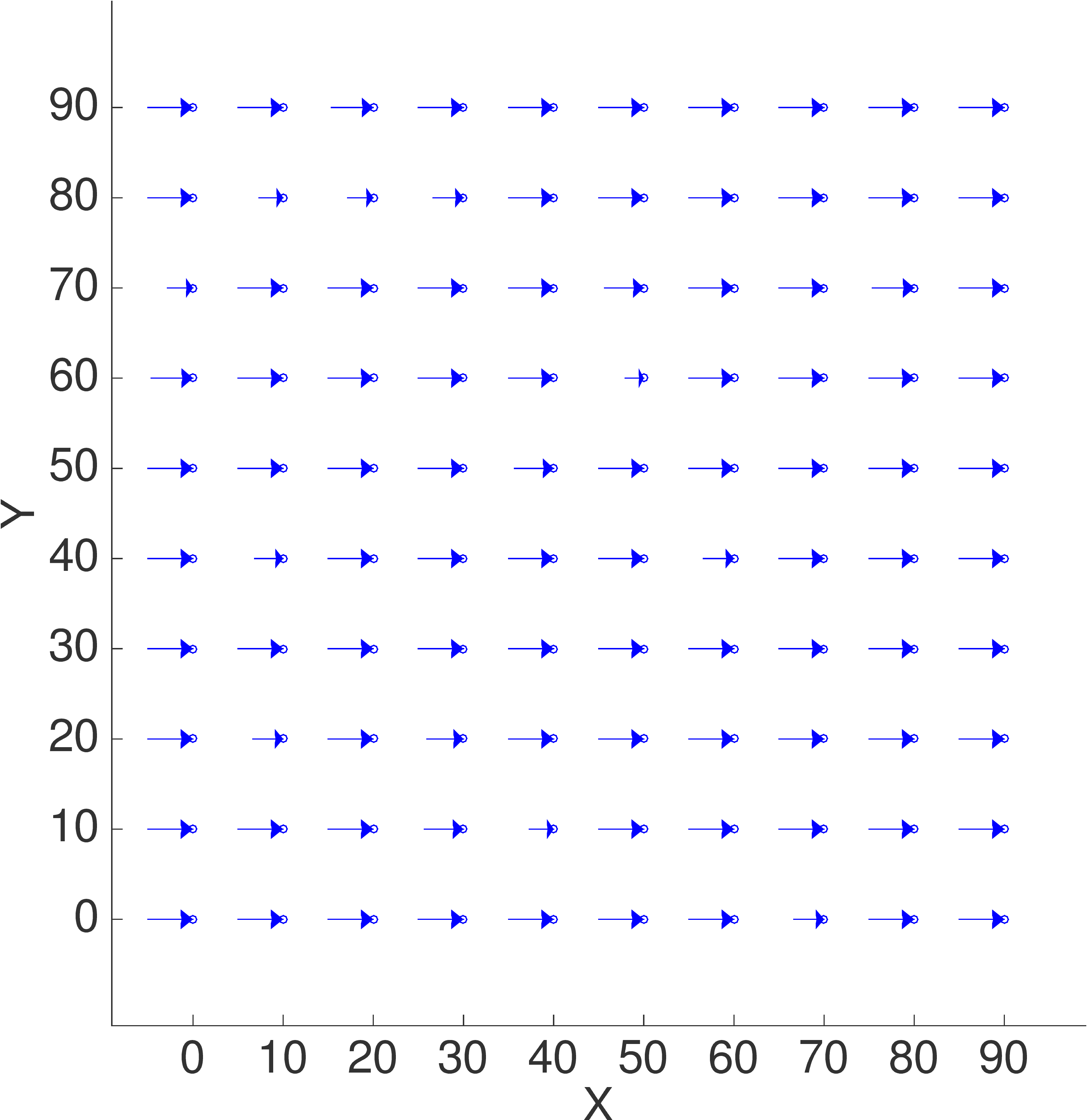}
}
\centerline{(d) Type 2 vel.\ for $M=1$ \hspace*{1.0cm} (e) Type 2 vel.\ for $M=2$ \hspace*{1.0cm} (f) Type 2 vel.\ for $M=4$  }
%
\caption{Results for advection experiments with $10 \times 10$ grid points and different temporal resolutions ($M=1,2,4$).   Velocity estimates (d,e,f) are scaled to $\Delta t=5$ sec and are identical with/without intra edges, since no intra edges were found.
\label{pure_advection_fig_1}}
\end{figure*}

\begin{figure*}
\centerline{ 
\includegraphics[width=3.5cm,angle=0]{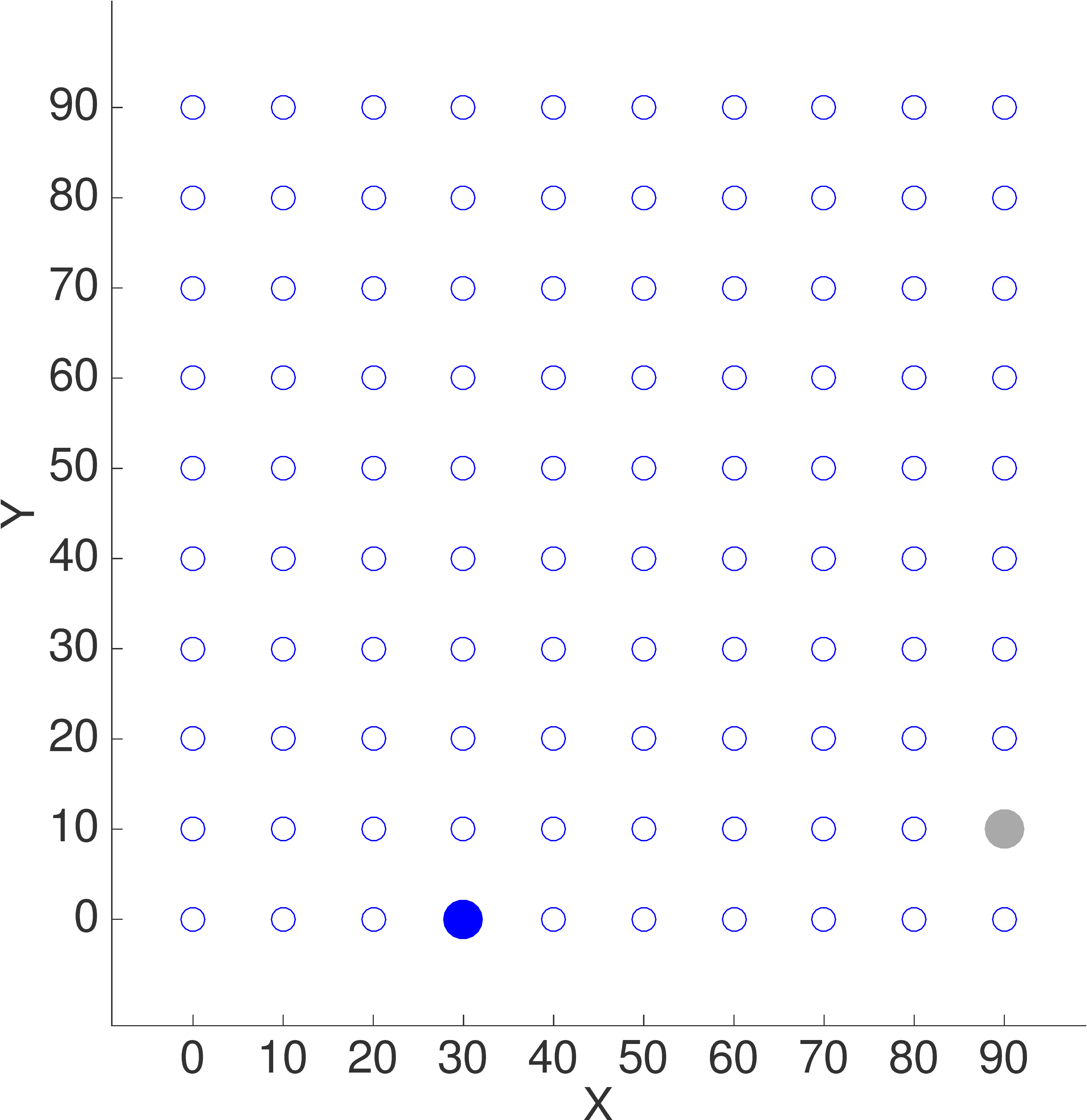}
\hspace*{0.5cm}
\includegraphics[width=3.5cm,angle=0]{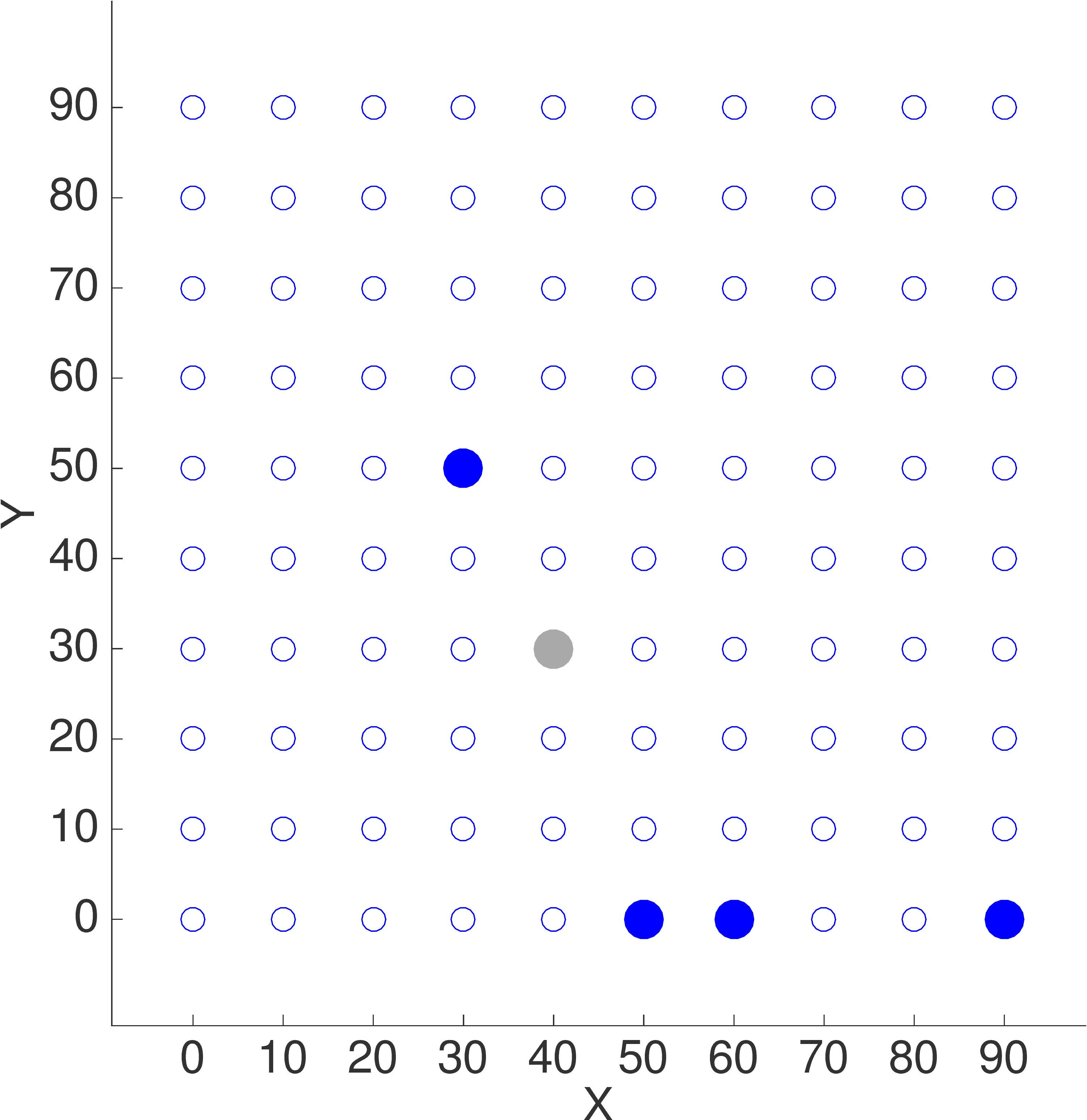}
\hspace*{0.5cm}
\includegraphics[width=3.5cm,angle=0]{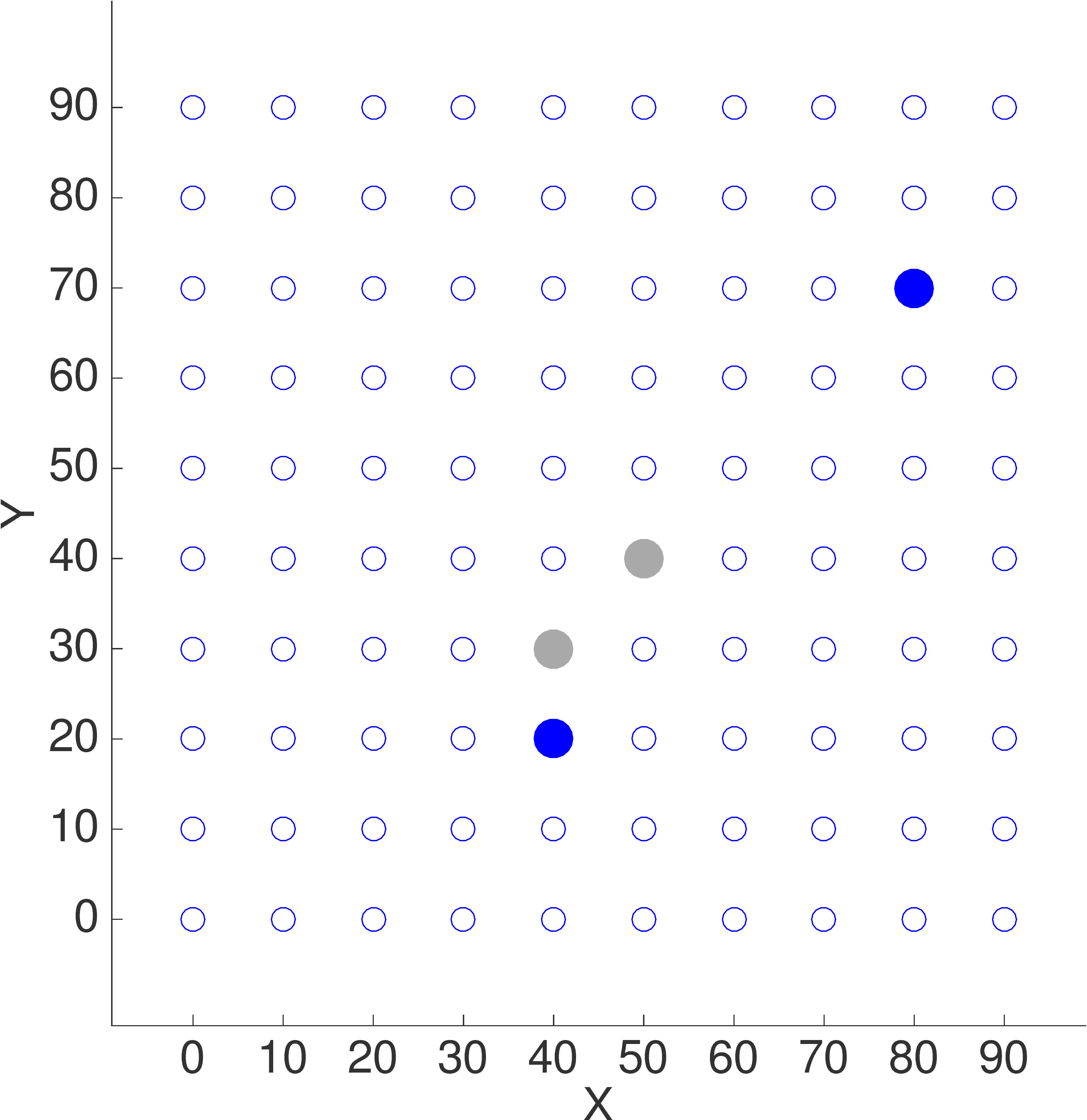}
\hspace*{0.5cm}
\includegraphics[width=3.5cm,angle=0]{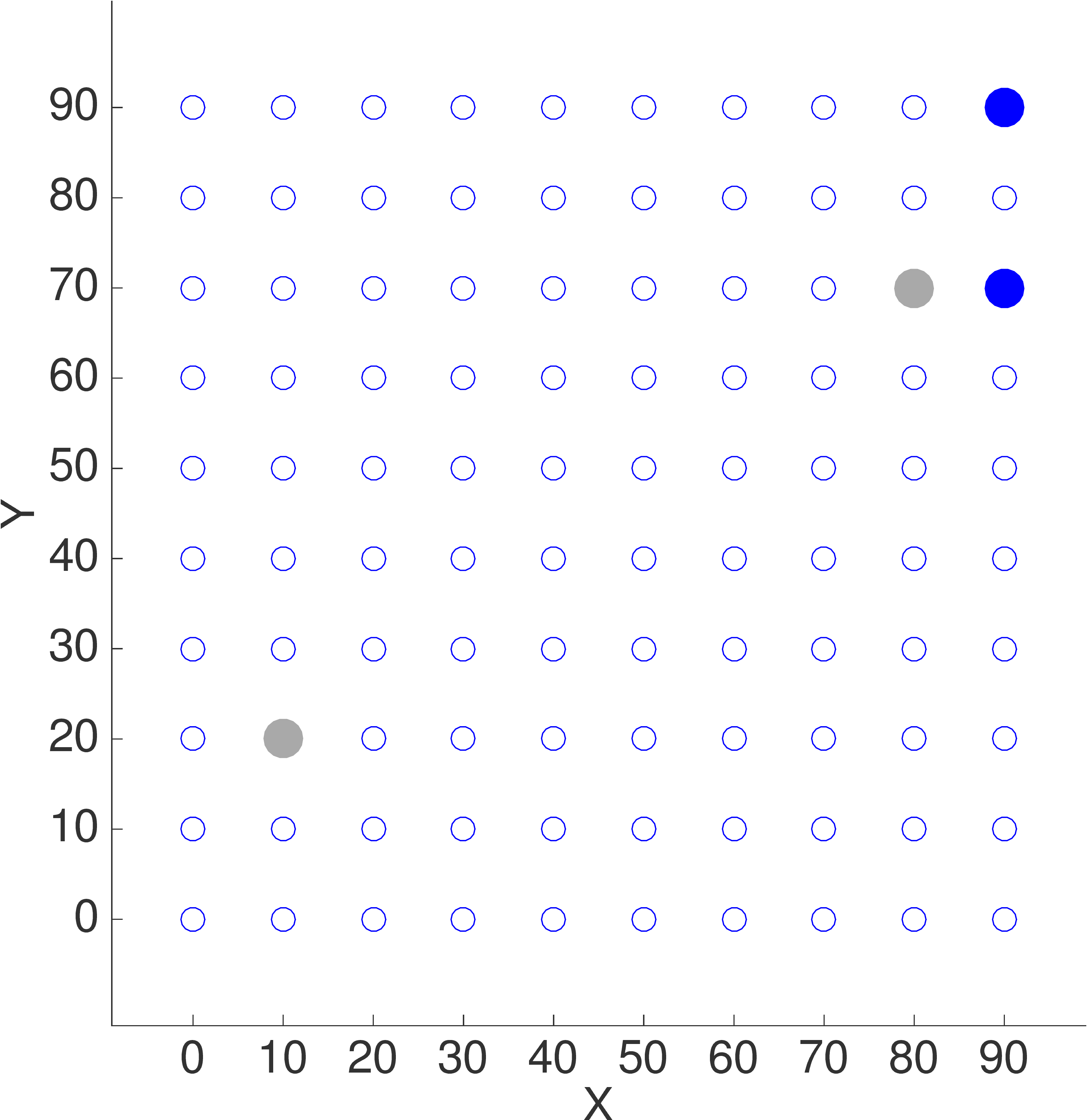}
}
\centerline{$T = \Delta t$  \hspace*{2.5cm} $T = 2 \Delta t$  \hspace*{2.5cm} $T = 3 \Delta t$ \hspace*{2.5cm} $T = 4 \Delta t$}
\vspace*{0.3cm}
\centerline{ 
\includegraphics[width=3.5cm,angle=0]{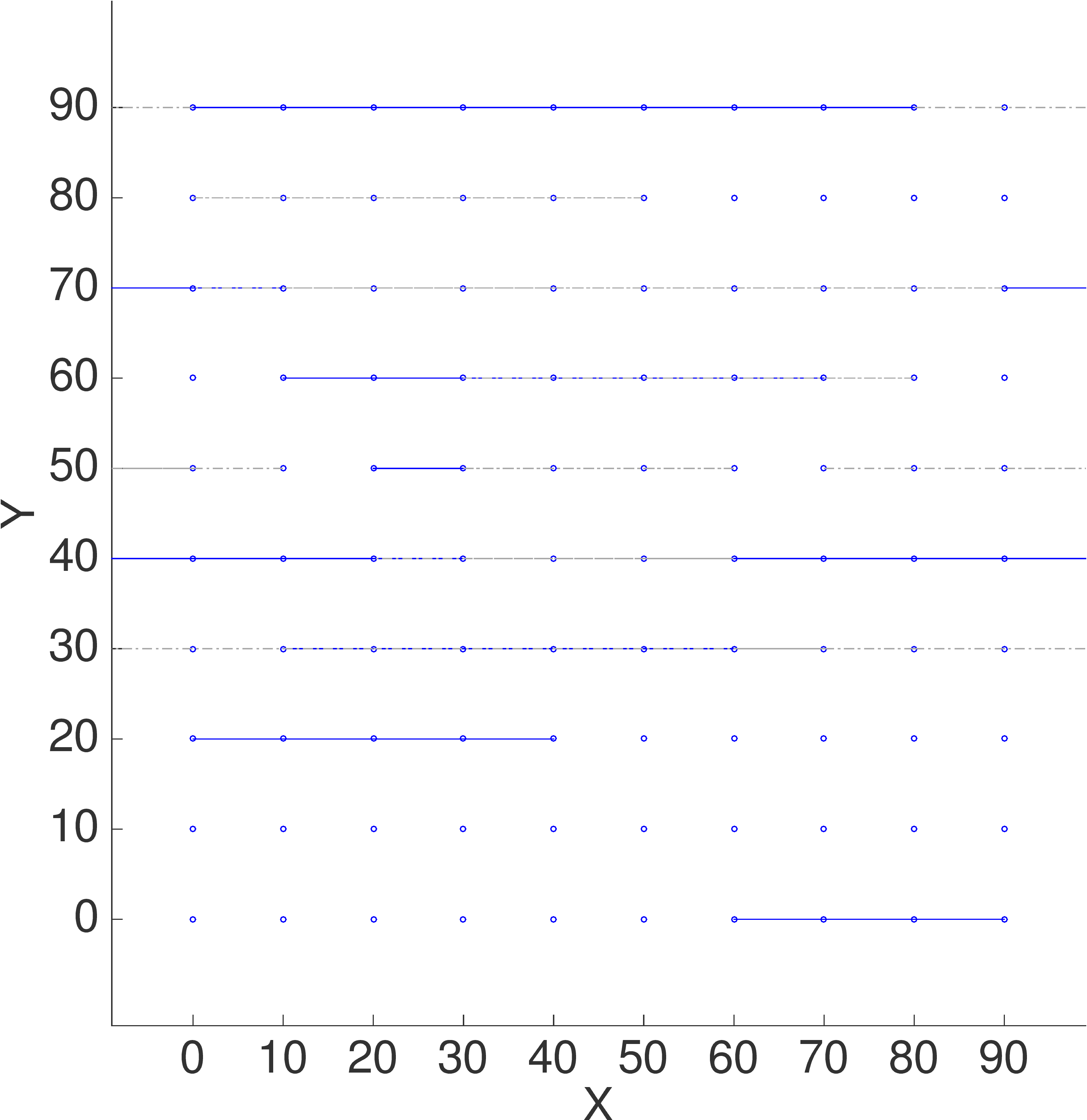}
\hspace*{0.5cm}
\includegraphics[width=3.5cm,angle=0]{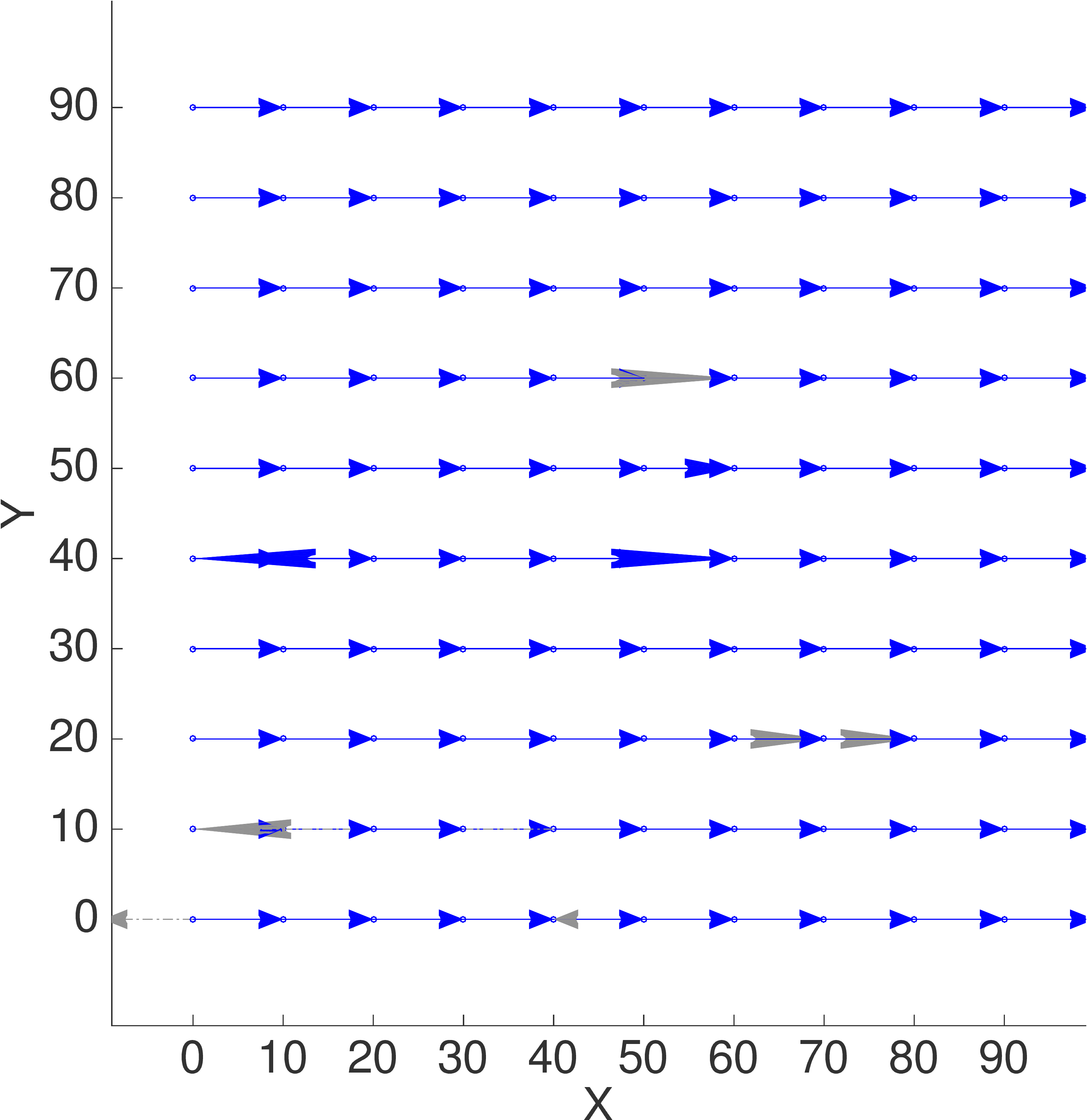}
\hspace*{0.5cm}
\includegraphics[width=3.5cm,angle=0]{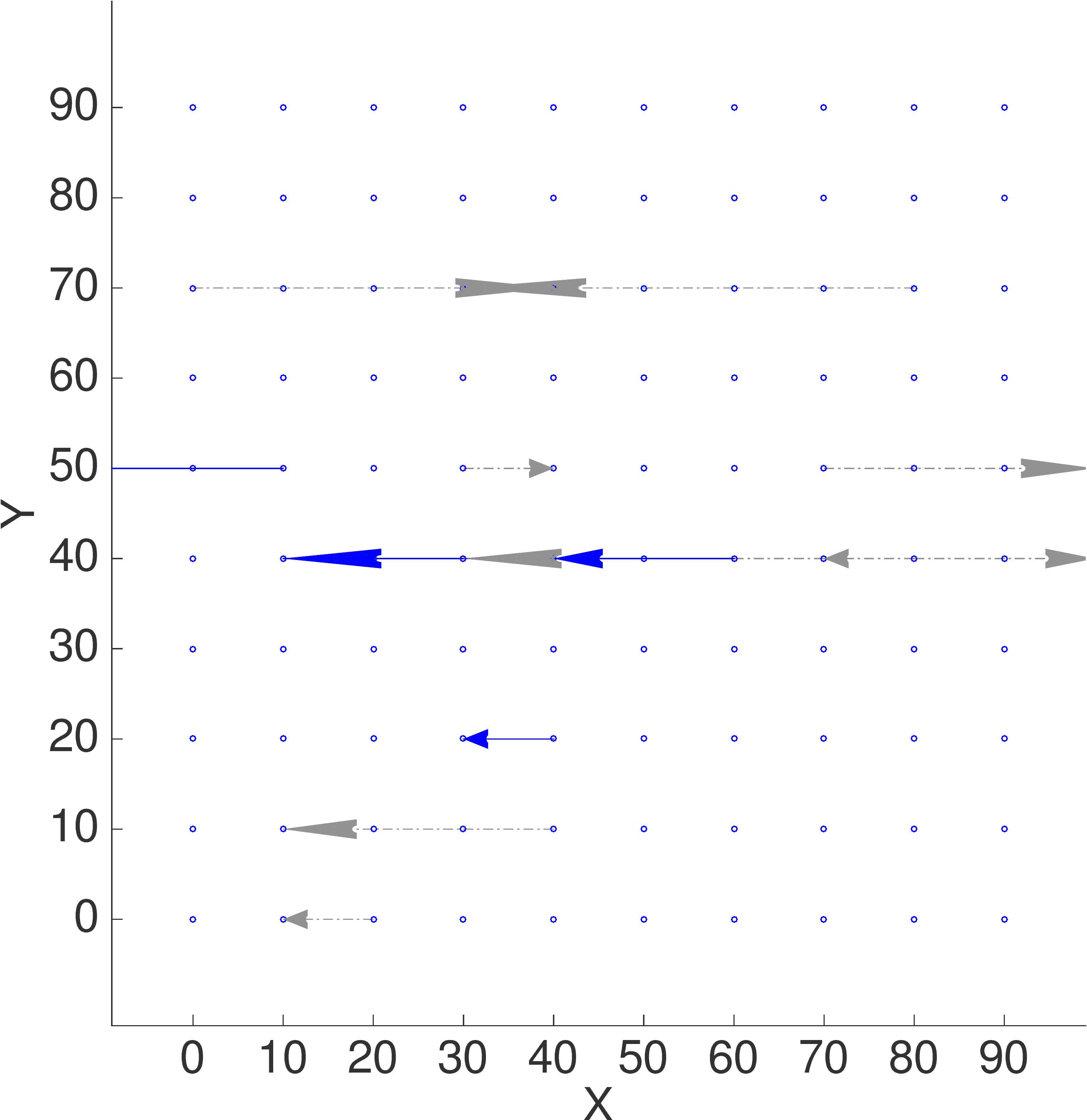}
\hspace*{0.5cm}
\includegraphics[width=3.5cm,angle=0]{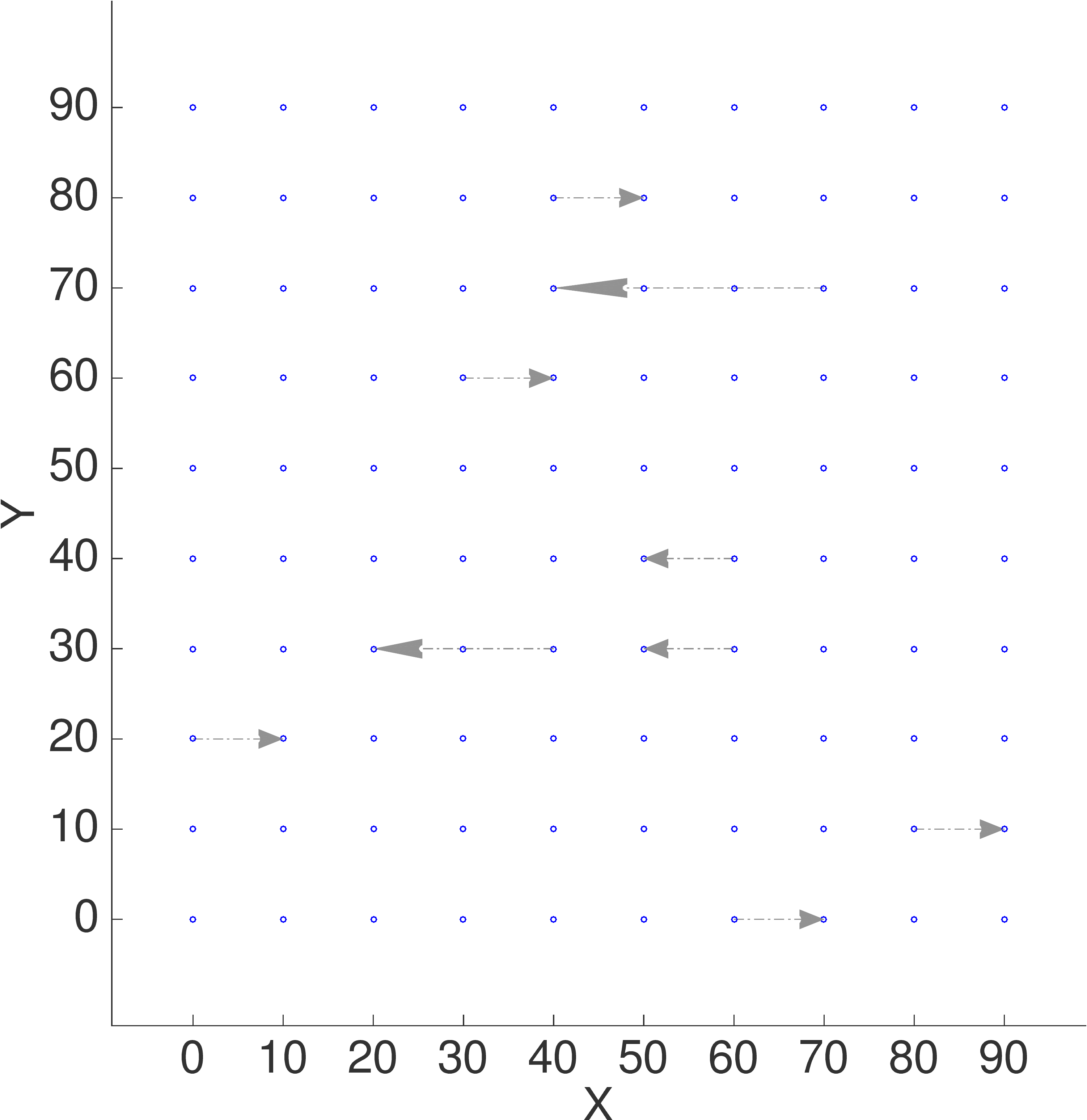}
}
\centerline{$T = 0$  \hspace*{2.5cm} $T = \Delta t$  \hspace*{2.5cm} $T = 2 \Delta t$  \hspace{2.5cm} $T = 3 \Delta t$}
\centerline{(a) Intra (top) and inter (bottom) edges for single-point noise}
\vspace*{0.8cm}
\centerline{ 
\includegraphics[width=3.5cm,angle=0]{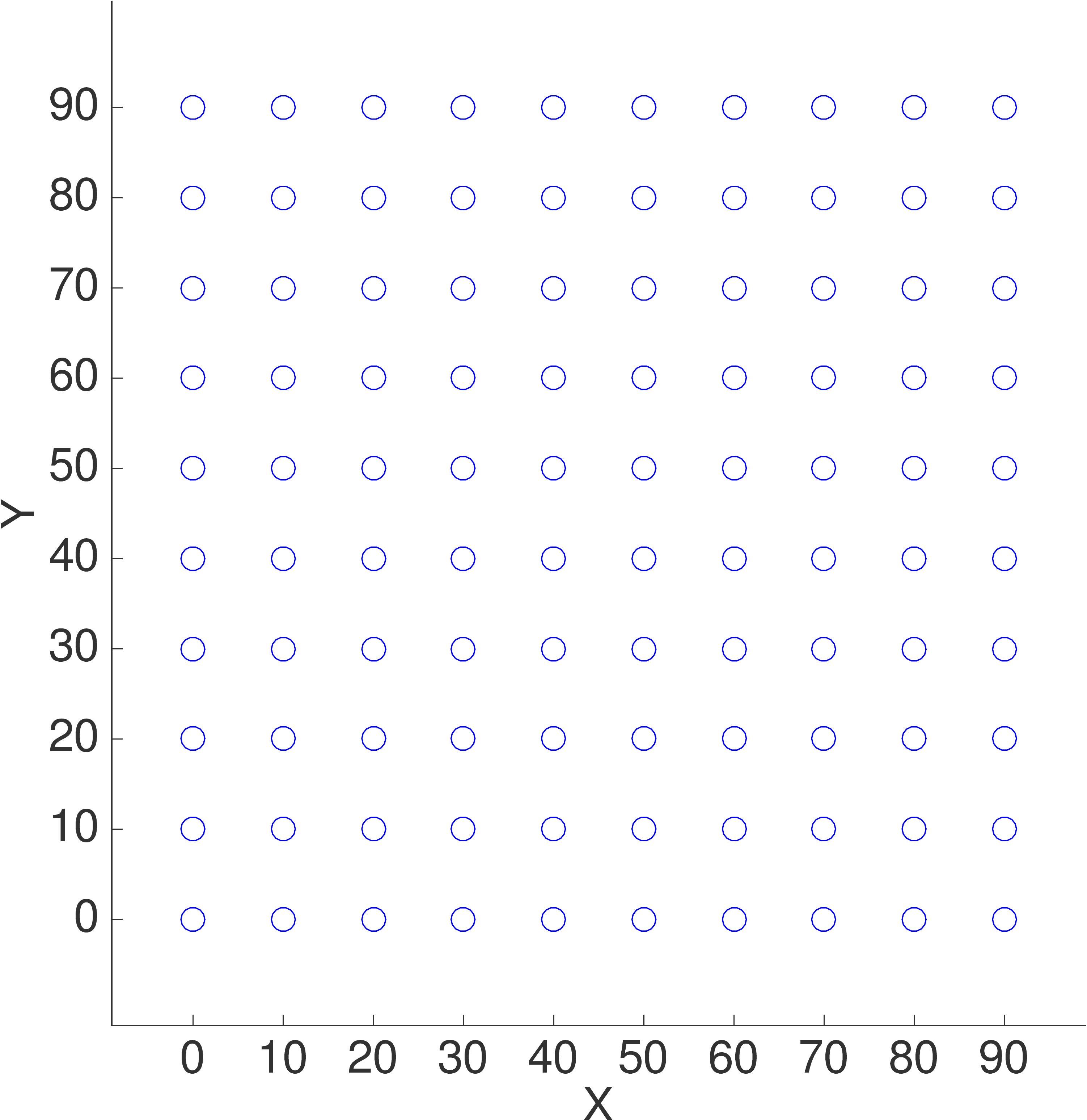}
\hspace*{0.5cm}
\includegraphics[width=3.5cm,angle=0]{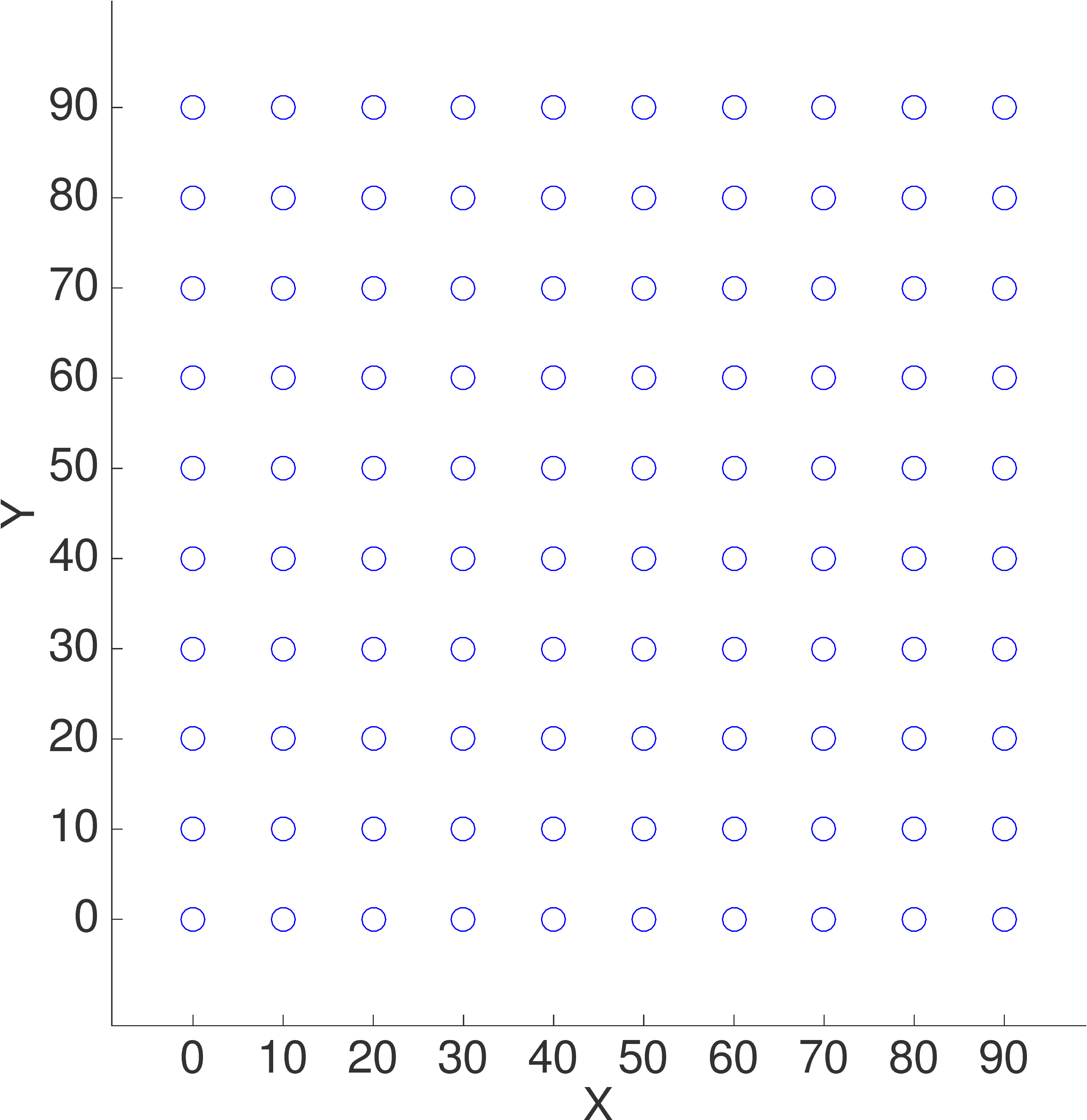}
\hspace*{0.5cm}
\includegraphics[width=3.5cm,angle=0]{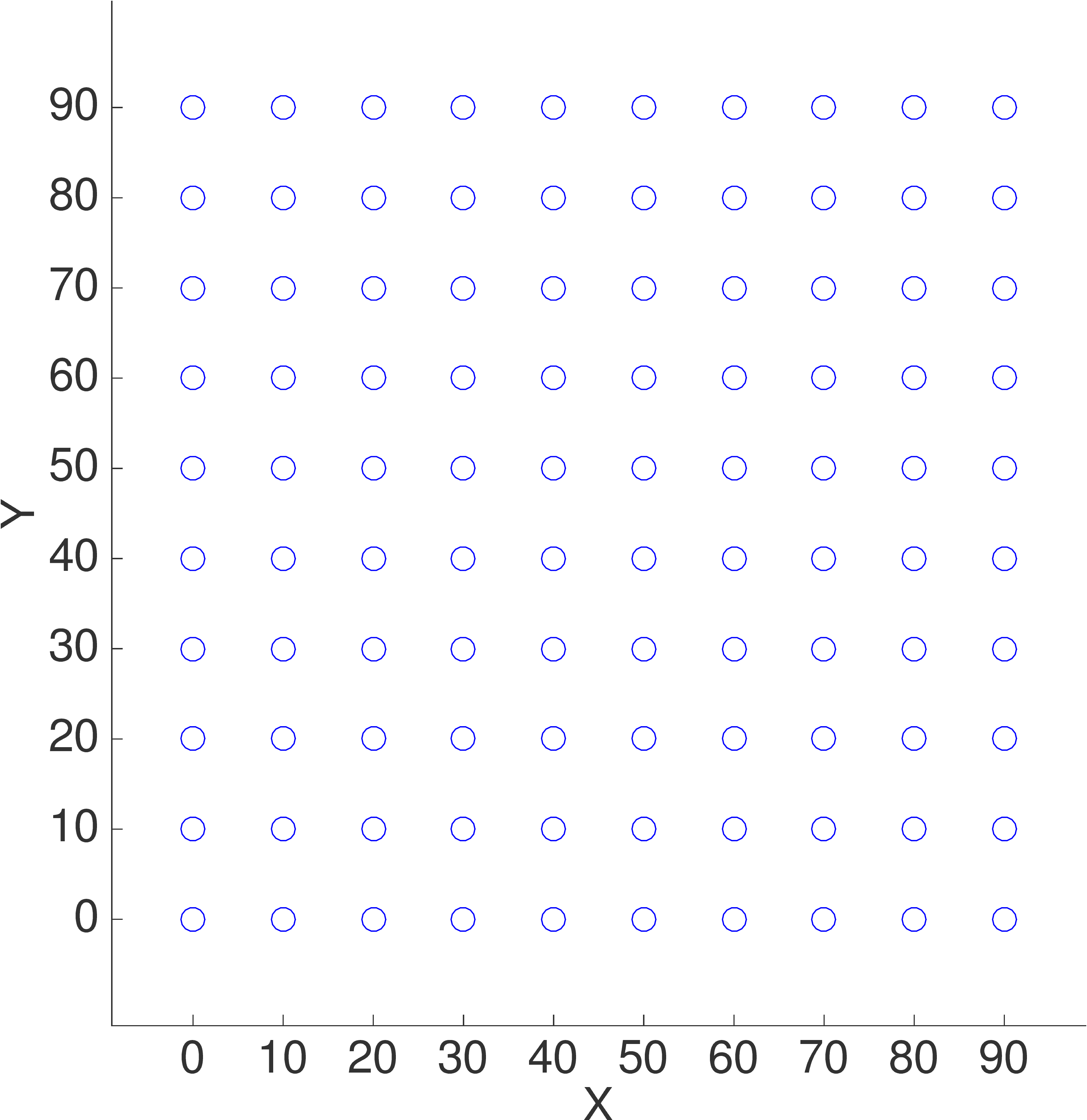}
\hspace*{0.5cm}
\includegraphics[width=3.5cm,angle=0]{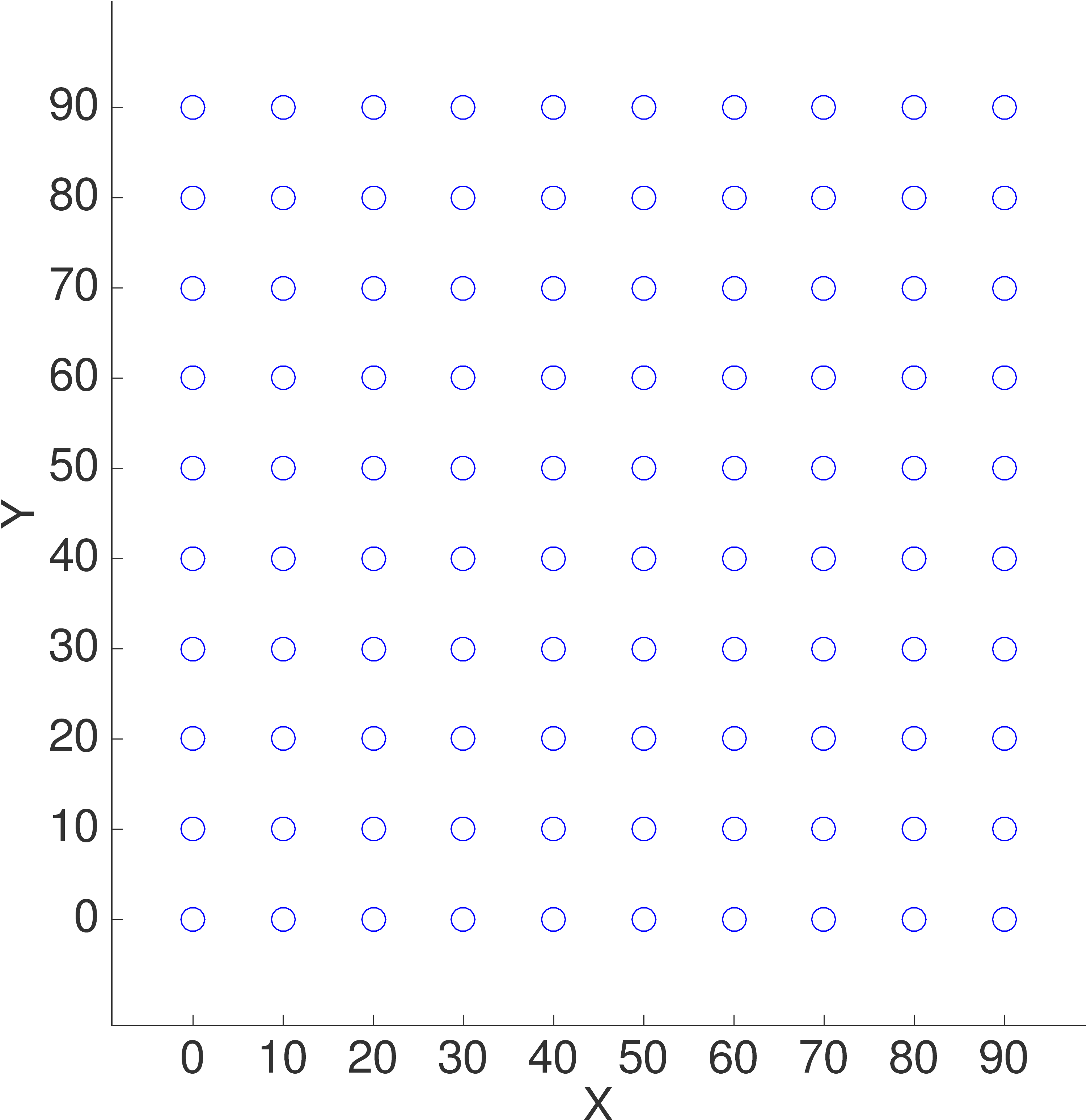}
}
\centerline{$T = \Delta t$  \hspace*{2.5cm} $T = 2 \Delta t$  \hspace*{2.5cm} $T = 3 \Delta t$ \hspace*{2.5cm} $T = 4 \Delta t$}
\vspace*{0.3cm}
\centerline{ 
\includegraphics[width=3.5cm,angle=0]{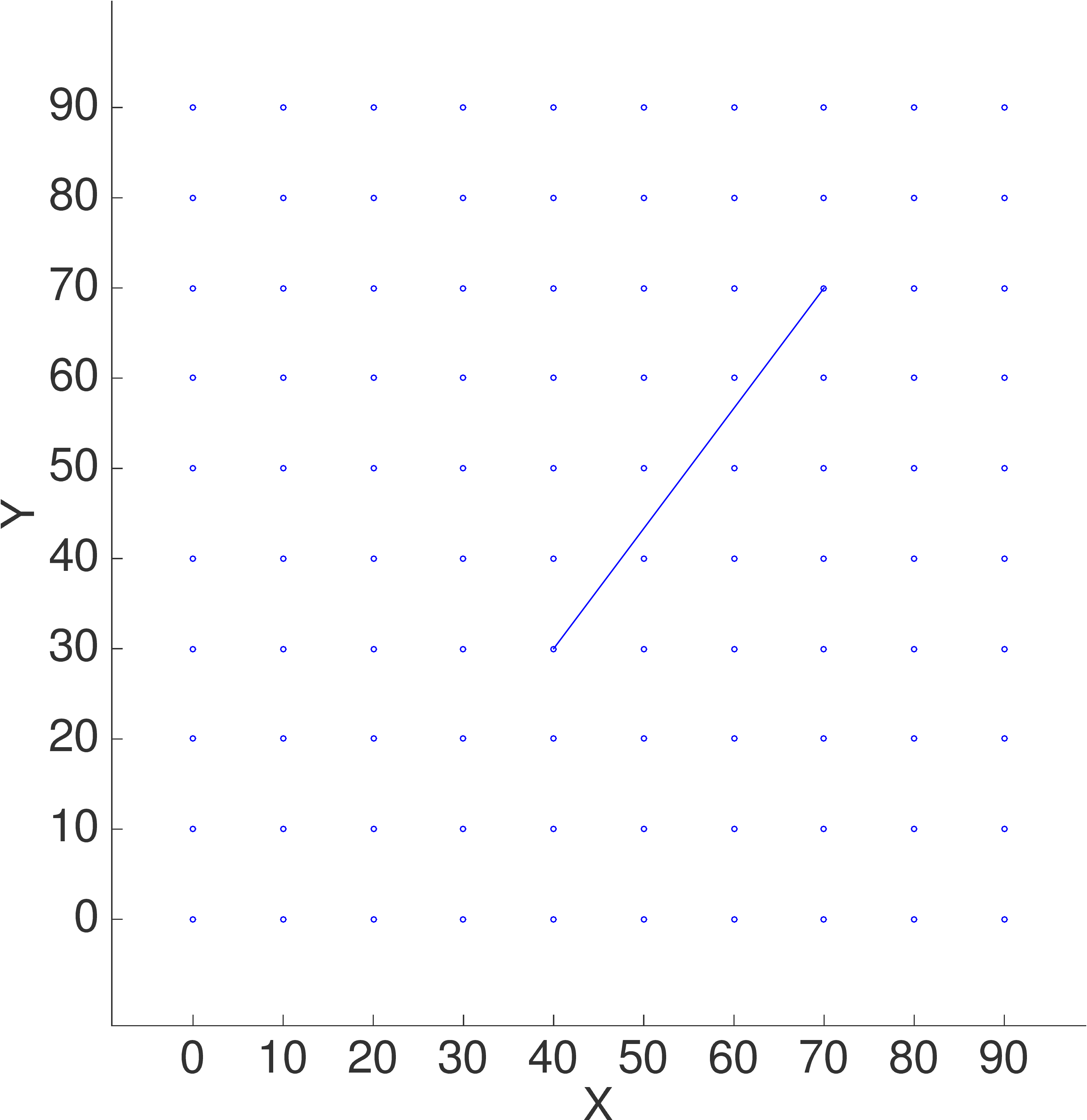}
\hspace*{0.5cm}
\includegraphics[width=3.5cm,angle=0]{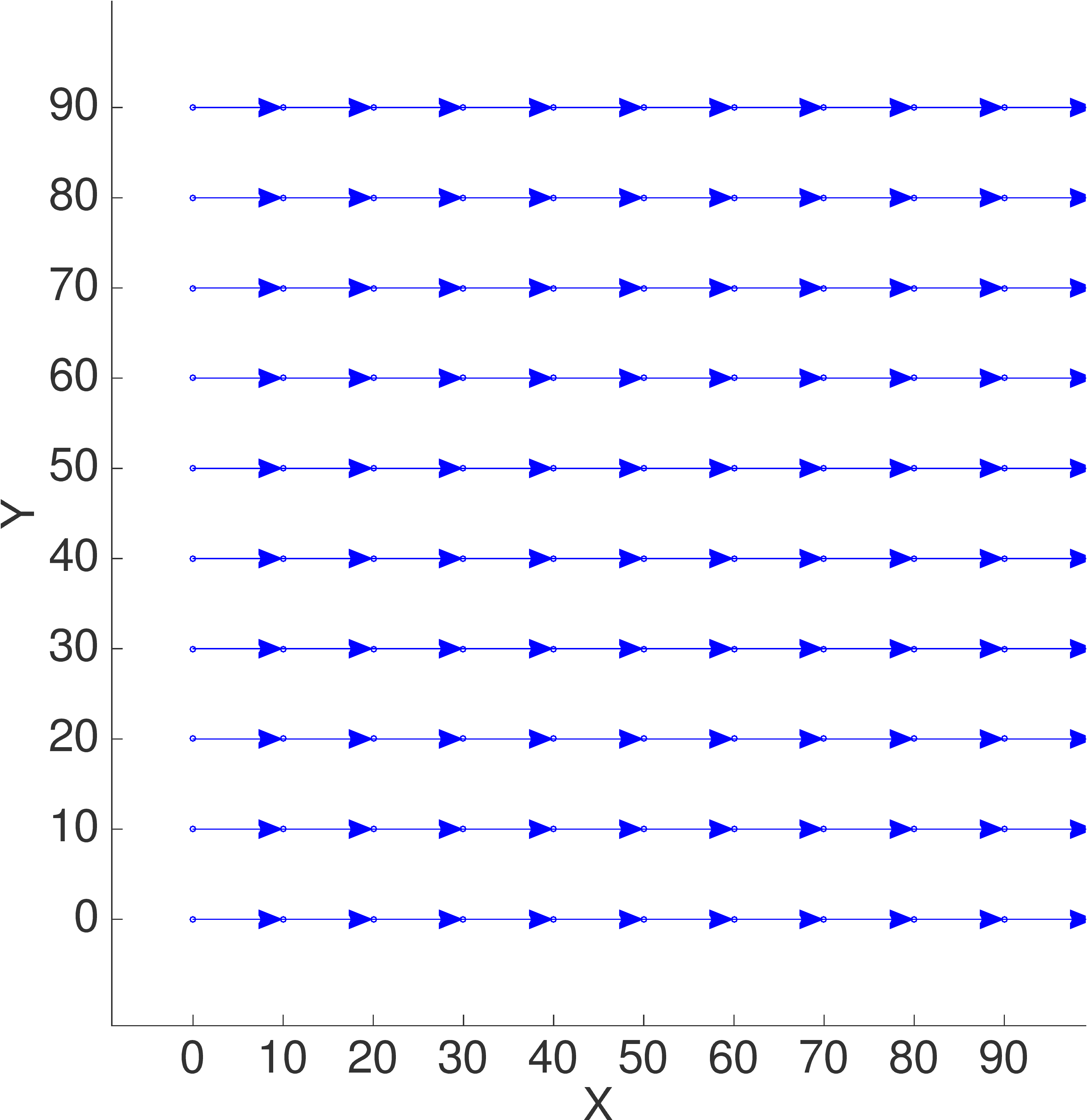}
\hspace*{0.5cm}
\includegraphics[width=3.5cm,angle=0]{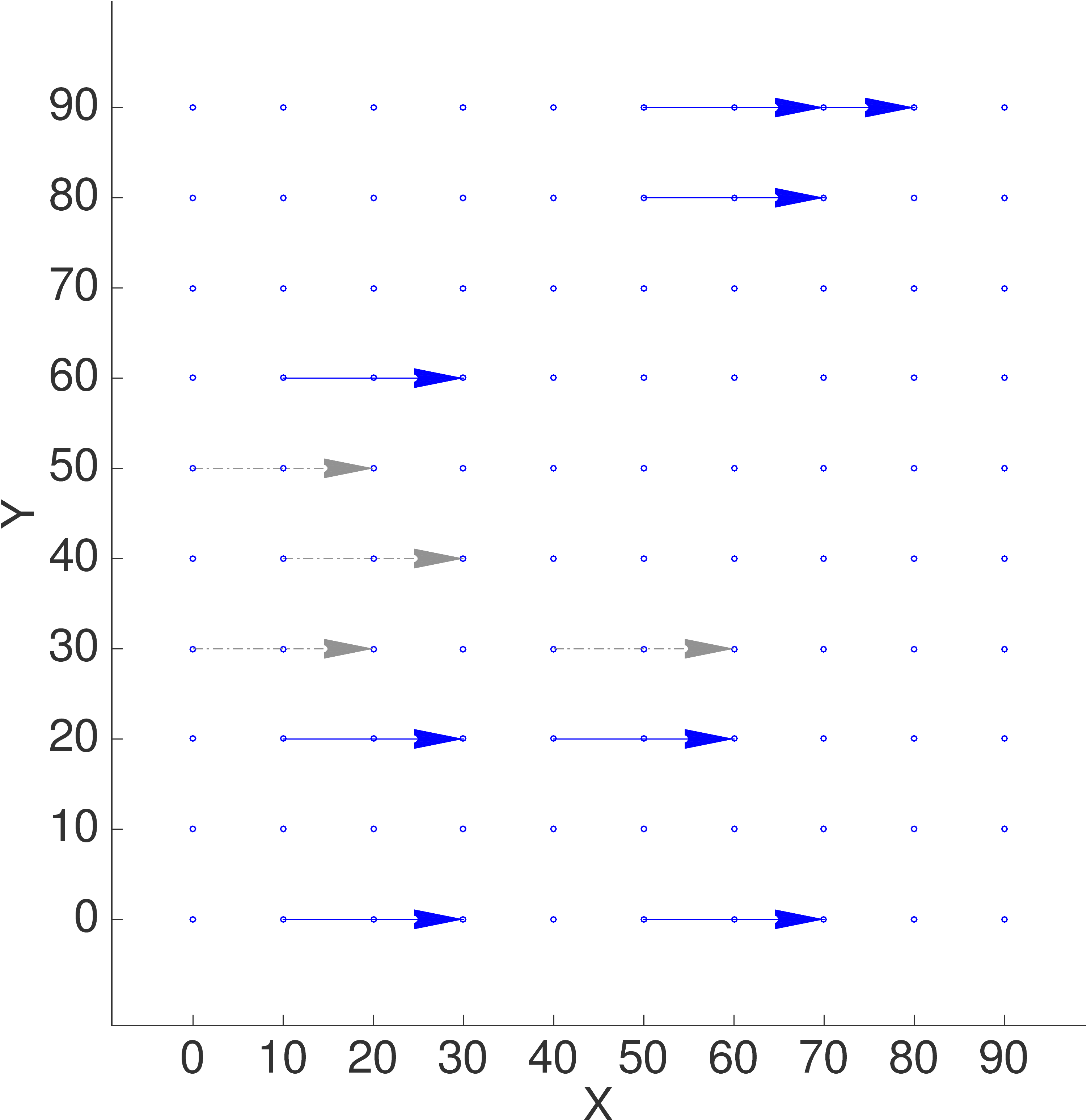}
\hspace*{0.5cm}
\includegraphics[width=3.5cm,angle=0]{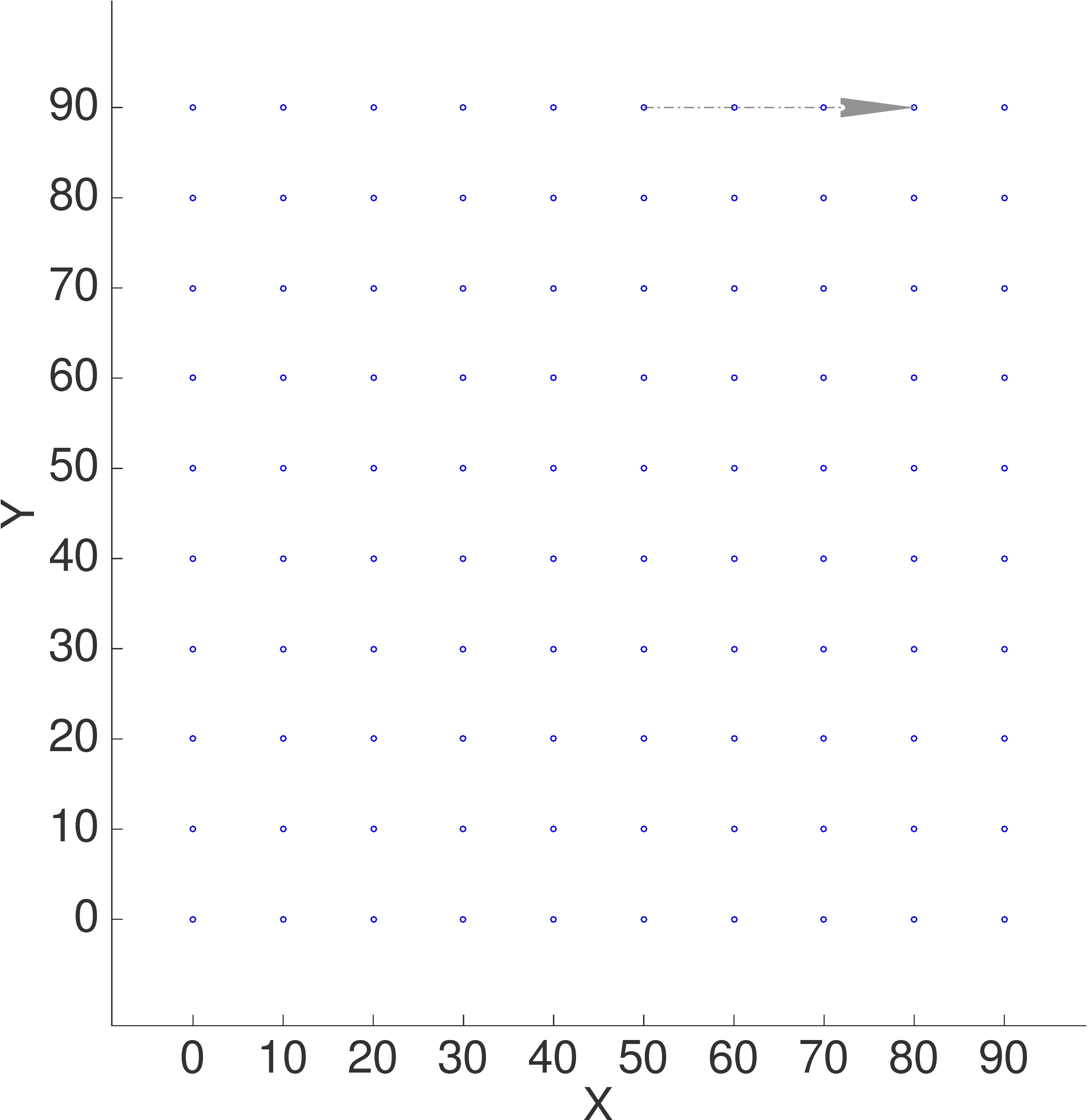}
}
\centerline{$T = 0$  \hspace*{2.5cm} $T = \Delta t$  \hspace*{2.5cm} $T = 2 \Delta t$  \hspace{2.5cm} $T = 3 \Delta t$}
\centerline{(b) Intra (top) and inter (bottom) edges for all-point noise}
\caption{Inter and intra edges for pure advection experiment, same parameters as in Fig.\ \ref{pure_advection_fig_1}(a), but using single-point or all-point noise instead of single-point peaks.
\label{pure_advection_noise_fig}}
\end{figure*}

\begin{figure*}
\centerline{ 
\includegraphics[width=5.0cm,angle=0]{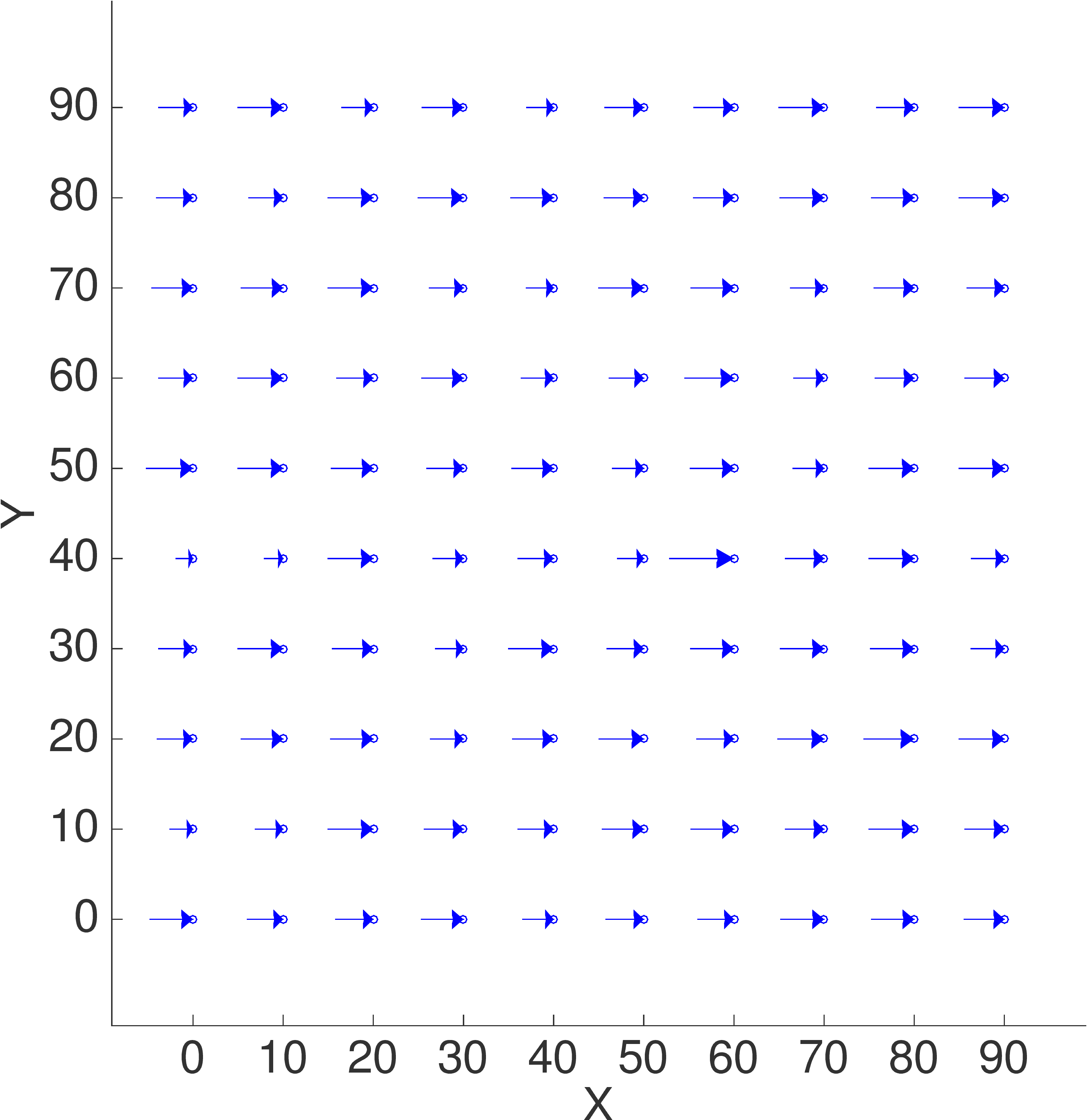}
\hspace*{2.0cm}
\includegraphics[width=5.0cm,angle=0]{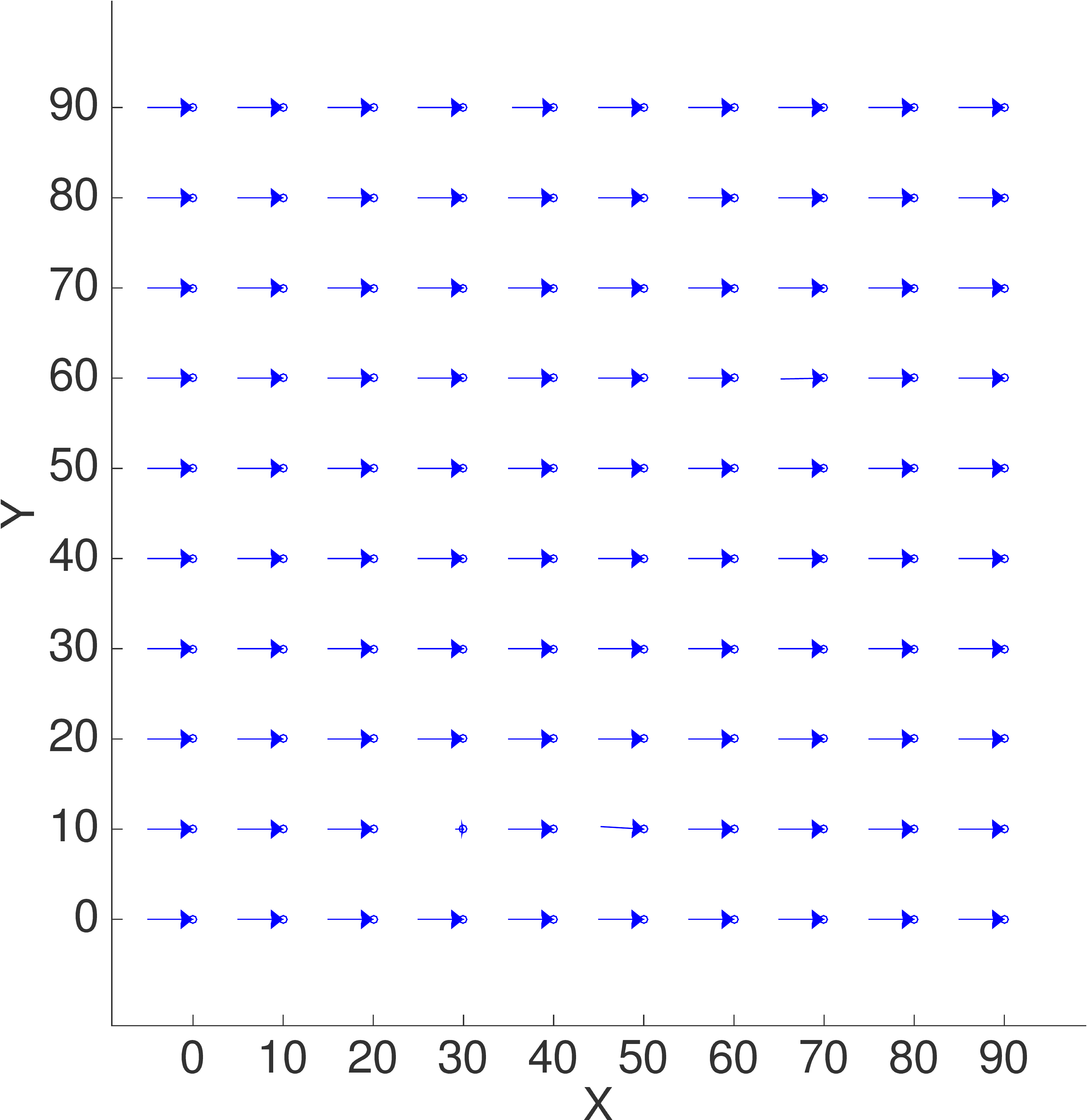}
}
\centerline{(a) Velocity estimate for single-point noise \hspace*{0.5cm} (b) Velocity estimate for all-point noise}
\caption{Type 2 velocity estimate for pure advection with continuous noise instead of 
peak initial conditions.  Type 1 (not shown) and Type 2 are practically 
indistinguishable, since only very few intra connections exist.   
\label{pure_advection_noise_vel_fig}}
\end{figure*}

Now we focus on tracking information flow in the pure advection scenario, i.e.\ 
we use $\kappa_x=\kappa_y=0$ and $C=1.0$.
However, those parameters yield signals that never decay, and generate data 
that mimics purely deterministic relationships, which the causal discovery algorithm
cannot handle.  Since this is not a realistic scenario anyway, and we use it only 
for theoretical analysis, we did not change the algorithm and instead 
add a tiny amount of prior noise to the signal at each point at every time step
(normally-distributed noise of small amplitude ($\sigma=0.1$), while the IC peaks have a large magnitude of $500$ units).
Typical results using Message Type 1 are shown in Fig.\ \ref{pure_advection_fig_1},
while results for Message Types 2 \& 3 are shown in 
Figures \ref{pure_advection_noise_fig} and \ref{pure_advection_noise_vel_fig}.

{\bf When using Message Type 1 (IC peak)}, we found that\\
(1) There are no intra edges.\\
(2) There are no concurrent inter edges.  This even holds for $M\ge 10$ (not shown here).\\
(3) The great majority of inter edges occur for $T=\Delta t$.  
The length of connection is one grid point per time step, i.e.\ 1 grid point for $T=\Delta t$, 2 grid points for $T=2 \Delta t$, etc.  This even holds for $T \ge 10 \Delta t$ (results not shown here).  This length is expected, since we use $C=1$, i.e.\ signals 
travel exactly one grid point per time step in this scenario.
Note that for $T= 6 \Delta t$ in Fig.\ \ref{pure_advection_fig_1}(b),
the length of the connections is 6 grid points, but 
since each row only has 10 grid points, it {\it appears} 
as if locations connect 4 points to the right, rather than 6 points to the right.
Also note, that the asymmetric inter edges that occur for $T> \Delta t$, 
are all due to the fact that even for Message Type 1 we use a small amount of noise 
(see previous paragraph).
Otherwise we would not see these longer edges, which describe an echo effect, since 
they basically duplicate the direction and speed of the edges found for $T=\Delta t$. 
\\
(4) Type 1 and Type 2 velocity estimates are identical, since there are no intra edges.  They are very good for M=1, 2, with a few incorrect magnitudes for M=4.

{\bf When using Message Types 2 \& 3 (prior noise)}, we found that\\
(1) The overall signature with prior noise is quite similar to the ones with the strong initial conditions, especially for all-point noise.\\
(2) Results for all-point noise are considerably better than for single-point noise.  The reason is probably that with all-point noise the algorithm has more non-zero signals to work with at any time.
\\
(3) No intra edges appear for all-point, a few for single-point noise.  Likewise single-point noise results have more stray inter edges, while all-point noise has few.


{\bf Summary:} In the pure advection experiments there are very few intra connections and the 
nonconcurring inter edges are dominant, as expected.  Velocity plots show decent estimates
of the original advection field for all message types used.


\subsection{Results for simple scenario combining advection and diffusion}
\label{adv_plus_diff_sec}

Now we briefly consider what happens if both advection and diffusion are present.
Since diffusion is present, there is no need to add noise when using single peak initial 
conditions. However, we encountered issues with numerical stability with this combination 
and thus chose a lower value, $C=0.7$, to make sure the numerical simulations are stable.  
This also adds additional diffusion in the simulation, thus enhancing the effect of diffusion.
Furthermore, it reduces the signal speed to 0.7 grid points per time step.

\begin{figure*}
\centerline{ 
\includegraphics[width=3.5cm,angle=0]{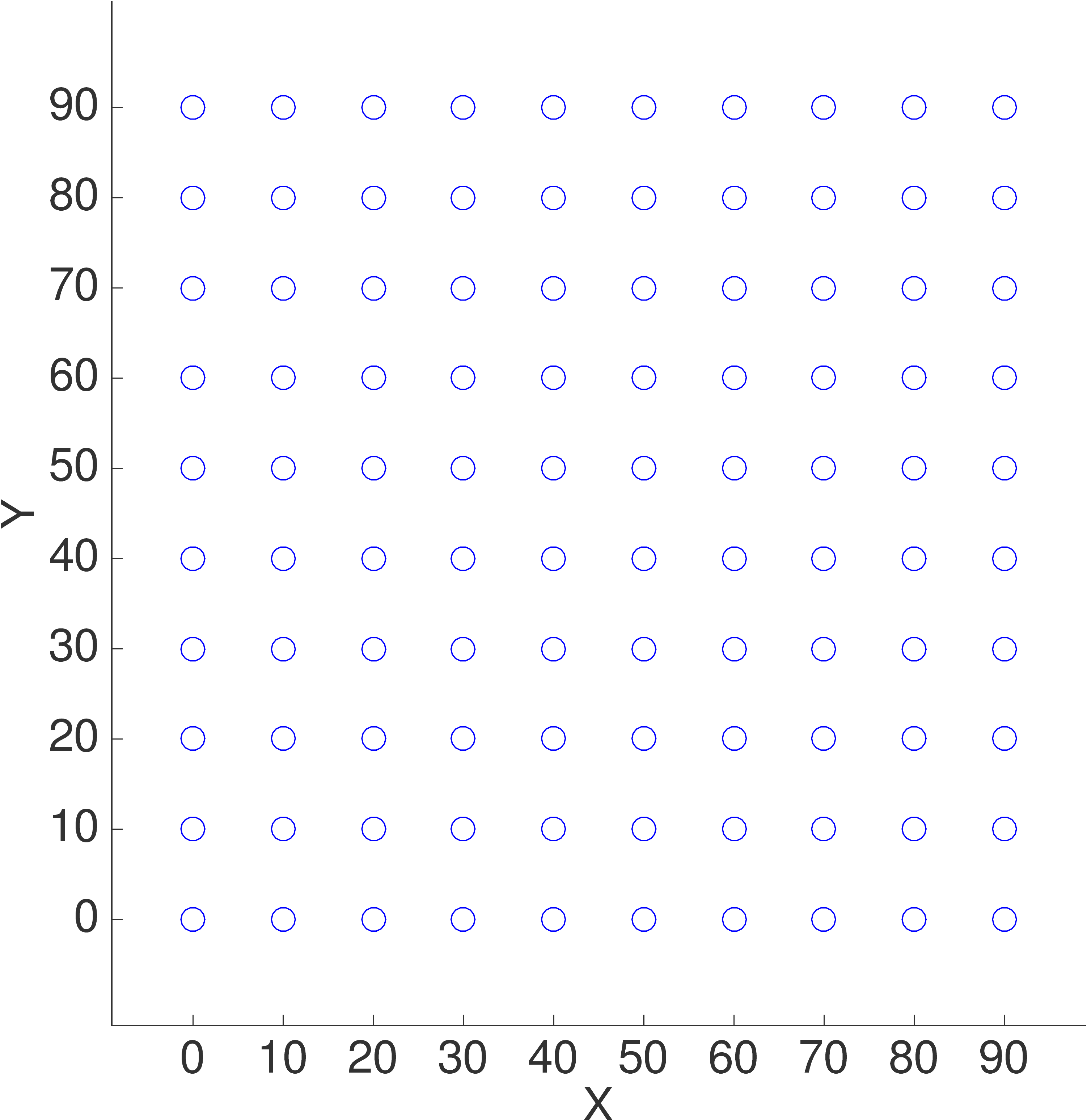}
\hspace*{0.5cm}
\includegraphics[width=3.5cm,angle=0]{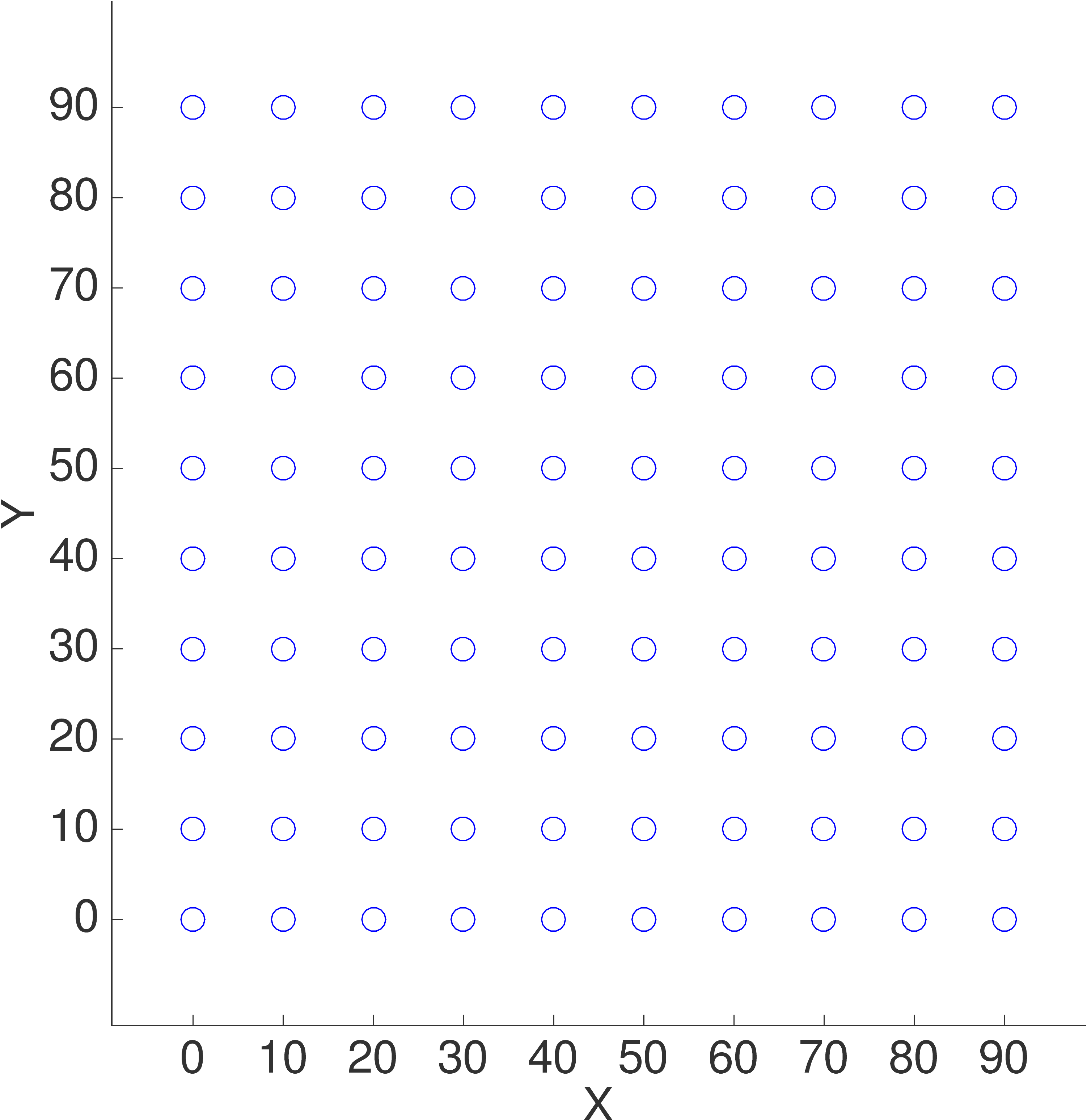}
\hspace*{0.5cm}
\includegraphics[width=3.5cm,angle=0]{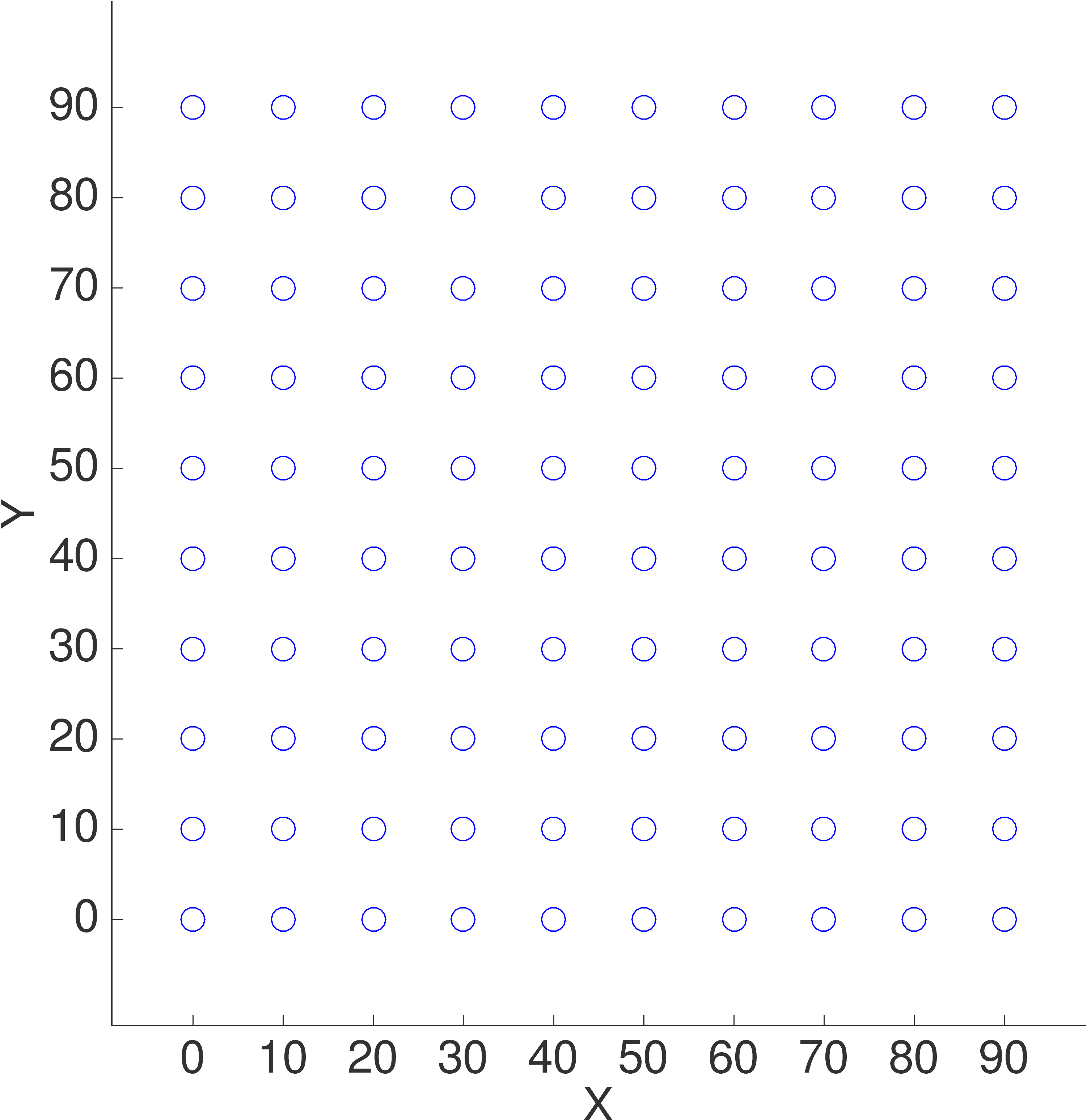}
\hspace*{0.5cm}
\includegraphics[width=3.5cm,angle=0]{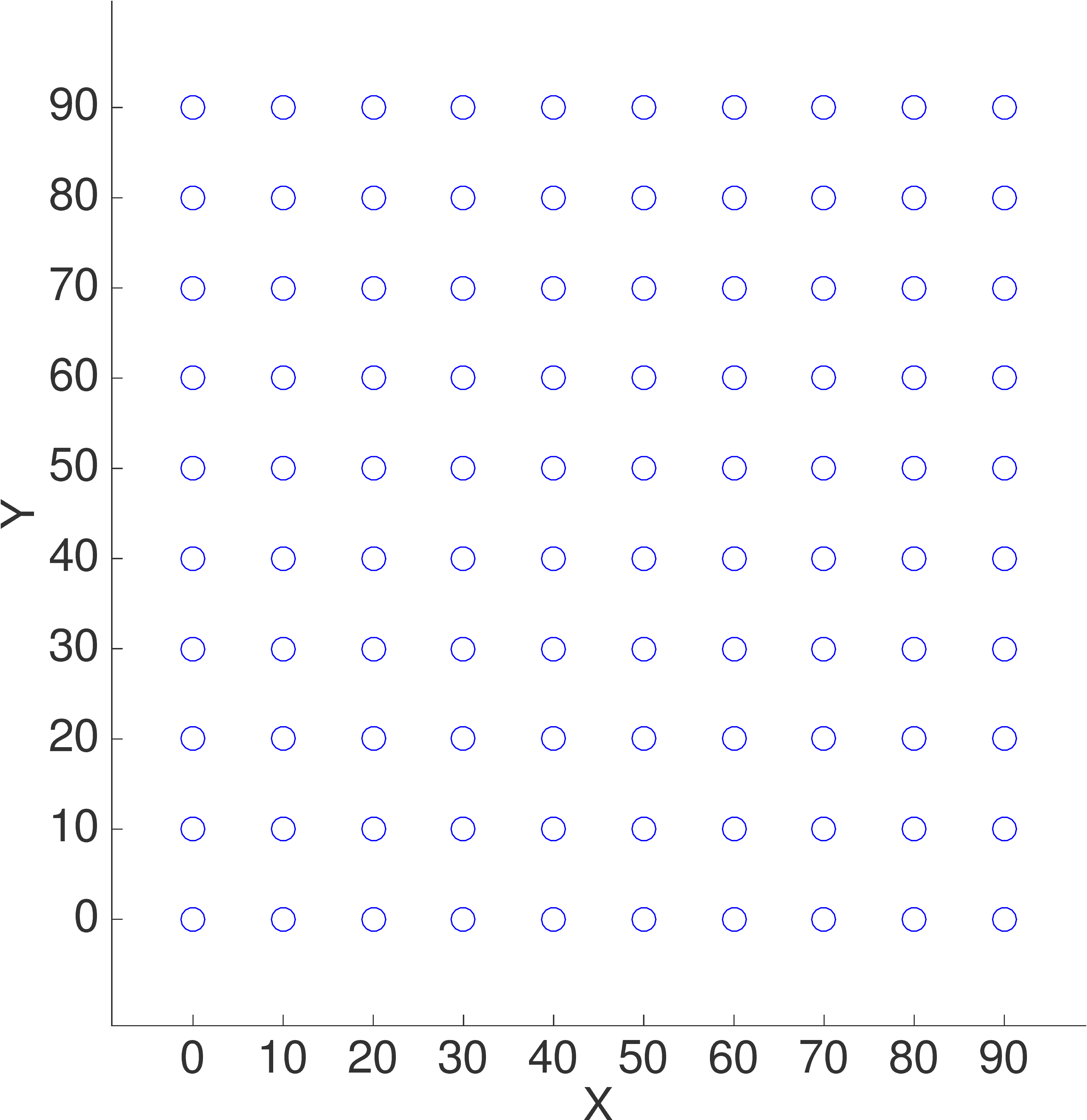}
}
\centerline{$T = \Delta t$  \hspace*{2.5cm} $T = 2 \Delta t$  \hspace*{2.5cm} $T = 3 \Delta t$ \hspace*{2.5cm} $T = 4 \Delta t$ }
\centerline{(a) Intra edges}
\vspace*{0.3cm}
\centerline{ 
\includegraphics[width=3.5cm,angle=0]{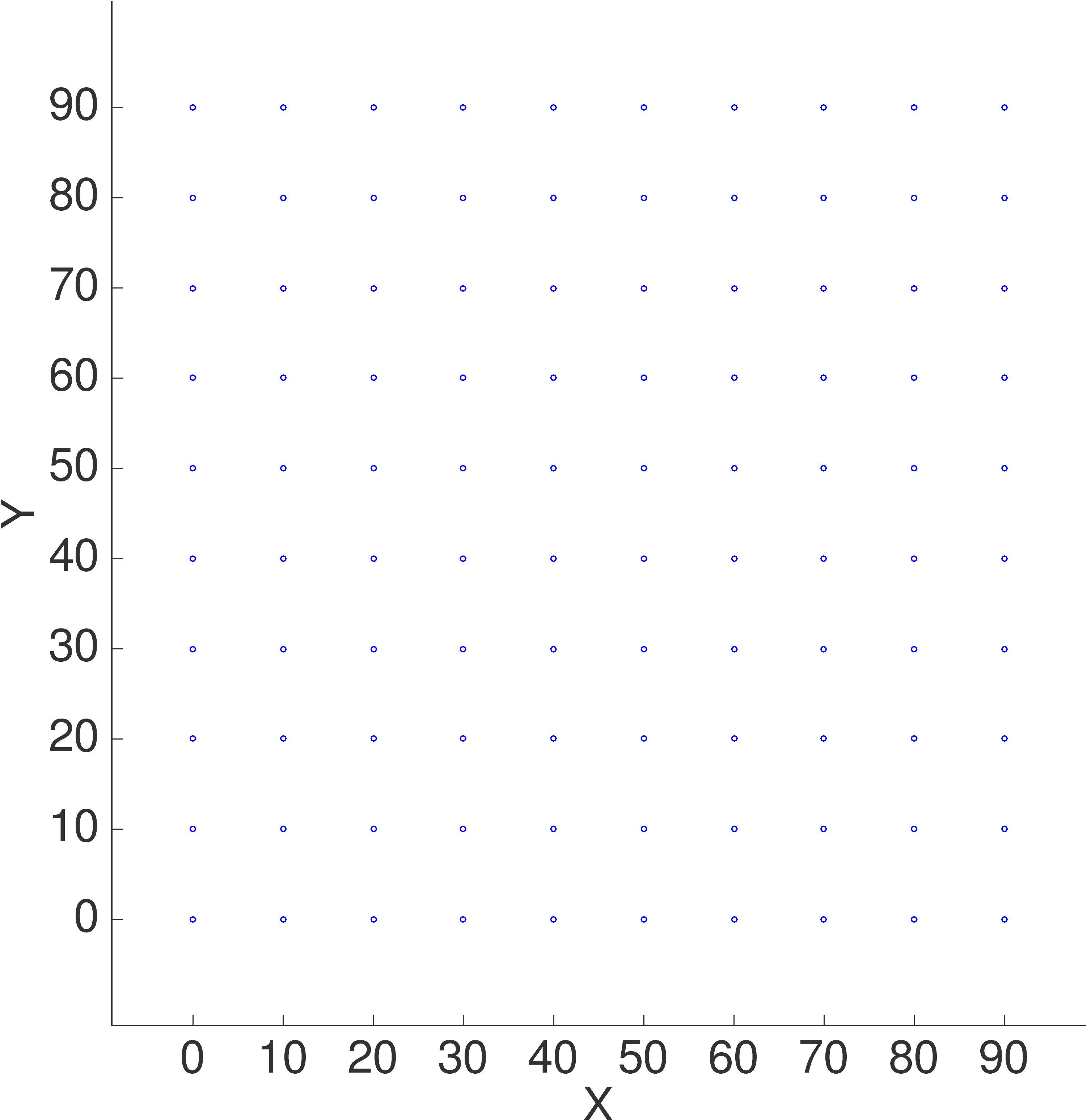}
\hspace*{0.5cm}
\includegraphics[width=3.5cm,angle=0]{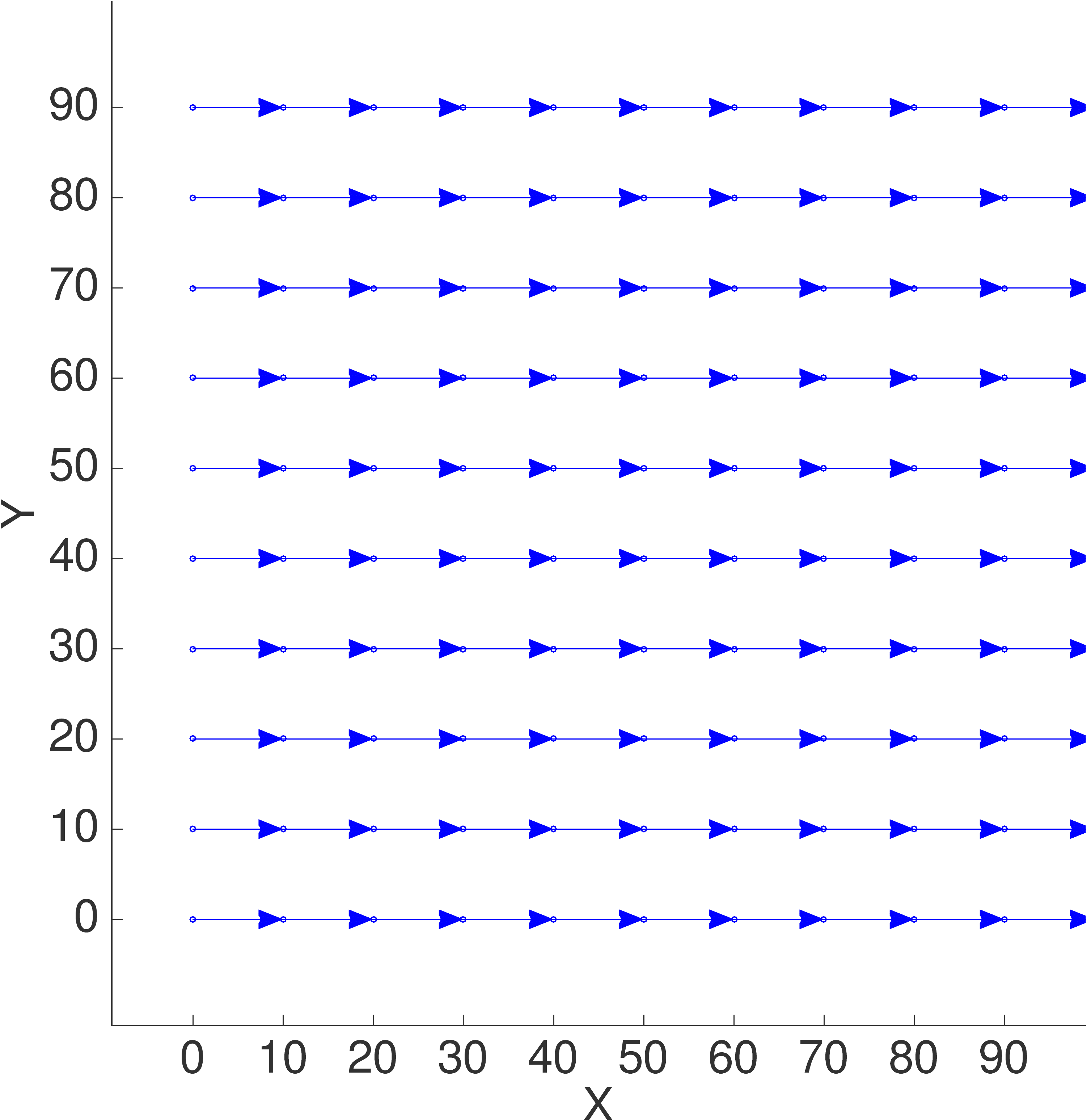}
\hspace*{0.5cm}
\includegraphics[width=3.5cm,angle=0]{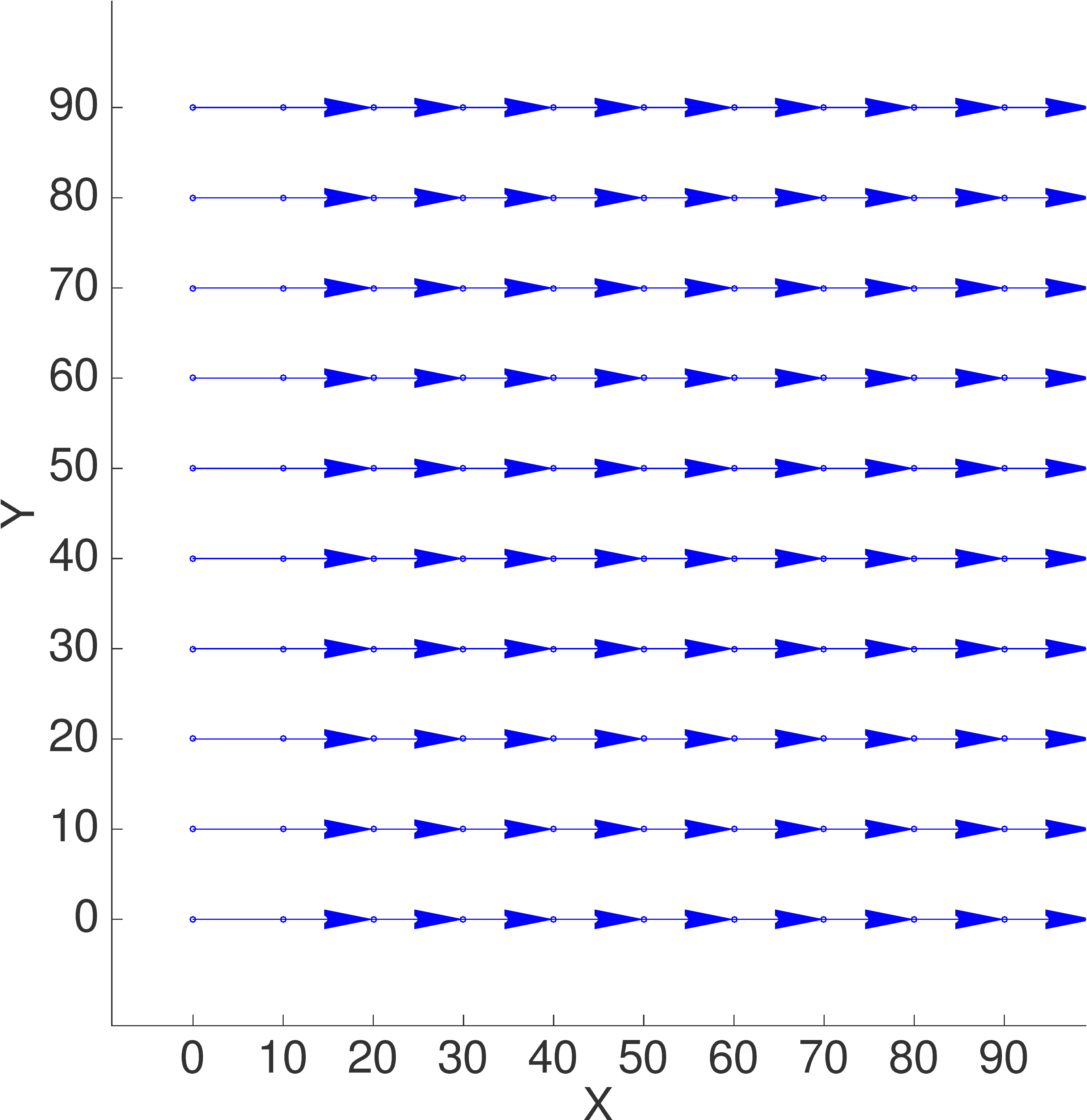}
\hspace*{0.5cm}
\includegraphics[width=3.5cm,angle=0]{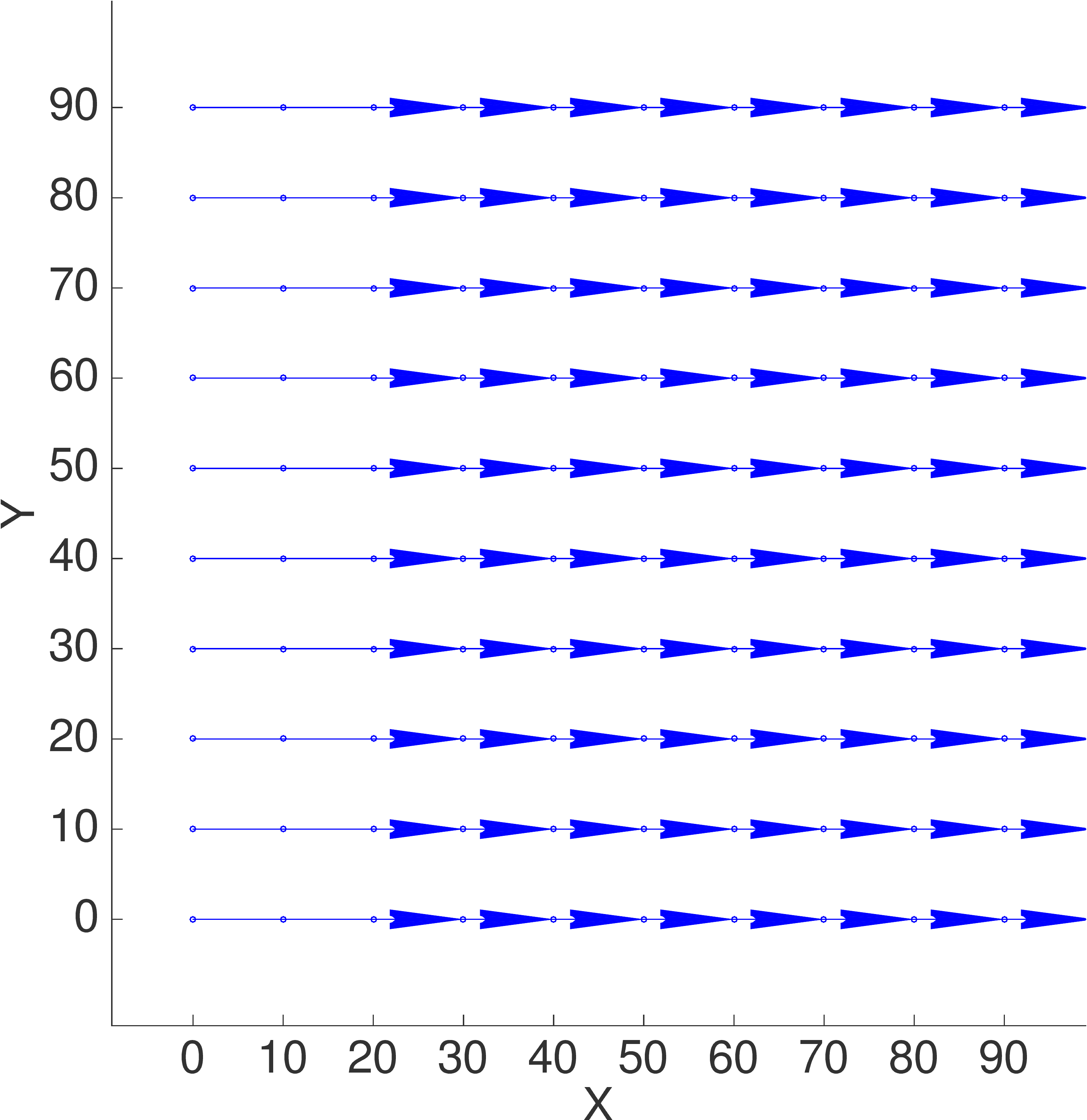}
}
\centerline{$T = 0$  \hspace*{2.5cm} $T = \Delta t$  \hspace*{2.5cm} $T = 2 \Delta t$ \hspace*{2.5cm} $T = 3 \Delta t$}
\centerline{(b) Inter edges}
\vspace*{0.3cm}
\centerline{ 
\includegraphics[width=5.0cm,angle=0,clip]{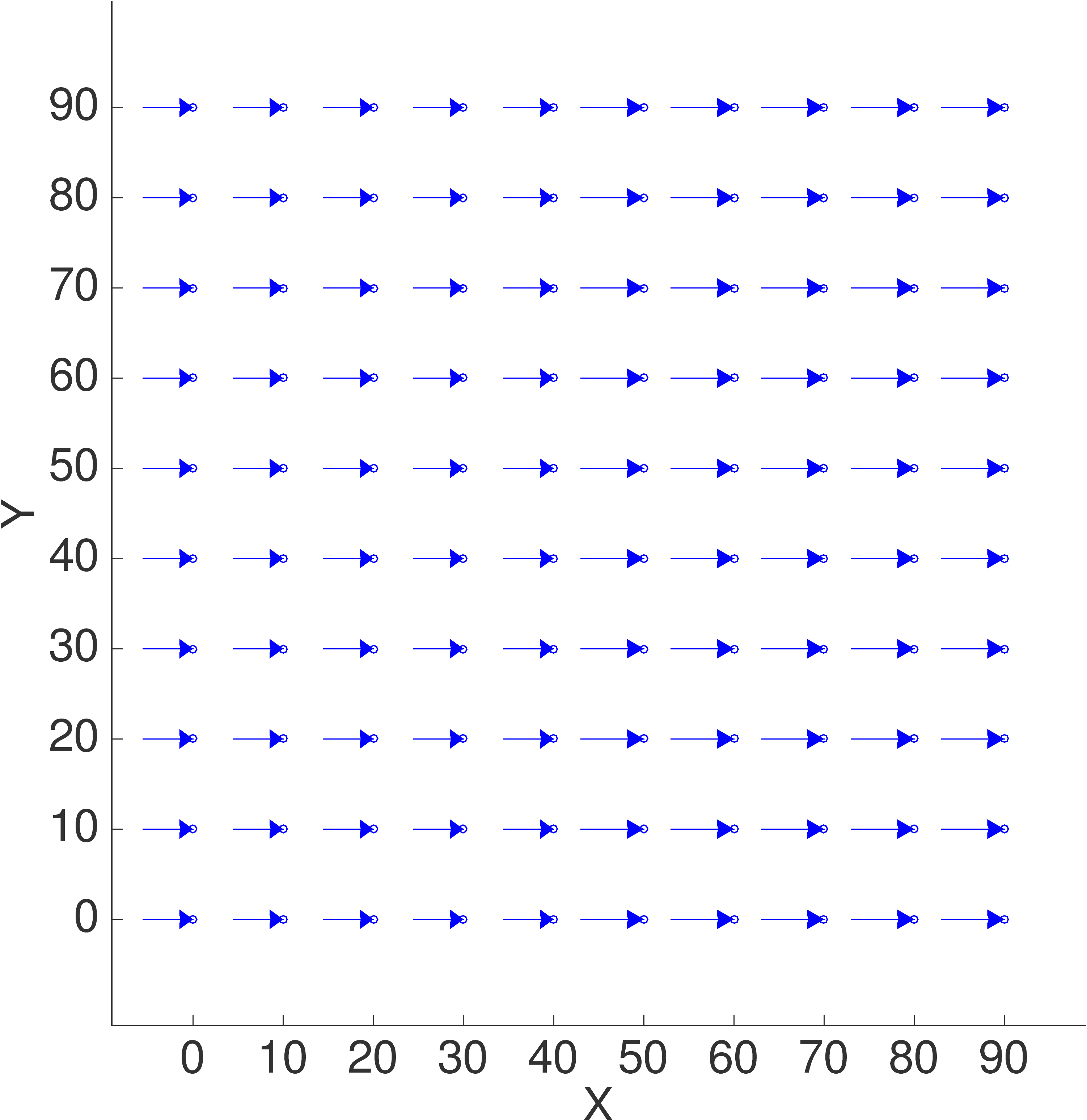}
}
\centerline{(c) Type 1 Velocity Estimate (Type 2 looks identical)}
\caption{Results for simple scenario with advection and diffusion combined, same parameters as in Fig.\ \ref{pure_advection_fig_1}, but using diffusion term, $M=1$ and $C=0.7$.
\label{adv_and_diff_combined_fig}}
\end{figure*}

Fig.\ \ref{adv_and_diff_combined_fig} shows results for Message Type 1.
We found that 
the results are very similar to the ones observed for advection only, which means that the advection process is clearly dominant.  
In fact no intra edges are found, in spite of the presence of the diffusion, plus the additional numerical diffusion from choosing $C=0.7$.
The primary difference between these results (Fig.\ \ref{adv_and_diff_combined_fig}) and the ones for pure advection (Fig. \ref{adv_and_diff_combined_fig}) 
is that there are now also many inter edges for $T> \Delta t$.  
These are again {\it echo} edges, repeating the edge direction and speed for $T=\Delta t$ 
at larger time spans.  In this case they are due to diffusion.
The directions in the velocity estimates (Type 1 and Type 2 velocity estimates are identical, since there are no intra edges) appear to be correct, 
but the magnitudes are inflated, due to the {\it echo} edges, which contribute the same 
information multiple times.
In summary, the large number of echo edges is the most interesting fact here. 
Whenever that happens one may consider to only use the inter edges up to $T=\Delta t$
to generate velocity plots.



%
%



\section{Simulation Results for Three Complex Scenarios}
\label{complex_scenarios_sec}

Now we define three more complex scenarios to test the
structure learning from complex spatio-temporal data.
%
%
The key feature of these scenarios is that the advection velocity fields 
no longer align with the horizontal or vertical direction of the grid
and that their magnitude also varies by location.
Thus these scenarios are suitable to test the effects of 
(1) varying orientation of arrow direction, i.e.\ no longer aligning with the grid;
(2) varying magnitude of arrows.
Furthermore, each scenario has some additional features that make it interesting 
for testing certain effects.

\begin{figure*}
\centerline{ 
\includegraphics[width=7.5cm,angle=0,clip]{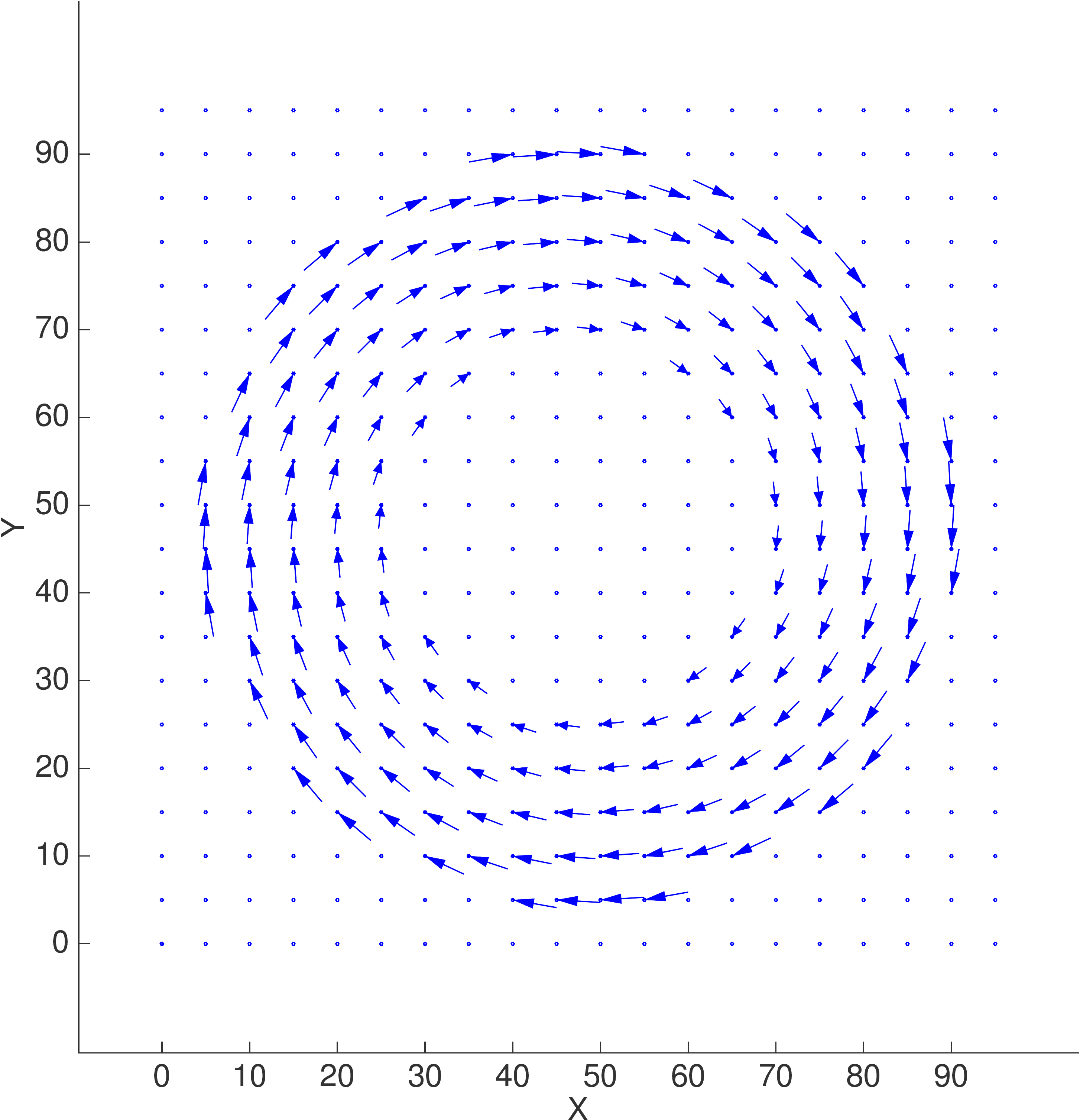}
\hspace*{0.5cm}
\includegraphics[width=7.5cm,angle=0,clip]{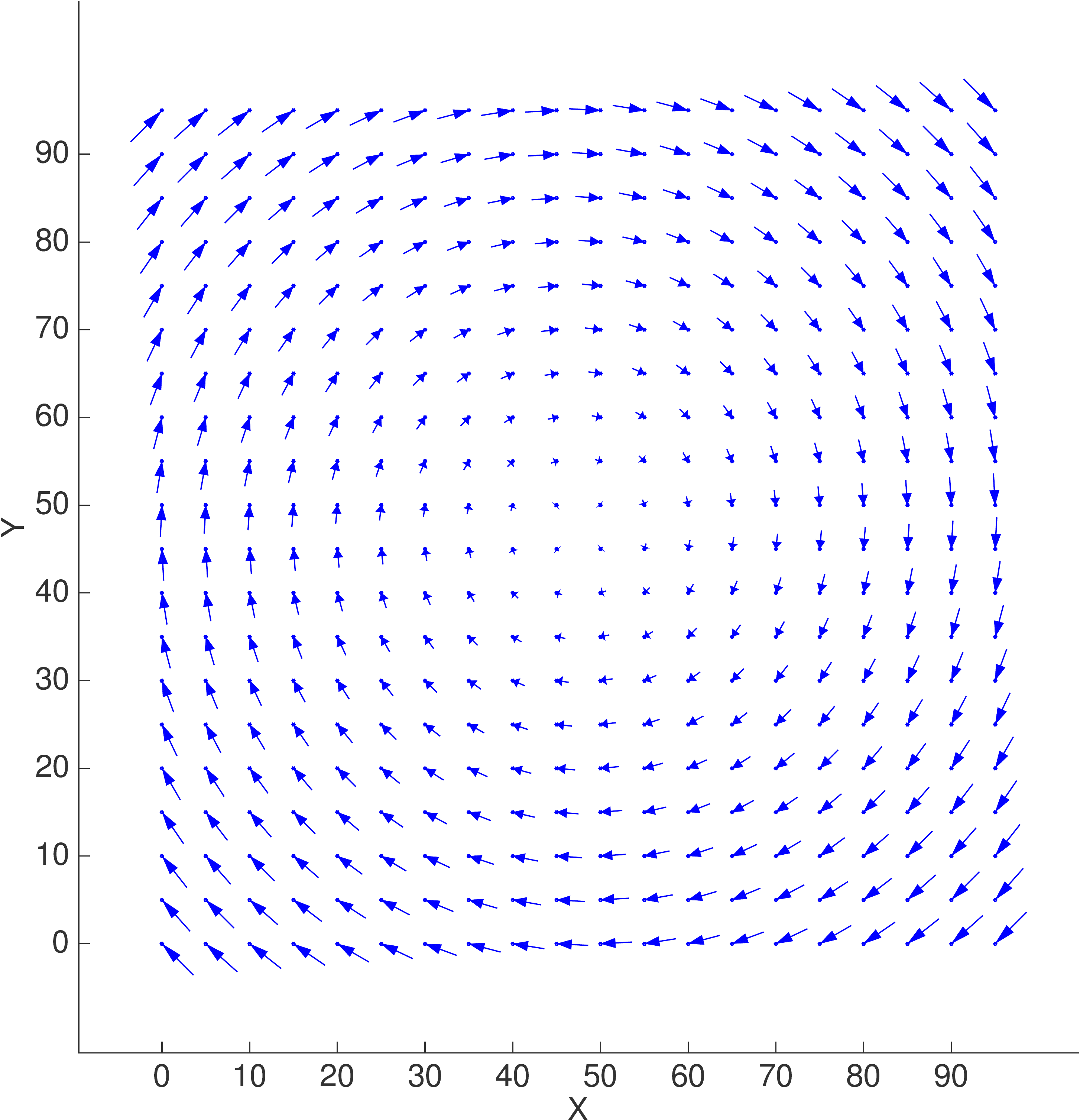}
}
\centerline{(a) Scenario 1 \hspace*{5.0cm} (b) Scenario 2}
\vspace*{0.5cm}
\centerline{ 
\includegraphics[width=7.5cm,angle=0,clip]{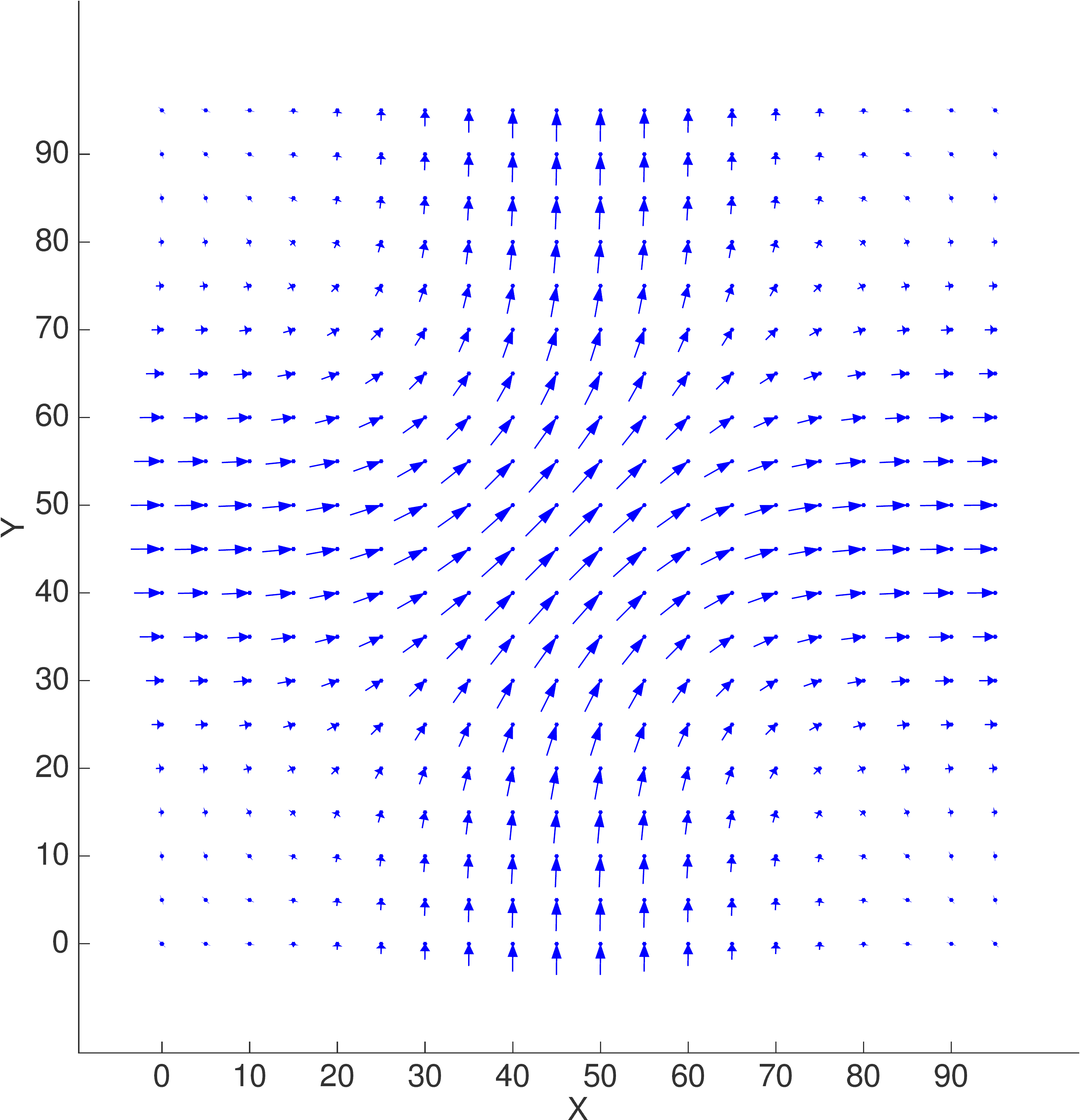}
}
\centerline{(c) Scenario 3}
\caption{Advection Velocity Fields for Scenarios 1, 2 and 3.
  \label{scenario_1_to_3_advection_vel}}
\end{figure*}

{\bf Scenario 1:}
The advection for Scenario 1 is defined by the rotating ring flow 
shown in Figure \ref{scenario_1_to_3_advection_vel}(a).
The special feature of Scenario 1 is that the velocities are set to exactly zero in large areas, 
namely anywhere inside and outside of the ring. 
Thus Scenario 1 is designed to also test the effect of large areas with zero 
advection velocity, i.e.\ pure diffusion in those areas.

{\bf Scenario 2:}
The advection for Scenario 2 is defined by the rotating vector field 
shown in Figure \ref{scenario_1_to_3_advection_vel}(b).
%
%
Note that the magnitude of the vectors is proportional to the distance from the center, 
i.e.\ velocities vanish near the center and are largest at the boundaries.  
Furthermore, the periodic boundary conditions
cause strong discontinuities of the velocity directions near the boundaries.  
For example, at the center of the upper boundary of Fig.\ \ref{scenario_1_to_3_advection_vel}(b)
the velocities are straight to the right,
while at the center of the bottom boundary the velocities are straight to the left, 
i.e.\ the velocities at these wrap-around neighboring points are exactly opposite of each other, mimicking the occurrence of {\it counter currents}.  
The same effect occurs at the centers of the right and left boundary.
Thus there is a 180 degree angle between the velocity vectors at wrap-around 
at the centers of all four boundary lines.  
This angle decreases to about 90 degrees towards the four corners, thus it remains 
significant.  Thus Scenario 1 is designed to test the effect of counter currents 
and other abrupt changes in velocity direction.

{\bf Scenario 3:}
The advection for Scenario 3 is motivated by the cross current velocity field 
used by \cite{MoReMaKu:2014} to test their correlation networks.
Scenario 3 emulates two crossing currents as shown in 
Figure \ref{scenario_1_to_3_advection_vel}(c). 
%
One current flows from left to right, the other from bottom to top,
with velocities increasing exponentially toward the center of the grid.
Even though velocities are small outside of the two main currents, 
there are no true zero velocities in this case.  Furthermore, in contrast to 
Scenario 1 the directions in Scenario 3 are all consistent at the boundaries
(no sudden changes).

{\bf Other parameters for Scenarios 1-3:}
Each grid consists of $20 \times 20 = 400$ points.  
%
As before, the advection velocity field is scaled 
for each scenario so that the maximal velocity is $1$ $m/s$ and 
diffusion is set to $\kappa_x = \kappa_y = 1$ $m^2/s$.
The numerical stability parameter is chosen as $C=0.5$, which yields significant 
additional diffusion.
All results reported here are for Message Type 1 (IC peak) and no prior noise is used.
%


\subsection{Results for complex scenarios when concurrent edges are allowed}



%
\begin{figure*}
\centerline{ 
\includegraphics[width=4.5cm,angle=0]{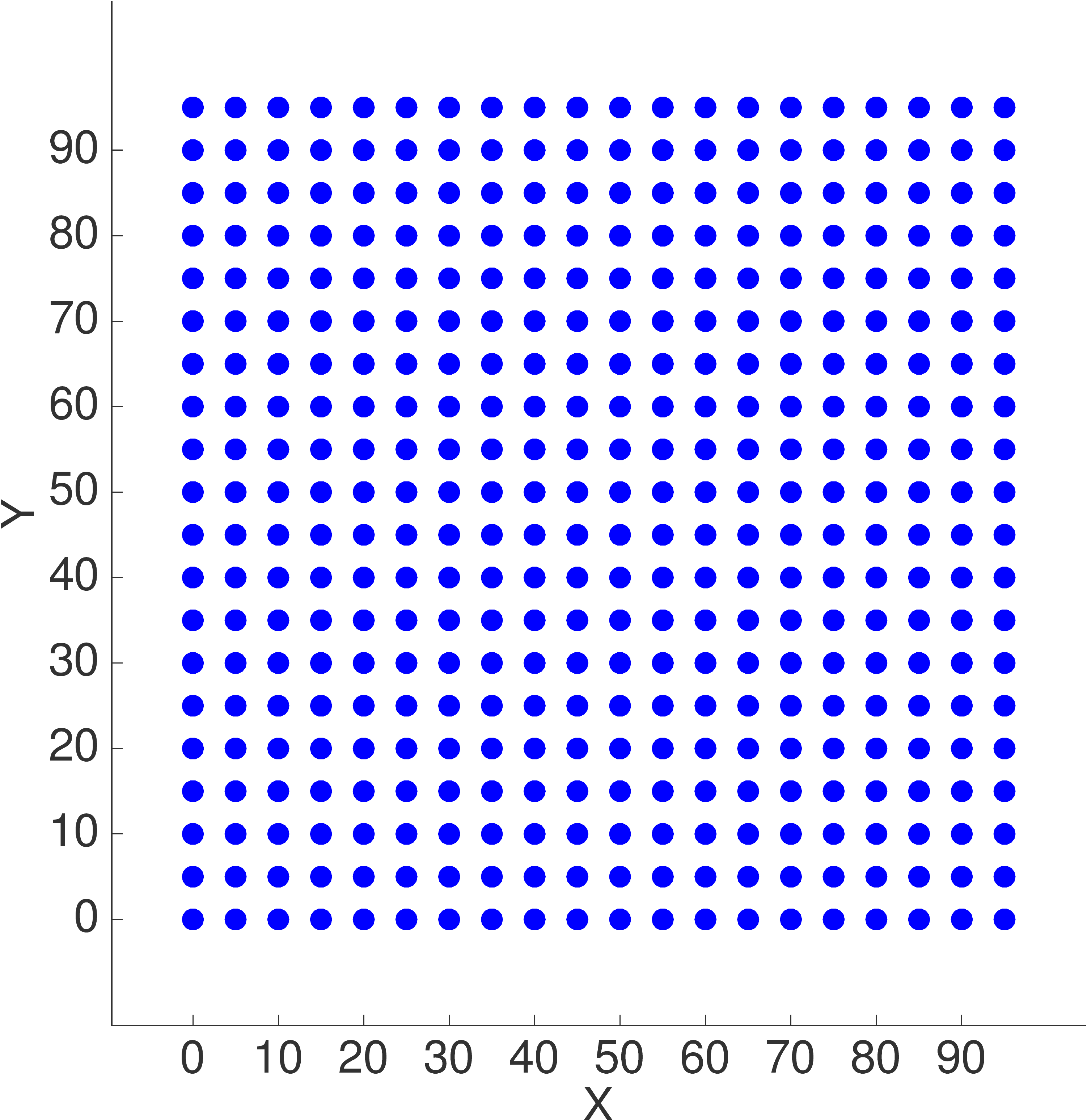}
\hspace*{0.5cm}
\includegraphics[width=4.5cm,angle=0]{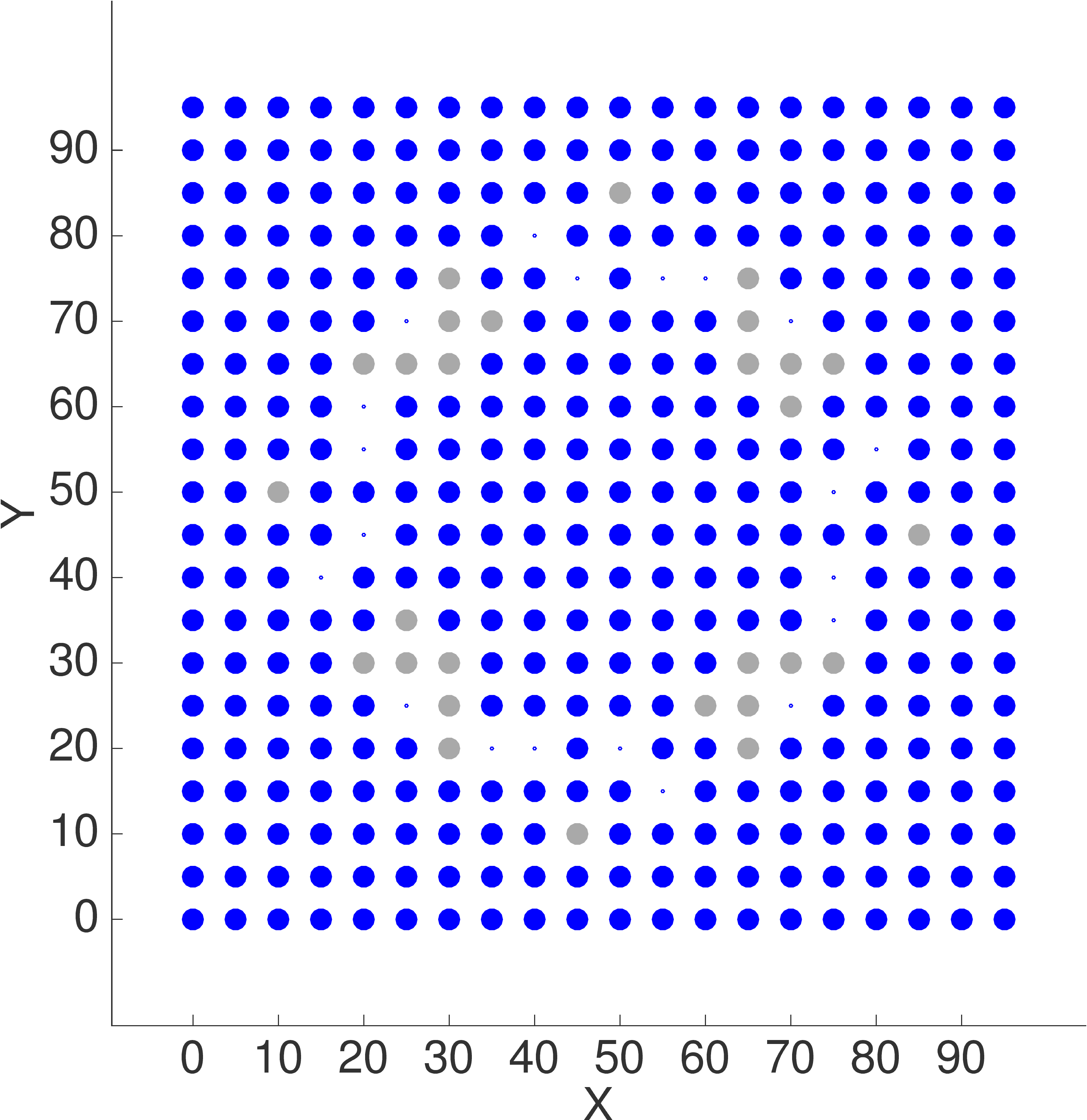}
\hspace*{0.5cm}
\includegraphics[width=4.5cm,angle=0]{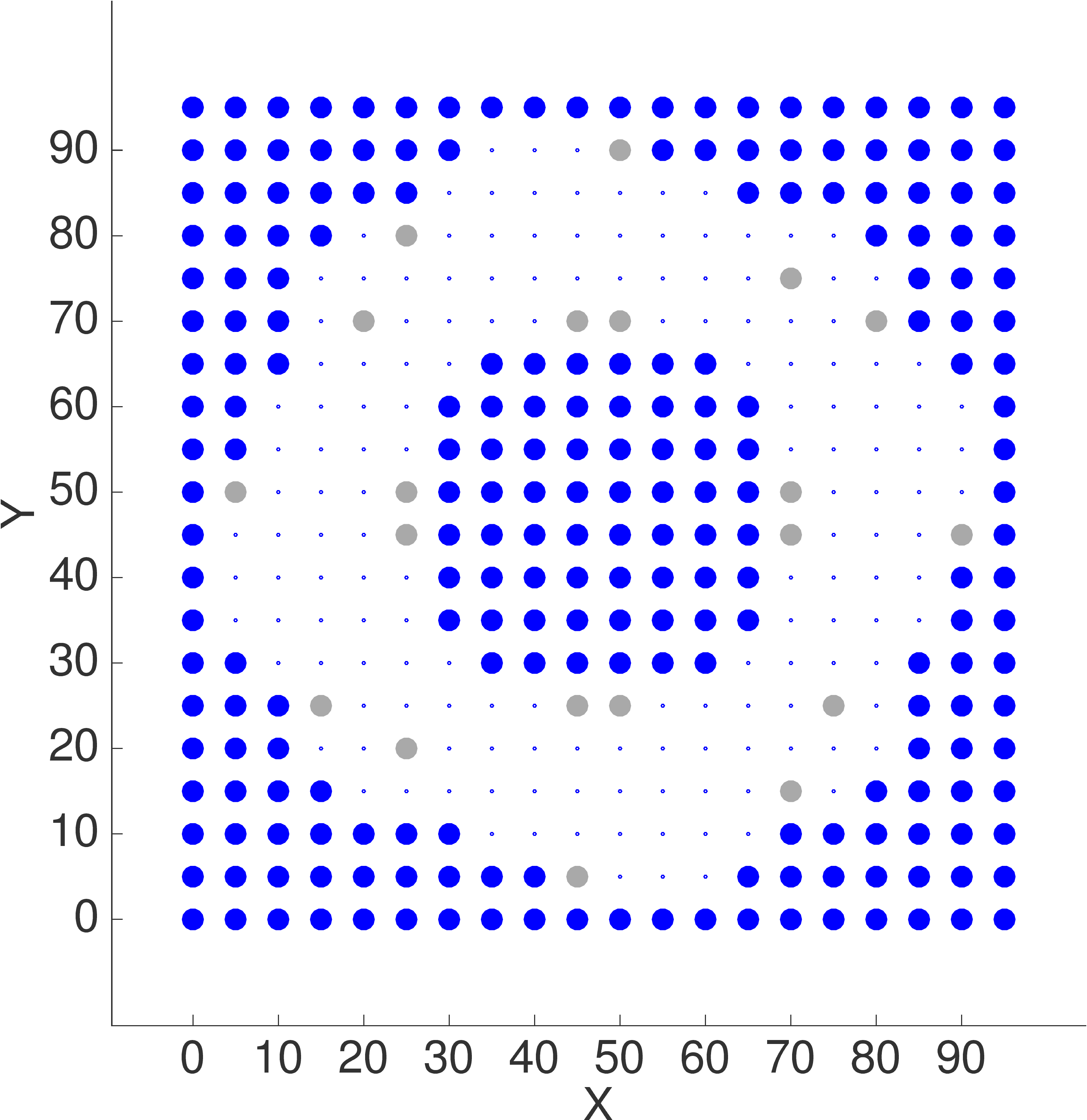}
}
\centerline{$T = \Delta t$  \hspace*{3.5cm} $T = 2 \Delta t$  \hspace*{3.5cm} $T = 3 \Delta t$ }
\centerline{(a) Intra edges}
\vspace*{0.3cm}
\centerline{ 
\includegraphics[width=4.5cm,angle=0]{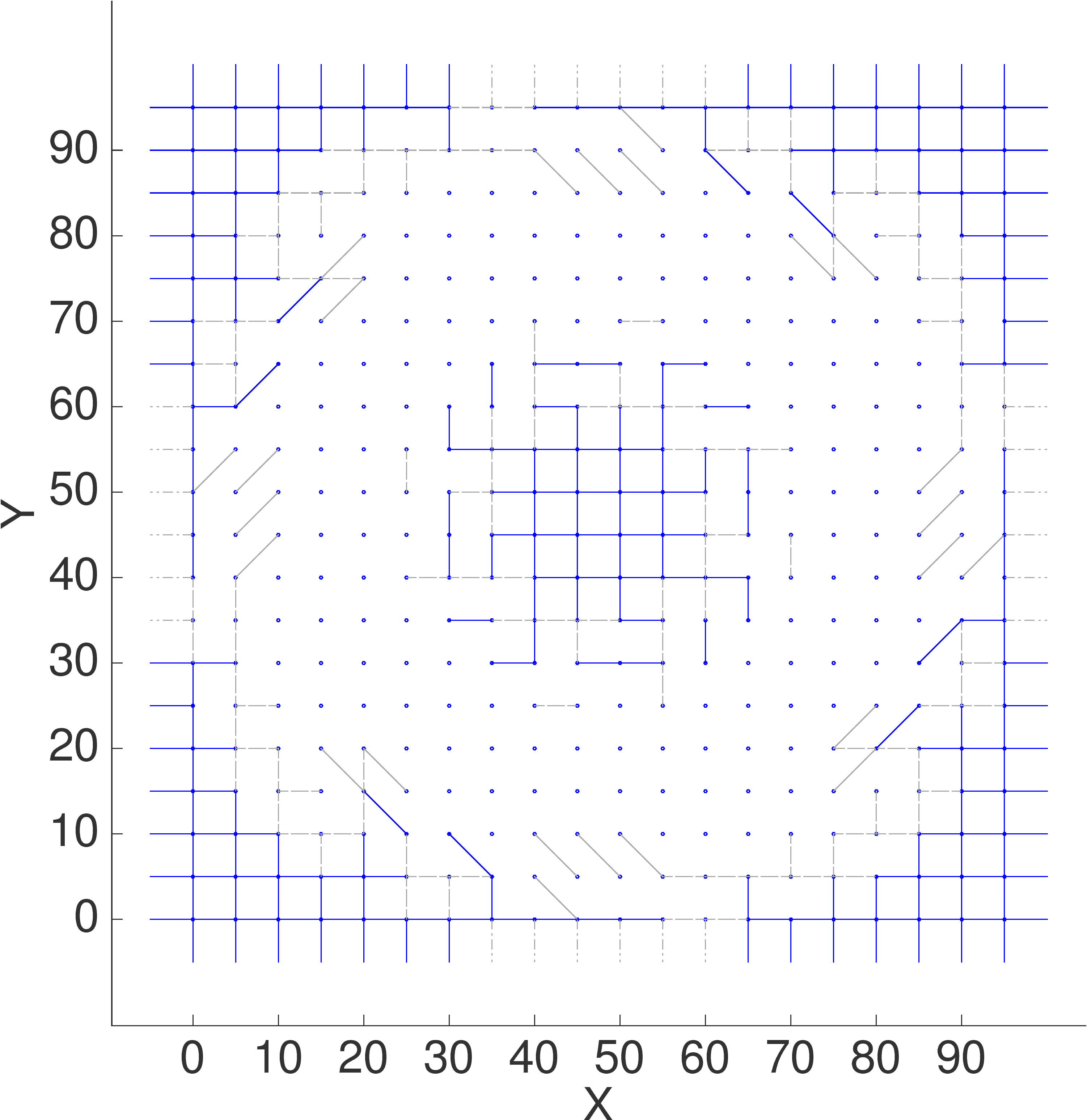}
\hspace*{0.5cm}
\includegraphics[width=4.5cm,angle=0]{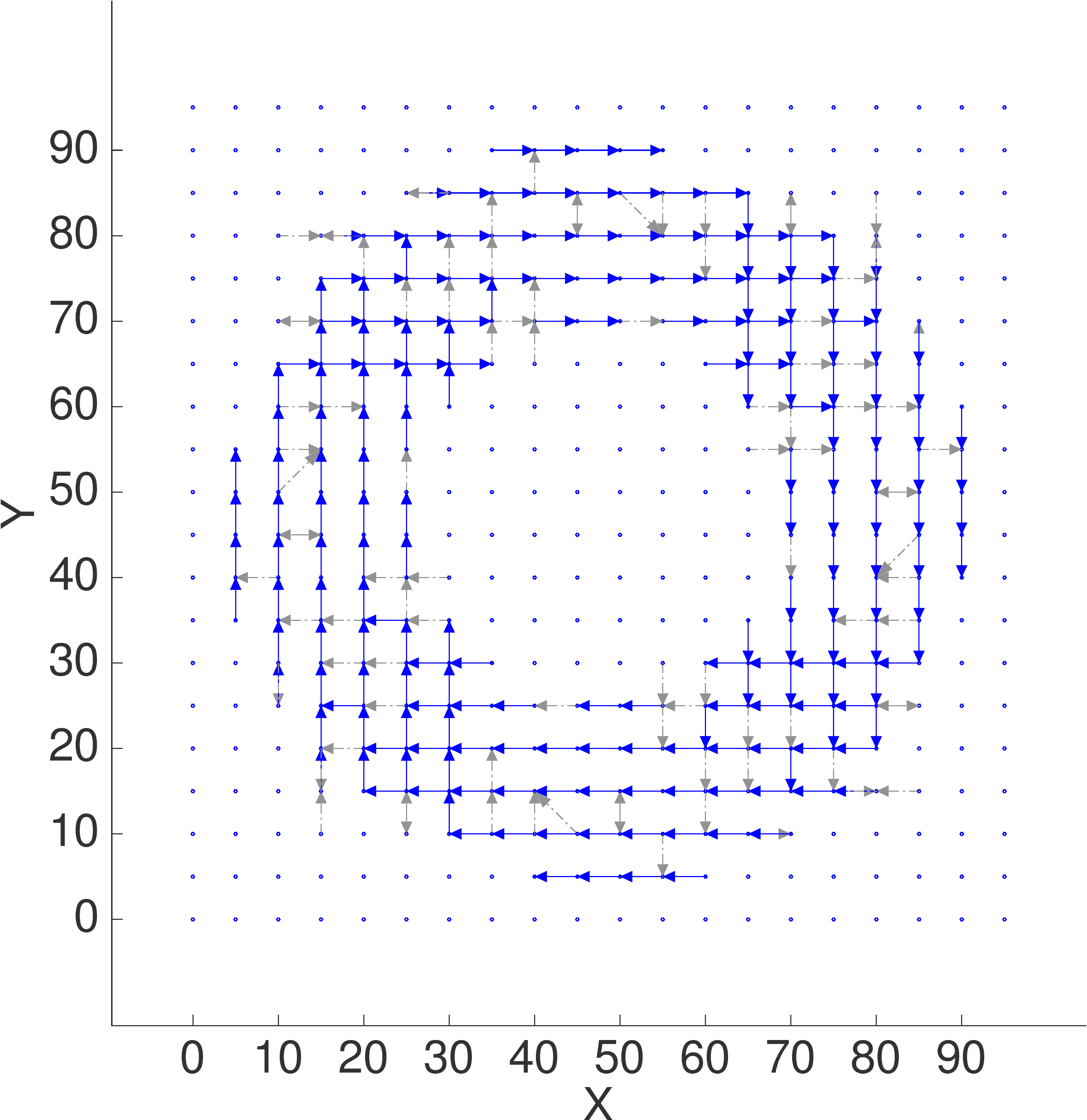}
\hspace*{0.5cm}
\includegraphics[width=4.5cm,angle=0]{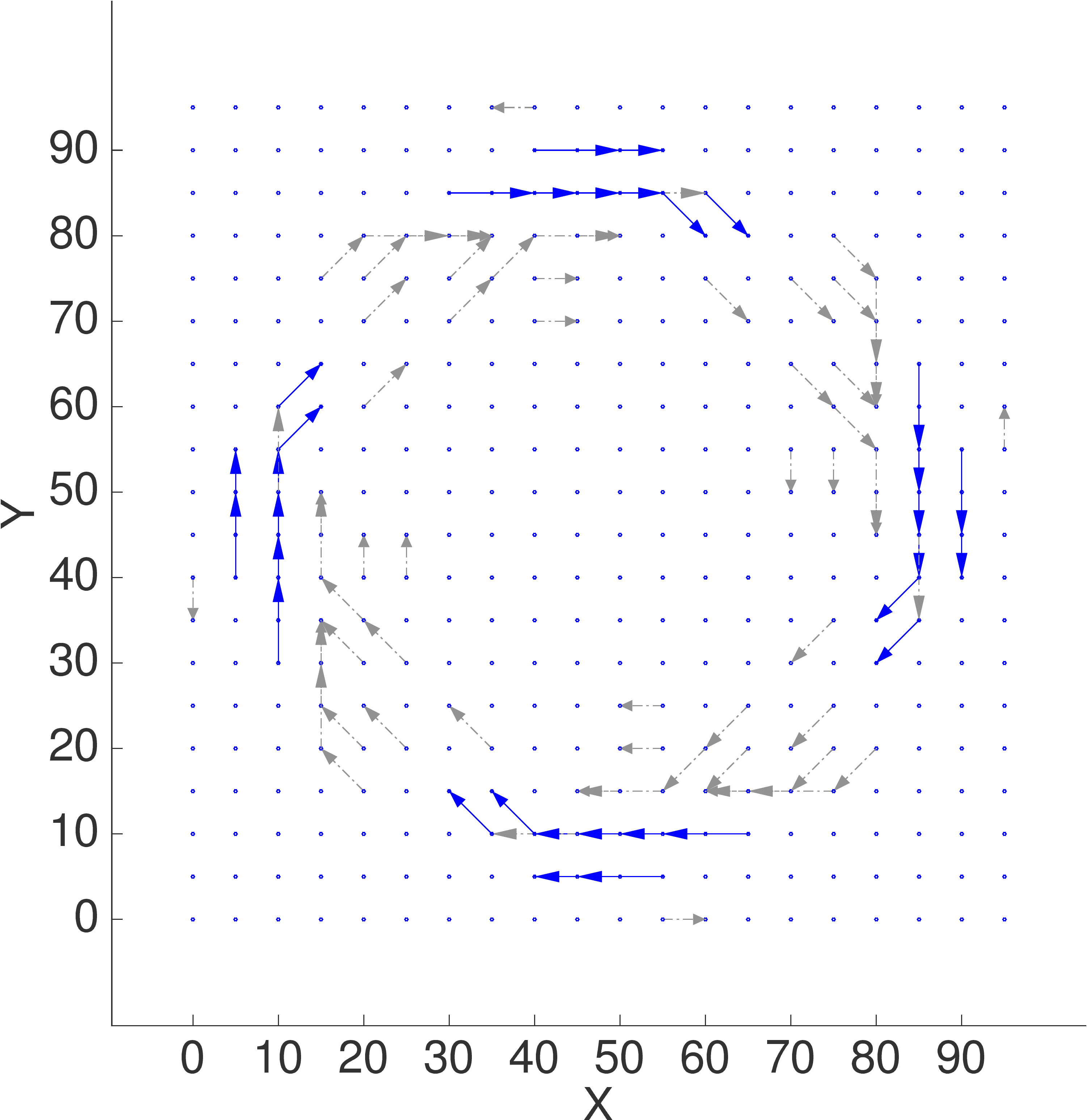}
}
\centerline{$T = 0$  \hspace*{3.5cm} $T = \Delta t$  \hspace*{3.5cm} $T = 2 \Delta t$}
\centerline{(b) Inter edges}
\vspace*{0.3cm}
\centerline{ 
\includegraphics[width=7.9cm,angle=0,clip]{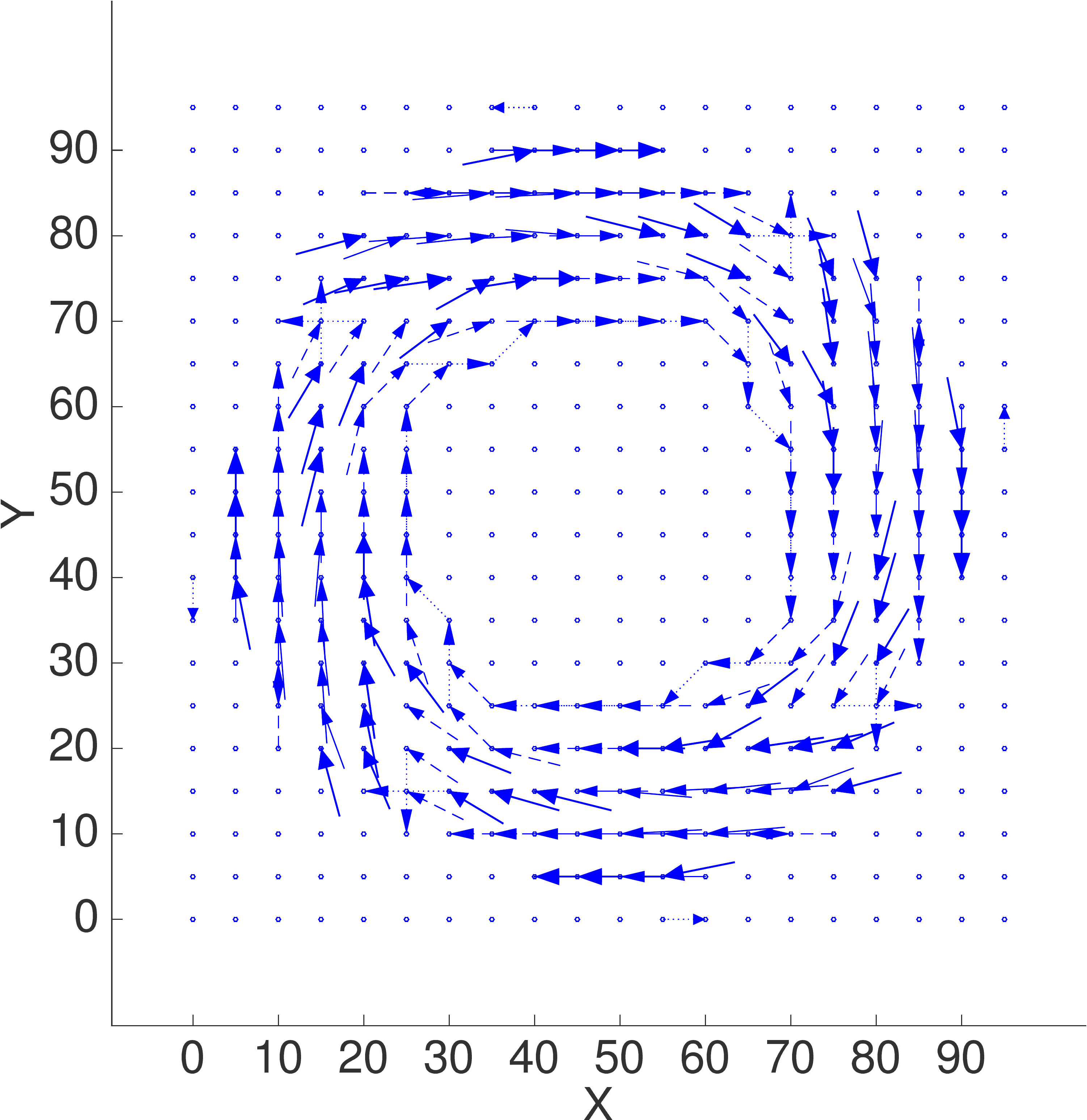}
\hspace*{0.2cm}
\includegraphics[width=7.9cm,angle=0,clip]{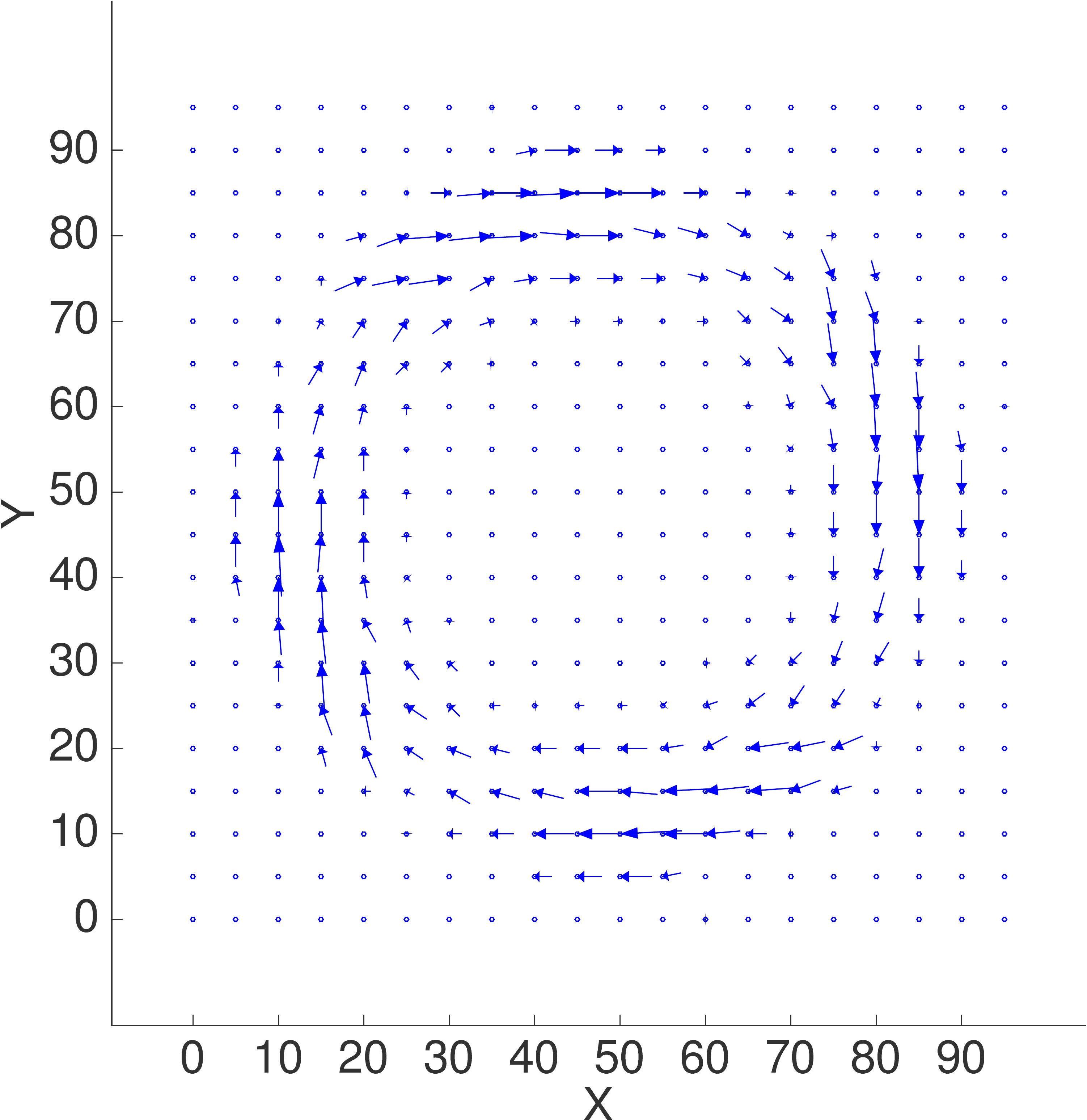}
}
\centerline{(c) Velocity estimated {\it without} intra edges \hspace*{1.0cm} (d) Velocity estimated {\it with} intra edges}
\caption{Results for Scenario 1 with concurrent edges allowed.
\label{scenario_1_with_concurrent_fig}}
\end{figure*}
%

{\bf Results for Scenario 1 (Fig.\ \ref{scenario_1_with_concurrent_fig}):}\\
(1) There are more intra edges than expect for $T=\Delta t$.  Starting at $T=3 \Delta t$ we see the pattern we expected, namely intra edges only where advection velocity is low.\\
(2) Inter edges for $T \ge \Delta t$ show good approximation of the original advection field, with most of the edges present for $T=\Delta t$.  Likewise the velocity estimates give a good indication of the shape of the original advection field.\\
3) Concurrent inter edges (T=0) occur in all locations where diffusion dominates (areas with zero advection velocity). No surprise there.

\begin{figure*}
\centerline{ 
\includegraphics[width=4.5cm,angle=0]{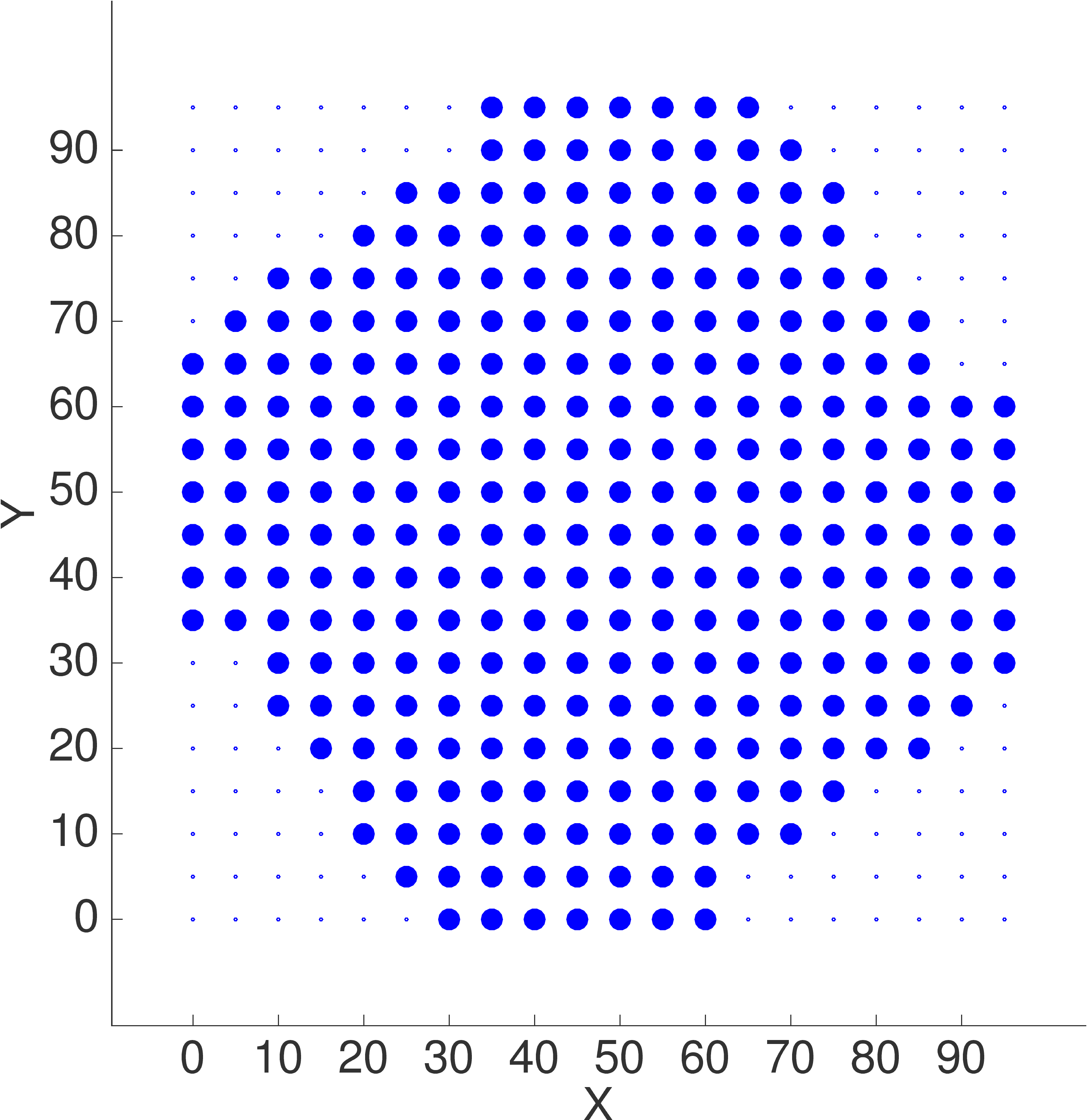}
\hspace*{0.5cm}
\includegraphics[width=4.5cm,angle=0]{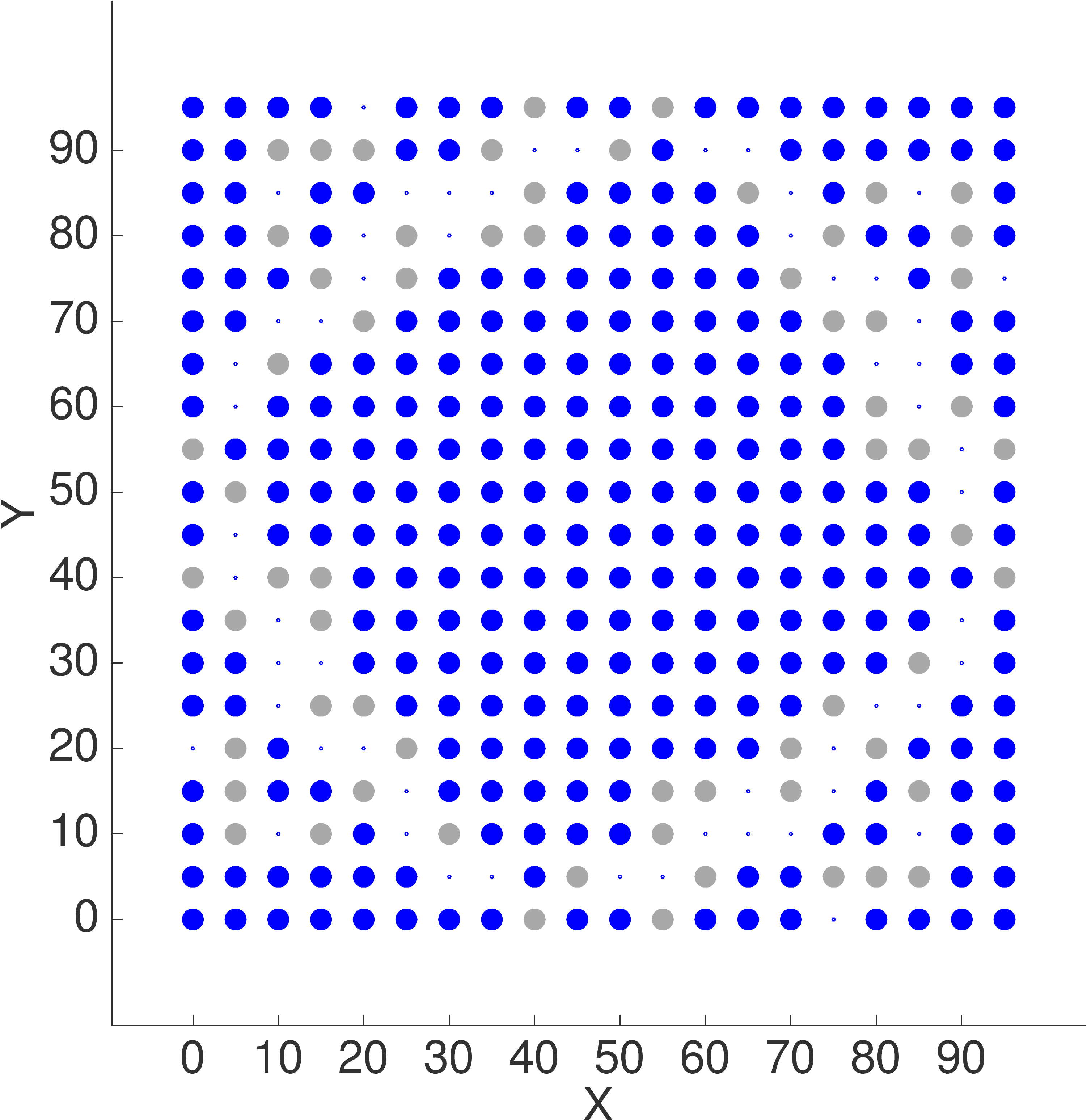}
\hspace*{0.5cm}
\includegraphics[width=4.5cm,angle=0]{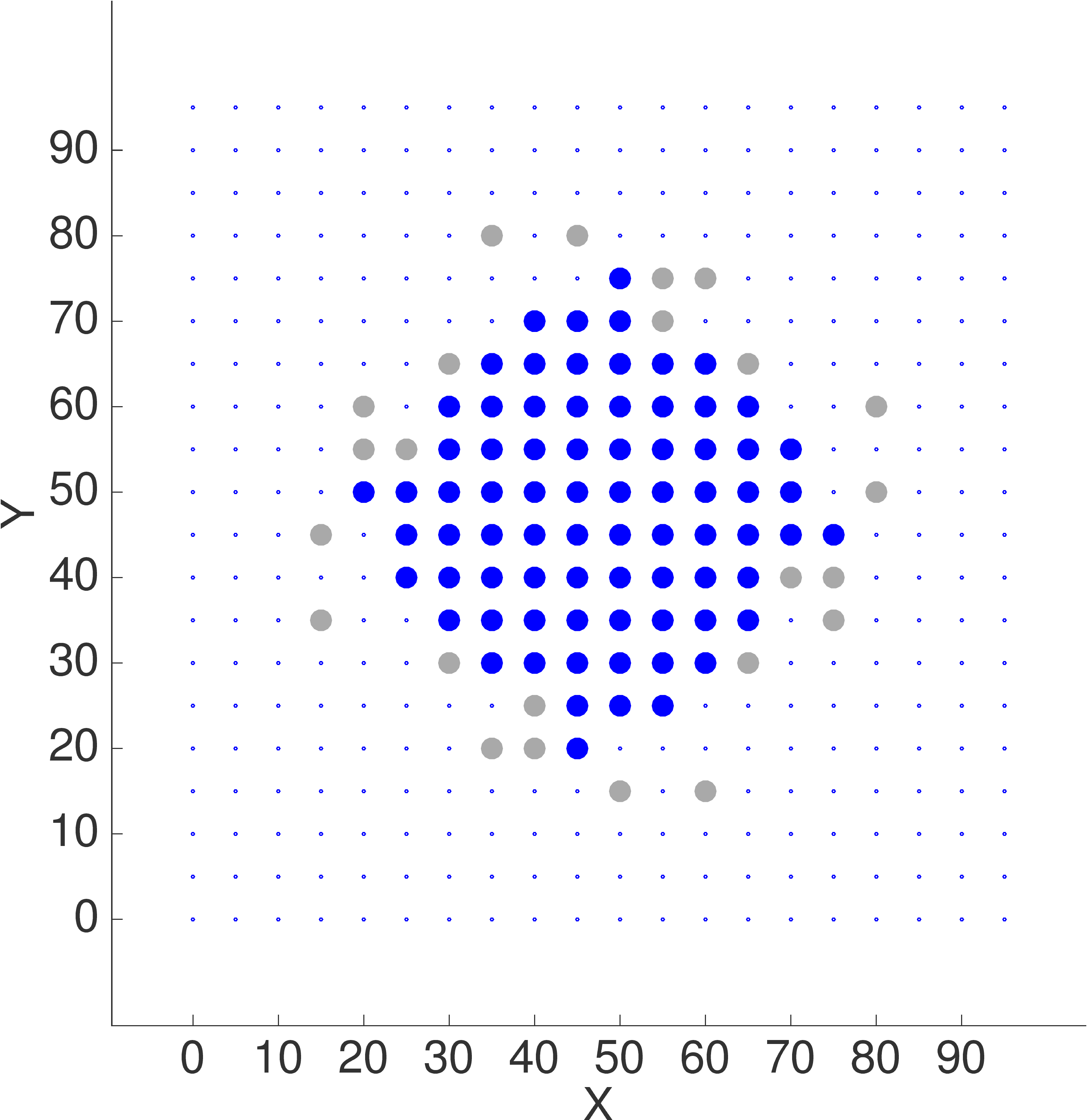}
}
\centerline{$T = \Delta t$  \hspace*{3.5cm} $T = 2 \Delta t$  \hspace*{3.5cm} $T = 3 \Delta t$ }
\centerline{(a) Intra edges}
\vspace*{0.3cm}
\centerline{ 
\includegraphics[width=4.5cm,angle=0]{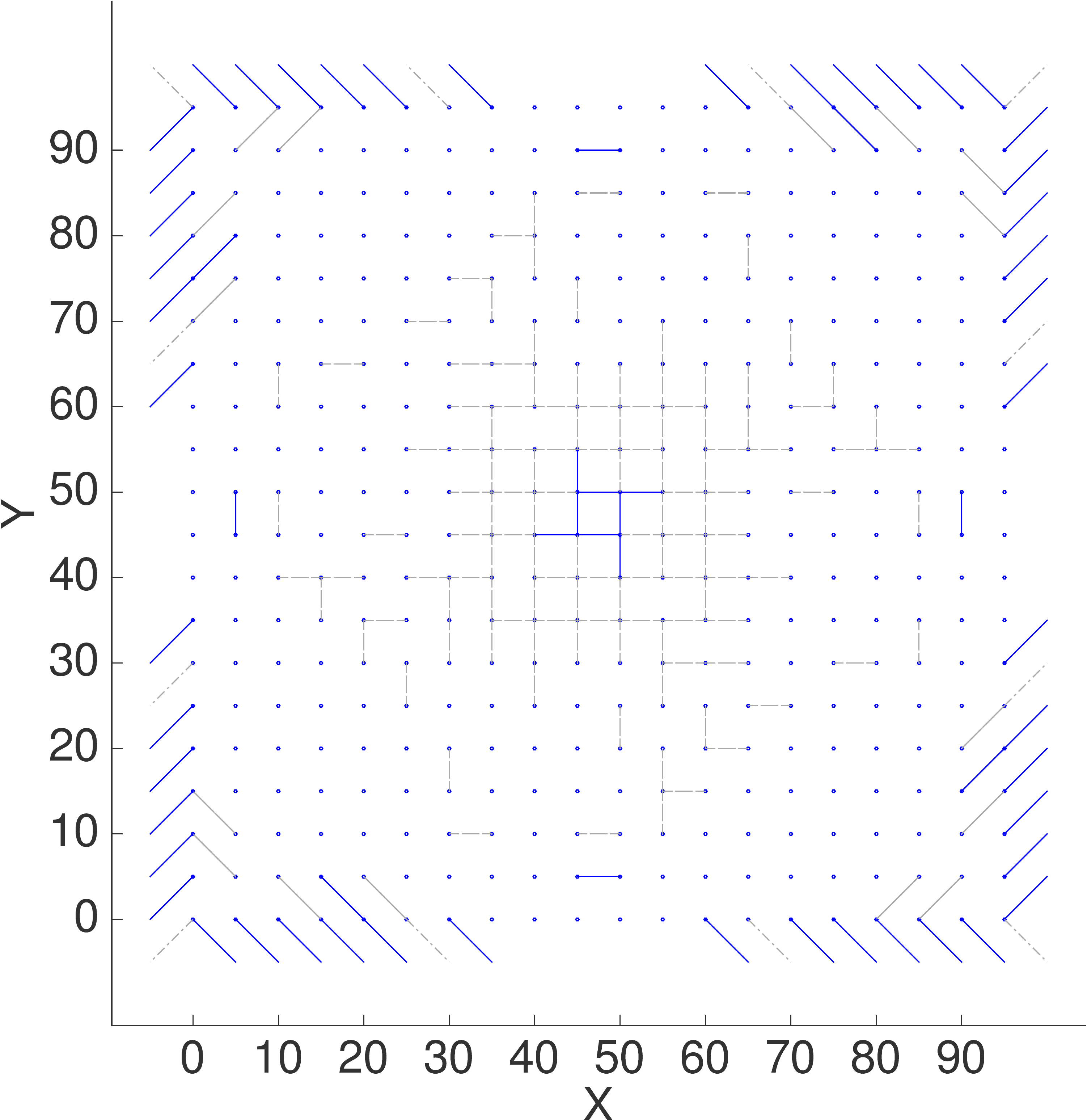}
\hspace*{0.5cm}
\includegraphics[width=4.5cm,angle=0]{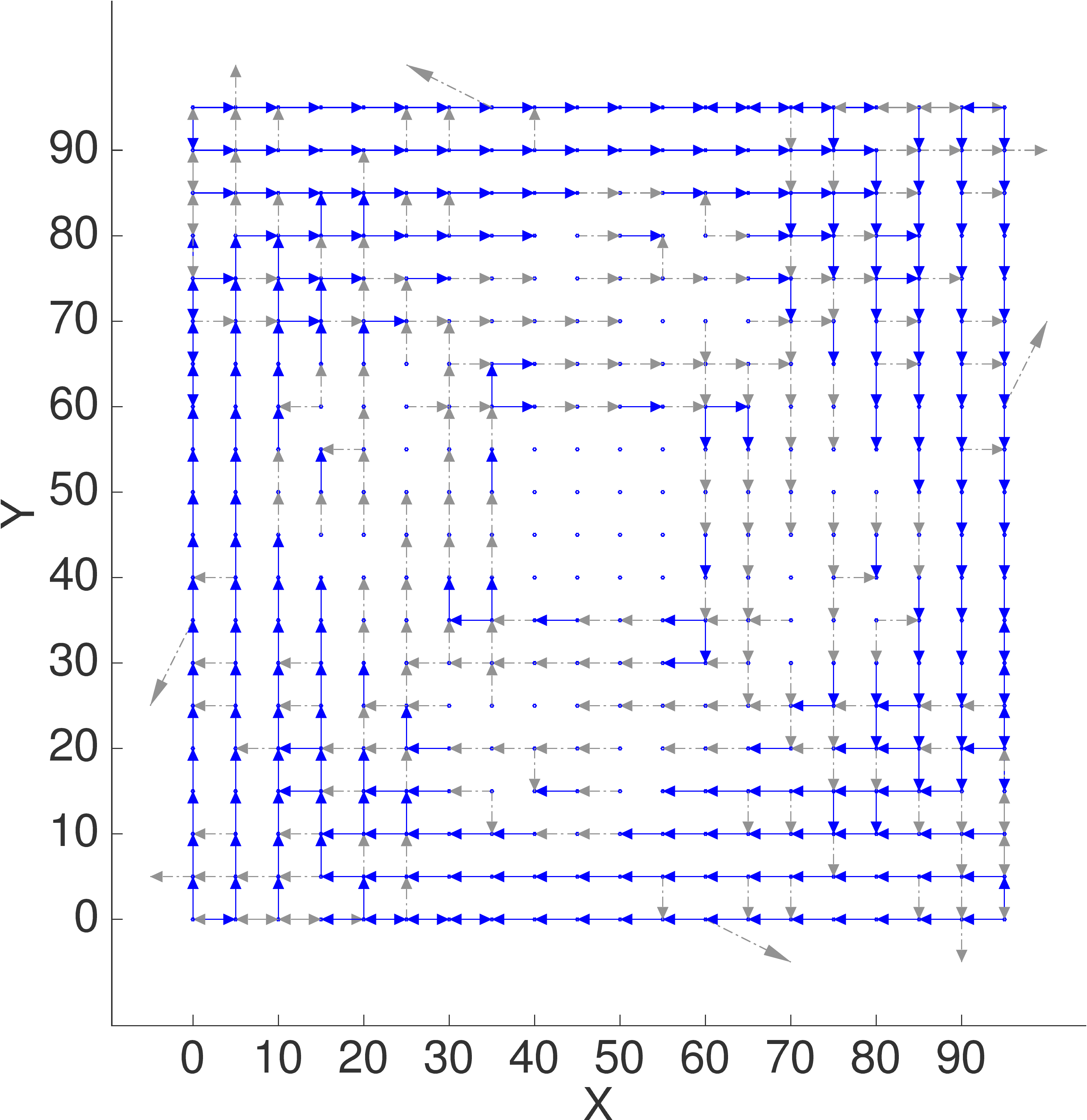}
\hspace*{0.5cm}
\includegraphics[width=4.5cm,angle=0]{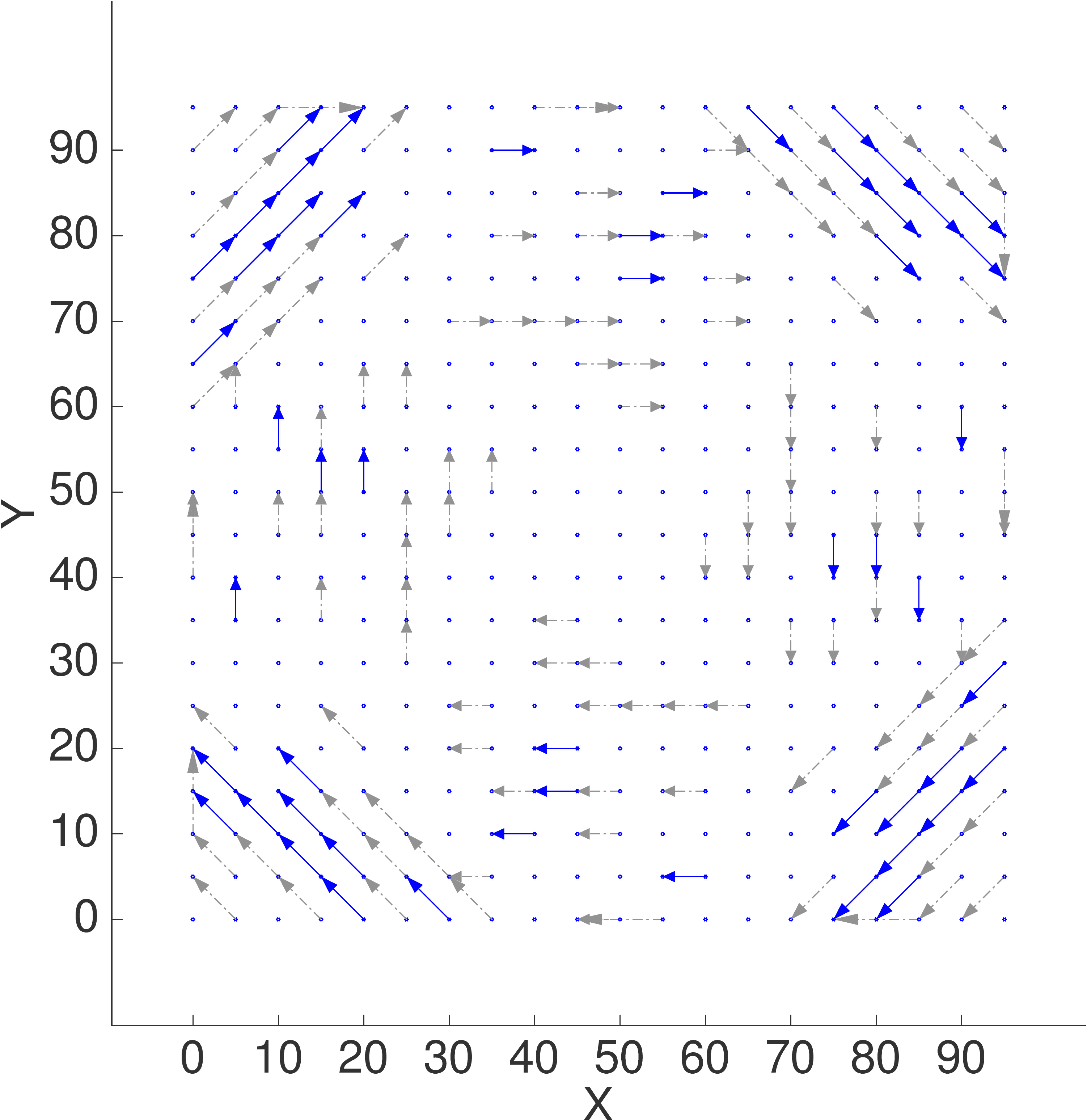}
}
\centerline{$T = 0$  \hspace*{3.5cm} $T = \Delta t$  \hspace*{3.5cm} $T = 2 \Delta t$}
\centerline{(b) Inter edges}
\vspace*{0.3cm}
\centerline{ 
\includegraphics[width=7.9cm,angle=0,clip]{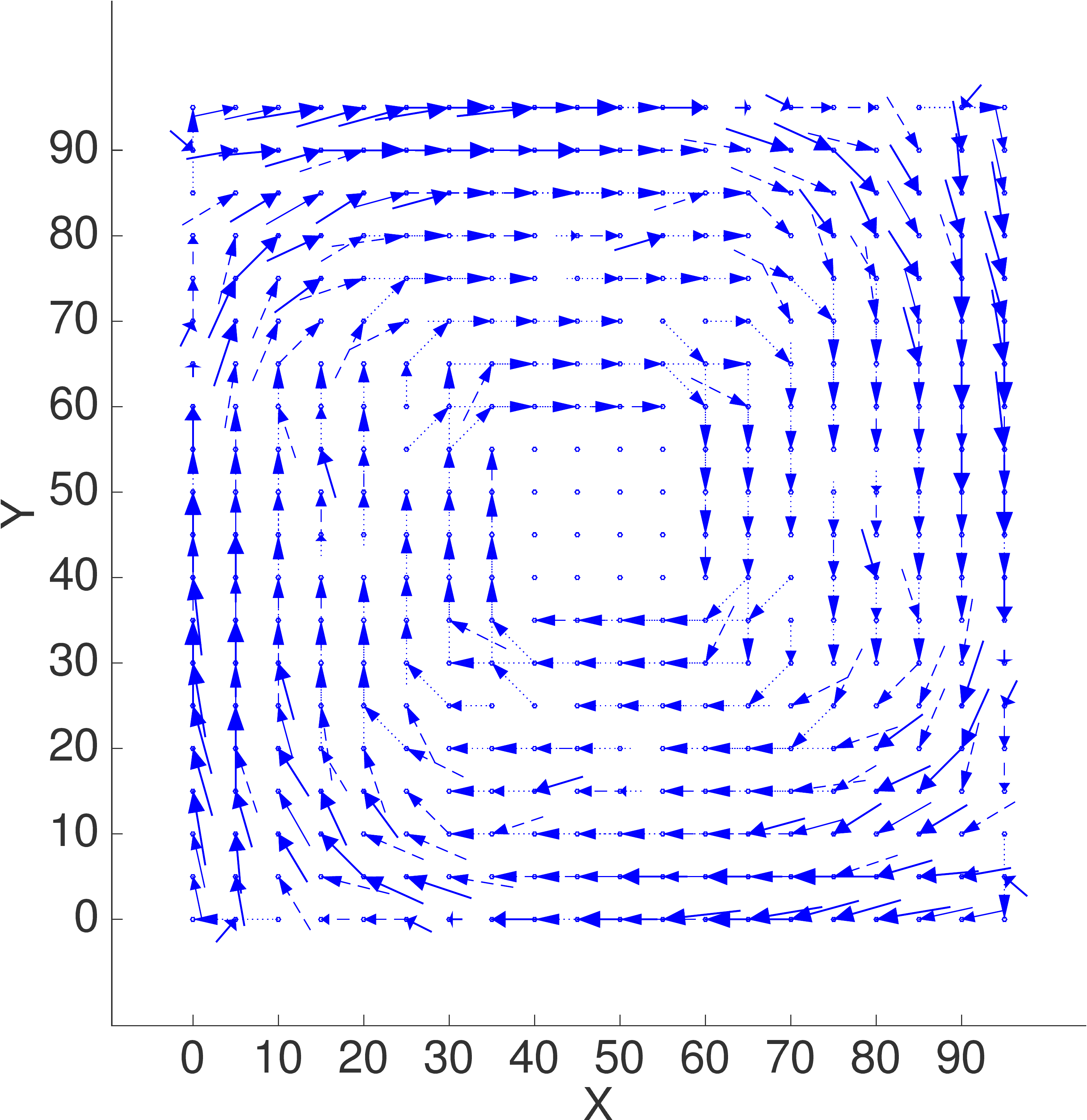}
\hspace*{0.2cm}
\includegraphics[width=7.9cm,angle=0,clip]{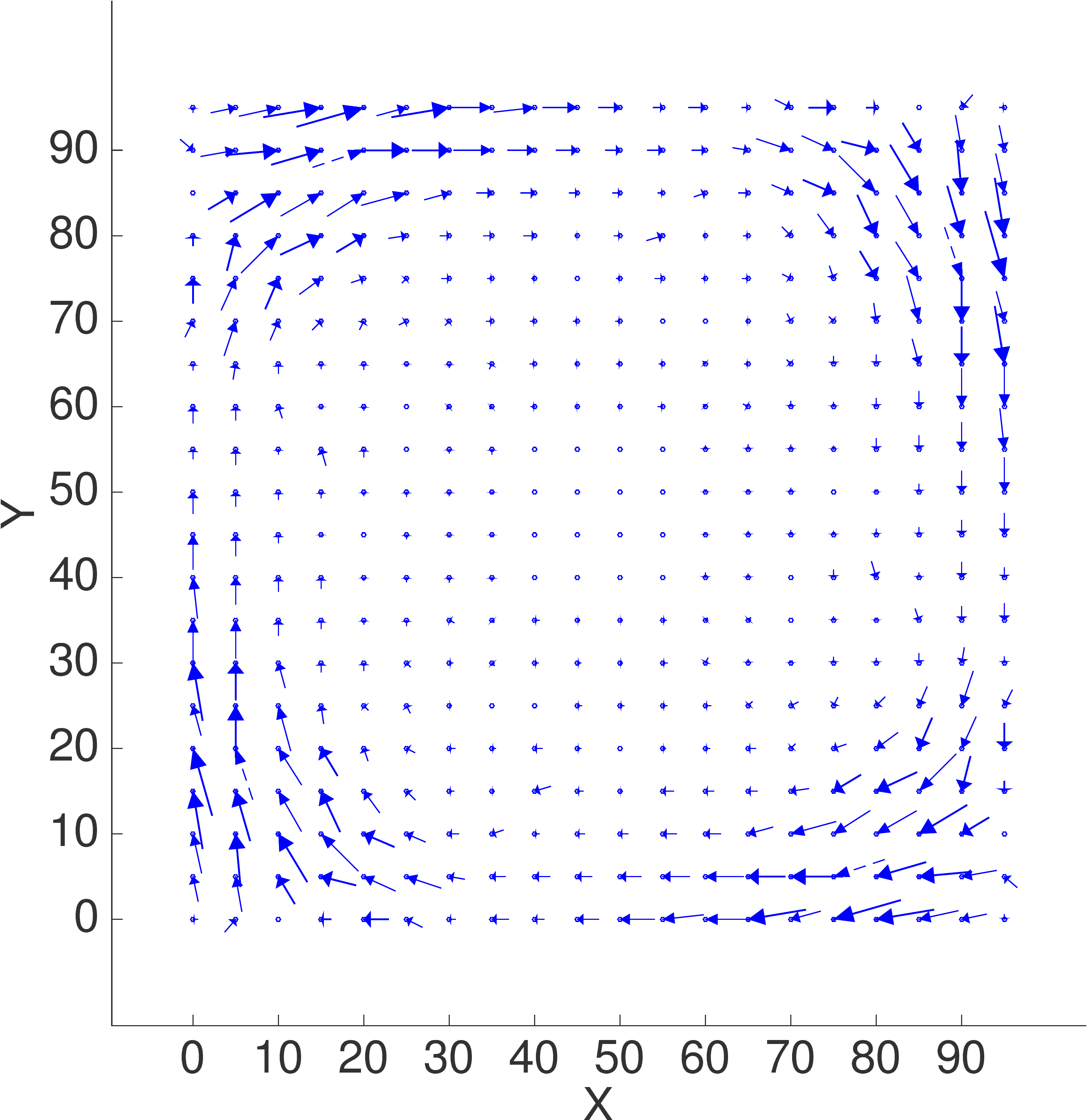}
}
\centerline{(c) Velocity estimated {\it without} intra edges \hspace*{1.0cm} (d) Velocity estimated {\it with} intra edges}
\caption{Results for Scenario 2 with concurrent edges allowed.
\label{scenario_2_with_concurrent_fig}}
\end{figure*}

\noindent
{\bf Results for Scenario 2 (Fig.\ \ref{scenario_2_with_concurrent_fig}):}\\
(1) Intra edges occur where advection velocity is smallest, as expected.\\
(2) Inter edges for $T= \Delta t$ capture most of the advection velocities, but one can see the increasing difficulty of using a rectangular grid to represent diagonal velocity vectors.\\
(3) The velocity plot of Type 1 (without intra edges) gives a good indication of the original velocity field, but (1) with directions distorted to align with the grid, and (2) with inflated magnitudes.   The velocity plot of Type 2 (taking intra edges into account) is better at detecting the high speed advection areas, but drops the connections in the low velocity areas almost completely.\\
(4) The concurrent inter edges are very interesting (left-most panel in Fig.\ \ref{scenario_2_with_concurrent_fig}).  The connections towards the center are expected, because diffusion is dominant there.  However, the connections toward the boundary are a new effect.  These are probably due to the contradictory velocities at the boundaries of the advection field of Scenario 2, and will be discussed more later.

 
%
%

%
%
%
\begin{figure*}
\centerline{ 
\includegraphics[width=4.5cm,angle=0]{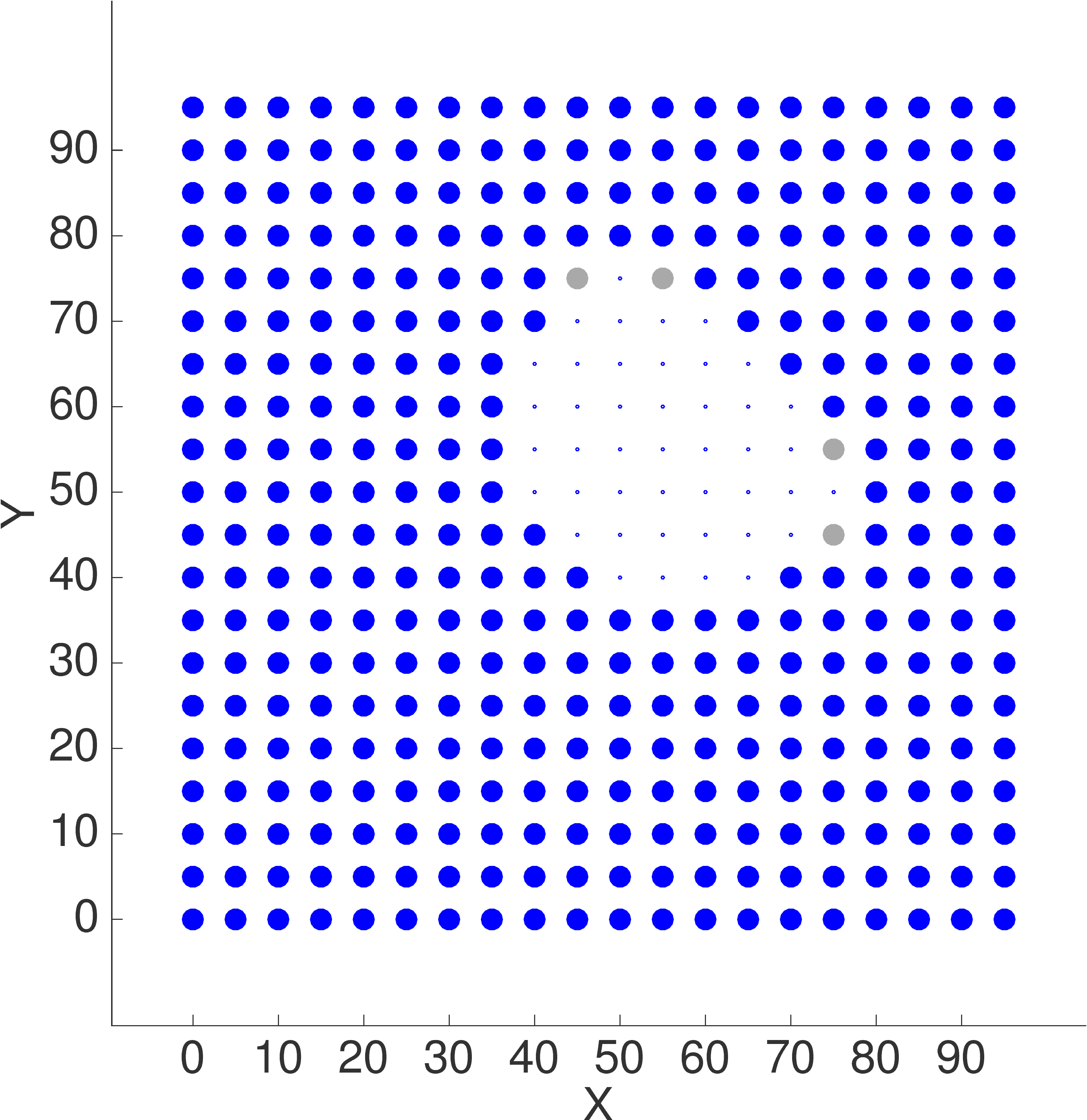}
\hspace*{0.5cm}
\includegraphics[width=4.5cm,angle=0]{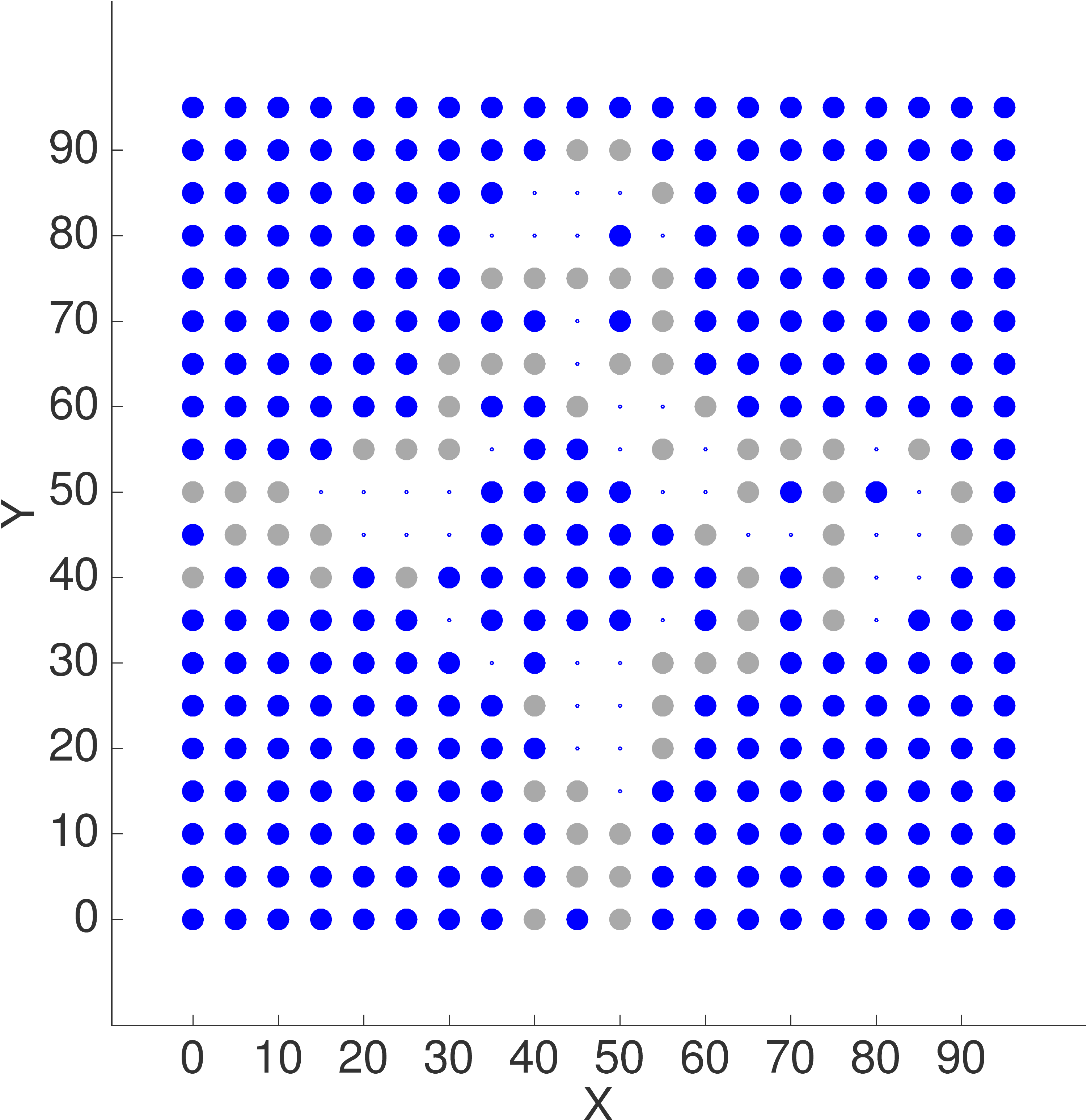}
\hspace*{0.5cm}
\includegraphics[width=4.5cm,angle=0]{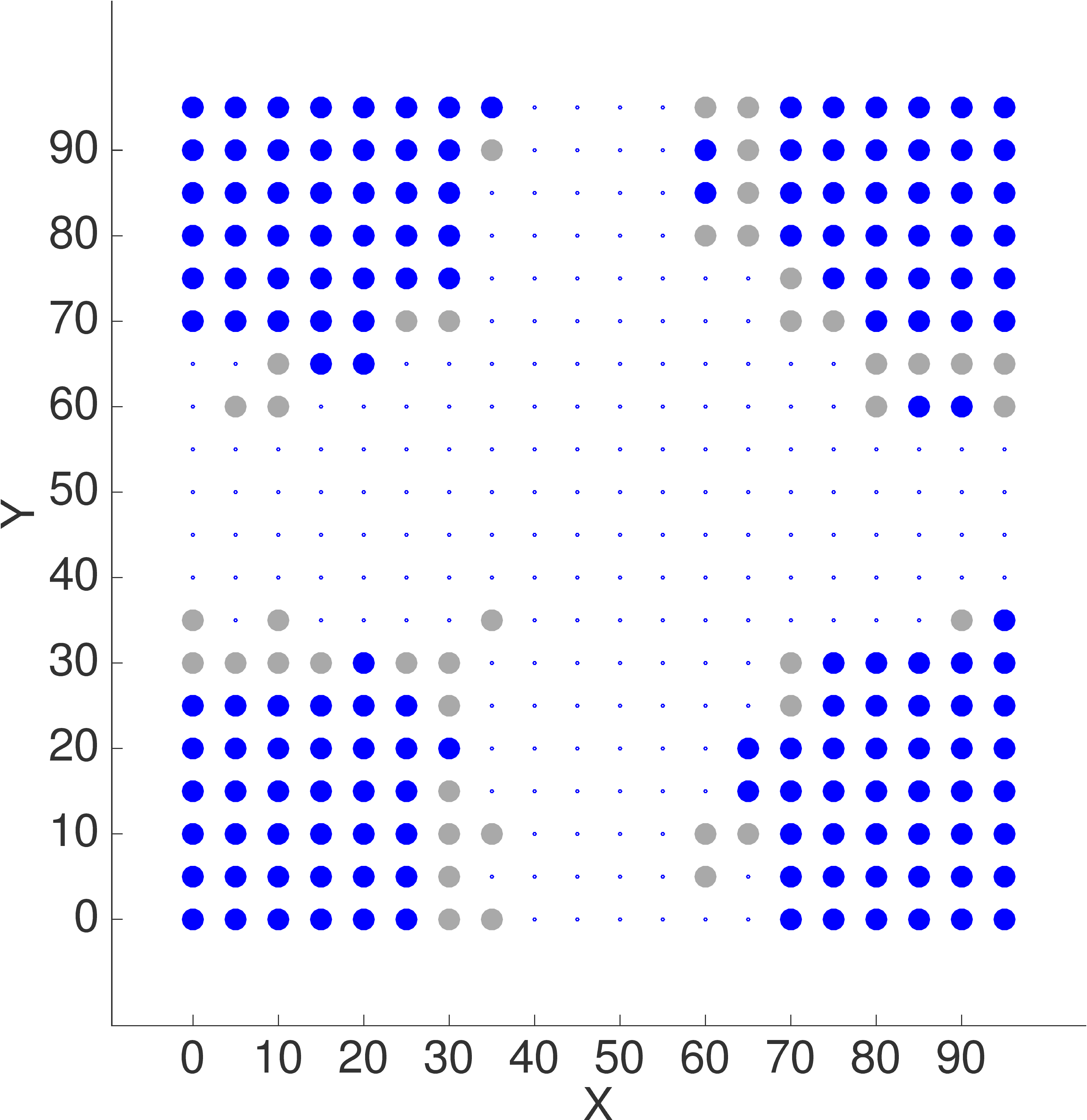}
}
\centerline{$T = \Delta t$  \hspace*{3.5cm} $T = 2 \Delta t$  \hspace*{3.5cm} $T = 3 \Delta t$ }
\centerline{(a) Intra edges}
\vspace*{0.3cm}
\centerline{ 
\includegraphics[width=4.5cm,angle=0]{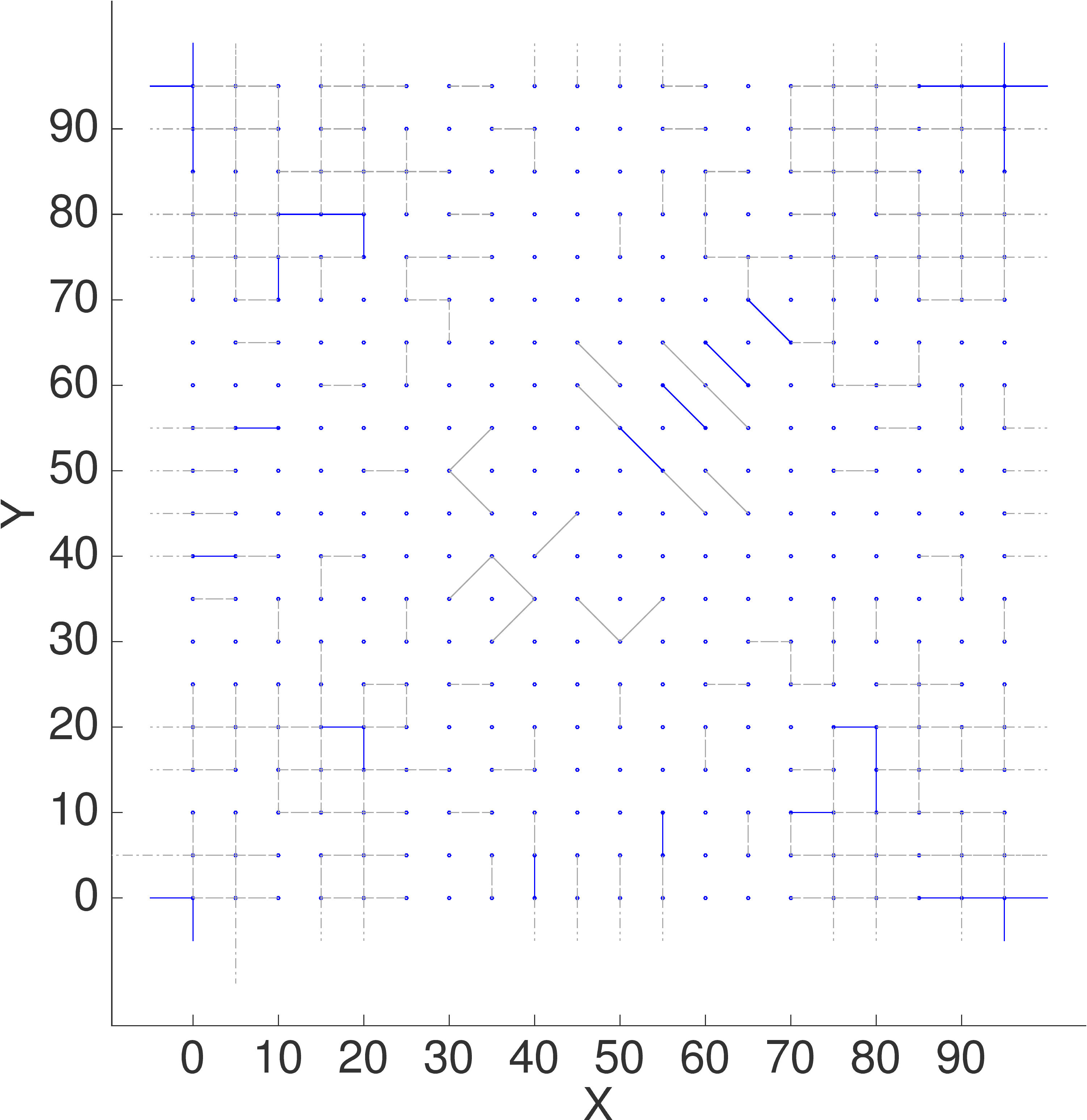}
\hspace*{0.5cm}
\includegraphics[width=4.5cm,angle=0]{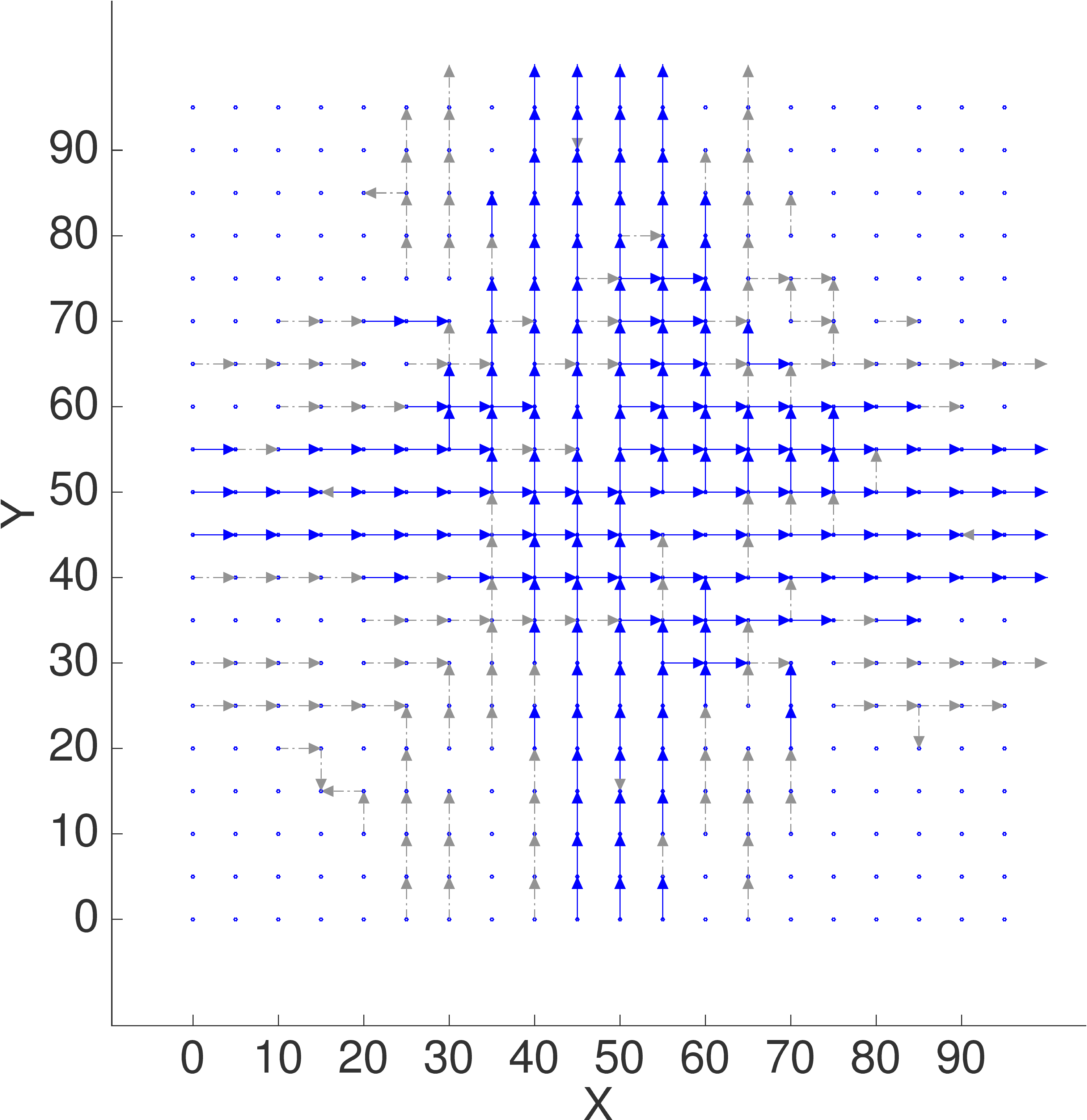}
\hspace*{0.5cm}
\includegraphics[width=4.5cm,angle=0]{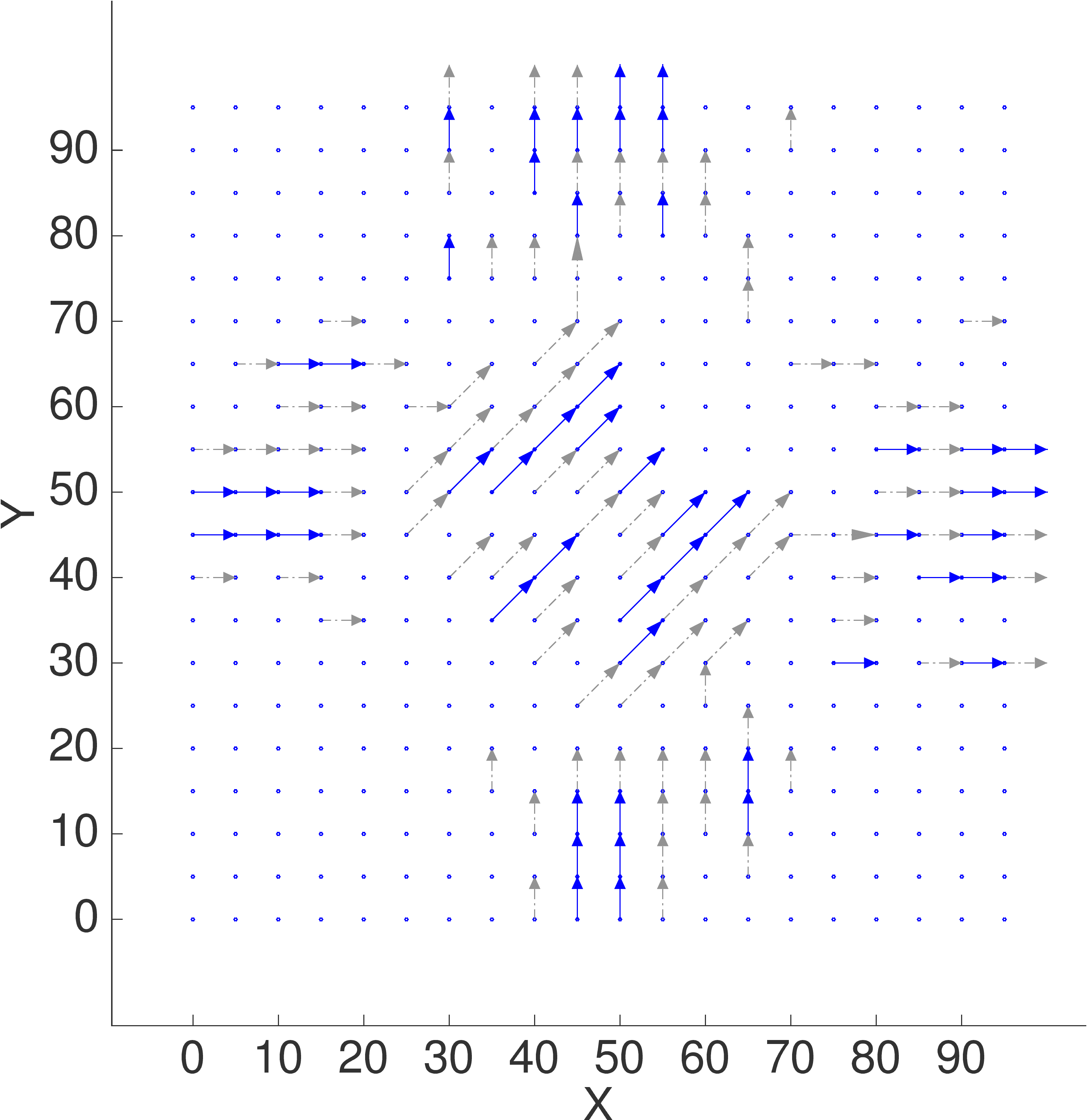}
}
\centerline{$T = 0$  \hspace*{3.5cm} $T = \Delta t$  \hspace*{3.5cm} $T = 2 \Delta t$}
\centerline{(b) Inter edges}
\vspace*{0.3cm}
\centerline{ 
\includegraphics[width=7.9cm,angle=0,clip]{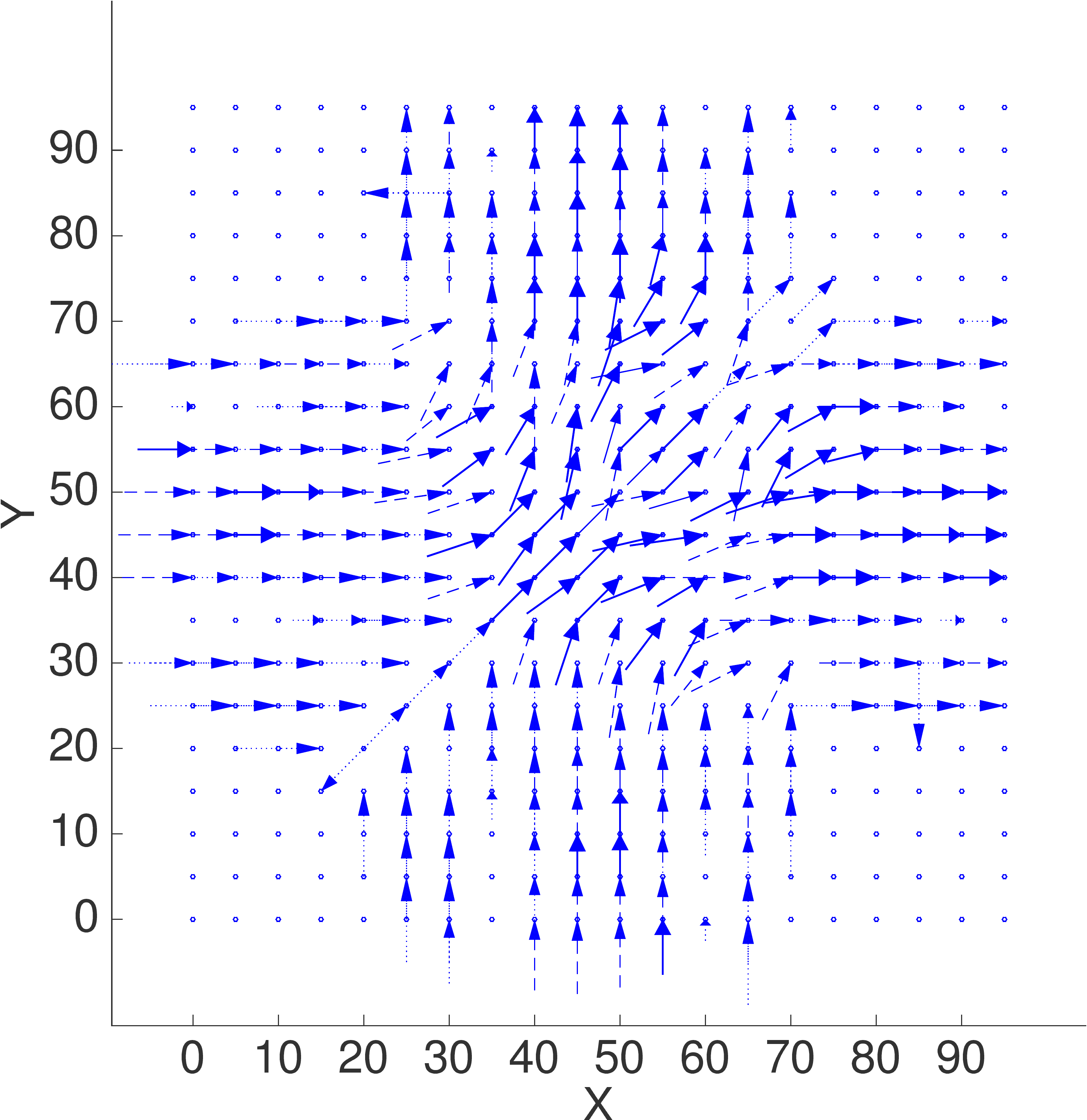}
\hspace*{0.2cm}
\includegraphics[width=7.9cm,angle=0,clip]{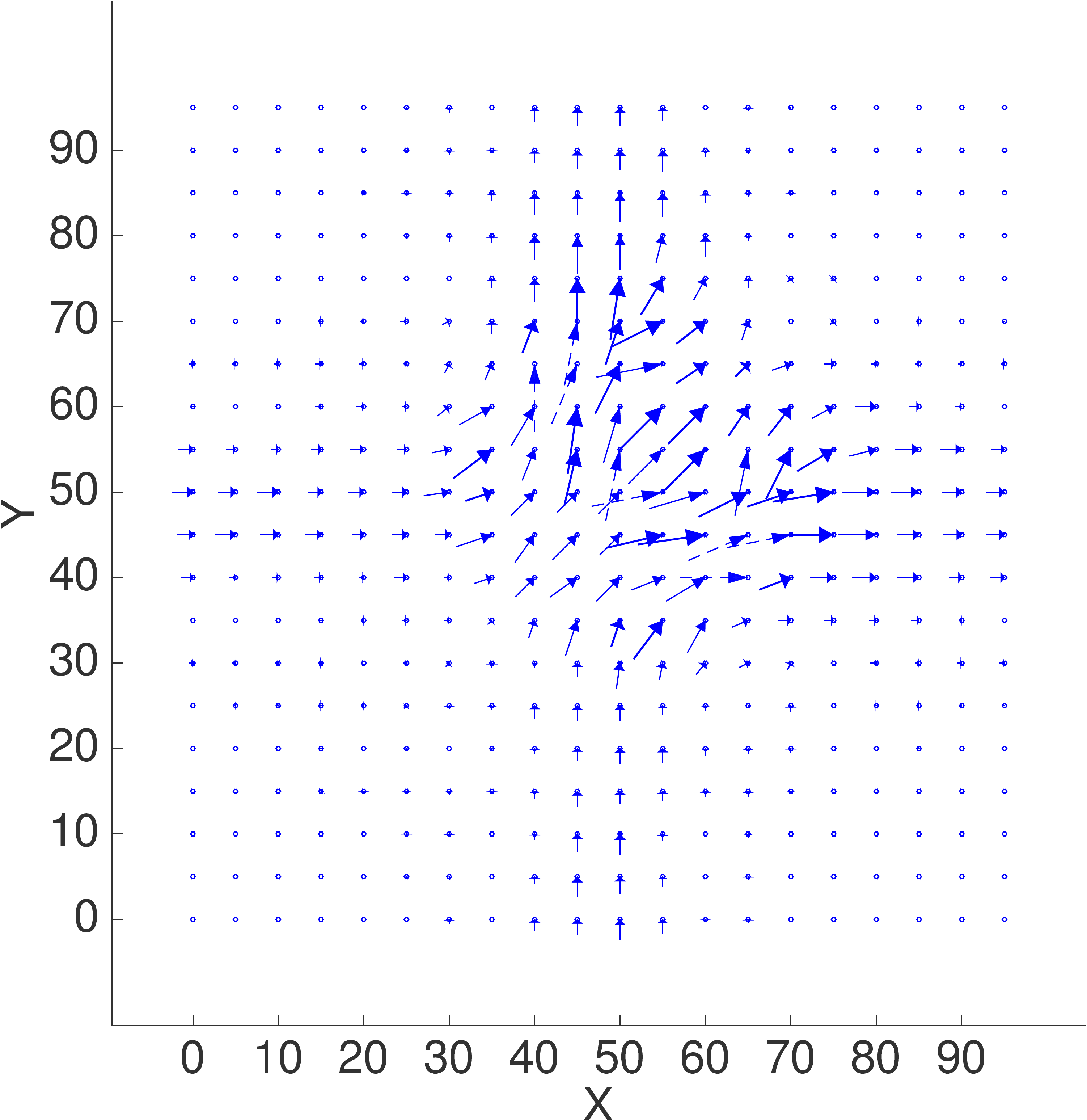}
}
\centerline{(c) Velocity estimated {\it without} intra edges \hspace*{1.0cm} (d) Velocity estimated {\it with} intra edges}
\caption{Results for Scenario 3 with concurrent edges allowed.
\label{scenario_3_with_concurrent_fig}}
\end{figure*}
%
%

\noindent
{\bf Results for Scenario 3 (Fig.\ \ref{scenario_3_with_concurrent_fig}):}\\
(1) Just as in Scenario 1, there are more intra edges than expected for $T=\Delta t$, with the pattern we expected only emerging for $T=3 \Delta t$.\\
(2) The inter edges for $T= \Delta t$ again capture the original advection field, but with some problems in some areas where the velocity is diagonal to the grid.\\
(3) There are again interesting concurrent edges (left-most panel in Fig.\ \ref{scenario_3_with_concurrent_fig}(b)), not only where advection velocity is small, but also near the center, which is exactly where the other inter edges have problems representing some diagonal edges.\\
(4) The difference between Velocity 1 and 2 plots is again significant.  Just as in previous cases, the Velocity 1 plot is more sensitive, i.e.\ catches more edges, but the amplitudes are inflated, while the Velocity 2 plot is better at estimating velocity magnitudes, but misses some of the smaller connections.



A lesson learned:
Velocity 1 plots are indeed better to detect connections that have lower speeds, i.e.\ they have much 
higher sensitivity.  
Velocity 2 plots are better at identifying the fastest connections and estimating actual speeds.  
By definition the directions in both plots are always the same, just the magnitudes change.

\subsection{Results for complex scenarios when concurrent edges are excluded}

Concurrent edges play an odd role, because they are the only edges that are not directed, 
thus provide less useful information about the underlying physical mechanisms,
and cannot be included in the velocity estimates. 
In this section we consider what happens if we exclude the concurrent edges a priori, 
i.e.\ if we run the causal discovery algorithm with the prior knowledge that edges
within the same time slice are forbidden (see Appendix A for details on how to do that).
How do the results change?
Does a priori excluding concurrent edges result in 
significant new inter edges for $T>0$ and, if so, what do those new edges look like?

\begin{figure*}
\centerline{ 
\includegraphics[width=7.9cm,angle=0,clip]{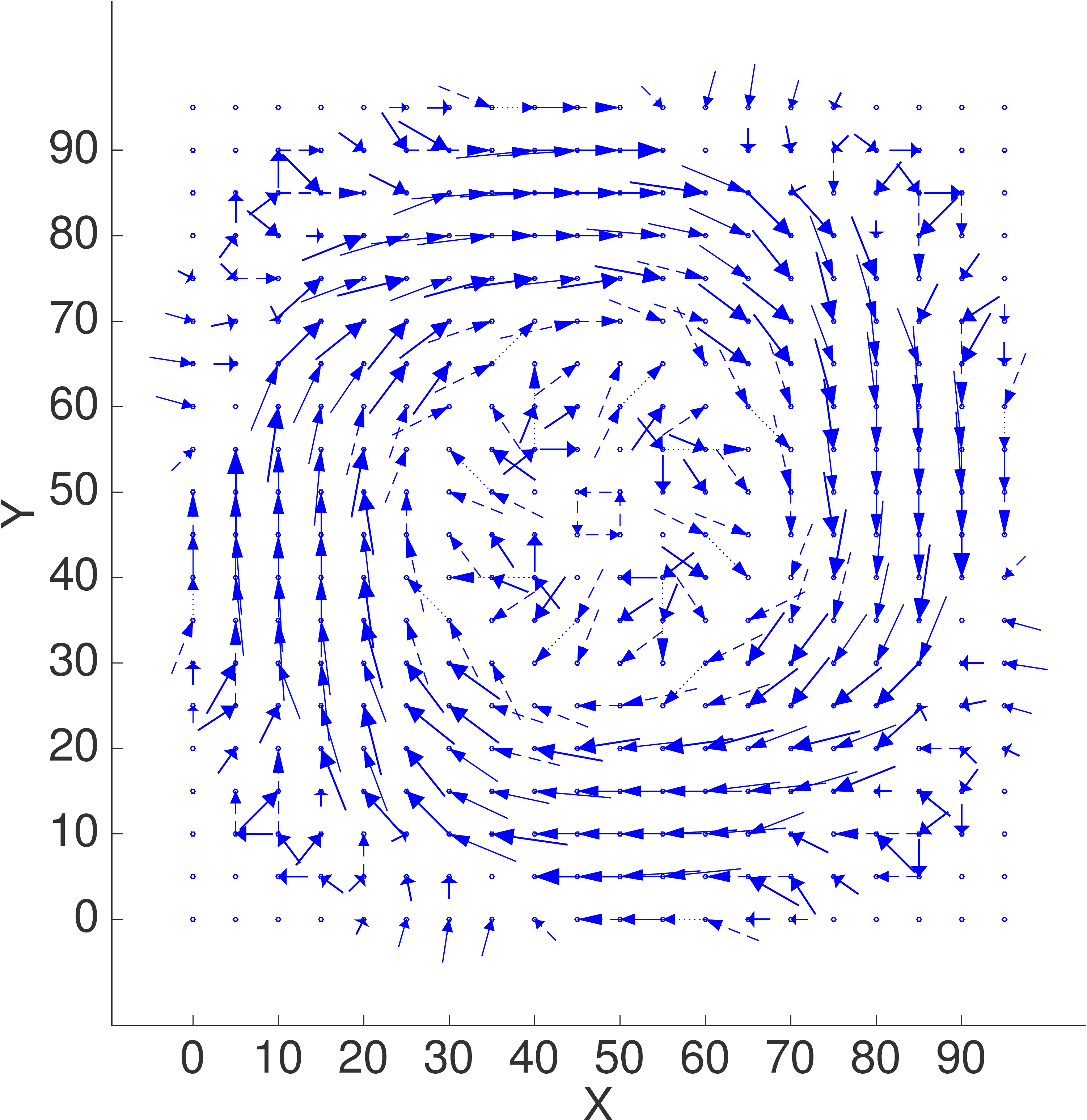}
\hspace*{0.2cm}
\includegraphics[width=7.9cm,angle=0,clip]{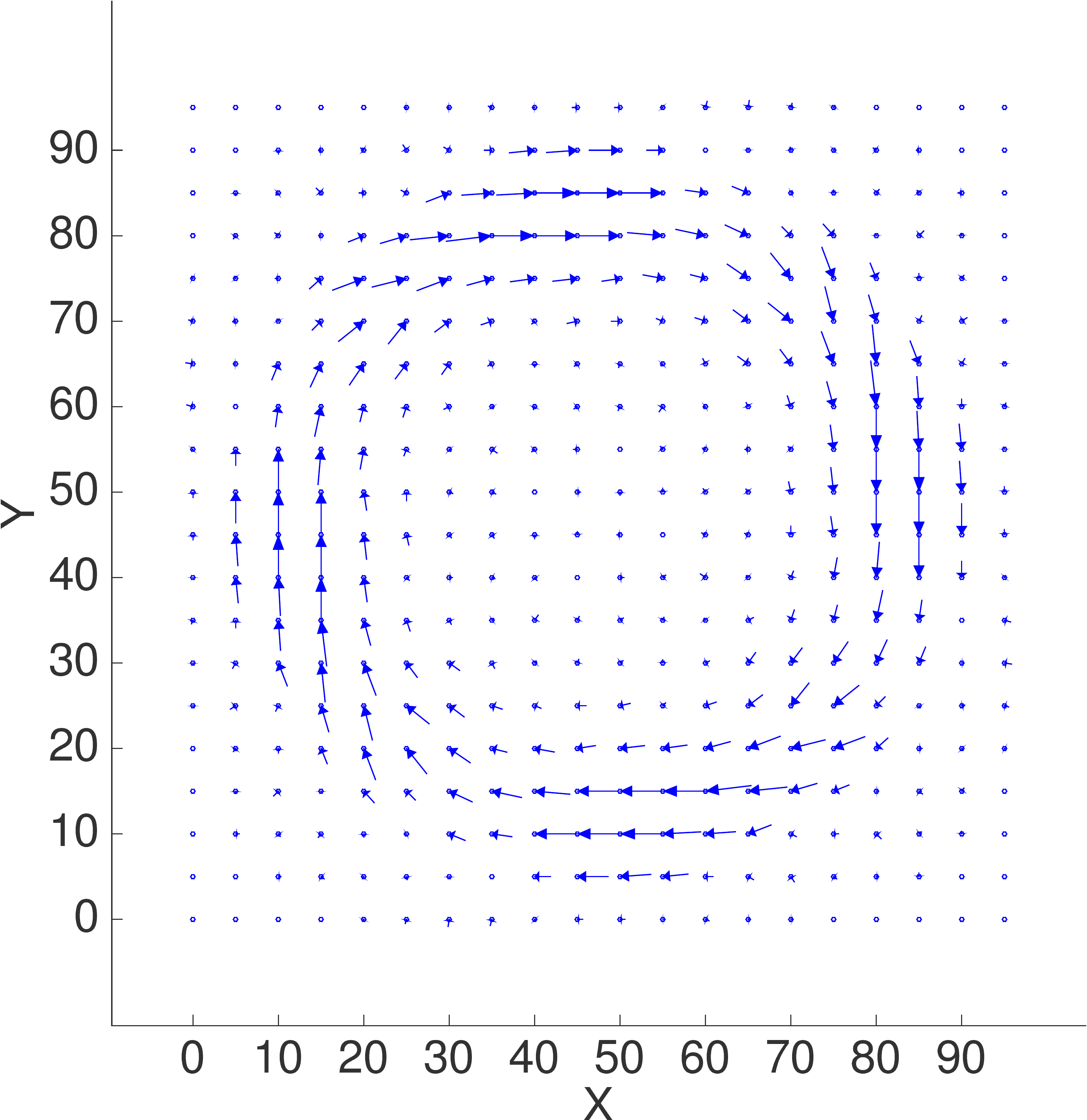}
}
\centerline{(a) Velocity estimated {\it without} intra edges \hspace*{1.0cm} (b) Velocity estimated {\it with} intra edges}
\caption{Results for Scenario 1 when concurrent edges are a priori forbidden.
\label{scenario_1_no_concurrent_fig}}
\end{figure*}

\begin{figure*}
\centerline{ 
\includegraphics[width=7.9cm,angle=0,clip]{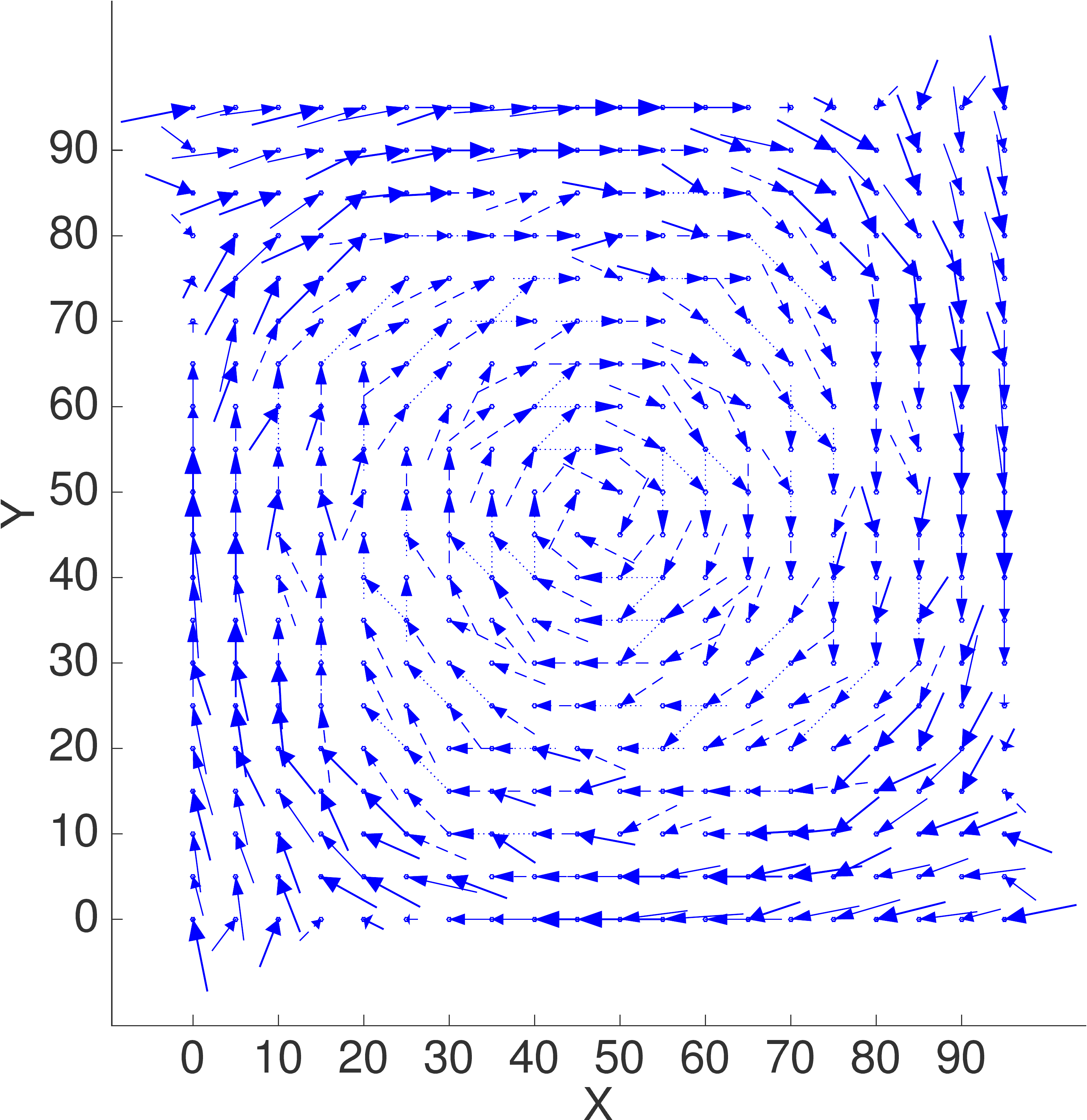}
\hspace*{0.2cm}
\includegraphics[width=7.9cm,angle=0,clip]{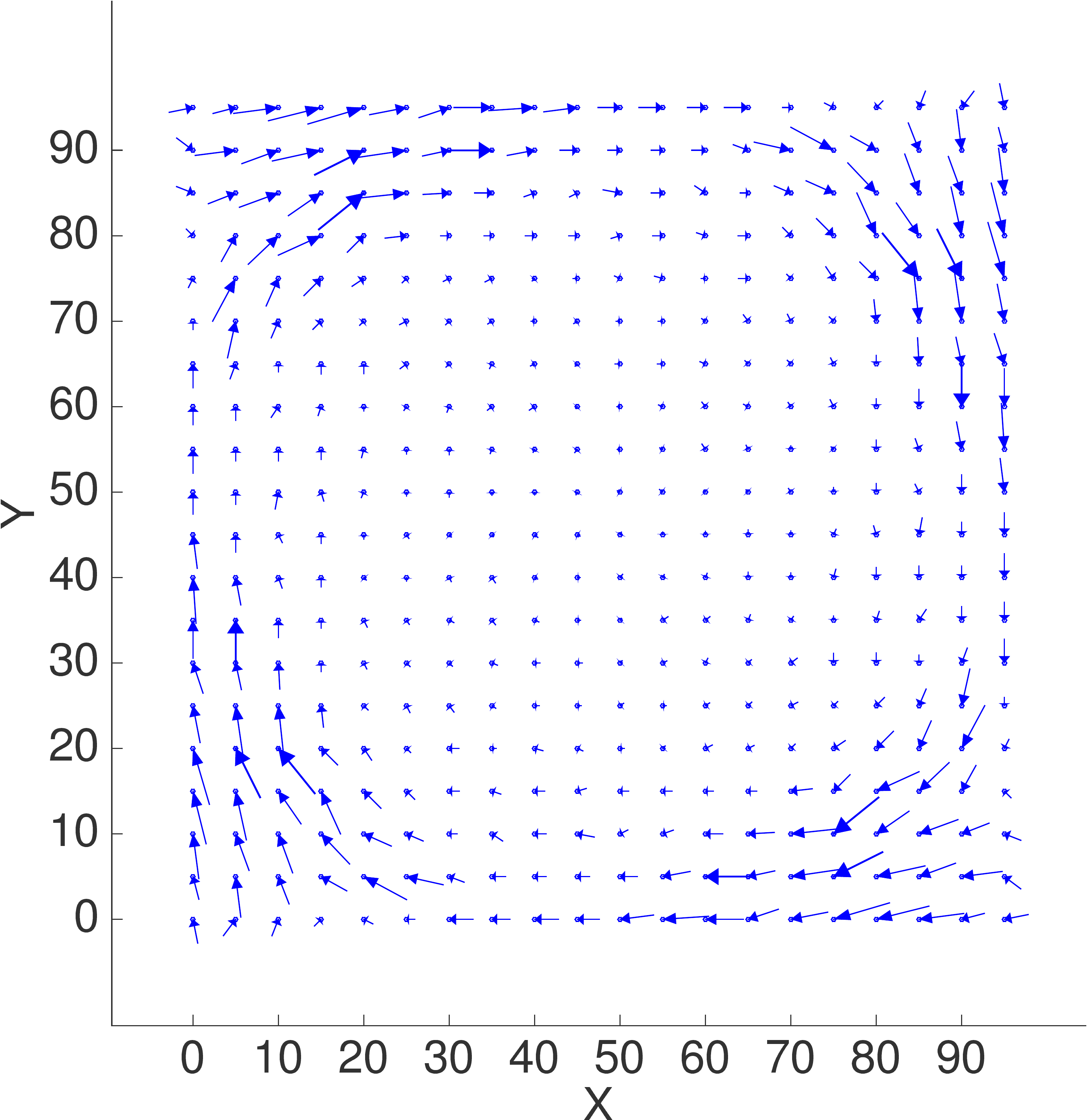}
}
\centerline{(a) Velocity estimated {\it without} intra edges \hspace*{1.0cm} (b) Velocity estimated {\it with} intra edges}
\caption{Results for Scenario 1 when concurrent edges are a priori forbidden
\label{scenario_2_no_concurrent_fig}}
\end{figure*}

\begin{figure*}
\centerline{ 
\includegraphics[width=7.9cm,angle=0,clip]{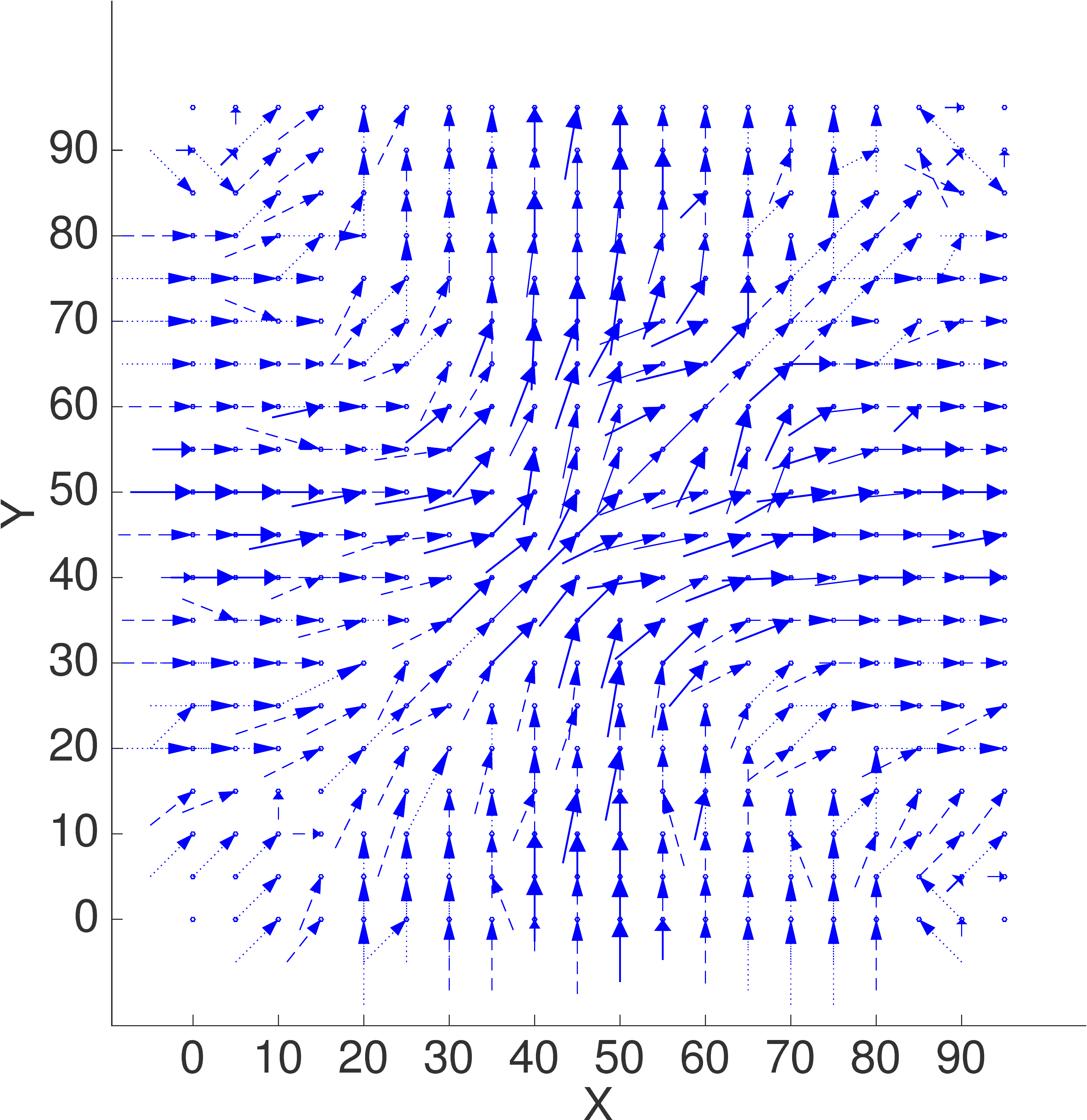}
\hspace*{0.2cm}
\includegraphics[width=7.9cm,angle=0,clip]{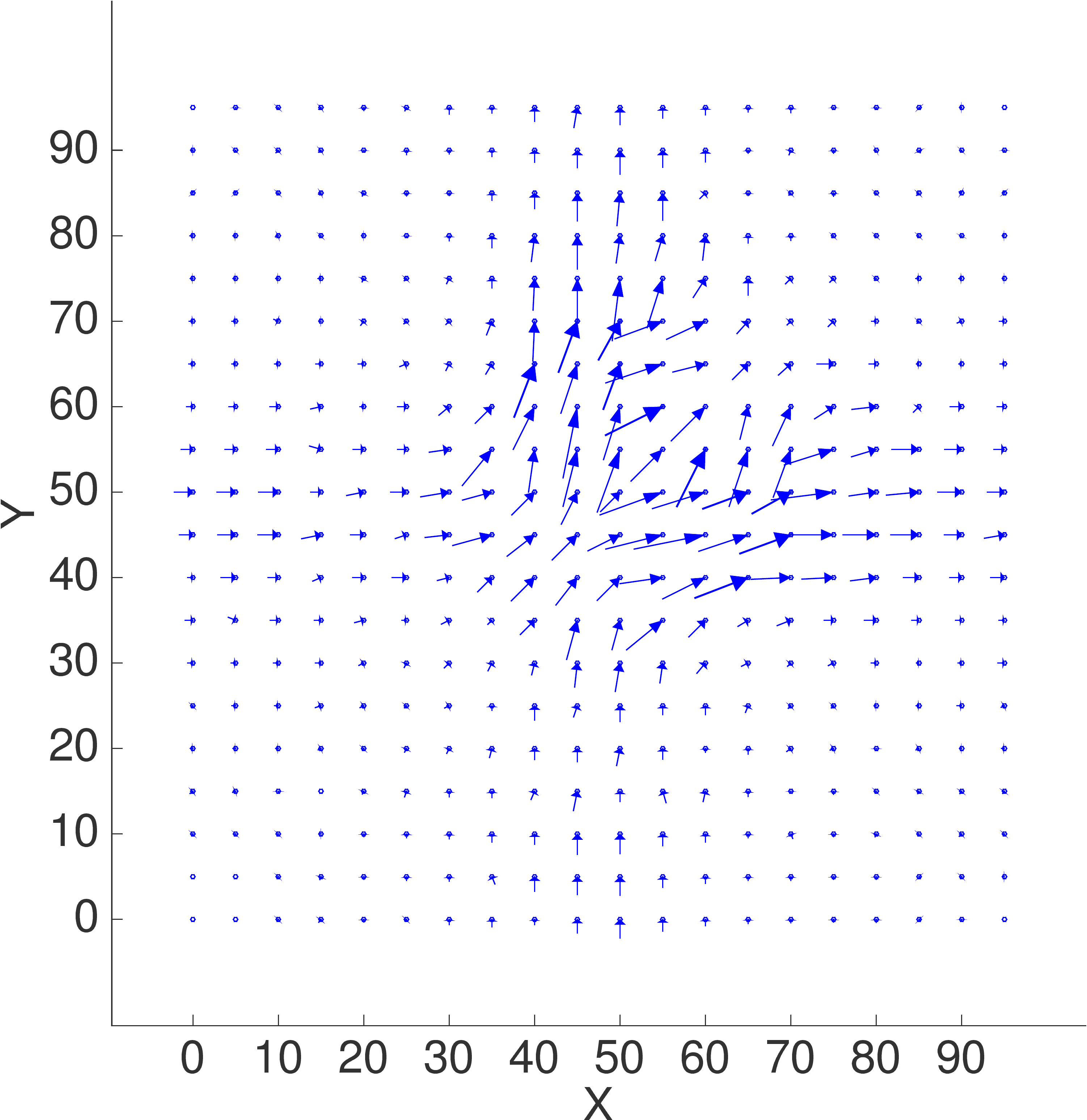}
}
\centerline{(a) Velocity estimated {\it without} intra edges \hspace*{1.0cm} (b) Velocity estimated {\it with} intra edges}

\caption{Results for Scenario 3 when concurrent edges are a priori forbidden.
\label{scenario_3_no_concurrent_fig}}
\end{figure*}

For brevity, here we only show results for the velocity estimates.
Figures \ref{scenario_1_no_concurrent_fig} to \ref{scenario_3_no_concurrent_fig}
show the velocity estimates for Scenarios 1-3 when concurrent edges are a priori excluded. 
These can be directly compared to Figs.\ \ref{scenario_1_with_concurrent_fig} - \ref{scenario_3_with_concurrent_fig}
where concurrent edges are allowed.
For Type 1 velocity estimates we see that new edges appear primarily in areas 
where previously concurrent edges were present.  
For Scenarios 2 and 3, excluding the concurrent edges a priori seems to result in an improvement, 
since it increases the sensitivity of the Type 1 plots, 
i.e.\ we now pick up connections corresponding to even lower advection velocities.  
For Scenario 1, however, it seems that the sensitivity is {\it too high}, 
namely creating connections where there should not be any, 
because the original advection velocities are truly zero.  
Scenario 1 is the only one with advection velocities truly zero anywhere, so this is 
likely the reason we see this negative effect only there.
In contrast, for Type 2 velocity estimates the results when concurrent edges are excluded are almost indistinguishable from those when concurrent edges were included.  This indicates that the concurrent edges were often weak and only occurred in areas where intra edges are dominant, thus once we take intra edges into account, there is no significant difference between the corresponding results.


\subsection{Results for complex scenarios when signal speed is increased}


%
\begin{figure*}
\centerline{ 
\includegraphics[width=4.5cm,angle=0]{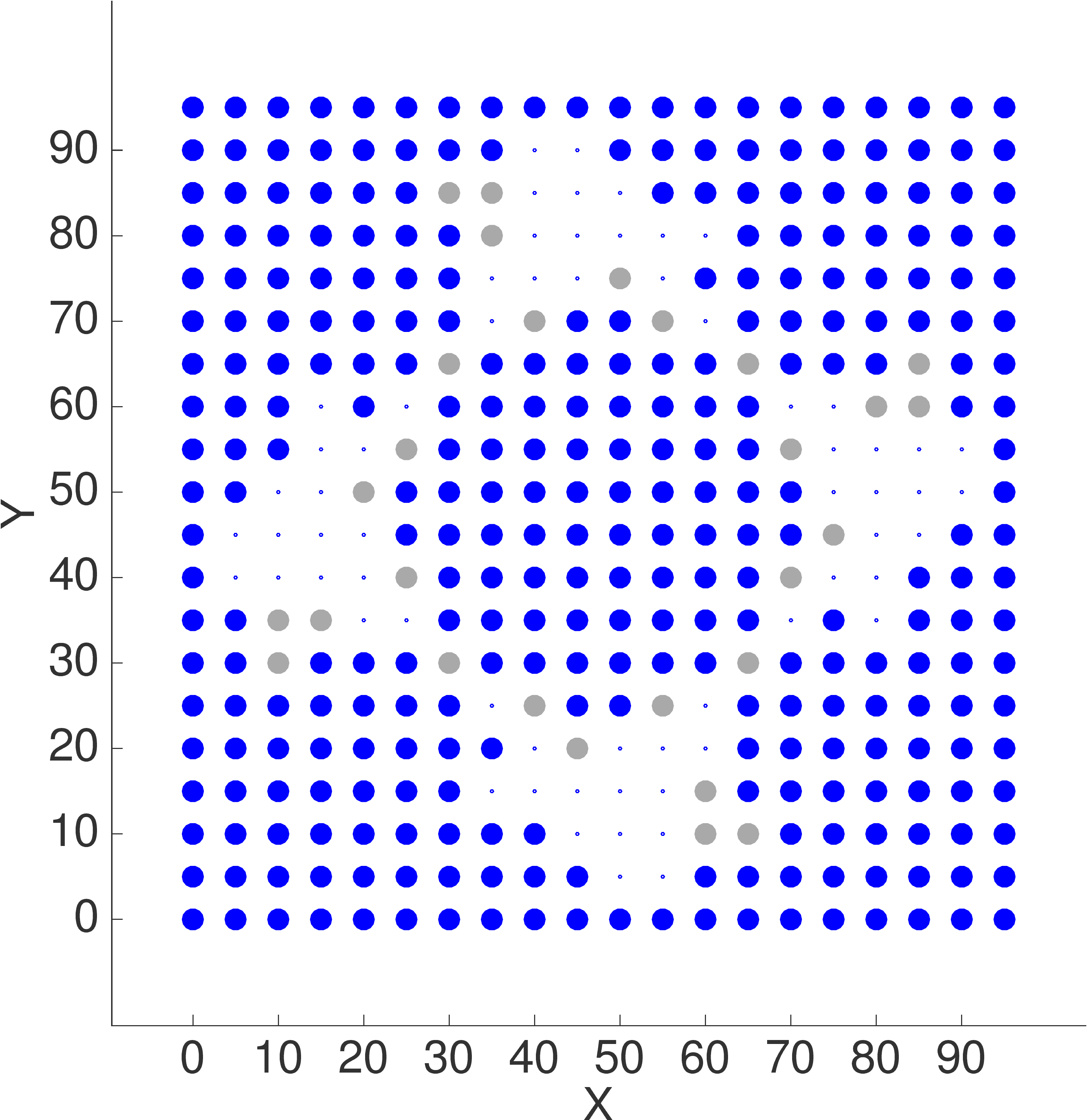}
\hspace*{0.5cm}
\includegraphics[width=4.5cm,angle=0]{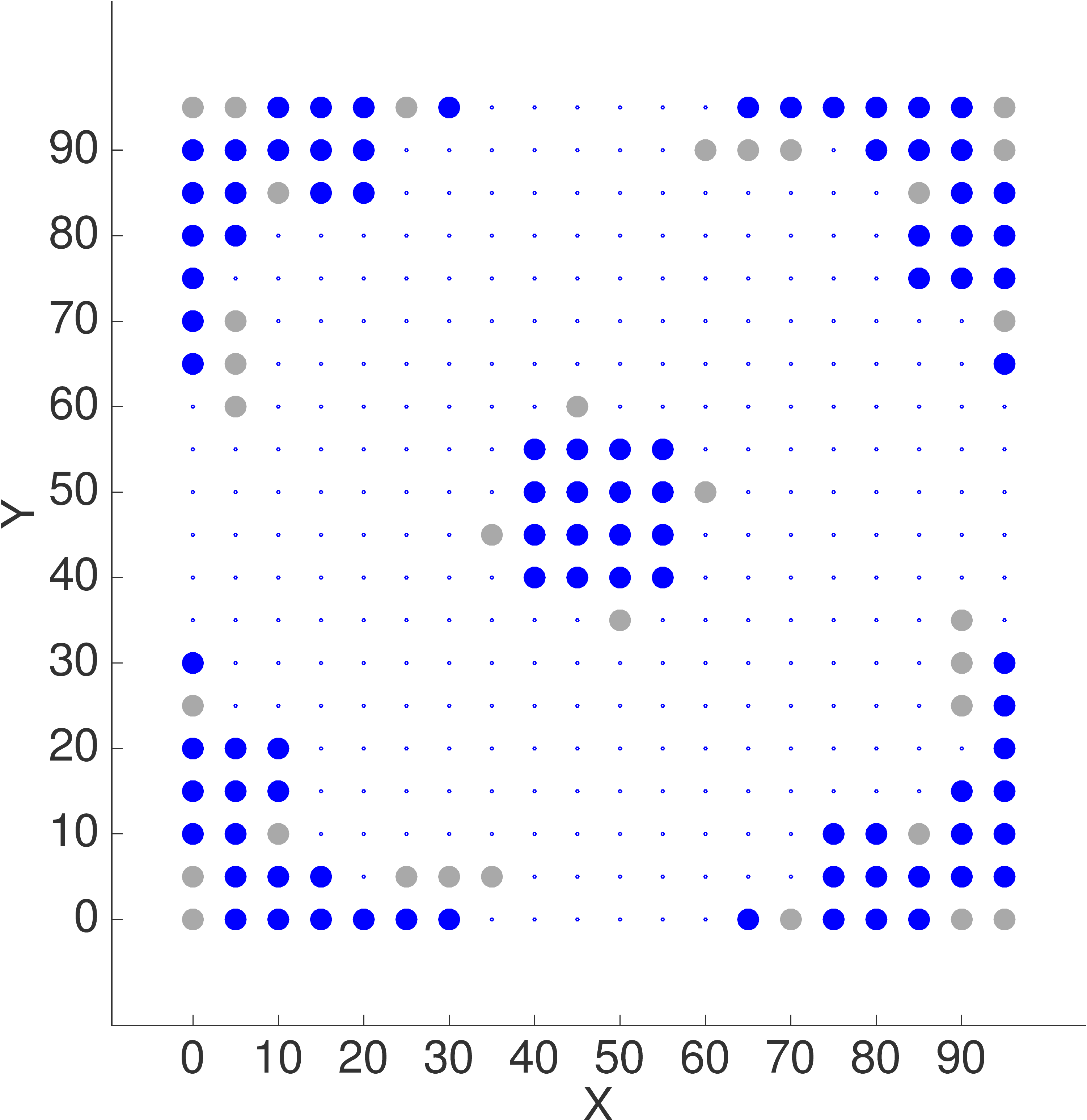}
\hspace*{0.5cm}
\includegraphics[width=4.5cm,angle=0]{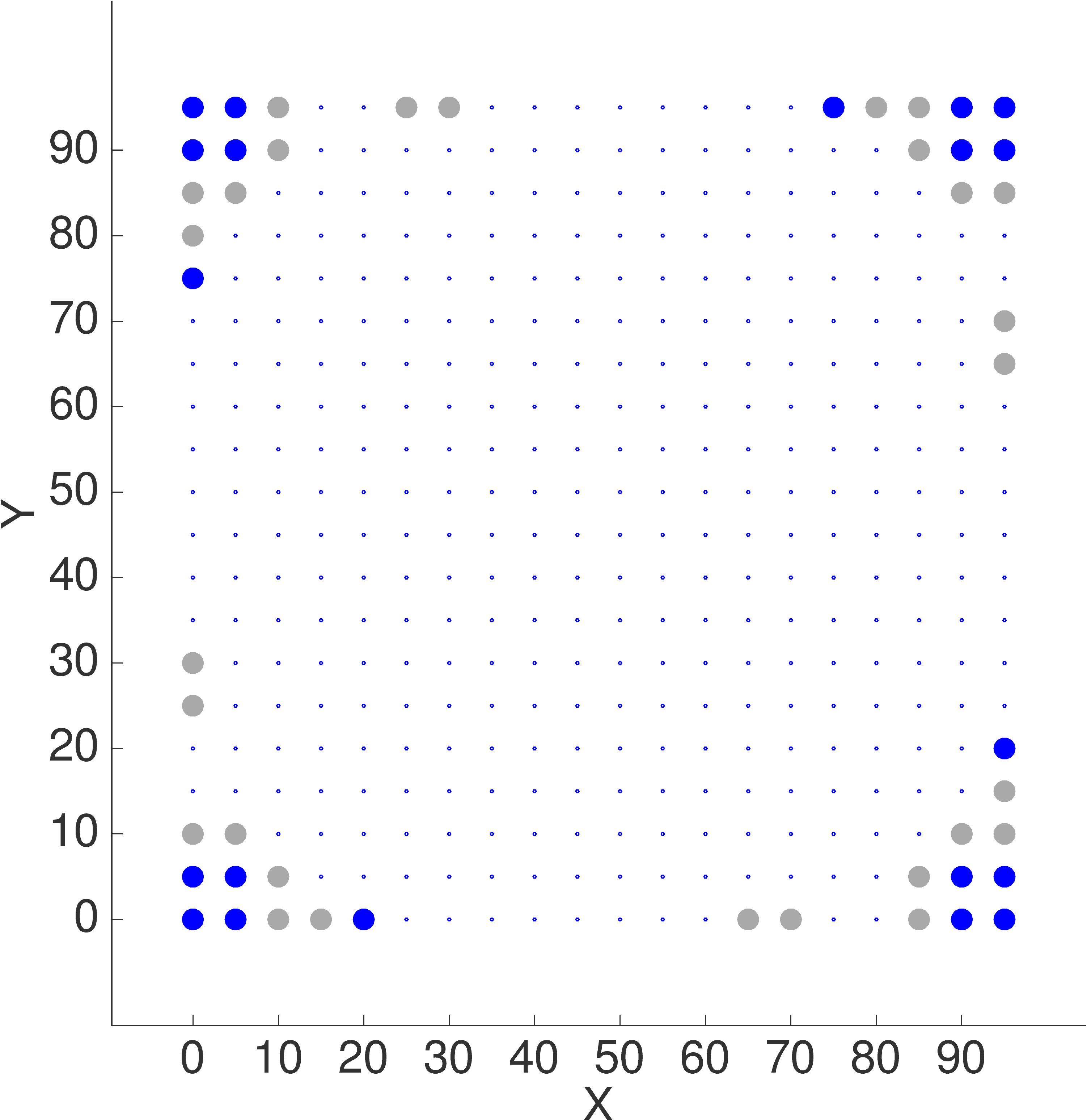}
}
\centerline{$T = 4 \Delta t$  \hspace*{3.5cm} $T = 8 \Delta t$  \hspace*{3.5cm} $T = 12 \Delta t$ }
\centerline{(a) Intra edges}
\vspace*{0.3cm}
\centerline{ 
\includegraphics[width=4.5cm,angle=0]{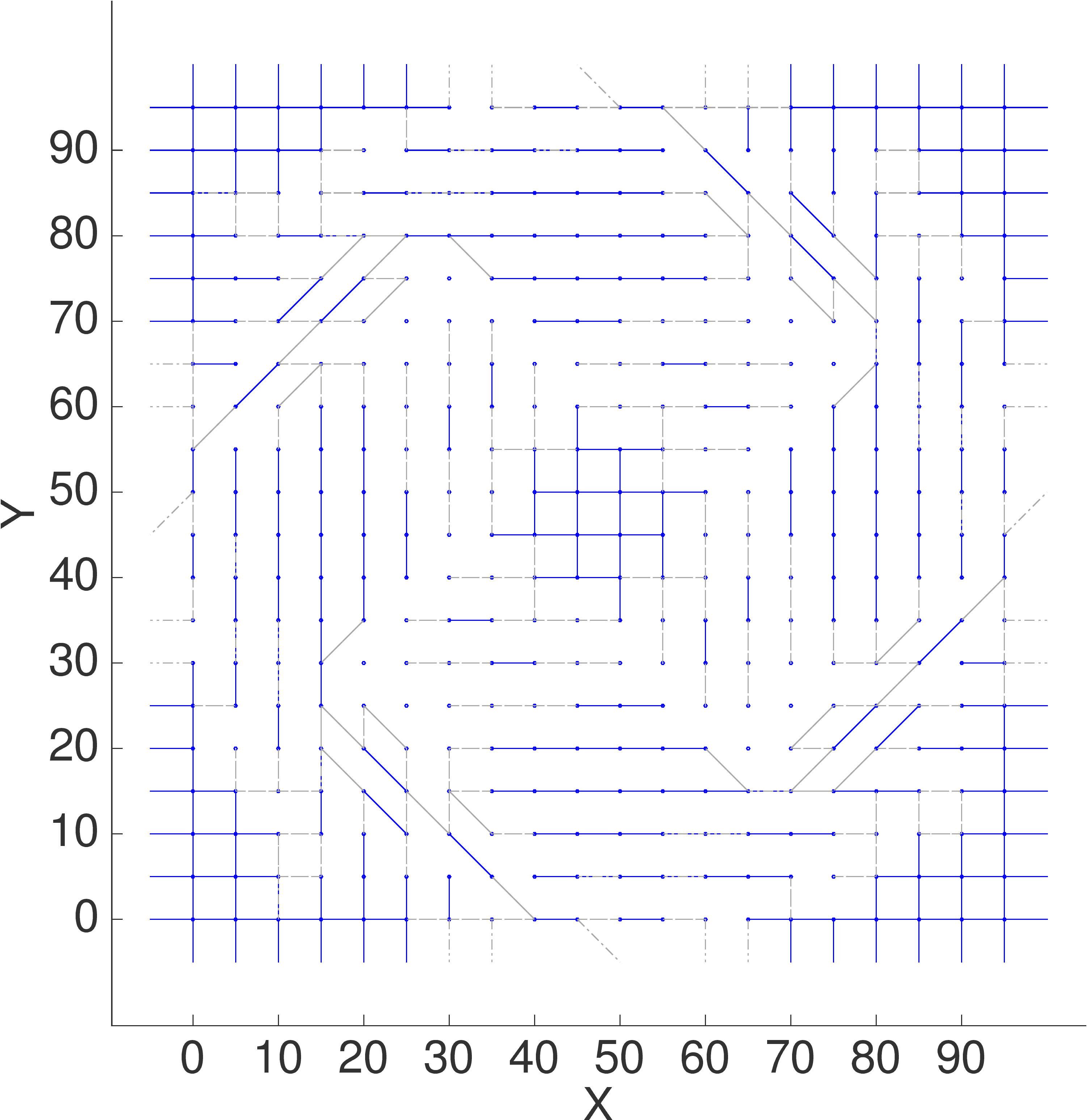}
\hspace*{0.5cm}
\includegraphics[width=4.5cm,angle=0]{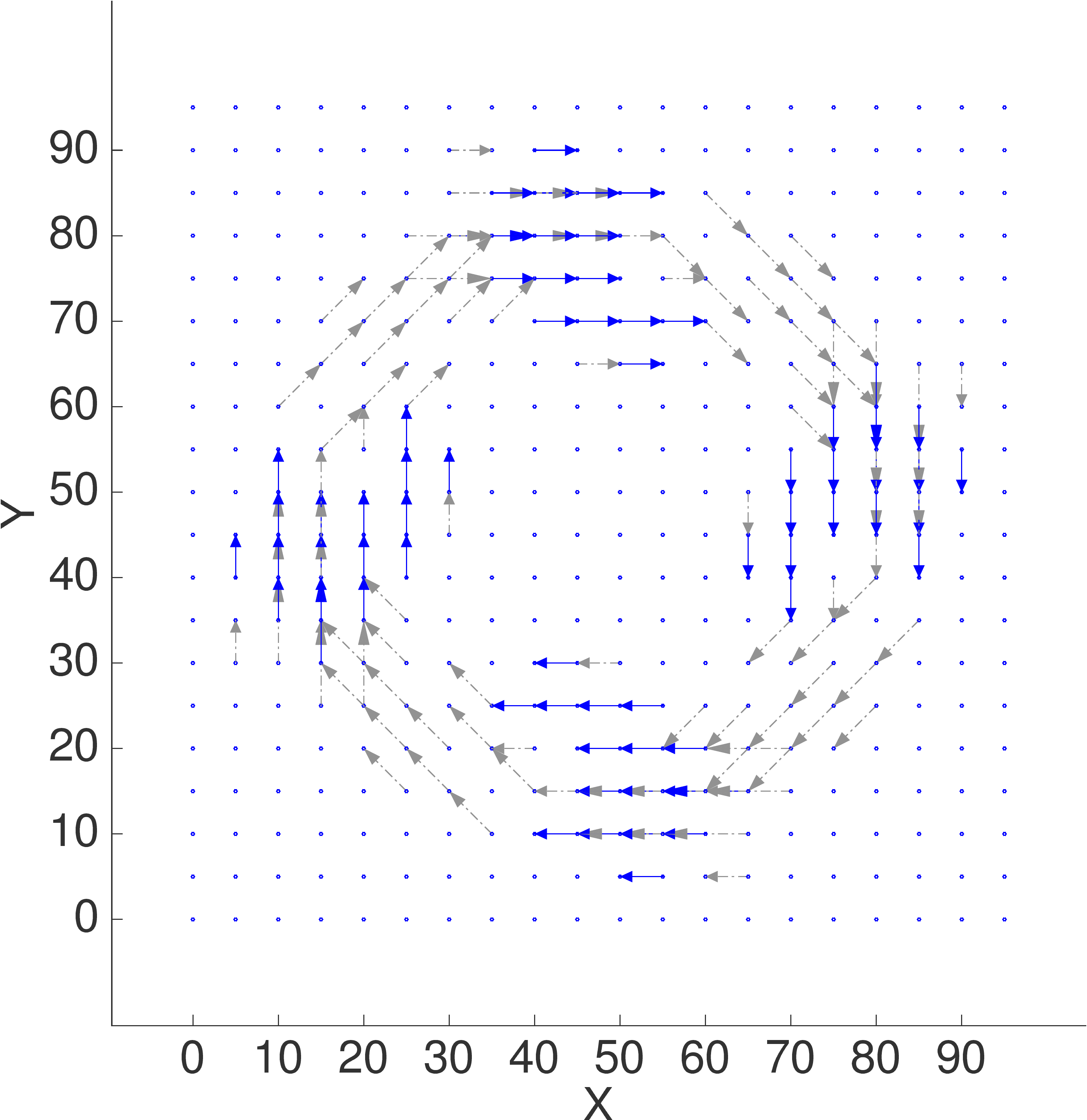}
\hspace*{0.5cm}
\includegraphics[width=4.5cm,angle=0]{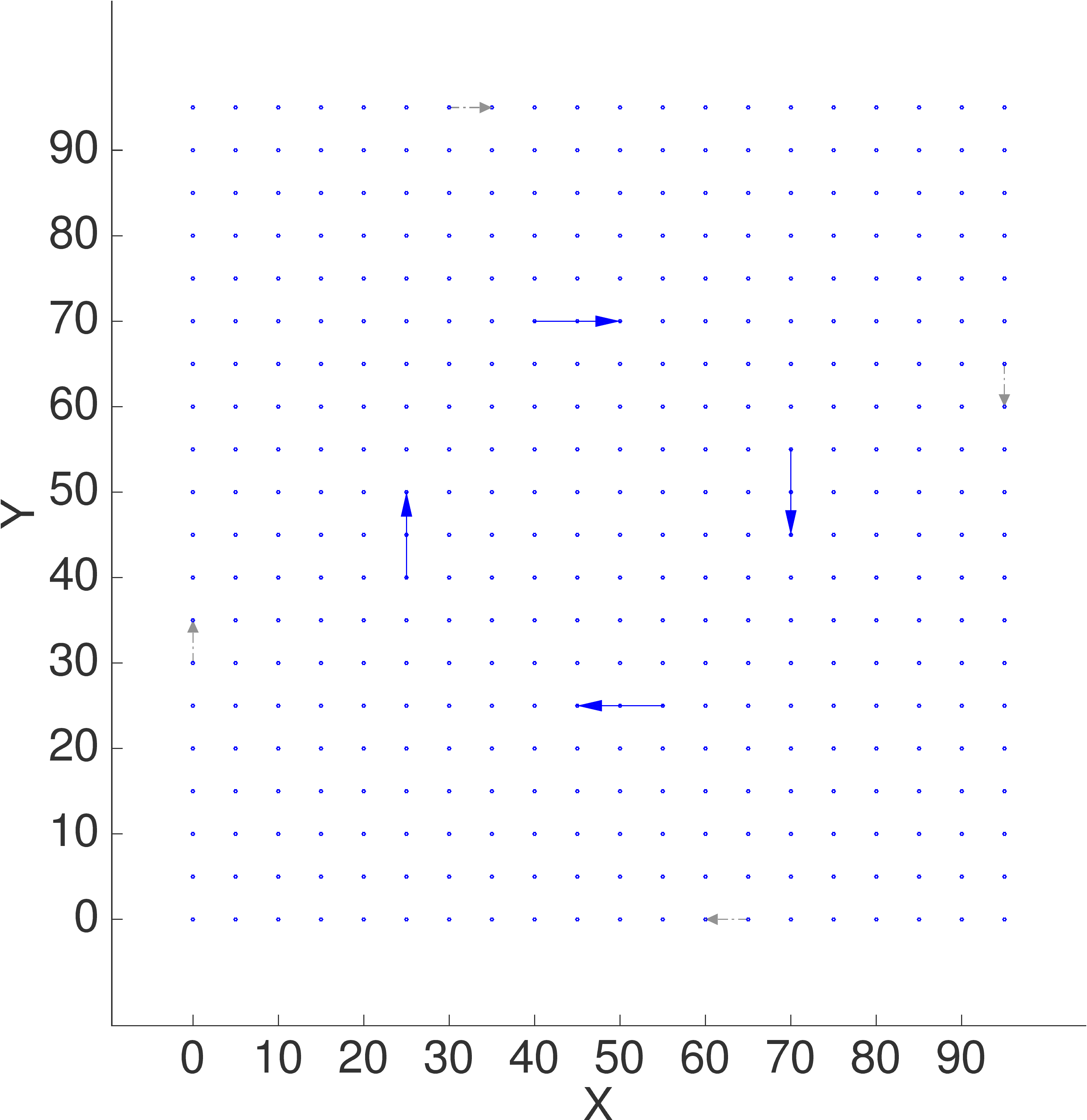}
}
\centerline{$T = 0$  \hspace*{3.5cm} $T = 4 \Delta t$  \hspace*{3.5cm} $T = 8 \Delta t$}
\centerline{(b) Inter edges}
\vspace*{0.3cm}
\centerline{ 
\includegraphics[width=7.9cm,angle=0,clip]{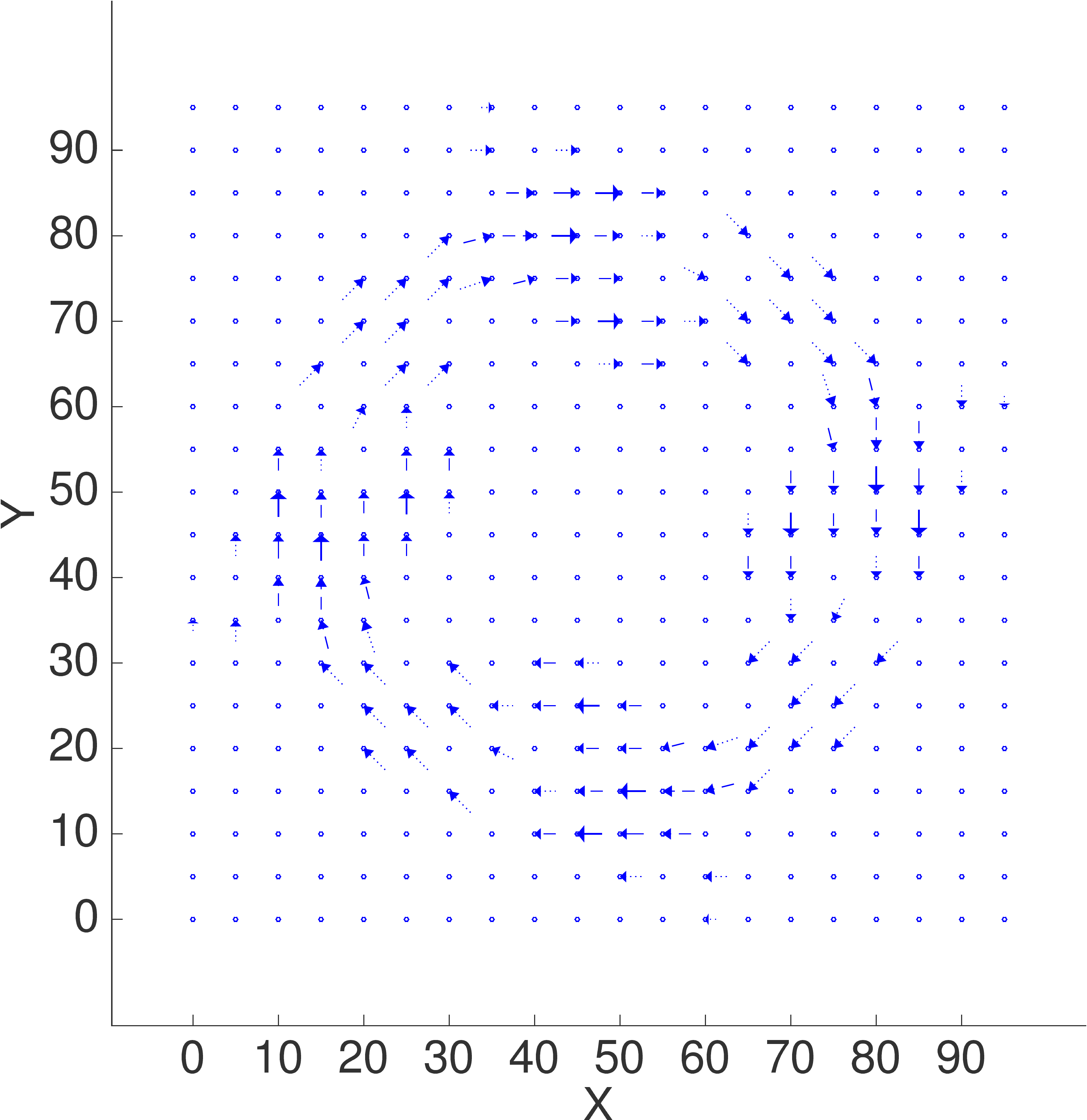}
\hspace*{0.2cm}
\includegraphics[width=7.9cm,angle=0,clip]{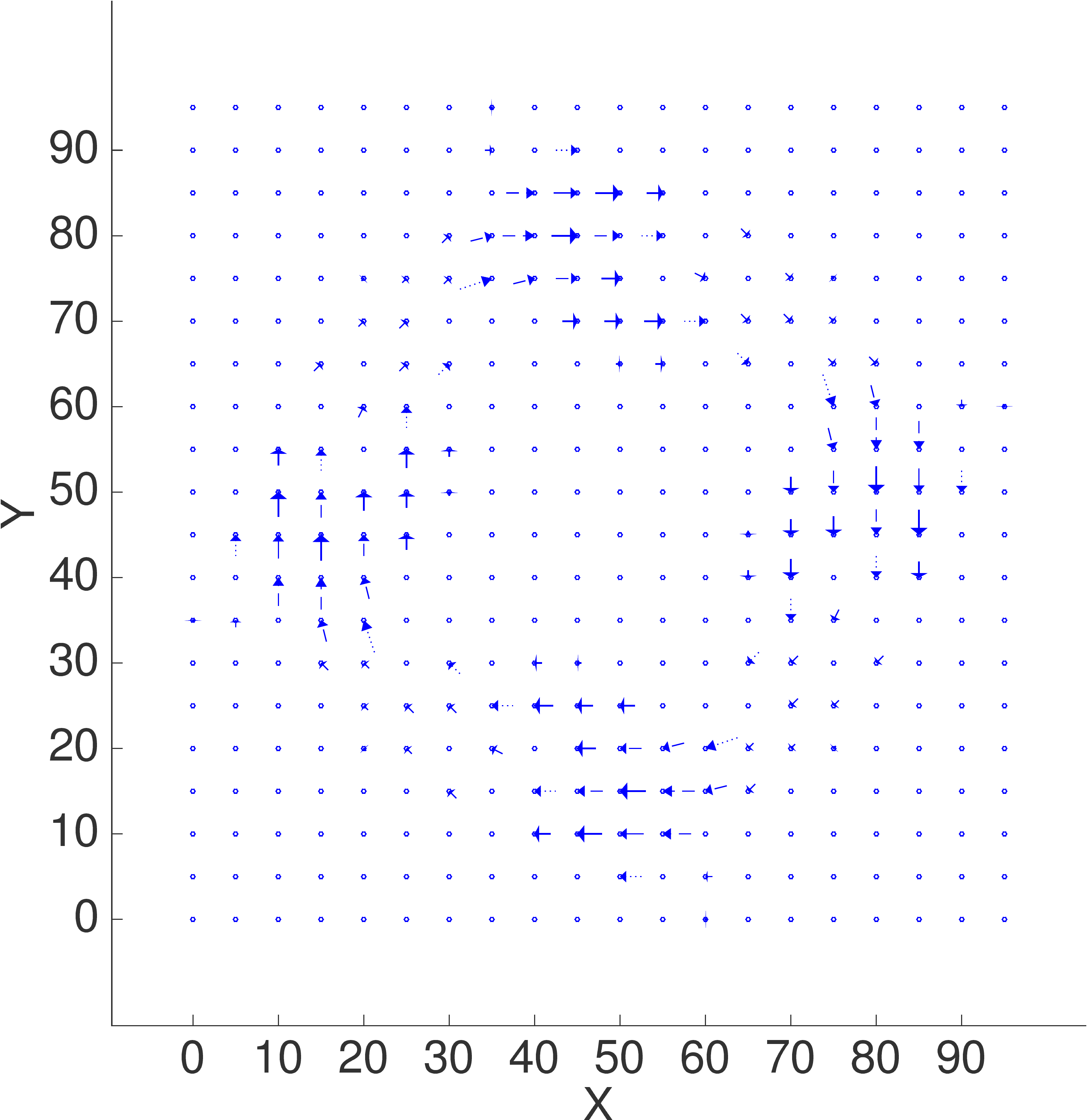}
}
\centerline{(c) Velocity estimated {\it without} intra edges \hspace*{1.0cm} (d) Velocity estimated {\it with} intra edges}
\caption{Results for Scenario 1 with concurrent edges allowed and $M=4$.
\label{scenario_2_M_4_with_concurrent_fig}}
\end{figure*}
%

%
\begin{figure*}
\centerline{ 
\includegraphics[width=4.5cm,angle=0]{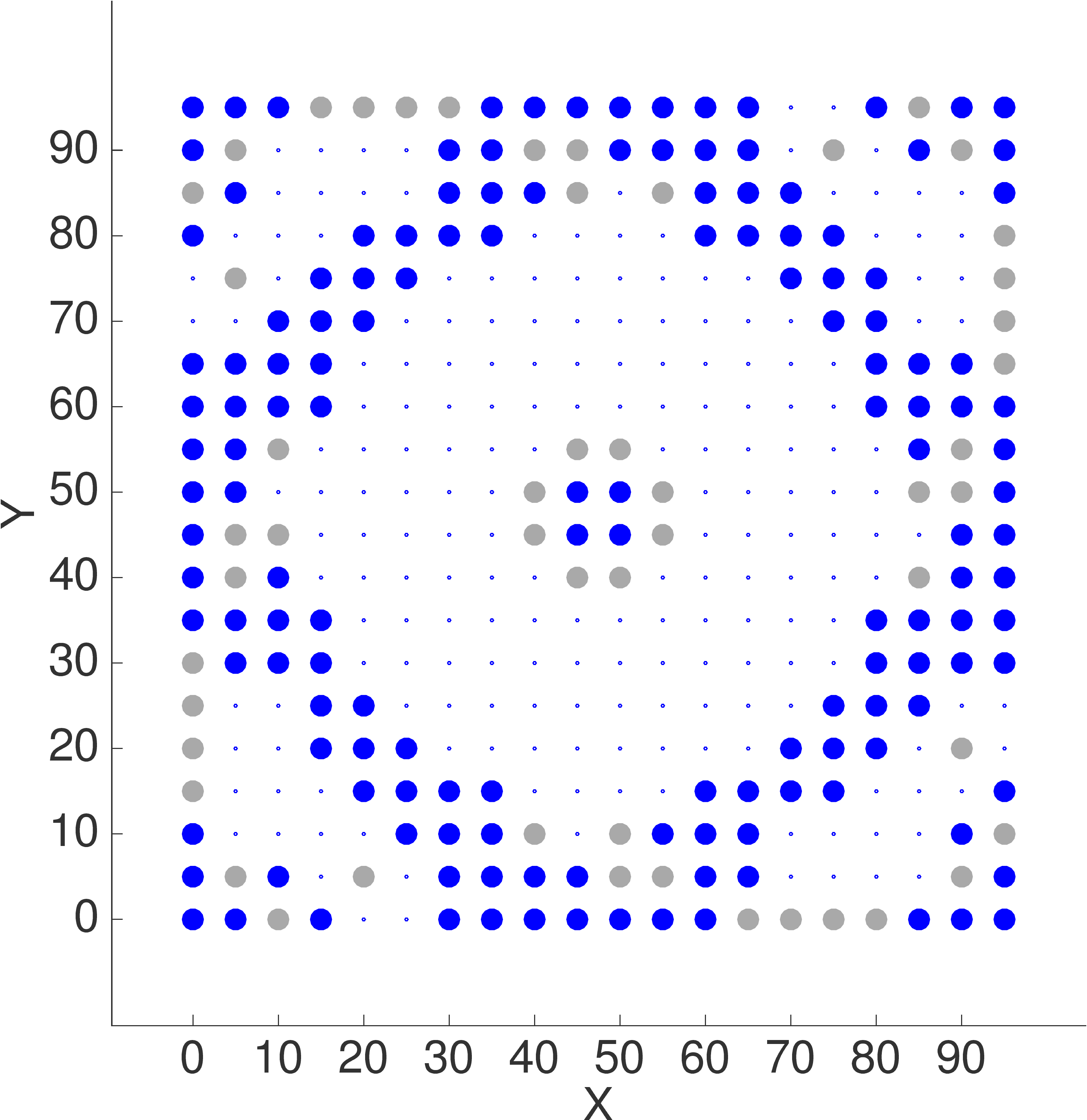}
\hspace*{0.5cm}
\includegraphics[width=4.5cm,angle=0]{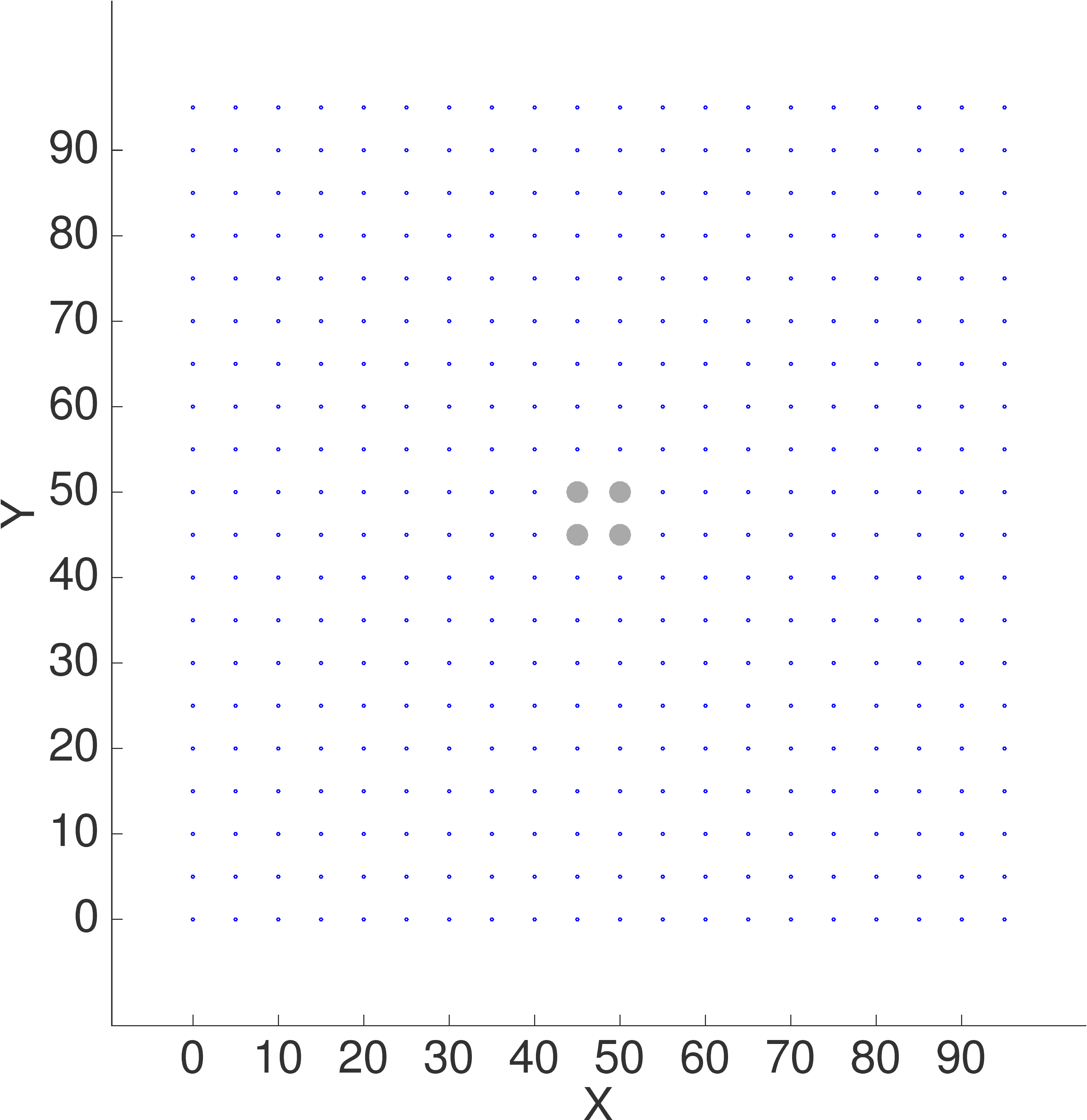}
\hspace*{0.5cm}
\includegraphics[width=4.5cm,angle=0]{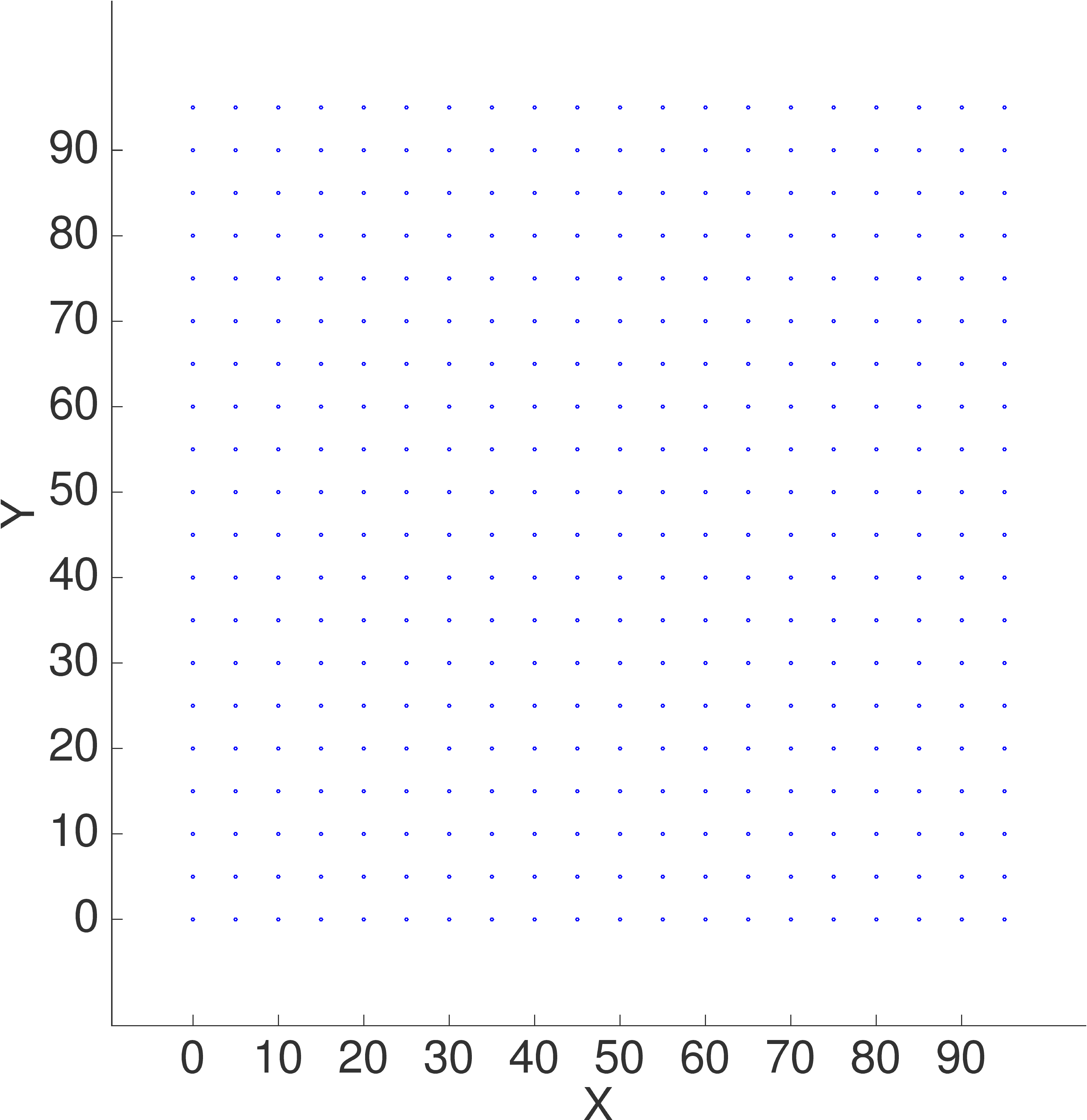}
}
\centerline{$T = 10 \Delta t$  \hspace*{3.5cm} $T = 20 \Delta t$  \hspace*{3.5cm} $T = 30 \Delta t$ }
\centerline{(a) Intra edges}
\vspace*{0.3cm}
\centerline{ 
\includegraphics[width=4.5cm,angle=0]{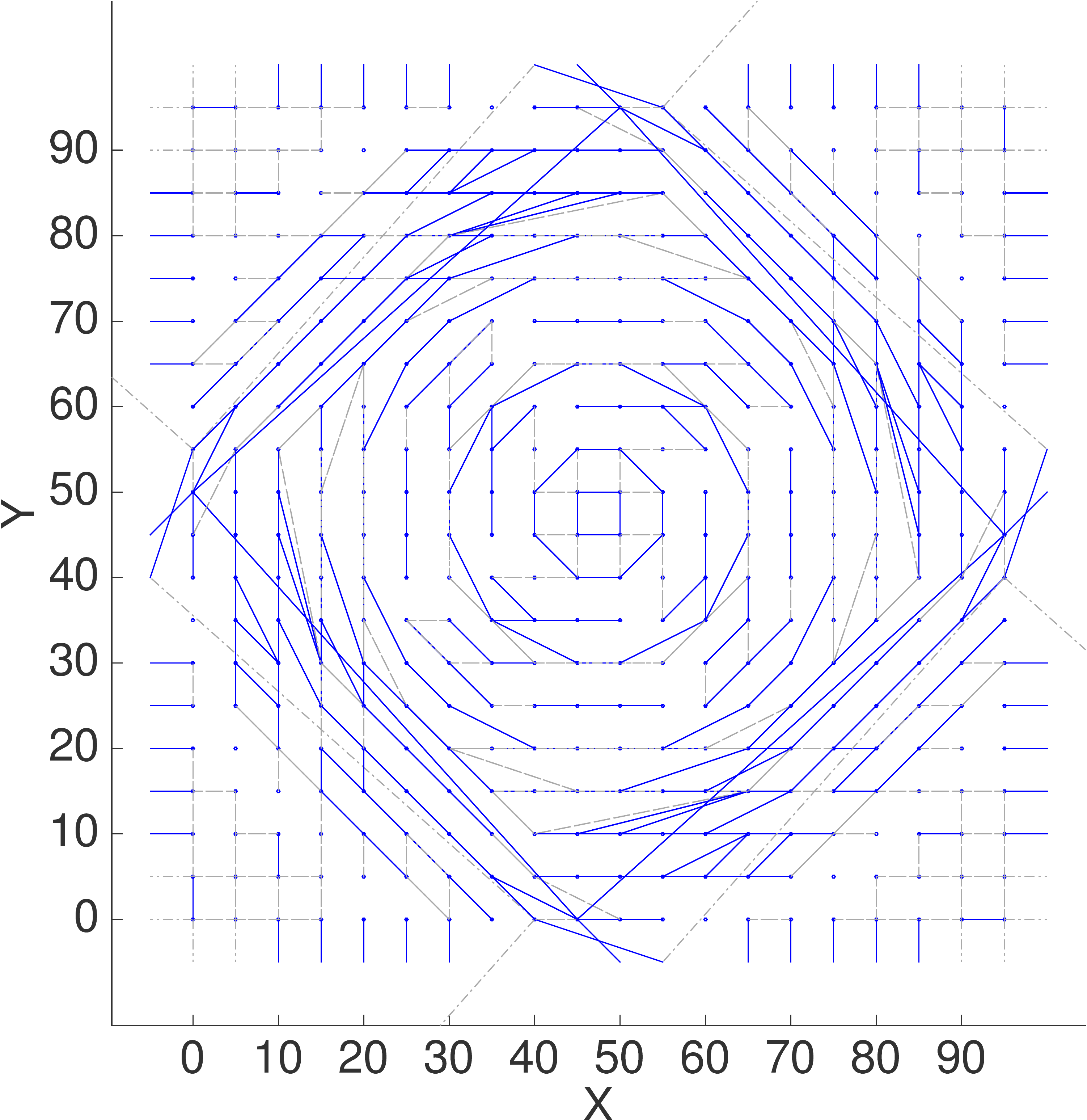}
\hspace*{0.5cm}
\includegraphics[width=4.5cm,angle=0]{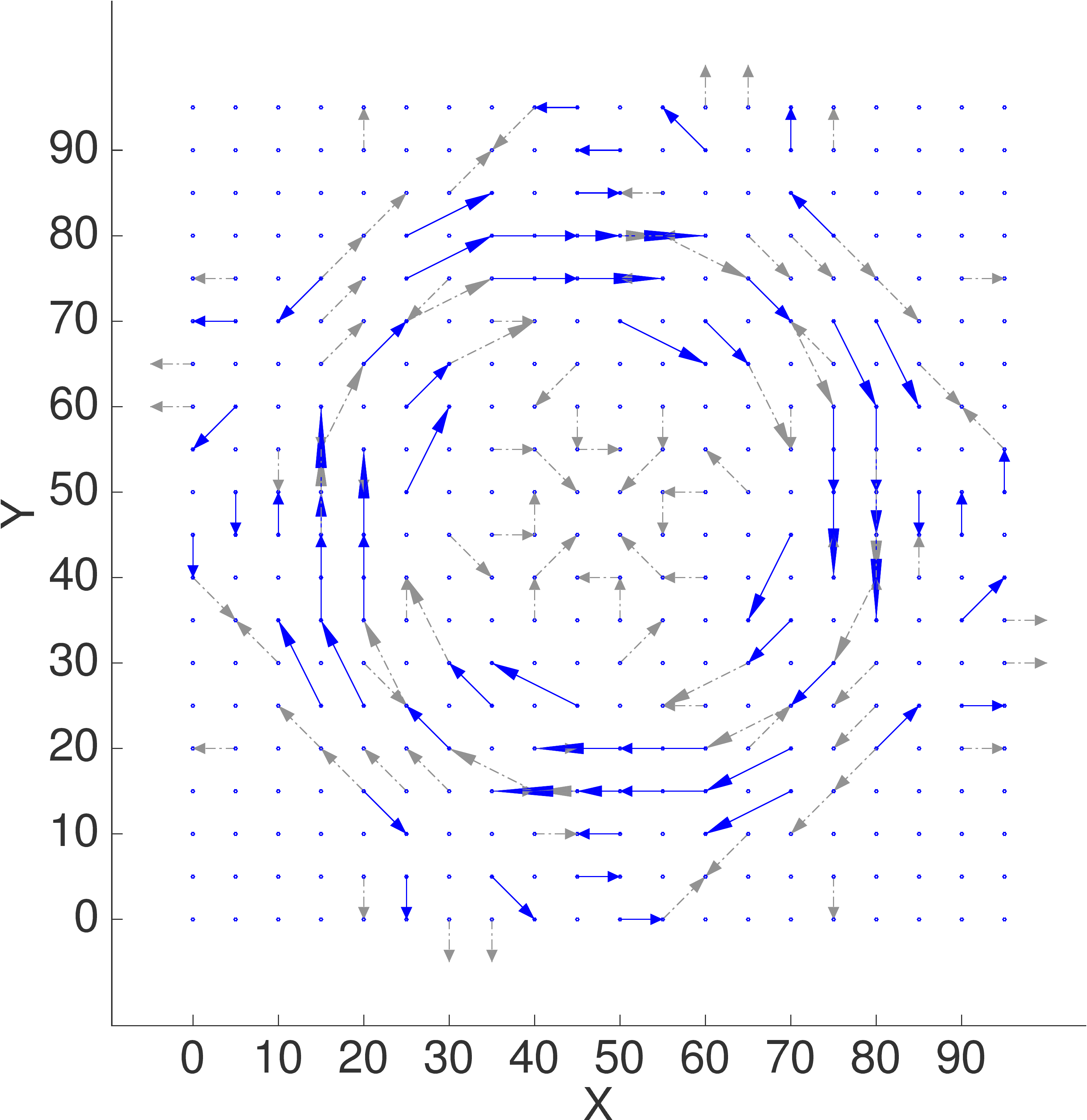}
\hspace*{0.5cm}
\includegraphics[width=4.5cm,angle=0]{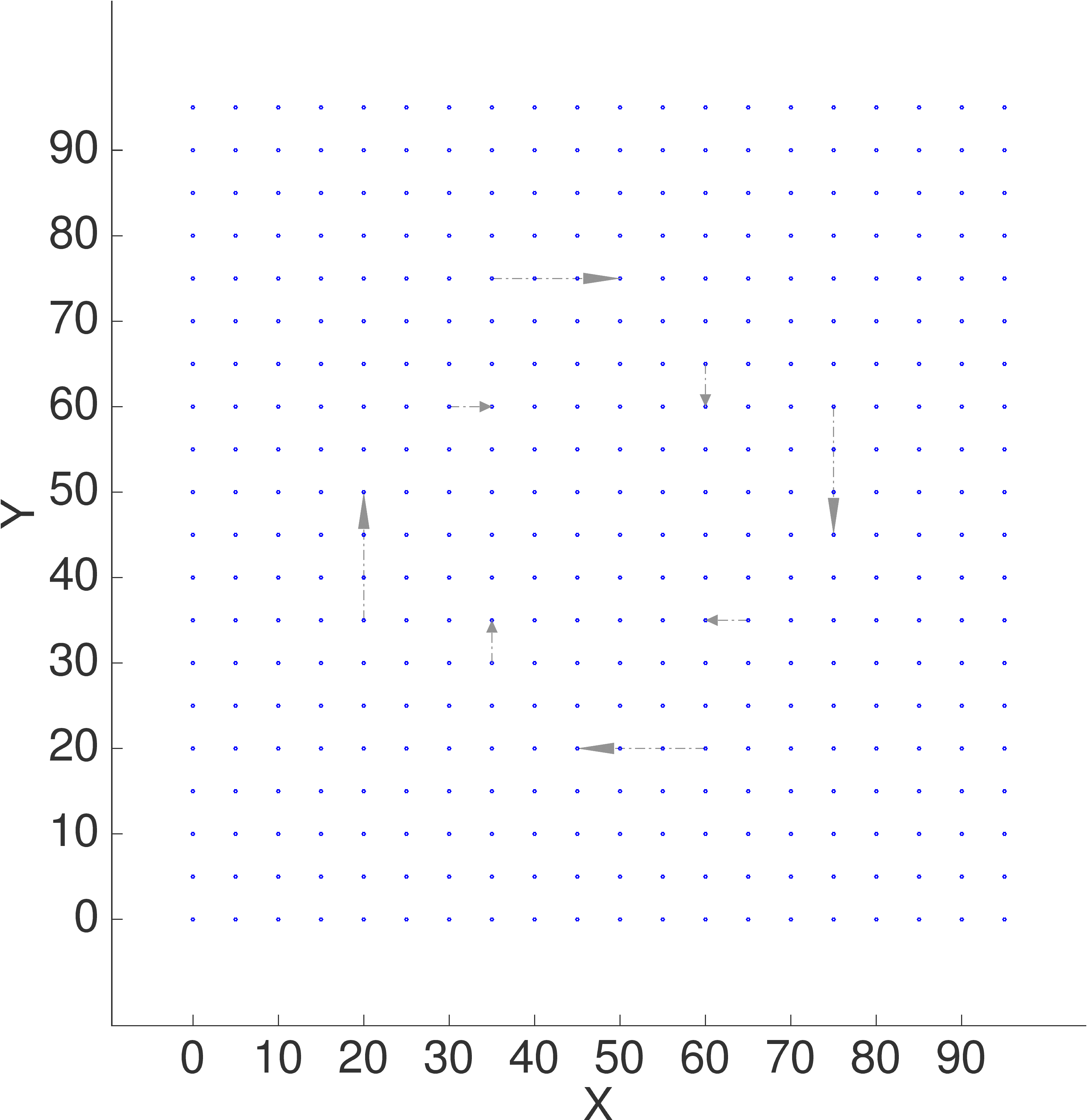}
}
\centerline{$T = 0$  \hspace*{3.5cm} $T = 10 \Delta t$  \hspace*{3.5cm} $T = 20 \Delta t$}
\centerline{(b) Inter edges}
\vspace*{0.3cm}
\centerline{ 
\includegraphics[width=7.9cm,angle=0,clip]{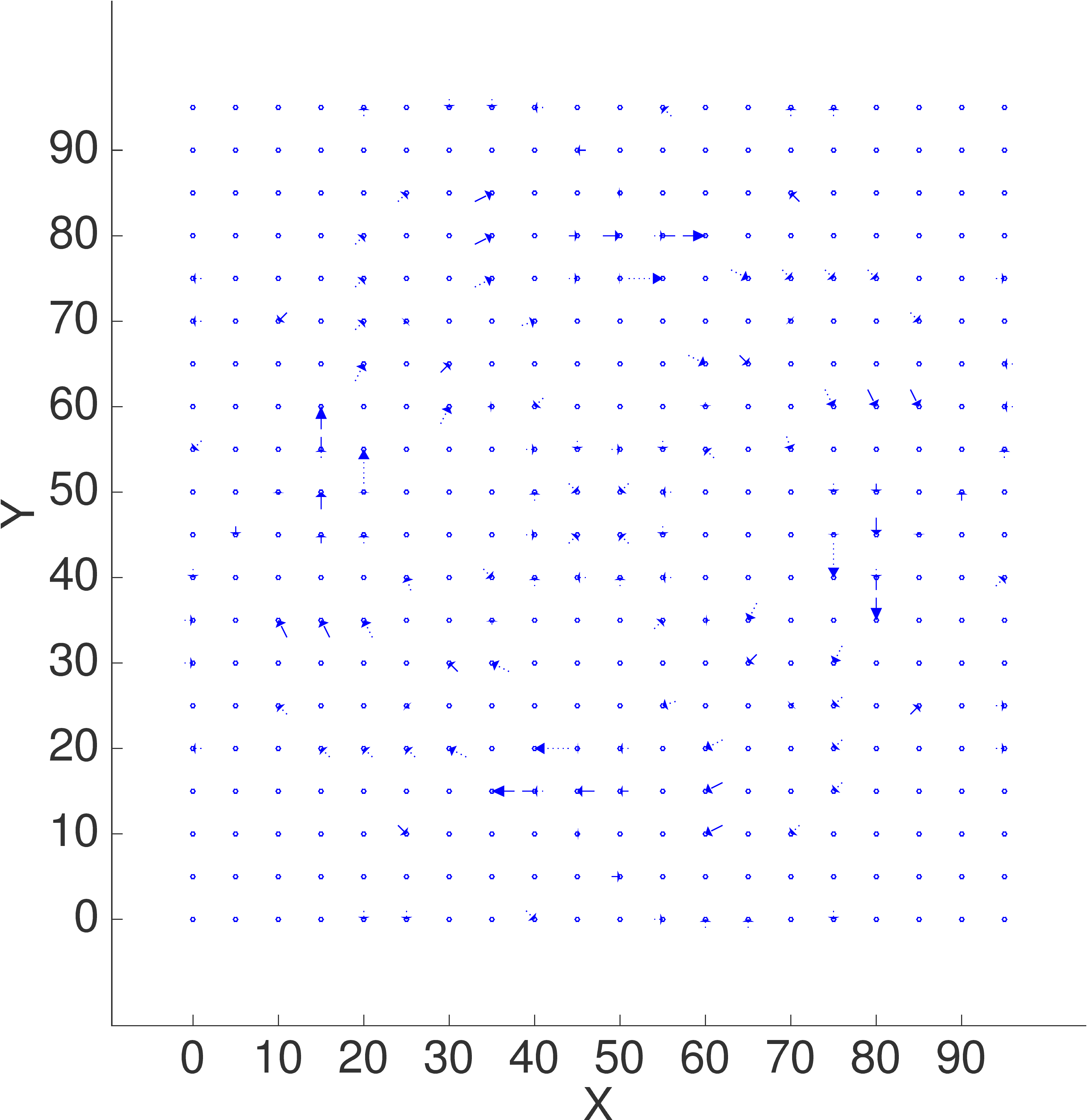}
\hspace*{0.2cm}
\includegraphics[width=7.9cm,angle=0,clip]{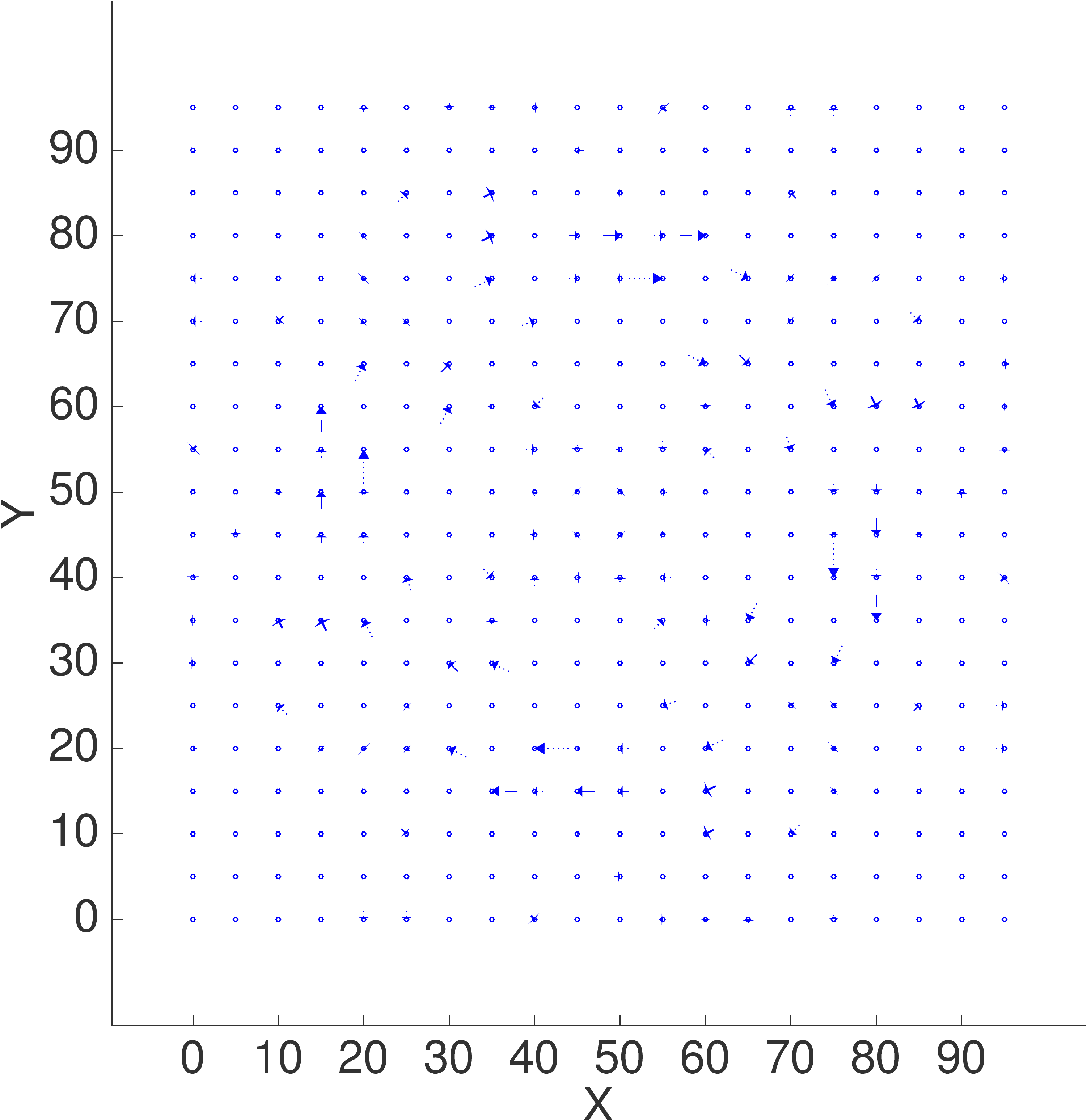}
}
\centerline{(c) Velocity estimated {\it without} intra edges \hspace*{1.0cm} (d) Velocity estimated {\it with} intra edges}
\caption{Results for Scenario 1 with concurrent edges allowed and $M=10$.
\label{scenario_2_M_10_with_concurrent_fig}}
\end{figure*}
%

The last set of results discussed here briefly investigates
what happens in the complex scenarios if we increase signal speed, 
i.e.\ if signals travel {\it several grid points per time step}. 
Since numerical stability prevents us from increasing the travel speed beyond one grid point
per time step in the numerical calculations, 
we instead reduce the temporal resolution {\it after} the numerical calculations
are finished, by keeping only data for every $M$th time step of the numerical calculations.
Figures \ref{scenario_2_M_4_with_concurrent_fig} and \ref{scenario_2_M_10_with_concurrent_fig} show the results for Scenario 1 for $M=4$ and $M=10$,
which can be compared directly to the results for $M=1$ shown in Fig.\ \ref{scenario_1_with_concurrent_fig}.
It is clear that for increasing $M$ the number of intra edges declines rapidly.
Much more interesting, however, is the drastic change to the concurrent 
inter edges. For $M=1$ the concurrent edges only occur in areas with zero advection velocity.
On the other extreme, for $M=10$, many of the concurrent edges actually represent high speed advection connections - an effect we had expected to see, 
but were unable to confirm in simulation results until now!
For $M=4$ we see a mixture of effects, i.e.\ some of the concurrent edges are due to diffusion (recognizable from the connectivity along the grid lines to all neighbors) 
and some represent high-speed advection (recognizable by their alignment with the pattern 
for directed intra edges).
Since concurrent edges are not included in the velocity estimates, as information from directed
inter edges ($T>0$) move to concurrent inter edges ($T=0$) for increasing $M$, 
that information no longer makes
it into the velocity plots.  Thus the resulting velocity estimates are getting weaker, a bit for $M=4$, and much more so for $M=10$.
In summary, these results demonstrate the varying roles of concurrent inter edges, and 
the importance of not simply dropping concurrent edges without analyzing first 
which roles they play and what they may tell us about the underlying dynamical mechanisms,
especially in cases where the velocity estimates are extremely sparse.

\section{Computational Time}
\label{computational_effort_sec}

Almost as interesting as the actual connectivity results for the different scenarios
presented in the previous sections 
is the {\it tremendous} difference in 
calculation time just based on the input data and parameter choices.
To demonstrate this Table \ref{calculation_time_table} shows results for the simple scenarios of pure 
diffusion and pure advection discussed in Section \ref{simple_scenarios_sec}.
In all cases we have the same number of variables (100 grid points $\times$ 20 tiers = 2,000 graph nodes), and identical temporal constraints 
(an effect cannot occur before its cause, i.e.\ we allow concurrent edges).
The {\it only difference} is whether the simulation data comes from diffusion or 
advection and what type of messages we feed in, so purely the physics of the problem.

As seen in Table \ref{calculation_time_table}, causal discovery is always drastically 
faster (at least 50 times) for advection data than for diffusion data.
Note for advection how quickly the number of remaining edges drop down to just a few thousands
by the time the CI tests of 
order\footnote{Order refers here to the order of the conditional 
	independence tests used in the {\it PC stable} algorithm, i.e.\ the number of 
	nodes we condition on in the conditional independence tests.  
	({\it PC} as well as {\it PC stable} perform the test in increasing order, i.e.\ first 
	all independence tests of order 0 are performed, then 1, then 2, etc.)}
0 are completed, i.e.\ by just using simple pair-wise correlation.  
This means that for advection the algorithm very quickly 
narrows down the possibilities to a small set of connections, 
which drastically reduces computational cost. 
In contrast the diffusion case shows a very different behavior.  
Since signals diminish quickly with diffusion,
the algorithm is not able to narrow down the possibilities early on 
to a small set of connections.  
Even when using IC peak signals there are still over 100,000 edges left after all order 1 tests are completed, while the final number of edges is 
6,300\footnote{Note 
	that the final order considered is actually a poor indication
	of computational effort.  In one advection case the highest order considered is 22 and 
	the calculations require less than 14 minutes, while in one diffusion 
	case the highest order is 12 and the algorithm still requires over 12 hours.}.
In summary the results clearly demonstrate that advection is much easier to track
than diffusion, both in terms of time and robustness to signal strength.
Namely, advection signals are quick to be found and can be found even if no strong 
signals are sent into the system, only noise.

\begin{table}
\begin{center}
\begin{tabular}{|p{1.6cm}|p{3.0cm}||r|r|r|p{1.4cm}|}
\hline
{\bf Type} & {\bf Input signal} & {\bf Order} & {\bf \# CI tests}  & {\bf \# edges left} & {\bf Total} \\
 &  & of CI tests & performed & after completion & {\bf time} \\
\hline\hline
Diffusion & IC peak & 0 & 1,999,000 & 1,927,800 &  \\
\cline{3-5} & (Message Type 1) & 1 & 1,313,986,814  & 101,100 & \\
\cline{3-5} & & 2 & 198,823,873 & 22,650 & \\
\cline{3-5} & & 3 &  21,759,968 & 9,150 & \\
\cline{3-5} &  & 8 & & 6,300 & 1.7 min \\
\hline\hline
Diffusion & Single-point noise & 0 & 1,999,000 & 1,996,774 &  \\
\cline{3-5} & (Message Type 2) & 1 &  3,432,381,850 & 1,120,533 & \\
\cline{3-5} & & 2 & 266,615,933,633 & 86,128 & \\
\cline{3-5} & & 3 & 12,085,420,716 & 16,417 & \\
\cline{3-5} &  & 13 & & 12,949 & 5.6 hrs \\
\hline\hline
Diffusion & All-point noise & 0 & 1,999,000 & 1,961,485 &  \\
\cline{3-5} & (Message Type 3) & 1 & 3,386,500,452 & 1,336,309 & \\
\cline{3-5} & & 2 & 483,241,883,491 & 164,025 & \\
\cline{3-5} & & 3 & 58,994,333,551 & 31,271 & \\
\cline{3-5} &  & 12 & & 12,225 & 12.1 hrs \\
\hline
\\
\hline
Advection & IC peak (w.\ small noise) & 0 & 1,999,000 & 19,000 &  \\
\cline{3-5} & (Message Type 1) & 1 & 245,912 & 1,969 & \\
\cline{3-5} & & 2 & 65 & 1,969 & \\
\cline{3-5} & & 3 & 0 & 1,969 & \\
\cline{3-5} &  & 3 & & 1,969 & $< 1$ sec \\
\hline\hline
Advection & Single-point noise & 0 & 1,999,000 & 145,481 &  \\
\cline{3-5} & (Message Type 2) & 1 &  5,838,246  & 4,138 & \\
\cline{3-5} & & 2 & 33,847 & 3,065 & \\
\cline{3-5} & & 3 & 15,493  & 3,039 & \\
\cline{3-5} &  & 9 & & 3,027 & 1 sec \\
\hline\hline
Advection & All-point noise & 0 & 1,999,000 & 633,393 & \\
\cline{3-5} & (Message Type 3) & 1 & 26,330,092  & 3,334 & \\
\cline{3-5} & & 2 & 801,173 & 3,195 & \\ 
\cline{3-5} & & 3 &  4,243,287 & 3,090 & \\
\cline{3-5} &  & 22 & & 2,817 & 13.8 min \\
\hline
\end{tabular}
\end{center}
\caption{Computational time of {\it PC stable} for pure diffusion and pure advection.  Listed are the number of conditional independence tests performed by {\it PC stable} for $order=0,1,2,3$ and how many edges
are left at the end of those tests.  For each experiment the last row lists the maximal order 
considered by {\it PC stable}, the final number of edges of the skeleton and the total time 
to calculate the skeleton. \label{calculation_time_table}}
\end{table}

Finally, Table \ref{calculation_time_concurrent_table} compares some results 
for different scenarios with or without concurrent edges allowed. 
For that comparison 
we are interested in both the computational effort and the 
final number of edges, which indicates the model complexity.
Excluding concurrent edges means that we include {\it more} prior knowledge, 
namely that nodes may not be connected to other nodes within the same time slice.
One may think that this additional knowledge, which limits the edges that can be considered,
would always reduce computational time. 
However, over the years we have often found the opposite to be true.  
Namely, if the prior knowledge forbids edges that are crucial to the model, 
then those edges may be replaced by a large number of substitute edges and
the final model is more complex, i.e.\ containing a larger number of connections.
Furthermore, while lower computational time indicates a simpler model, which is usually preferred,
one has to be more careful about interpreting the final number of edges. 
An increase in the number of edges may indicate either an unnecessary increase 
in complexity, or it may indicate an increased sensitivity, i.e.\ that more 
valid connections are picked up.  
Which one is the case needs to be determined in each case. 
Table \ref{calculation_time_concurrent_table} lists the number of edges and computational 
time for several different scenarios.  For the case where concurrent edges are allowed,
the numbers in parentheses split the total number of edges into 
the number of concurrent plus nonconcurrent edges.
For the pure advection scenario ($M=1$) the results are identical whether concurrent edges
are allowed or not, since no concurrent edges are found either way.  This is also true
for $M=2$ and $M=4$ (not shown in table).
For Complex Scenarios 1-3 the computational effort and number of edges are extremely 
similar when concurrent edges are allowed vs.\ forbidden.  
A possible explanation is that 
all complex scenarios are dominated by advection at most grid points,
thus behave more like pure advection scenarios. 
The most interesting case is pure diffusion. 
While there is not much difference for $M=1$, 
computational time starts to increase when concurrent edges are forbidden
for $M=2$ and becomes overwhelming for $M=4$.  (In fact, the calculations 
for $M=4$ never finished - we stopped them after 2 weeks.)
However, while computational time is much higher when concurrent edges are forbidden,
we did not see any case (yet) where 
the actual number of edges increases drastically - in fact so far the total number of 
edges was always lower, with the number of non-concurrent edges going up slightly.  
It just seems to take longer - in some cases prohibitively longer - 
to calculate the model when concurrent edges are forbidden. 
However, one needs to keep in mind that the computational time is likely 
to be quite different for other methods, such as 
Gaussian models or Granger causal methods.
Clearly, more research needs to be done on the advantages and disadvantages
of including concurrent edges in the modeling process - and on the exact role 
of the concurrent edges in the final model.


\begin{table}
\begin{center}
\begin{tabular}{|l||r|r||r|r|}
\hline
 & \multicolumn{2}{c||}{\bf Concurrent allowed} & \multicolumn{2}{c|}{\bf Concurrent forbidden}\\
\cline{2-5}
{\bf Scenario}    & {\bf \# edges} & {\bf time} & {\bf \# edges} & {\bf time}\\
    & (conc. + non-conc.) &  &  & \\
\hline
\hline
Diffusion (M=1) & 6,300 (800 + 5,500) & 96 sec & 5,700 & 106 sec
\\ \hline
Diffusion (M=2) & 8,400 (1,400 + 7,000) & 2.6 min & 8,200 & 31.4 min
\\ \hline
Diffusion (M=4) & 6,000 (1,500 + 4,500) & 14.2 min & $\le$ 9,500 & $\ge$ 14 days
\\ \hline \hline
Advection (M=1) & 1,969 (0 + 1.969) & $<$ 1 sec & 1,969 & $<$ 1 sec
\\ \hline \hline
Complex Scenario 1 & 24,050 (2,742 + 21,308) & 1.7 hrs & 22,520 & 1.6 hrs
\\ \hline
Complex Scenario 2 & 14,304 (2,424 + 11,880) & 1.4 hrs & 13,392 & 1.3 hrs 
\\ \hline
Complex Scenario 3 & 18,696 (2,438 + 16,258) & 1.7 hrs & 17,943 & 1.6 hrs
\\ \hline
\end{tabular}
\end{center}
\caption{Computation time and final number of edges when concurrent edges
are allowed vs.\ forbidden.  All results are for single peak initial conditions.
(Given in parentheses is the number of concurrent, plus the number of nonconcurrent
edges.) \label{calculation_time_concurrent_table}}
\end{table}

\section{Discussion and Future Work}
\label{future_sec}

The scenarios provided in this article provide a very rich playground to test a variety of 
structure learning algorithms on spatio-temporal data.
The properties of the graph structure to be discovered varies tremendously 
with the physical process chosen to generate the data (e.g.\ diffusion vs.\ advection),
but also with algorithm choices, such as whether to allow concurrent edges or not.
It will be very interesting to compare results for alternative methods of structure learning
for these scenarios, such as Granger causal models and Gaussian models.
In particular we hope that comparing results from Granger, Gaussian and Graphical models
will help us gain a better intuitive understanding of the inherent differences
between those algorithms, in particular of their strengths and weaknesses for 
such geoscience applications.  Some of the algorithms are expected to be much faster than
the graphical model approach used here, 
but are their results comparable in quality to the ones obtained here?  
Is their sensitivity, accuracy and robustness better or worse
than for the algorithm discussed here?
Are some of the alternatives maybe better for some types of processes 
(e.g.\ high-speed signals or maybe scenarios where concurrent edges can be ignored) and worse for others?
To the best of our knowledge the three types of algorithms have never been rigorously compared 
for structure learning from spatio-temporal data 
and these scenarios provide a perfect platform for comparison. 

One of the most important lessons we learned about the interpretation 
of the results from the current approach concerns the concurrent edges. 
The {\it different roles that concurrent edges can play} in this context 
are fascinating and clearly deserve further study in the future. 
Concurrent inter edges can occur for reasons we did not realize before. 
We had assumed that concurrent inter edges always represent extremely fast connections, but they also occur for other reason, for example when diffusion is dominant, when there are abrupt changes in neighboring advection velocities, or other cases in which it is very hard to model the advection by regular inter connections.  
Fortunately, the signatures for the concurrent edges look different for the above cases, a fact  we can use to distinguish between them:\\
(1) Concurrent edges representing connections with very high velocity stand out by occurring only in one direction at each point and aligning with the general patterns seen for inter edges for $T>0$.\\ 
(2) Concurrent edges representing diffusion connect each such location to all of its closest neighbors in the grid.  These are easy to spot.\\
(3) Other concurrent edges are usually weak, and do not align in direction with the inter connections for $T>0$.

Applying this new knowledge to the graphs we obtained from real-world data (Figure \ref{Fekete_800_fig}), we see that the great majority of concurrent edges identified there
are likely to be due to diffusion.  This matches expert knowledge about the atmosphere, which is quite diffusive, especially in the lower troposphere and in the boundary layer due to the prevalence of mechanically and thermally excited turbulent eddies.
Thus we have finally solved the two-year old mystery of the many concurrent edges that occur when using this real-world atmospheric data, especially for increased spatial resolution. 

We also learned more about the primary factors affecting the quality of results,
namely (1) directions and (2) signal speed.
Connections with directions along the horizontal/vertical grid lines are easiest to identify and represent, while other connections may be distorted in direction to align with the grid, or may be represented by combinations of connections, which are harder to interpret.  In other words, the grid introduces significant bias.
(Interestingly, in our real-world applications the bias is less prominent, since we use Fekete grids that are based primarily on hexagons, rather than rectangles, thus provide 6 direction angles, rather than 4, to choose from.)
%
Concerning speed of connections, 
it is easiest to identify signals with a speed of at least one grid point per time step.  
Signals slower than that may not show up in the inter edges at all, while 
extremely fast signals may show up as concurrent inter edges, rather than regular inter edges, i.e.\ they have no direction associated with them. 
%
That being said, we were very pleased with the overall results, as the method 
is very capable of identifying the primary patterns of the advection velocity fields.
So far there is no single plot though that represents all the results for given data.  
Velocity plots are good in some cases, especially when connection density is high, while the original intra and inter plots are particularly important when connection density is low.  
In the latter case one may not even want to consider velocity plots. 
If velocity plots are considered, then Type 1 velocity plots provide very high sensitivity, useful 
to identify all connections and provide good direction estimates, while 
Type 2 velocity plots are better at identifying the actual speed of connections.

The simulation framework has proven to be invaluable to learn about the typical 
information flow signatures of the specific processes of advection and diffusion, which will
help us in interpreting the results obtained from real-world data in the future.
Even though much more research needs to be
done on specific interpretation guidelines, these new-found insights
already provide a foundation to the use and interpretation 
of spatio-temperal structure learning for new geoscience applications.

\remove{
We plan to extend the simulation framework to including other types of dynamical processes 
in the future. 
We also hope that by making all of the data sets available to other researchers
we provide a much needed   
benchmark to compare the results of different types of spatio-temporal structure learning algorithms
for processes similar to real-world geoscience data.   
}

%% file: Causal_discovery_testbed_TEXT_acknowledge.tex

\acks{Support for this work is provided by two grants of the NSF Climate and Large-Scale Dynamics (CLD) program, namely Grant AGS-1147601 awarded to Yi Deng, and a collaborative grant (AGS-1445956 and AGS-1445978) awarded to both authors.}


%% file: Causal_discovery_testbed_TEXT_appendix.tex
\newpage

\appendix

\section*{Appendix A: How to Generate Temporal Models}

This section briefly explains the approach 
first introduced by 
\cite{ChDaGl:2005}, 
which allows us to incorporate time explicitly in the model by 
adding lagged variables to the model,
i.e.\ it is a method to learn a dynamic graphical model, rather than a static one.
We briefly outline it here, since 
this approach does not seem to be widely known, 
and certainly has not yet received the attention it deserves.

Given time series data for $N$ variables, $X_1, \ldots, X_N$, the basic approach is as follows:
\begin{enumerate}
  \setlength{\leftmargin=-1.5in}  
\item
  \setlength{\leftmargin=-1.5in}  
   Choose the distance, $D$, between time slices, e.g. $D= 3$ time steps.
\item
   Choose number of time slices to include, $S$.  
\item
   Define lagged variables for all $i=1, \ldots, N$ and $s \in [0,S-1]$:\\ 
           $\qquad X_i^s = X_i  \mbox{ lagged by $s$ time slices}$.\\
   This results in a total of $S \cdot N$ variables, which form the {\it nodes} of
   the temporal graph.
\item
   Add temporal constraints: Causes can only occur before or at the same time as their effect, i.e.\
   $X_i^s \mbox{ can be a cause of }  X_j^t  \mbox{ only if }  s \le t.$ 
\end{enumerate}
When setting up the temporal constraints in Step 4
we can choose to either {\it allow} or to {\it a priori exclude} concurrent edges.
To allow concurrent edges we use the constraints as described above.
To a priori exclude concurrent edges we simply replace the condition $s \le t$ by 
$s < t$.
Using this procedure we can express the temporal model with $N$ time series 
variables and $S$ time slices 
as a standard static problem with $(N \cdot S)$ variables, plus temporal constraints.
The temporal constraints can be incorporated in {\it PC stable} as prior knowledge, so that
the standard algorithms can now be used to provide a temporal model.
The price we pay for this though is much higher computational complexity, 
since we are now dealing with $(N \cdot S)$, rather than $N$ variables.
\remove{
It is a little known fact that use of the lagged variables violates one of the 
assumptions of constraint-based
structure learning, 
namely that the probability distributions should be independent of each other 
(this is necessary for faithfulness).
So far it seems that this violation does not affect the method at all, since
it works well even in spite of this violation.
Nevertheless, this issue should be studied further,
to understand why it works so well and to ensure that this is always the case. 
However, that topic is beyond the scope of this paper. [CHECK WITH ARINDAM - DOES HIS RECENT PAPER SHED SOME LIGHT ON THIS?]
}
Finally, as discussed in (\cite{EbDe:2014ICMLA}), 
proper initialization of the first time slices 
is a critical issue, but one that can be resolved easily by calculating the model 
for more time slices than needed and then discarding the first few time slices in the results.


\section*{Appendix B: Advection-Diffusion Equations and Their Numerical Implementation}

This section briefly discusses the partial differential equations governing the 
advection-diffusion process and the numerical implementation we use.
For more general information on advection and diffusion, 
see any textbook on heat and mass transfer, 
such as \cite{BeLaInDe:2011}, or for a book solely focusing on advection and diffusion
see \cite{bennett:2012book}. 
For a shorter introduction check out 
the excellent course material provided by 
Dietmar Muller at the University of Sydney
for a self-contained tutorial on the advection diffusion equation 
and its implementation in Matlab
(materials for Practices 4 and 5 at\\ {\small
\verb+ftp://www.geosci.usyd.edu.au/pub/dietmar/GEOS3104/Pracs/+}).

Let us start out with the one-dimensional version, which is simpler,
and then generalize it to two dimensions.
The advection diffusion equation in one dimension can be described by the following
partial differential equation:
\begin{equation}
    \frac{\partial f}{\partial t} + V(x) \frac{\partial f}{\partial x} = \kappa_x \frac{\partial^2 f}{\partial x^2},
\end{equation}
where $f(x,t)$ can be interpreted as the temperature of the fluid at location $x$
over time,
$\kappa_x$ is the diffusion coefficient, and $V(x)$ is the scalar velocity of the fluid.
The first term, $\left( \frac{\partial f}{\partial t} \right)$ describes the change of temperature 
at any location over time.  The second term, $\left( V(x_n) \frac{\partial f}{\partial x} \right)$, 
is the advection term, describing the {\it sideways motion} of the signal due to advection.  
Lastly, the term $\left( \kappa_x \frac{\partial^2 f}{\partial x^2} \right)$ is the diffusion term,
describing the {\it spreading} of the signal due to diffusion. 

Likewise, the two-dimensional version of the advection-diffusion equation is as follows:
\begin{equation}
    \frac{\partial f}{\partial t} + \left( V_x \frac{\partial f}{\partial x} 
   + V_y \frac{\partial f}{\partial y} \right)
= \left( \kappa_x \frac{\partial^2 f}{\partial x^2} + \kappa_y \frac{\partial^2 f}{\partial y^2} \right),
\label{adv_dif_pde}
\end{equation}
where $V_x$ and $V_y$ are the $x$ and $y$-component, respectively,
of the velocity, ${\bf V}(x,y)$, which is now a vector.
%

{\bf Numerical implementation:} 
For the numerical implementation we use what is known as the 
{\it First Order Upwind Scheme}, which again is explained first 
for the one-dimensional case.
We calculate the temperature, $f_{i+1}(x_n)$, 
from $f$ at the previous time step using an explicit first-order {\it upwind} scheme, 
as follows:
\begin{eqnarray}
   f_{i+1}(x_n) &=& f_i(x_n) - V(x_n) \; \Delta t \; \frac{\partial f_i(x_n)}{\partial x} + \kappa_x \; \frac{\partial^2 f_i(x_n)}{(\partial x)^2},
\label{advection_1D_eq}
\end{eqnarray}
where the partial derivatives are calculated as 
\begin{eqnarray}
   \frac{\partial f_i(x_n)}{\partial x} = 
      \left\{ 
		\left( f_i(x_n) - f_i(x_{n-1}) \right) / \Delta x
		\quad \mbox{if} \; V_x(x_n) \ge 0
        \atop 
		\left( f_i(x_{n+1})- f_i(x_n) \right) / \Delta x
		\quad \mbox{otherwise.} \hspace*{0.5cm}
        \right.
   \label{partial_upwind_1D_eq}
\end{eqnarray}
and
\begin{eqnarray}
   \frac{\partial^2 f_i(x_n)}{(\partial x)^2} = 
	\frac{ f_i(x_{n-1}) - 2 f_i(x_n) + f_i( x_{n+1}) } { (\Delta x )^2 } 
\end{eqnarray}
The partial derivative in Eq.\ (\ref{partial_upwind_1D_eq})
is defined such that when the velocity is positive, i.e.\ 
the signal is moving to the right in the grid,  
the temperature difference to the previously visited (further left) 
grid point at $x_{n-1}$ is used.  
Likewise, for motion to the left, the previously visited (further right) grid point at
$x_{n+1}$ is used.  Thus we are always using a value for the partial derivative
that is {\it upwind} 
(or {\it upstream}) from the direction the signal is traveling in.

%
The two-dimensional version of the upwind scheme,
where $f(x_j,y_k)$ denotes the temperature at grid points with 
coordinates $(x_j,y_k)$,
is 
\begin{eqnarray}
    f_{i+1}(x_j,y_k) &=& 
      f_i(x_j,y_k) 
      - \left( V_x(x_j,y_k) \; \Delta t \right) \; \frac{\partial f_i(x_j,y_k)}{\partial x}
      - \left( V_y(x_j,y_k) \; \Delta t \right) \; \frac{\partial f_i(x_j,y_k)}{\partial y}
\nonumber\\
      & & + \; \kappa_x \; \frac{\partial^2 f_i(x_n)}{(\partial x)^2} 
      + \kappa_y \; \frac{\partial^2 f_i(x_n)}{(\partial y)^2},
\end{eqnarray}
where, analogous to the one-dimensional upward scheme, the partial derivatives are estimated as follows
\begin{eqnarray}
   \frac{\partial f_i(x_j,y_k)}{\partial x} =  
        \left\{ 
		\left( f_i(x_j,y_k) - f_i(x_{j-1},y_k) \right) / \Delta x
		\quad \mbox{if} \; V_x(x_j,y_k) \ge 0
        \atop 
		\left( f_i(x_{j+1},y_k)-f_i(x_j,y_k) \right) / \Delta x
		\quad \mbox{otherwise.} \hspace*{1.1cm}
        \right.
\\[0.3cm]  
   \frac{\partial f_i(x_j,y_k)}{\partial y} =  
        \left\{ 
		\left( f_i(x_j,y_k) - f_i(x_j,y_{k-1}) \right) / \Delta y
		\quad \mbox{if} \; V_y(x_j,y_k) \ge 0
        \atop 
		\left( f_i(x_j,y_{k+1})-f_i(x_j,y_k) \right) / \Delta y
		\quad \mbox{otherwise.} \hspace*{1.1cm}
        \right.
\end{eqnarray}
and
\begin{eqnarray}
   \frac{\partial^2 f_i(x_j,y_k)}{(\partial x)^2} &=& 
	\frac{ f_i(x_{j-1},y_k) - 2 f_i(x_j,y_k) + f_i(x_{j+1},y_k) } {( \Delta x )^2}, 
\\
   \frac{\partial^2 f_i(x_j,y_k)}{(\partial y)^2} &=& 
	\frac{ f_i(x_j,y_{k-1}) - 2 f_i(x_j,y_k) + f_i(x_j,y_{k+1}) } {( \Delta y )^2}.
\end{eqnarray}

\section*{Appendix C: Simulation Parameters}

Parameters that are identical for all simulations are listed below, while
parameters specific to each scenario are listed in the corresponding sections.
Furthermore, data files for all scenarios, including lists of coordinates, 
advection velocity fields and 
resulting time series data, are provided on the supplemental website\\
({\small
\verb+http://www.engr.colostate.edu/~iebert/DATA_SETS_CAUSAL_DISCOVERY/+}).


The range of $x$ as well as $y$ values is always $[0,100]$ m.  Thus for a 
10x10 grid of squares
we have $\Delta x= \Delta y = 10$ m, 
for a 20x20 grid we have $\Delta x= \Delta y = 5$ m.
The default diffusion parameter is $\kappa = 1$ $m/s^2$ and 
the default maximal advection velocity norm is $V_{max}=1$ m/s.
For the causal discovery algorithm 
we use 20 tiers throughout all experiments in this article, 
i.e.\ for each original variable we create 20 lagged variables with lag
$0, \Delta t, \ldots, 19 \Delta t$.  Once results are obtained the first 2 time 
slices are discarded to assure proper initialization (see \cite{EbDe:2014ICMLA}).  
Regardless of which type of information we feed into the system, whether peak initial 
conditions or prior noise forcing, we perform one run for each grid point,
then move on to the next grid points, and at the end 
concatenate all runs in the data files.  (If one is interested in separating 
the runs in the data files, the length of each 
run is given by dividing the total 
number of samples divided by the number of grid points used.)
The number of samples per run is chosen such that the total number of samples 
- when using 20 tiers - is at least 5,000.
(Note that to get $N$ samples for 20 tiers per run, 
one needs $(N+19)$ non-lagged samples per run.)
The data files on the website contain the time series for all grid points, 
without the lagged variables, 
thus each file contains significantly more than 5,000 samples 
(the exact number depends on $M$).

For the Fisher-Z tests of the {\it PC stable} algorithm we use a significance level 
of $\alpha=0.05$.  That value is relatively low ($\alpha=0.1$ is often suggested 
as default value for such algorithms), and is thus a conservative estimate, i.e.\ 
increasing the value of $\alpha$ would yield more connections.
We have found, however, that increasing the value of $\alpha$, even to values as high as
$\alpha=0.5$, makes surprisingly little difference for the results, but often slows
the algorithm down immensely.  Thus we have found 
using $\alpha=0.05$ to be a good choice.

%% file: Causal_discovery_testbed_MAIN.bbl
\begin{thebibliography}{27}
\providecommand{\natexlab}[1]{#1}
\providecommand{\url}[1]{\texttt{#1}}
\expandafter\ifx\csname urlstyle\endcsname\relax
  \providecommand{\doi}[1]{doi: #1}\else
  \providecommand{\doi}{doi: \begingroup \urlstyle{rm}\Url}\fi

\bibitem[Arnold et~al.(2007)Arnold, Liu, and Abe]{ArLiAb:2007}
Andrew Arnold, Yan Liu, and Naoki Abe.
\newblock Temporal causal modeling with graphical granger methods.
\newblock In \emph{Proceedings of the 13th ACM SIGKDD international conference
  on Knowledge discovery and data mining}, pages 66--75. ACM, 2007.

\bibitem[Bennett(2012)]{bennett:2012book}
Ted Bennett.
\newblock \emph{Transport by Advection and Diffusion}.
\newblock Wiley, New York, 1st edition, 2012.
\newblock 640pp.

\bibitem[Bergman et~al.(2011)Bergman, Lavine, Incropera, and
  DeWitt]{BeLaInDe:2011}
Theodore~L. Bergman, Adrienne~S. Lavine, Frank~P. Incropera, and David~P.
  DeWitt.
\newblock \emph{Fundamentals of Heat and Mass Transfer}.
\newblock Wiley, New York, 7th edition, 2011.
\newblock 1076pp.

\bibitem[Chu et~al.(2005)Chu, Danks, and Glymour]{ChDaGl:2005}
Tianjiao Chu, David Danks, and Clark Glymour.
\newblock Data driven methods for nonlinear granger causality: Climate
  teleconnection mechanisms.
\newblock Technical Report CMU-PHIL-171, Dep. of Philos., Carnegie Mellon
  Univ., Pittsburgh, PA, 2005.

\bibitem[Colombo and Maathuis(2012)]{CoMa:2012}
Diego Colombo and Marloes~H Maathuis.
\newblock Order-independent constraint-based causal structure learning.
\newblock \emph{arXiv preprint arXiv:1211.3295}, 2012.

\bibitem[Colombo and Maathuis(2014)]{CoMa:2014}
Diego Colombo and Marloes~H Maathuis.
\newblock Order-independent constraint-based causal structure learning.
\newblock \emph{Journal of Machine Learning Research}, 15:\penalty0 3741--3782,
  2014.

\bibitem[Colombo et~al.(2012)Colombo, Maathuis, Kalisch, Richardson,
  et~al.]{Colombo:2012}
Diego Colombo, Marloes~H Maathuis, Markus Kalisch, Thomas~S Richardson, et~al.
\newblock Learning high-dimensional directed acyclic graphs with latent and
  selection variables.
\newblock \emph{The Annals of Statistics}, 40\penalty0 (1):\penalty0 294--321,
  2012.

\bibitem[Deng and Ebert-Uphoff(2014)]{DeEb:2014GRL}
Yi~Deng and Imme Ebert-Uphoff.
\newblock Weakening of atmospheric information flow in a warming climate in the
  community climate system model.
\newblock \emph{Geophysical Research Letters}, 41\penalty0 (1):\penalty0
  193--200, 2014.

\bibitem[Donges et~al.(2009)Donges, Zou, Marwan, and Kurths]{DoZoMaKu:2009}
J.F. Donges, Y.~Zou, N.~Marwan, and J.~Kurths.
\newblock The backbone of the climate network.
\newblock \emph{Europhysics Letters}, 87:\penalty0 48007 (6pp.), 2009.

\bibitem[Ebert-Uphoff and Deng(2012{\natexlab{a}})]{EbDe:2012GRL}
Imme Ebert-Uphoff and Yi~Deng.
\newblock A new type of climate network based on probabilistic graphical
  models: Results of boreal winter versus summer.
\newblock \emph{Geophysical Research Letters}, 39\penalty0 (19),
  2012{\natexlab{a}}.

\bibitem[Ebert-Uphoff and Deng(2012{\natexlab{b}})]{EbDe:2012JCLI}
Imme Ebert-Uphoff and Yi~Deng.
\newblock Causal discovery for climate research using graphical models.
\newblock \emph{Journal of Climate}, 25\penalty0 (17):\penalty0 5648--5665,
  2012{\natexlab{b}}.

\bibitem[Ebert-Uphoff and Deng(2014)]{EbDe:2014ICMLA}
Imme Ebert-Uphoff and Yi~Deng.
\newblock Causal discovery from spatio-temporal data with applications to
  climate science.
\newblock In \emph{13th International Conference on Machine Learning and
  Applications (ICMLA'14)}, page 8 pp., Detroit, MI, 12 2014.

\bibitem[Kalnay et~al.(1996)Kalnay, Kanamitsu, Kistler, Collins, Deaven,
  Gandin, Iredell, Saha, White, Woollen, et~al.]{Ka:1996}
Eugenia Kalnay, Masao Kanamitsu, Robert Kistler, William Collins, Dennis
  Deaven, Lev Gandin, Mark Iredell, Suranjana Saha, Glenn White, John Woollen,
  et~al.
\newblock The ncep/ncar 40-year reanalysis project.
\newblock \emph{Bulletin of the American meteorological Society}, 77\penalty0
  (3):\penalty0 437--471, 1996.

\bibitem[Kistler et~al.(2001)Kistler, Collins, Saha, White, Woollen, Kalnay,
  Chelliah, Ebisuzaki, Kanamitsu, Kousky, et~al.]{Ki:2001}
Robert Kistler, William Collins, Suranjana Saha, Glenn White, John Woollen,
  Eugenia Kalnay, Muthuvel Chelliah, Wesley Ebisuzaki, Masao Kanamitsu, Vernon
  Kousky, et~al.
\newblock The ncep-ncar 50-year reanalysis: Monthly means cd-rom and
  documentation.
\newblock \emph{Bulletin of the American Meteorological society}, 82\penalty0
  (2):\penalty0 247--267, 2001.

\bibitem[Koller and Friedman(2009)]{KoFr:2009}
Daphne Koller and Nir Friedman.
\newblock \emph{Probabilistic Graphical Models - Principles and Techniques}.
\newblock MIT Press, 1st edition, 2009.

\bibitem[Molkenthin et~al.(2014)Molkenthin, Rehfeld, Marwan, and
  Kurths]{MoReMaKu:2014}
Nora Molkenthin, Kira Rehfeld, Norbert Marwan, and J{\"u}rgen Kurths.
\newblock Networks from flows-from dynamics to topology.
\newblock \emph{Scientific reports}, 4, 2014.

\bibitem[Neapolitan(2003)]{Ne:2003}
R.~E. Neapolitan.
\newblock \emph{Learning Bayesian Networks}.
\newblock Prentice Hall, 2003.

\bibitem[Pearl(1988)]{Pearl:1988}
Judea Pearl.
\newblock \emph{Probabilistic Reasoning in Intelligent Systems: Networks of
  Plausible Interference}.
\newblock Morgan Kaufman Publishers, San Mateo, CA, revised second printing
  edition, 1988.

\bibitem[Runge(2014)]{Runge:2014dis}
Jakob Runge.
\newblock \emph{Detecting and quantifying causality from time series of complex
  systems}.
\newblock PhD thesis, Humboldt-University Berlin, Berlin, Germany, 8 2014.
\newblock Available at
  http://edoc.hu-berlin.de/dissertationen/runge-jakob-2014-08-05/PDF/runge.pdf.

\bibitem[Spirtes and Glymour(1991)]{SpGl:1991}
Peter Spirtes and Clark Glymour.
\newblock An algorithm for fast recovery of sparse causal graphs.
\newblock \emph{Social science computer review}, 9\penalty0 (1):\penalty0
  62--72, 1991.

\bibitem[Spirtes et~al.(1993)Spirtes, Glymour, and Scheines]{SGS:1993}
Peter Spirtes, Clark Glymour, and Richard Scheines.
\newblock Causation, prediction and search.
\newblock \emph{Lecture Notes in Statistics}, 1993.

\bibitem[Steinhaeuser et~al.(2010)Steinhaeuser, Chawla, and
  Ganguly]{StChGa:2010}
Karsten Steinhaeuser, Nitesh~V. Chawla, and Auroop~R. Ganguly.
\newblock Complex networks in climate science: progress, opportunities and
  challenges.
\newblock In \emph{Proceedings 2010 Conference on Intelligent Data
  Understanding}, pages 16 -- 26, 2010.

\bibitem[Tsonis and Roebber(2004)]{TsRo:2004}
A.A. Tsonis and P.J. Roebber.
\newblock The architecture of the climate network.
\newblock \emph{Physics A: Statistical and Theoretical Physics}, 333:\penalty0
  497--504, February 2004.

\bibitem[Tsonis et~al.(2008)Tsonis, Swanson, and Kravtsov]{TsSwKr:2007}
A.A. Tsonis, K.L. Swanson, and S.~Kravtsov.
\newblock A new dynamical mechanism for major climate shifts.
\newblock \emph{Physical Review Letters}, 100\penalty0 (22):\penalty0
  228502--1--4, 2008.

\bibitem[Tsonis et~al.(2006)Tsonis, Swanson, and Roebber]{TsSwRo:2006}
Anastasios~A. Tsonis, Kyle~L. Swanson, and Paul~J. Roebber.
\newblock What do networks have to do with climate?
\newblock \emph{Bulletin- American Meteorological Society}, 87\penalty0
  (5):\penalty0 585--596, 2006.

\bibitem[Yamasaki et~al.(2008)Yamasaki, Gozolchiani, and Havlin]{YaGoHa:2008}
K.~Yamasaki, A.~Gozolchiani, and S.~Havlin.
\newblock Climate networks around the globe are significantly affected by {El
  Ni\~no}.
\newblock \emph{Physical Review Letters}, 100\penalty0 (2):\penalty0
  228501--1--4, June 2008.

\bibitem[Zerenner et~al.(2014)Zerenner, Friederichs, Lehnertz, and
  Hense]{ZeFrLeHe:2014}
Tanja Zerenner, Petra Friederichs, Klaus Lehnertz, and Andreas Hense.
\newblock A gaussian graphical model approach to climate networks.
\newblock \emph{Chaos: An Interdisciplinary Journal of Nonlinear Science},
  24\penalty0 (2):\penalty0 023103, 2014.

\end{thebibliography}
